\documentclass[11pt]{article}

\usepackage[preprint]{acl}

\usepackage{times}
\usepackage{latexsym}

\usepackage[T1]{fontenc}

\usepackage[utf8]{inputenc}

\usepackage{microtype}

\usepackage{inconsolata}

\usepackage{graphicx}

\usepackage[dvipsnames,table]{xcolor}
\usepackage{colortbl}
\usepackage{microtype}
\usepackage{subcaption}
\usepackage{booktabs} %
\usepackage{pifont}
\usepackage{wrapfig}
\usepackage{url}
\usepackage{xspace}
\usepackage{multicol}
\usepackage{multirow}
\usepackage{latexsym}
\usepackage{enumitem}
\usepackage{caption}
\usepackage{hyphenat}
\usepackage{float}
\usepackage{fontawesome5}
\usepackage{makecell}
\usepackage{listings}
\usepackage{amsmath}
\usepackage{amssymb}
\usepackage{mathtools}
\usepackage{amsthm}
\usepackage{bm}
\usepackage{cuted}

\usepackage[capitalize,noabbrev]{cleveref}

\usepackage[most,skins,theorems]{tcolorbox}

\usepackage[textsize=tiny]{todonotes}

\definecolor{CaseBlue}{RGB}{28,82,150}
\definecolor{OkGreen}{RGB}{40,167,69}
\definecolor{WarnOrange}{RGB}{227,170,0}
\definecolor{BadRed}{RGB}{200,45,38}
\definecolor{LightGray}{RGB}{246,246,246}

\newcommand{\datasetname}{GSM8K-V\xspace}

\title{
\makebox[0pt][r]{%
  \smash{\raisebox{-0.5cm}{%
    \includegraphics[width=1.25cm,height=1.25cm,keepaspectratio]{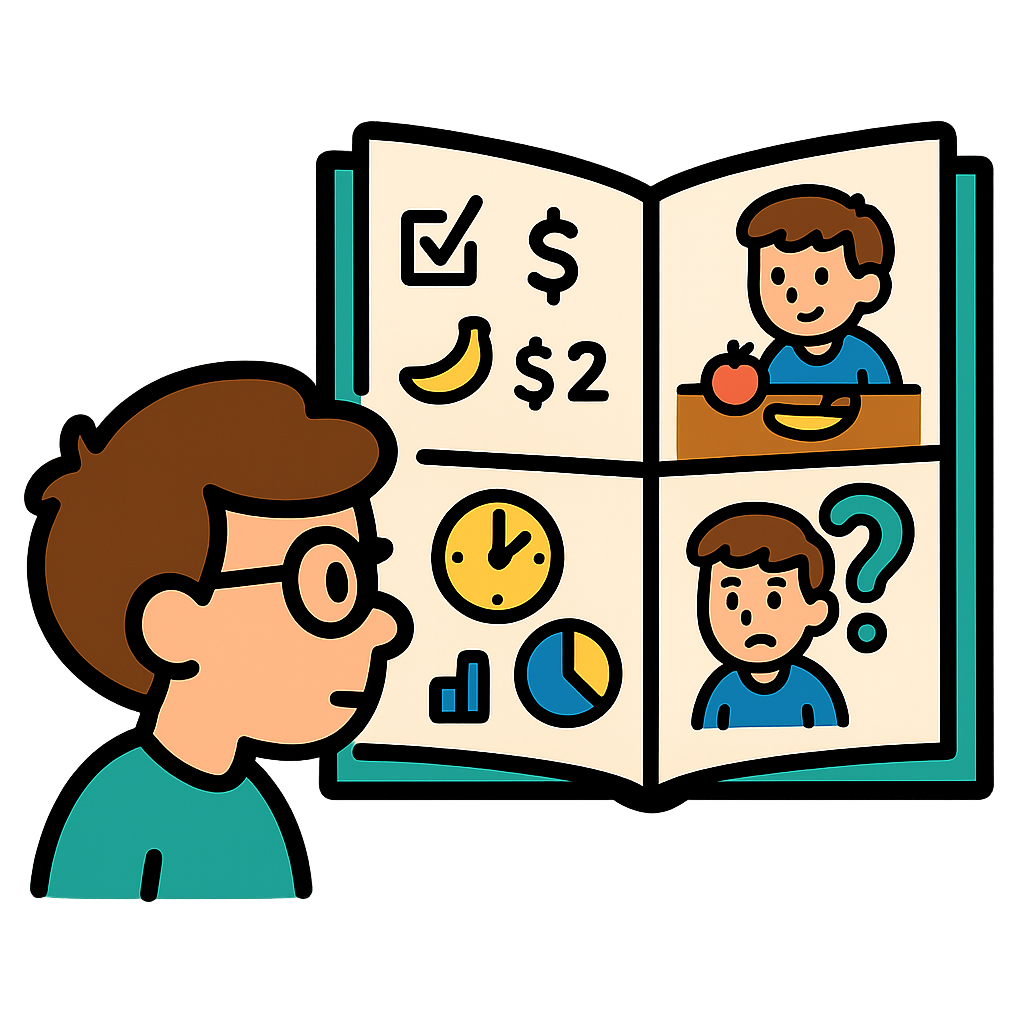}%
  }}%
  \hspace{0.15cm}%
}
\datasetname: Can Vision Language Models Solve Grade School \\
Math Word Problems in Visual Contexts?
}

\author{
  \textbf{Fan Yuan\textsuperscript{1}}\thanks{Equal contribution.},
  \textbf{Yuchen Yan\textsuperscript{1}}\footnotemark[1],
  \textbf{Yifan Jiang\textsuperscript{1}},
  \textbf{Haoran Zhao\textsuperscript{1}},
\\
  \textbf{Tao Feng\textsuperscript{1}},
  \textbf{Jinyan Chen\textsuperscript{1}},
  \textbf{Yanwei Lou\textsuperscript{1}},
  \textbf{Wenqi Zhang\textsuperscript{1}},
\\
  \textbf{Weiming Lu\textsuperscript{1}},
  \textbf{Jun Xiao\textsuperscript{1}},
  \textbf{Yueting Zhuang\textsuperscript{1}},
  \textbf{Yongliang Shen\textsuperscript{1}}\thanks{Corresponding author.}
\\
  \textsuperscript{1}Zhejiang University
\\
  \texttt{\{leofyfan,yanyuchen,syl\}@zju.edu.cn}
  \\[0.3em]
\faGithub\,
\href{https://github.com/ZJU-REAL/GSM8K-V}{\textbf{GitHub}}
\quad
\faGlobe\,
\href{https://zju-real.github.io/GSM8K-V}{\textbf{Project}}
\quad
\faDatabase\,
\href{https://huggingface.co/datasets/ZJU-REAL/GSM8K-V}{\textbf{Data}}
\\[1em]
}

\begin{document}
\maketitle
\begin{abstract}
Mathematical reasoning is a key capability for vision-language models (VLMs), yet current benchmarks mainly evaluate text-based or explicitly symbolic visual inputs.
It remains unclear whether VLMs can reason mathematically when information must be perceived and inferred from images rather than read from explicit symbols.
We introduce GSM8K-V, a benchmark transforming GSM8K into multi-image sequences with semantic equivalence preserved.
By mapping text-based problems into visual form via an automated pipeline and human verification, we curate 1,319 high-quality samples.
In GSM8K-V, quantities must be extracted through visual perception, and reasoning chains must be reconstructed by integrating implicit cues across scenes. 
Evaluation of 34 VLMs reveals a striking modality gap: while most models exceed 90\% on text, the best model achieves only 59\% on GSM8K-V, far below the 91\% human accuracy. Notably, models enhanced for visual math reasoning show no improvement on GSM8K-V despite large gains on existing benchmarks, confirming that it evaluates a distinct capability. Error analysis shows that the primary bottleneck lies in Implicit Visual Inference Error (IVIE), where models fail to recover visual semantics that are implied rather than explicitly stated. Our code and data are released at \url{https://github.com/ZJU-REAL/GSM8K-V}.
\end{abstract}

\begin{figure*}[t]
  \centering
  \begin{minipage}[t]{0.59\linewidth}
    \centering
    \includegraphics[width=\linewidth]{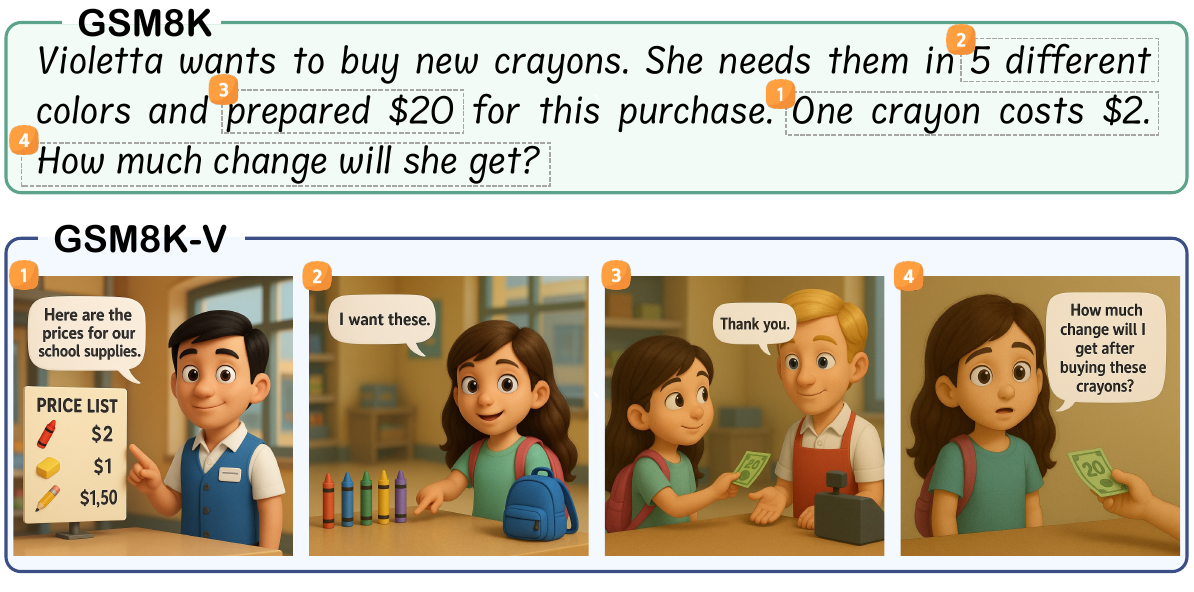}
  \end{minipage}
  \hfill
  \begin{minipage}[t]{0.40\linewidth}
    \centering
    \includegraphics[width=\linewidth]{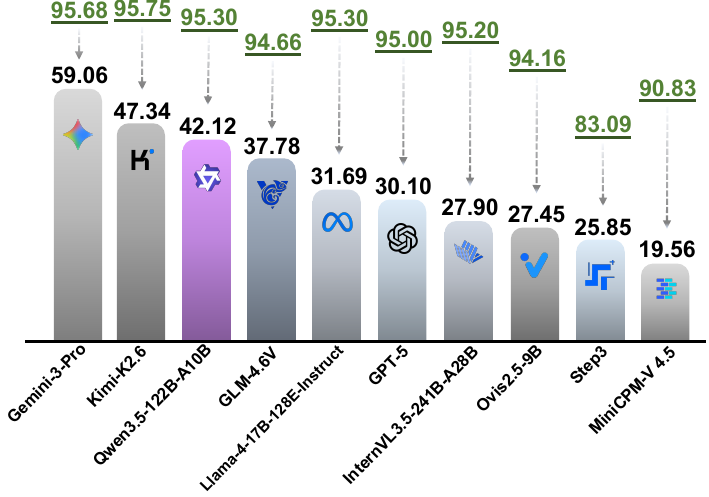}
  \end{minipage}

  \caption{Left: An example of converting a GSM8K text problem into a GSM8K-V multi-image item, where key semantics are visually grounded across images to support the same arithmetic reasoning from visual context. Right: VLM performance (\%) comparison between GSM8K-V and GSM8K. Accuracy for GSM8K is indicated by the green-colored text, while the bar chart represents GSM8K-V accuracy, illustrating the performance gap.}
  \label{fig:intro_and_performance}
\end{figure*}

\section{Introduction}

Vision-language models (VLMs) have achieved remarkable progress in multimodal understanding and reasoning~\citep{openai2024gpt4o,comanici2025gemini,wang2025internvl35,team2026glm45v}. On text-based mathematical reasoning benchmarks such as GSM8K~\citep{cobbe2021training}, state-of-the-art models now achieve over 95\% accuracy, approaching human-level performance. This near-saturation raises a critical question: \textbf{\emph{Do these models truly understand mathematical reasoning, or have they primarily learned to exploit the explicit symbolic structure of text?}}

Language serves as a high-level abstraction of the world~\citep{li2024llms,lupyan2018language}. A textual math problem like ``Violetta needs 5 crayons at \$2 each with \$20 budget'' explicitly encodes all quantities and relations through symbols. In visual scenarios, however, the same semantic content must be conveyed differently: the number of crayons must be perceived by counting, the price read from a signboard, and the purchase action inferred from scene transitions (Figure~\ref{fig:intro_and_performance}). This fundamental distinction between \emph{explicit textual encoding} and \emph{implicit visual encoding} of mathematical information remains largely unexplored.

Existing visual mathematical benchmarks have made significant strides, yet they predominantly evaluate \emph{domain-specific} and \emph{classroom-style} reasoning over \emph{structured visual inputs}. 
MathVista~\citep{lu2023mathvista} and MathVerse~\citep{zhang2025mathverse} focus on geometry diagrams and algebraic plots; MATHVision~\citep{wang2024measuring,wang2025mathcodervl} targets competition-level problems with formal visual statements; VideoMathQA~\citep{rasheed2025videomathqa} assesses instructional video comprehension. These benchmarks share a common characteristic: the visual content serves primarily as a \emph{carrier of explicit symbolic information} (e.g., geometric primitives, function graphs, labeled diagrams) that models must recognize and then reason about symbolically. 
Consequently, they do not test \emph{cross-scene integration} or reasoning over \emph{implicitly encoded} visual information distributed across multiple images.

To address these limitations, we introduce \textbf{GSM8K-V}, a purely visual counterpart of GSM8K that tests whether VLMs can perform mathematical reasoning when information is implicitly encoded across multiple images. As illustrated in Figure~\ref{fig:intro_and_performance}, each text problem is transformed into a multi-image sequence where mathematical information must be extracted through visual perception (counting objects, reading signboards, interpreting gestures) and integrated across scenes that encode actions and state changes implicitly.
By preserving exact semantic equivalence with GSM8K, we enable direct comparison: the same mathematical content, but requiring visual perception and cross-scene integration rather than symbol reading.
GSM8K-V offers three key evaluation capabilities:
\begin{itemize}[leftmargin=15pt, topsep=1pt, itemsep=1pt]
\item \textbf{Implicit visual reasoning.} Models must infer quantities through counting, deduce actions from scene transitions (e.g., a money exchange implies purchase), and extract relations visually suggested but never explicitly stated.
\item \textbf{Cross-scene integration.} Information is distributed across multiple images, requiring models to track entities, accumulate evidence, and maintain coherent visual reasoning.
\item \textbf{Controlled semantic equivalence.} Identical mathematical content in text and visual forms enables direct comparison, isolating whether failures stem from visual understanding or mathematical reasoning.
\end{itemize}

We construct GSM8K-V through a three-stage pipeline: (1) decomposing problems into structured mathematical elements distributed across scenes; (2) generating scene descriptions via a tripartite schema of objects, actions, and compositions; and (3) rendering images with GPT-Image-1~\citep{openai2025gptimage1} followed by rigorous human verification. This yields 1,319 samples with 5,343 images.

Evaluation of 34 VLMs reveals that cross-scene integration and implicit visual reasoning pose substantial challenges (Figure~\ref{fig:intro_and_performance}): while 27 of 34 models exceed 90\% on text-based GSM8K, the best model (Gemini-3-Pro) achieves only 59.06\% on GSM8K-V, with most falling below 30\%. Human annotators achieve 91.15\%, confirming that the visual encoding preserves solvability. Crucially, VL-Rethinker-7B, a model enhanced for visual math reasoning via reinforcement learning, improves substantially on MathVista (+6.7\%), MathVerse (+7.9\%), and MathVision (+7.2\%), yet shows \emph{no gain} on GSM8K-V (-0.6\%), demonstrating that our benchmark targets a distinct capability. Error analysis identifies \emph{Explicit Perception Error (EPE)} and \emph{Implicit Visual Inference Error (IVIE)} as dominant failure modes, rather than OCR or arithmetic mistakes. Our contributions are threefold:
\begin{itemize}[leftmargin=15pt,itemsep=0.3ex,parsep=0ex,topsep=0.5ex]
\item We introduce \datasetname, a visual counterpart of GSM8K that evaluates cross-scene integration and implicit visual reasoning under semantic equivalence with the original text problems.
\item We develop a reproducible pipeline for transforming text-based math problems into multi-scene visual representations.
\item We show through systematic evaluation and error analysis that Implicit Visual Inference Error is the primary bottleneck for VLMs on \datasetname, providing actionable diagnostics for advancing VLMs reasoning capabilities.
\end{itemize}

\section{Related Work}

\subsection{Mathematical Reasoning of VLMs}

Reasoning is a core dimension of intelligence, and multimodal reasoning provides a stringent test of whether models can connect visual evidence with symbolic and logical inference.
Recent Vision-Language Models (VLMs), including GPT-4o~\citep{openai2024gpt4o}, Gemini-3-Pro, Gemini-2.5-Pro~\citep{comanici2025gemini}, Qwen3-VL~\citep{bai2025qwen3vl}, GLM-4.5V~\citep{team2026glm45v}, LLaMA 4~\citep{meta2024llama4}, Ovis 2.5~\citep{lu2025ovis25}, and MiniCPM-V~\citep{yao2024minicpmv}, have shown rapid progress.
Prior work also improves mathematical reasoning by refining intermediate reasoning steps~\citep{yan2026mathfimerenhancingmathematicalreasoning}.
Nevertheless, visual mathematical reasoning remains challenging because it requires grounding symbolic relations in visual contexts, integrating multimodal information, and performing multi-step inference~\citep{hendrycks2021measuring,cobbe2021training}.
Recent work has studied modality preferences in MLLMs~\citep{ZhangModality2026}, modality-dependent reasoning failures~\citep{xu-etal-2026-seeing}, contextual reasoning~\citep{zhengcontext}, structure-aware multimodal reasoning for complex visual layouts~\citep{ZhuTable2026}, reasoning consistency across interconnected knowledge and multi-step inference~\citep{zhang2026towards}, and reasoning enhancement through chain-of-thought supervision and reinforcement learning~\citep{zhang2025improve,chen2025mintcot,shao2024deepseekmath,deepseek-ai2026deepseekr1,dai2025sgrpo,dai2025stable,wang2025reinforcement}. In contrast, \datasetname focuses on implicit visual inference and cross-scene integration for recovering arithmetic evidence from visual narratives.

\subsection{Mathematical Benchmarks for VLMs}

A variety of vision-based benchmarks has been proposed to evaluate mathematical reasoning in visual contexts. 
Early work focused on geometry and multiple-choice reasoning~\citep{lu2021intergpsa,chen2021geoqa}, while recent benchmarks broaden both task scope and evaluation methodology. 
MathVista~\citep{lu2023mathvista} unifies diverse math-vision tasks, MATH-Vision~\citep{wang2024measuring,wang2025mathcodervl} introduces competition-level visual math problems, MathVerse~\citep{zhang2025mathverse} probes genuine visual understanding by redistributing information between text and diagrams, and MM-MATH~\citep{sun2024mmmath} emphasizes process-level evaluation. 
DynaMath~\citep{zou2025dynamath} evaluates robustness through perturbed variants, Omni-MATH~\citep{gao2024omni} increases difficulty with Olympiad-level problems, VideoMathQA~\citep{rasheed2025videomathqa} studies reasoning over video sequences, and VisuLogic~\citep{xu2025visulogic} targets systematic visual reasoning across multimodal tasks.
Synthetic visual reasoning benchmarks have also been constructed over abstract and everyday scenarios~\citep{zhang2024multimodal}.
Despite these advances, many benchmarks rely on explicit symbolic carriers such as geometric primitives, coordinate plots, diagrams, or instructional videos. 
In contrast, \datasetname renders GSM8K problems into everyday multi-scene visual form while preserving exact semantic equivalence, enabling direct comparison between text-based reasoning and visual reasoning that depends on perception, implicit visual inference, and cross-scene integration.

\section{Benchmark Construction}
\label{sec:benchmark}

\begin{figure*}[t]
    \centering
    \includegraphics[width=0.95\linewidth]{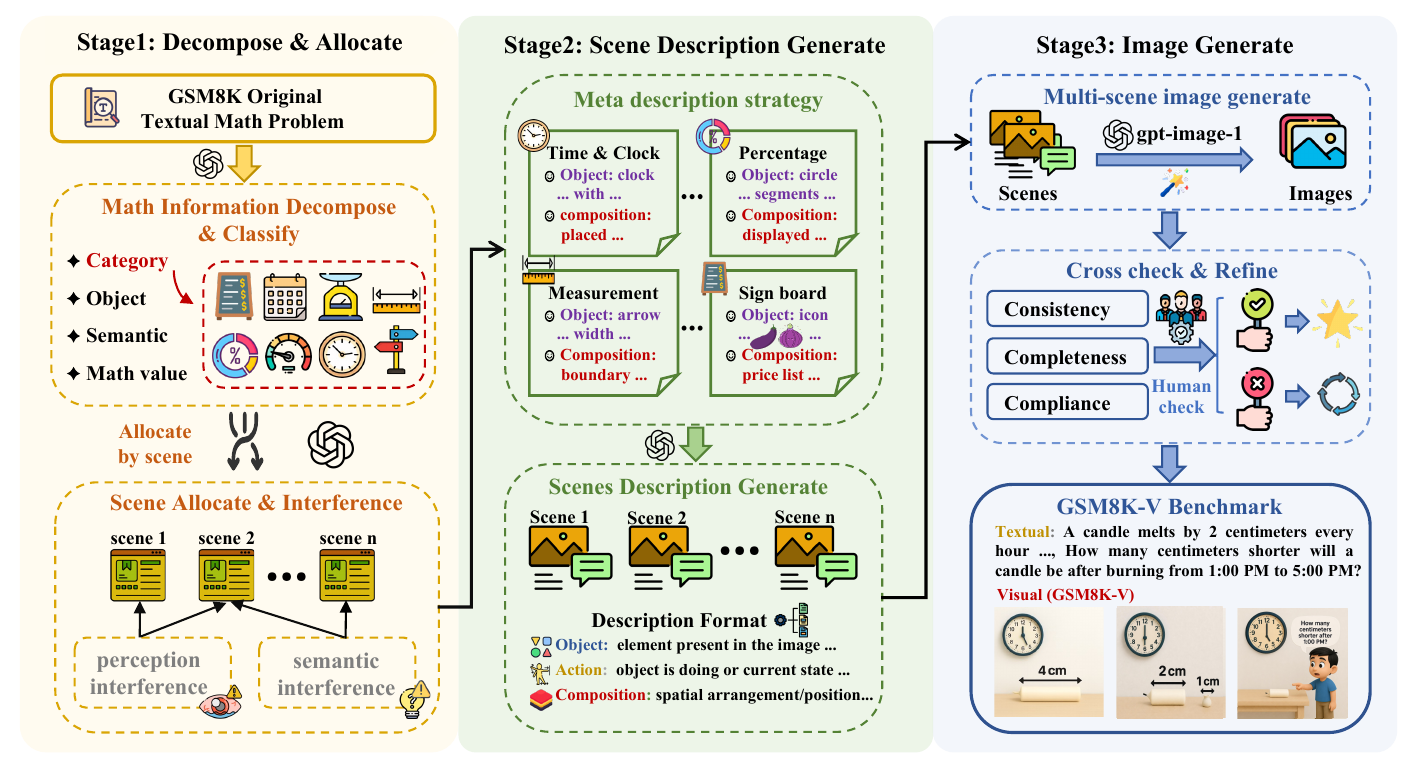}
    \caption{The benchmark construction pipeline used in \datasetname. This pipeline serves as a systematic framework for processing text-based problems, generating scene descriptions, and producing visual representations.}
    \label{fig:pipeline}
\end{figure*}

We propose \textbf{GSM8K-V}, a benchmark presenting real-world daily math problems through multiple images, systematically converted from the textual GSM8K benchmark. The overall pipeline is illustrated in Figure~\ref{fig:pipeline}. Our core idea is to transform each textual problem into multiple visually expressed scenes. Specifically, the pipeline consists of three steps. For each sample in GSM8K, we first decompose its mathematical information and allocate them across different scenes (Section~\ref{sec:method_step1}). Then we design prompts according to the category to generate scene descriptions (Section~\ref{sec:method_step2}). Finally, we use GPT-Image-1~\citep{openai2025gptimage1} to generate scene images, followed by human validation and refinement to obtain the final benchmark (Section~\ref{sec:method_step3}). 
Appendix~\ref{appendix:image_generator_robustness} presents an ablation study on different T2I generation models to assess the robustness of the generated benchmark.

\subsection{Problem Decomposition and Allocation}
\label{sec:method_step1}

\paragraph{Math information decomposition and classification.} 
Each GSM8K sample contains multiple pieces of mathematical information that must be faithfully preserved during visual conversion. 
We employ GPT-4.1~\citep{openai2025gpt41} to parse each problem into structured triples $(\textit{object}, \textit{math value}, \textit{semantic})$, where \textit{object} denotes the described entity, \textit{math value} specifies its numerical attribute, and \textit{semantic} captures its contextual role. 
We further categorize these mathematical elements into 13 typical classes, which are rendered in distinct visual forms to preserve the original mathematical information while aligning with real-world scenarios. 
Details are provided in Appendix~\ref{appendix:math_information_classification}.

\paragraph{Scene allocation.} 
Since a single problem often contains multiple pieces of mathematical information, we allocate them across different scenes. 
This allocation follows three guiding principles:
\begin{itemize}[leftmargin=15pt,itemsep=0.3ex, parsep=0ex, topsep=0.5ex]
    \item \textit{Contextual grouping:} Information that refers to the same object type, spatial setting, or functional purpose is grouped into one scene.
    \item \textit{Final isolation:} The final scene contains only the question itself, serving as the reasoning target without additional numerical or visual data.
    \item \textit{Atomic fidelity:} No values are inferred or merged. Only atomic math facts from Step 1 are retained to prevent premature reasoning during scene construction.
\end{itemize}
We employ GPT-4.1 to perform this allocation, with further details provided in Appendix~\ref{appendix:assign_category}.

\paragraph{Multi-dimensional interference.} 
GSM8K-V introduces controlled interference not to simply increase difficulty, but to make the original text-based math problems better reflect realistic visual scenarios. 
Specifically, \textit{perception interference} adds irrelevant but visually salient objects to simulate real-world visual noise, while \textit{semantic interference} introduces contextually related distractors. Details are provided in Appendix~\ref{appendix:controlled_interference}.
These designs require VLMs to distinguish task-relevant evidence from natural environmental clutter. 
As shown in Section~\ref{subsec:error_analysis}, interference-related errors such as Entity Identity Confusion (EIConf) account for only a small fraction of failures, while the main bottleneck lies in implicit visual inference.

\subsection{Scene Description Generation}
\label{sec:method_step2}

\paragraph{Meta description strategy definition.} 
To enhance the quality of image generation, we first construct high-quality scene descriptions in textual form. Based on the mathematical information introduced in Section~\ref{sec:method_step1}, we design a series of meta description strategies for each type of mathematical information.
Each mathematical information category is mapped to the corresponding meta description strategy, defined through pre-specified prompts (e.g. \emph{action}, \emph{time \& clock}, \emph{percentage}, \emph{measurement}, \emph{sign board \& icon}). These strategies provide high-level templates that ensure the mathematical content is faithfully represented in the visual modality. The complete set of templates is documented in Appendix~\ref{appendix:meta_description_strategy}.  

\paragraph{Scene description generation.} 
Based on the selected meta strategy, we use GPT-4.1 to produce a structured scene description for each allocated unit of mathematical information. Each description follows a tripartite schema $(\textit{object}, \textit{action}, \textit{composition})$:  
\begin{itemize}[leftmargin=15pt,itemsep=0.3ex, parsep=0ex, topsep=0.5ex]
    \item \textit{Object}: the concrete entities that must appear in the image and carry mathematical information.  
    \item \textit{Action}: the state or activity that defines how an object is presented and conveys semantic cues.  
    \item \textit{Composition}: the spatial arrangement or positional relations among all elements.  
\end{itemize}
This structured representation enforces consistency across scenes and provides explicit guidance for subsequent image generation.

\subsection{Image Generation}
\label{sec:method_step3}

\paragraph{Multi-scene image generation.} 
With the generated structured scene descriptions in Section ~\ref{sec:method_step2}, we generate multi-scene images with the GPT-Image-1 model~\citep{openai2025gptimage1}. The specific generation prompts for each category are provided in Appendix~\ref{appendix:image_generation_prompts}, ensuring reproducibility of the visual pipeline. For each scene, we generate a $1024 \times 1024$ pixel image.

\paragraph{Human cross check and refinement.} 
Human cross check is a key step for ensuring the quality and reliability of \datasetname. 
To verify each generated problem--image pair, all images undergo a dual human cross-validation process conducted by fully trained annotators. 
This iterative procedure is guided by three principles:
\begin{itemize}[leftmargin=15pt,itemsep=0.3ex, parsep=0ex, topsep=0.5ex]
    \item \textit{Consistency}: visual scenes must preserve the entities, quantities, and constraints in the original text, ensuring semantic equivalence.  
    \item \textit{Completeness}: all information needed for solving the problem must be visually accessible, without requiring additional assumptions.  
    \item \textit{Compliance}: images must satisfy safety and formatting requirements, including clear identities and legible numerical symbols.  
\end{itemize}
Whenever violations are identified, the images are refined. This generation-verification loop is repeated until every problem–image pair meets the established requirements. Detailed protocols of the dual cross-checking process are provided in Appendix~\ref{appendix:human_verification}.

\begin{figure}[t]
    \centering
    \includegraphics[width=0.95\linewidth]{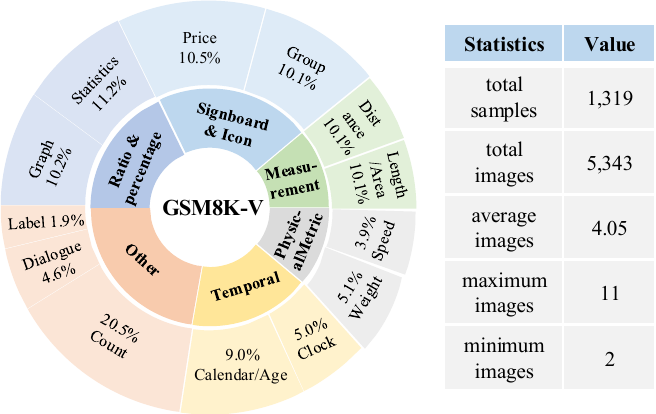}
    \caption{Data distribution and core statistics of GSM8K-V. The left visualizes the category distribution of \datasetname, the right provides benchmark statistics.}
    \label{fig:benchmark_statistics}
\end{figure}

\paragraph{Benchmark statistics.} 
\datasetname consists of 1,319 samples and 5,343 images, requiring models to process an average of 4 images per case (ranging from 2 to 11). As shown in Figure~\ref{fig:benchmark_statistics}, \datasetname covers 6 broad categories and 13 fine-grained sub-categories, providing a diverse and comprehensive suite for testing mathematical reasoning in complex, multi-scene visual environments.

\section{Evaluation Results}
\label{sec:evaluation}

\subsection{Evaluation Setup}
\label{sec:evaluation_setup}

We evaluate a representative selection of both closed-source and open-source VLMs on the GSM8K-V benchmark.
The closed-source models include \texttt{Gemini-2.5-Pro, Gemini-3-Pro}~\citep{comanici2025gemini}, \texttt{GPT-5}~\citep{singh2025openai}, \texttt{GPT-4o}~\citep{openai2024gpt4o}, and \texttt{QVQ-Max-Latest}~\citep{qwen2025qvqmax}, while the open-source models cover the \texttt{Llama-4} series~\citep{meta2024llama4}, \texttt{InternVL3.5} series~\citep{wang2025internvl35}, \texttt{Qwen3.6} series~\citep{qwen2026qwen36a3b}, \texttt{Qwen3.5} series~\citep{qwen2026qwen35}, \texttt{Ovis2.5} series~\citep{lu2025ovis25}, as well as \texttt{Step-3}~\citep{stepfun2025step3}, \texttt{Kimi-K2.6}~\citep{moonshot2026kimik26}, \texttt{MiniCPM}~\citep{openbmb2026minicpmv46}, and \texttt{GLM-4.6V}~\citep{zai2025glm46v_hf}, which represent a diverse set of strong VLMs. For closed-source models, evaluation is conducted 
through their official APIs, whereas for open-source models, inference is performed using the 
\textit{vLLM} framework~\citep{kwon2023efficient}. Details of the hyperparameter configurations for each model are provided in Appendix~\ref{appendix:Evaluation Setup} to ensure consistency and reproducibility.

We further evaluate models on both text-only GSM8K and its visual counterpart, \datasetname. 
For \datasetname, models are prompted to reason step by step over the images and output only the final integer answer. 
Detailed evaluation modes and prompt templates are provided in Appendix~\ref{appendix:Evaluation Modes}. 
We also report human accuracy by asking annotators to answer directly from the generated images.

\begin{table*}[t]
\centering
\small
\setlength{\tabcolsep}{5.5pt}        %
\begin{tabular}{lcccccc>{\columncolor[gray]{0.95}}cc}  %
\toprule
\textbf{Model} &
\multicolumn{7}{c}{\textbf{GSM8K-V}} &
\textbf{GSM8K} \\
\cmidrule(lr){2-8} \cmidrule(lr){9-9}

& \makecell{Signboard \\ \& Icon} 
& \makecell{Measu- \\ rement} 
& \makecell{Physical \\ Metric} 
& \makecell{Temp- \\oral} 
& \makecell{Ratio \& \\ Percentage} 
& \makecell{Other} 
& \makecell{Avg.} 
& \makecell{Avg.}  \\
\midrule
Human & \textbf{90.27} & \textbf{86.76} & \textbf{96.05} & \textbf{91.20} & \textbf{88.57} & \textbf{93.25} & \textbf{91.15} & - \\
\midrule

\rowcolor[rgb]{ .91,  .91,  .91} \multicolumn{9}{l}{\textit{Closed-Source VLMs}} \\
\midrule

Gemini-3-Pro & \textbf{66.18} & \textbf{63.11} & \textbf{62.18} & \textbf{57.30} & \textbf{56.54} & \textbf{54.34} & \textbf{59.06} & \textbf{95.68} \\

Gemini-2.5-Pro & \underline{52.94} & \underline{52.43} & \underline{56.30} & \underline{47.57} & \underline{42.05} & \underline{41.18} & \underline{46.93} & \underline{95.22} \\

GPT-5 & 41.54 & 46.60 & 26.89 & 28.65 & 21.91 & 24.93 & 30.10 & 95.00 \\
GPT-4o & 41.91 & 42.72 & 31.93 & 28.11 & 21.91 & 22.41 & 29.57 & 94.92 \\
QVQ-Max-Latest & 30.51 & 39.81 & 21.85 & 25.95 & 16.25 & 17.37 & 23.20 & 95.00 \\

\midrule
\rowcolor[rgb]{ .91,  .91,  .91}\multicolumn{9}{l}{\textit{Open-Source VLMs}} \\
\midrule

Llama-4-17B-16E-Instruct & 36.76 & 36.89 & 30.25 & 25.41 & 19.79 & 21.01 & 26.69 & 94.16 \\
Llama-4-17B-128E-Instruct & 42.28 &  41.75 & 28.57 & 34.59 & 22.61 & 27.45 &  31.69 & \underline{95.30} \\

InternVL3.5-8B & 16.91 & 24.27 & 10.92 & 8.65 & 5.30 & 7.56 & 10.77 & 90.67 \\
InternVL3.5-38B & 36.03 & 38.84 &  31.09 & 25.41 & 17.67 & 22.97 & 26.84 & 93.86 \\
InternVL3.5-30B-A3B & 29.41 & 37.86 &  21.85 &  25.41 & 13.43 & 19.89 &  22.82 & 93.18 \\
InternVL3.5-241B-A28B & 36.76 & 46.60 & 28.57 & 27.03 & 21.20 & 21.29 & 27.90 & 95.22 \\

Qwen3.6-35B-A3B & 38.44 & 39.46 & 44.49 & 32.56 & 24.82 & 24.60 & 30.71 & 94.16 \\

Qwen3.6-35B-A3B-Thinking & 39.73 & 40.78 & 45.98 & 33.65 & 25.65 & 25.43 & 31.74 & 94.09 \\

Qwen3.5-9B & 33.90 & 44.95 & 29.02 & 28.22 & 21.21 & 21.48 & 27.82 & 94.24 \\

Qwen3.5-27B & 43.04 & 42.42 & \textbf{49.08} & 42.18 & \underline{37.79} & 34.40 & 39.88 & 94.77 \\

Qwen3.5-122B-A10B & 50.87 & 46.21 & 39.58 & 36.42 & 35.15 & 34.76 & 40.08 & 94.47 \\

Qwen3.5-122B-A10B-Thinking & \underline{54.62} & \underline{49.34} & 42.48 & 39.08 & 36.21 & \underline{37.27} & \underline{42.12} & \underline{95.30} \\

Ovis2.5-2B & 19.12 & 27.18 & 15.13 & 13.51 & 6.01 & 10.64 & 13.50 & 86.28 \\
Ovis2.5-9B & 37.13 & 38.83 & 31.09 & 24.86 & 18.02 & 24.37 & 27.45 & 94.16 \\

Step3 & 34.19 & 34.95 & 33.61 & 33.51 & 15.19 & 18.77 & 25.85 & 83.09 \\

Kimi-K2.6 & \textbf{62.50 }& \textbf{51.35} & \underline{48.33} & \textbf{49.06} & \textbf{40.56} & \textbf{39.03} & \textbf{47.34} & \textbf{95.75} \\

MiniCPM-V-4.6 & 15.81 & 13.59 & 11.76 & 7.03 & 4.24 & 8.12 & 9.48 & 75.51 \\

MiniCPM-V-4.6-Thinking & 17.65 & 20.39 & 12.61 & 8.11 & 4.95 & 9.04 & 11.02 & 85.11 \\

GLM-4.6V & 48.29 & 46.00 & 35.09 & \underline{42.52} & 27.91 & 32.12 & 37.78 & 94.66 \\

\bottomrule
\end{tabular}
\caption{Performance (\%) comparison of vision-language models across six categories and overall average accuracy on GSM8k-V. The rightmost column shows the performance in text-only GSM8K.}
\label{tab:main_results}
\end{table*}

\subsection{Overall Performance}
We evaluate 34 mainstream VLMs and present the main evaluation results in Table~\ref{tab:main_results}, with the complete results reported in Appendix Table~\ref{tab:appendix_full_results}. From these results, we derive several key observations:

\paragraph{A stark modality gap under semantic equivalence.} As shown in Table~\ref{tab:main_results}, while textual reasoning on GSM8K is nearly saturated with 20 out of 24 models exceeding 90\% accuracy, performance collapses in the visual setting of \datasetname. The best-performing closed-source model, Gemini-3-Pro, drops from 95.68\% in text form to 59.06\% in visual form, while the top open-source model, Kimi-K2.6, achieves only 47.34\%. Since \datasetname maintains exact semantic equivalence with GSM8K, this dramatic discrepancy isolates the failure to visual perception and cross-modal mapping rather than mathematical logic. These results prove that VLM performance is heavily predicated on the explicit symbolic structure of text rather than a deep, modal-invariant understanding of mathematical reasoning.

\paragraph{Bottlenecks in implicit reasoning and cross-scene integration.} 
The results show that extracting quantities from implicit visual cues and integrating evidence across scenes remain major challenges for most VLMs. 
Performance is particularly low in categories requiring high-level semantic inference from transitions, such as \emph{Ratio \& Percentage} and \emph{Other}, where many models score below 20\%. By contrast, performance is relatively higher in the \emph{Measurement} category because these tasks typically involve fewer scene transitions and rely on more explicit visual cues. This category imposes lower demands on cross-scene integration, leading to a substantial performance gap between measurement and implicit reasoning for models. Notably, Gemini-3-Pro performs consistently well across most categories requiring implicit visual reasoning, yet its average accuracy remains significantly lower than the human level of 91.15\%. This suggests that while frontier models possess a foundational capacity for cross-scene tracking, their failure to fully reconstruct reasoning chains indicates that current visual architectures often struggle to preserve fine-grained numerical relations when distilling high-level semantic from distributed mathematical narratives.

\paragraph{Divergence between human and VLM reasoning patterns.} 
Humans achieve a robust average accuracy of $91.15\%$, outperforming Gemini-3-Pro ($59.06\%$), and remain balanced across categories, including the unstructured \emph{Other} category. 
In contrast, VLMs exhibit pronounced variability and category-dependence. They perform relatively better on \emph{Signboard \& Icon}, but drop sharply on categories such as \emph{Other} and \emph{Temporal}. 
This gap suggests that human reasoning relies on generalized semantic grounding, whereas current VLMs remain sensitive to visual formats and struggle with robust inference in implicit, unstructured visual contexts.

\section{Analysis}

In this section, we provide a detailed analysis of \datasetname. We first compare the \datasetname and other mathematical benchmark in Section~\ref{subsec:mathematical_reasoning_model_on_gsm8k-v}. We further conduct ablation studies on image organizational formats in Section~\ref{subsec:image_input_format_ablation}, image styles in Section~\ref{subsec:image_style_ablation}, modality validation in Section~\ref{subsec:modality_validation} and error analysis of \datasetname in Section~\ref{subsec:error_analysis}.

\begin{table*}[t]
  \centering
  \small
  \setlength{\tabcolsep}{8.5pt}
  \begin{tabular}{lcccc}
    \toprule
    \multicolumn{1}{c}{\textbf{Model}} &
    \makecell{\textbf{MathVista}\\\textbf{(testmini)}} &
    \makecell{\textbf{MathVerse}\\\textbf{(testmini)}} &
    \makecell{\textbf{MathVision}\textbf{(test)}} &
    \textbf{GSM8K-V} \\
    \midrule

    \makecell[l]{Qwen2.5-VL-7B-Instruct~\citep{bai2025qwen25vl}}
    & 68.2
    & 46.3
    & 25.1
    & 15.16 \\

    \makecell[l]{VL-Rethinker-7B~\citep{wangvlrethinker}}
    & 74.9 {\scriptsize(\textcolor{green!50!black}{+6.7})}
    & 54.2 {\scriptsize(\textcolor{green!50!black}{+7.9})}
    & 32.3 {\scriptsize(\textcolor{green!50!black}{+7.2})}
    & 14.56 {\scriptsize(\textcolor{red!70!black}{-0.60})} \\

    \makecell[l]{MM-Eureka-7B~\citep{meng2025mm}}
    & 73.0 {\scriptsize(\textcolor{green!50!black}{+4.8})}
    & 50.3 {\scriptsize(\textcolor{green!50!black}{+4.0})}
    & 26.9 {\scriptsize(\textcolor{green!50!black}{+1.8})}
    & 15.85 {\scriptsize(\textcolor{green!50!black}{+0.69})} \\

    \makecell[l]{OpenVLThinker-7B~\citep{deng2026openvlthinker}}
    & 70.2 {\scriptsize(\textcolor{green!50!black}{+2.0})}
    & 47.9 {\scriptsize(\textcolor{green!50!black}{+1.6})}
    & 25.3 {\scriptsize(\textcolor{green!50!black}{+0.2})}
    & 14.78 {\scriptsize(\textcolor{red!70!black}{-0.38})} \\

    \makecell[l]{R1-VL-7B~\citep{zhang2025r1}}
    & 74.3 {\scriptsize(\textcolor{green!50!black}{+6.1})}
    & 52.2 {\scriptsize(\textcolor{green!50!black}{+5.9})}
    & 28.2 {\scriptsize(\textcolor{green!50!black}{+3.1})}
    & 14.94 {\scriptsize(\textcolor{red!70!black}{-0.22})} \\

    \makecell[l]{R1-Onevision-7B~\citep{yang2025r1}}
    & 64.1 {\scriptsize(\textcolor{red!70!black}{-4.1})}
    & 46.4 {\scriptsize(\textcolor{green!50!black}{+0.1})}
    & 29.9 {\scriptsize(\textcolor{green!50!black}{+4.8})}
    & 13.57 {\scriptsize(\textcolor{red!70!black}{-1.59})} \\

    \midrule
    \textbf{Average Change}
    & \cellcolor{green!20}\textbf{+3.10}
    & \cellcolor{green!20}\textbf{+3.90}
    & \cellcolor{green!20}\textbf{+3.42}
    & \cellcolor{red!20}\textbf{-0.42} \\

    \bottomrule
  \end{tabular}

  \caption{
  Performance (\%) of Qwen2.5-VL-7B-Instruct and five reasoning-enhanced variants across mathematical benchmarks.
  Values in parentheses denote performance changes relative to the Qwen2.5-VL-7B-Instruct backbone.
  }
  \label{tab:qwen_reasoning_transfer}
\end{table*}

\begin{table*}[t]
\centering
\small
\setlength{\tabcolsep}{10.5pt}

\begin{tabular}{lcccccc}
\toprule
\multicolumn{1}{c}{\textbf{Model}} &
\makecell{\textbf{Single-}\\\textbf{Image}} &
\makecell{\textbf{Single-}\\\textbf{Image($r^*$)}} &
\makecell{\textbf{Multi-}\\\textbf{Image}} &
\makecell{\textbf{Multi-}\\\textbf{Image($r^*$)}} &
\makecell{\textbf{Cross-Scene}\\\textbf{Subset}} &
\makecell{\textbf{Cross-Scene}\\\textbf{Subset($r^*$)}} \\ 
\midrule

GPT-4o       & 27.45 & 25.85 & 29.57 & 26.54 & 25.11 & 11.42 \\
GLM-4.5V       & 26.76 & 24.29 & 28.28 & 26.91 & 28.31 & 11.87 \\
Qwen2.5-VL-72B & 20.47 & 18.35 & 21.91 & 18.12 & 20.55 & 8.68 \\
Qwen2.5-VL-7B  & 15.39 & 5.16  & 15.16 & 3.87  & 13.24 & 5.02 \\
\bottomrule
\end{tabular}
\caption{Performance (\%) under different input formats. $r^*$ denotes randomized image order.}
\label{tab:single_multi}
\end{table*}

\subsection{Comparison between \datasetname and Other Mathematical Benchmark}
\label{subsec:mathematical_reasoning_model_on_gsm8k-v}

To verify that GSM8K-V targets a different evaluation ability, we conduct performance comparison experiments based on five reasoning-enhanced models built on the same Qwen2.5-VL-7B-Instruct backbone~\citep{bai2025qwen25vl}: 
VL-Rethinker-7B~\citep{wangvlrethinker}, 
MM-Eureka-7B~\citep{meng2025mm}, 
OpenVLThinker-7B~\citep{deng2026openvlthinker}, 
R1-VL-7B~\citep{zhang2025r1}, and 
R1-Onevision-7B~\citep{yang2025r1}.
These models adopt different post-training strategies, including 
reinforcement learning, iterative SFT--RL, and other 
reasoning-oriented optimization methods.

As shown in Table~\ref{tab:qwen_reasoning_transfer}, 
reasoning post-training yields average improvements of 
+3.10, +3.90, and +3.42 points on 
MathVista, MathVerse, and MathVision, respectively.
In contrast, the average change on \datasetname is 
-0.42 points.
This transfer gap suggests that improvements obtained from current 
reasoning post-training objectives on existing visual-mathematical 
benchmarks do not directly translate into comparable gains on \datasetname.
This performance divergence provides empirical evidence that \textbf{\datasetname provides a complementary diagnostic dimension by emphasizing implicit visual inference and cross-scene 
integration}.
A detailed comparison between \datasetname and other benchmarks ~\citep{lu2023mathvista, zhang2025mathverse, wang2024measuring, rasheed2025videomathqa} is provided in Appendix~\ref{appendix:comparison_benchmark}.

\subsection{Cross-Scene Integration Analysis}
\label{subsec:image_input_format_ablation}

In practical scenarios, information is often distributed across multiple images with an underlying temporal or semantic order. 
To evaluate how scene organization affects model reasoning, we evaluate Single-Image, Multi-Image, and their randomized-order variants, denoted by $r^*$.
We further construct a Cross-Scene Subset of 219 samples from the \emph{Temporal} and \emph{Other} categories, where solving the problem strongly depends on the correct scene order. 

As shown in Table~\ref{tab:single_multi}, $r^*$ leads to consistent performance drops, with the effect being much stronger on the Cross-Scene Subset. 
For example, GPT-4o and GLM-4.5V drop from $25.11\%$ and $28.31\%$ to $11.42\%$ and $11.87\%$, respectively, indicating that even frontier models rely on coherent scene order when reasoning over distributed visual evidence. 
The degradation is also severe for Qwen2.5-VL models, suggesting weaker robustness to disrupted visual narratives. 
These results show that cross-scene integration is a key challenge in \datasetname, especially when mathematical relations must be inferred from ordered visual context. 
Details are provided in Appendix~\ref{appendix:single_multi}.

\subsection{Image Style Ablation}
\label{subsec:image_style_ablation}

\begin{table}[H]
\centering
\setlength{\tabcolsep}{6pt}
\small
\begin{tabular}{lccc}
\toprule
\textbf{Model} & \textbf{Pixar} & \textbf{Giphli} & \textbf{Real-World}\\
\midrule
GPT-4o            & 27.27 & 28.93 & 29.70\\
GLM-4.5V          & 28.10 & 26.45 & 25.74\\
Qwen2.5-VL-72B    & 18.18 & 17.36 & 17.82\\
Qwen2.5-VL-7B     & 12.40 & 12.40 & 11.88\\
\bottomrule
\end{tabular}
\caption{Performance (\%) under different image styles.}
\label{tab:image_styles}
\end{table}

Our GSM8K-V benchmark uses \emph{Pixar-Style} rendering for visual clarity and natural lighting. 
To assess style robustness, we further evaluate \emph{Ghibli-Style} and \emph{Real-world Style} variants. 
As shown in Table~\ref{tab:image_styles}, performance differences across styles are small, with no systematic degradation under photorealistic rendering. 
For example, GPT-4o performs comparably on Real-world and Pixar styles, and other models show similar fluctuations. 
These results indicate that reasoning performance is not strongly driven by visual style, supporting the cross-style generalizability of \datasetname. 
Side-by-side examples are provided in Appendix~\ref{appendix:image_styles}.

\begin{table*}[t]
\centering
\small
\setlength{\tabcolsep}{10pt}

\begin{tabular}{lccccc}
\toprule
\textbf{Model} & \textbf{Text only} & \textbf{Image only} & \textbf{OCR} & \textbf{OCR + Image} & \textbf{Caption} \\
\midrule
GPT-4o & 94.92 & 29.57 &
11.98 {\scriptsize(\textcolor{red!70!black}{-17.59})} &
33.06 {\scriptsize(\textcolor{green!50!black}{+3.49})} &
55.11 {\scriptsize(\textcolor{green!50!black}{+25.54})} \\

GLM-4.5V & 90.60 & 28.28 &
9.33 {\scriptsize(\textcolor{red!70!black}{-18.95})} &
32.22 {\scriptsize(\textcolor{green!50!black}{+3.94})} &
44.96 {\scriptsize(\textcolor{green!50!black}{+16.68})} \\

Qwen2.5-VL-72B & 94.09 & 21.91 &
9.63 {\scriptsize(\textcolor{red!70!black}{-12.28})} &
24.34 {\scriptsize(\textcolor{green!50!black}{+2.43})} &
47.31 {\scriptsize(\textcolor{green!50!black}{+25.40})} \\

Qwen2.5-VL-7B & 78.32 & 15.16 &
3.11 {\scriptsize(\textcolor{red!70!black}{-12.05})} &
16.45 {\scriptsize(\textcolor{green!50!black}{+1.29})} &
31.92 {\scriptsize(\textcolor{green!50!black}{+16.76})} \\
\bottomrule
\end{tabular}
\caption{Performance (\%) across input settings, with changes relative to the image-only baseline.}
\label{tab:modality_valid}
\end{table*}

\subsection{Modality Ablation}
\label{subsec:modality_validation}

\paragraph{OCR insufficiency and modality gap.}
Table~\ref{tab:modality_valid} shows that OCR-based inputs yield extremely low accuracy, confirming that text extraction cannot recover the implicit visual cues required by \datasetname. 
Although the text-only and image-only settings are semantically equivalent, models perform near-perfectly on text but drop sharply on images, indicating that failures mainly stem from visual understanding rather than mathematical reasoning. 
Caption-based inputs improve performance but remain far below text-only results, suggesting that the tripartite schema $(\textit{object}, \textit{action}, \textit{composition})$ in Section~\ref{sec:method_step2} supports image synthesis but does not fully preserve the implicit relations needed for reasoning. 
Overall, the persistent modality gap shows that VLMs still struggle to ground mathematical logic in complex visual narratives. 
Further details are provided in Appendix~\ref{appendix:modality_valid}.

\subsection{Error Analysis}
\label{subsec:error_analysis}

\begin{figure}[t]
    \centering
    \includegraphics[width=0.95\linewidth]{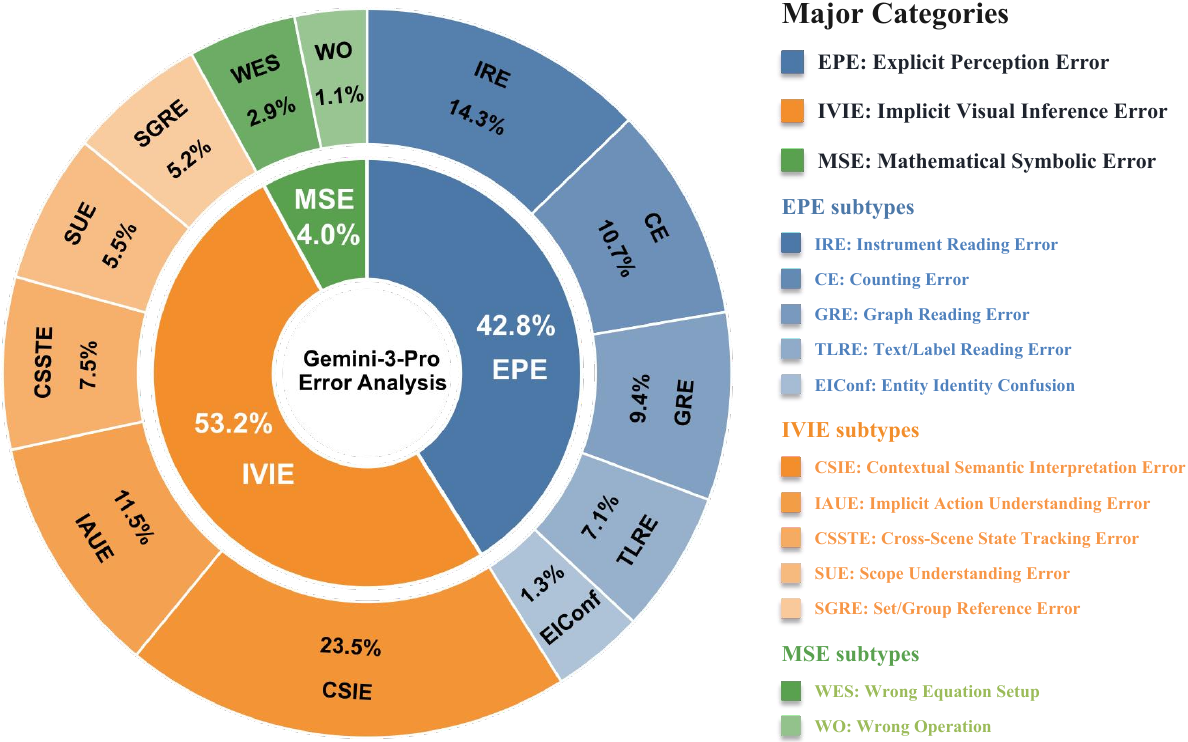}
    \caption{
    Fine-grained error distribution of Gemini-3-Pro on GSM8K-V. 
    The inner ring shows EPE, IVIE, and MSE, while the outer ring decomposes each category into fine-grained subtypes.
    }
    \label{fig:gemini_error}
\end{figure}

To understand why VLMs struggle on GSM8K-V, we conduct a fine-grained error analysis of model failures. 
We use Qwen3.5-122B-A10B~\citep{qwen2026qwen35} for first-pass error labeling with manual taxonomy auditing, and report Gemini-3-Pro results in Figure~\ref{fig:gemini_error}, with GPT-4o and Qwen2.5-VL-72B in Appendix~\ref{appendix:error_analysis}.

\paragraph{Explicit Perception Error (EPE).}
EPE refers to failures in reading explicit visual facts, such as object counts, text, instrument readings, graphs, and entity identities. 

\paragraph{Implicit Visual Inference Error (IVIE).}
IVIE refers to failures in recovering information implied by visual context, including actions, groupings, state changes, scope constraints, and cross-scene relations. 

\paragraph{Mathematical Symbolic Error (MSE).}
MSE refers to errors in equation construction or numerical computation. 

\paragraph{Key observations.}
As shown in Figure~\ref{fig:gemini_error}, IVIE is the dominant error category for Gemini-3-Pro, accounting for $53.2\%$ of the analyzed failures. 
EPE also remains substantial ($42.8\%$), showing that models still struggle to read explicit visual facts. 
By contrast, MSE accounts for only $4.0\%$ of failures, suggesting that most errors occur before the final symbolic calculation stage. 
Overall, the error distribution indicates that GSM8K-V primarily stresses implicit visual inference, where models must infer what visual evidence implies rather than merely perceive what is directly visible.

\section{Conclusion}

We introduce GSM8K-V, a benchmark designed to evaluate whether VLMs can perform mathematical reasoning when information is implicitly encoded across multiple images.
By converting GSM8K into a semantically equivalent visual counterpart, GSM8K-V isolates visual perception and cross-scene integration from mathematical computation. 
Evaluations on 34 VLMs reveal a clear modality gap, as text-based reasoning is nearly saturated whereas frontier models still struggle with implicit visual reasoning.
Error analysis identifies Implicit Visual Inference Error (IVIE) as the primary bottleneck, and models improved on structured math benchmarks show no gains on GSM8K-V. 
These results suggest that \datasetname provides a complementary diagnostic dimension for evaluating implicit visual inference and cross-scene integration, while sharing general visual-language capabilities with existing visual-mathematical benchmarks.

\section*{Limitations}
While \datasetname is designed as a controlled visual counterpart of GSM8K, its scope is limited to everyday arithmetic word problems with integer answers and does not cover broader mathematical domains such as geometry, algebra, proof, or scientific reasoning.
In addition, \datasetname is constructed using synthetic images under a 
controlled generation pipeline. Although our image-style and image-generator ablations suggest that the main findings are not specific to a particular rendering style or generator, synthetic scenes cannot fully reproduce the ambiguity, diversity, long-tail phenomena, and natural data distributions of real-world imagery. Therefore, \datasetname should be interpreted as a controlled diagnostic benchmark rather than a substitute for evaluation on naturally collected visual data.
Moreover, the benchmark relies on static multi-image sequences, whereas real-world multimodal agents often operate over continuous video streams and may use active perception, such as changing viewpoints or gathering additional information through interaction. Therefore, GSM8K-V should be viewed as a controlled benchmark for evaluating implicit visual reasoning and cross-scene integration, rather than a complete simulation of real-world embodied reasoning. Future work may extend this text-to-visual conversion paradigm to video-based settings and incorporate process-level annotations for more detailed reasoning diagnostics.

\section*{Ethical considerations}
In this paper, all evaluation samples in our constructed benchmark GSM8K-V are derived through a one-to-one mapping from the original GSM8K~\citep{cobbe2021training} benchmark, which has already undergone ethical review. Consequently, the content of GSM8K-V poses negligible ethical risks. 
During the human annotation process, we provided thorough training for all annotators. Their tasks involved only objective judgments, without engagement in subjective human preferences or exposure to ethical concerns. All annotators received fair compensation aligned with their workload. 
Furthermore, we collected feedback regarding their annotation experience, and they unanimously reported that they were able to accomplish the task satisfactorily and were appropriately rewarded.

\section*{Acknowledgments}
This work was supported by New Generation Artificial Intelligence-National Science and Technology Major Project (2025ZD0123100), National Natural Science Foundation of China (No. 62506332), "Pioneer" and "Leading Goose" R\&D Program of Zhejiang (NO. 2026C02A1223) and  the Key Research and Development Program of Ningbo (No. 2026Z003)

\bibliography{custom}

\definecolor{CaseBlue}{RGB}{28,82,150}
\definecolor{OkGreen}{RGB}{40,167,69}
\definecolor{WarnOrange}{RGB}{227,170,0}
\definecolor{BadRed}{RGB}{200,45,38}
\definecolor{LightGray}{RGB}{246,246,246}

\newcommand{\ie}{\textit{i.e.}\xspace}
\newcommand{\eg}{\textit{e.g.}\xspace}
\newcommand{\todoanon}[1]{{\color{red} [TODO: #1]}}
\newcommand{\torevise}[1]{{\color{red} [REVISE: #1]}}
\newcommand{\tocheck}[1]{{\color{red} [CHECK: #1]}}
\newcommand{\tofinish}[1]{{\color{red} [TO FINISH: #1]}}
\newcommand{\tocite}[1]{{\color{red} [TO CITE]}}
\newcommand{\revise}[2]{%
\textcolor{gray}{\st{#1}}~\textcolor{red}{#2}%
}

\newtcolorbox{casestudy}[1][]{
  enhanced, colback=white, colframe=CaseBlue, coltitle=CaseBlue,
  boxrule=0.8pt, arc=2pt, left=8pt, right=8pt, top=8pt, bottom=8pt,
  title={#1}, fonttitle=\bfseries, attach boxed title to top left={yshift*=-2mm},
  boxed title style={colback=white}, segmentation style={solid}
}

\newtcolorbox{subsecbox}[1][]{
  enhanced, colback=LightGray, colframe=black!15, boxrule=0.4pt, segmentation style={solid},
  arc=1.5pt, left=6pt, right=6pt, top=8pt, bottom=8pt, #1
}

\newtcolorbox{modelbox}[3][]{
  enhanced, colback=white, colframe=black!20, boxrule=0.5pt, arc=1.5pt,segmentation style={solid},
  left=7pt, right=7pt, top=8pt, bottom=6pt,
  overlay={
    \node[anchor=north west, xshift=4mm, yshift=0.5mm] at (frame.north west)
      {\includegraphics[height=13pt]{#3}};
  },
  title={\hspace*{12mm}\textbf{#1}}, fonttitle=\bfseries
}

\newtcolorbox{casefloatbox}[1][]{
  enhanced,
  unbreakable,
  colback=white,
  colframe=CaseBlue,
  coltitle=CaseBlue,
  boxrule=0.8pt,
  arc=2pt,
  left=8pt,
  right=8pt,
  top=8pt,
  bottom=8pt,
  title={#1},
  fonttitle=\bfseries,
  attach boxed title to top left={yshift*=-2mm},
  boxed title style={colback=white}
}

\newtcolorbox{subsecfloatbox}[1][]{
  enhanced,
  unbreakable,
  colback=LightGray,
  colframe=black!15,
  boxrule=0.4pt,
  arc=1.5pt,
  left=6pt,
  right=6pt,
  top=6pt,
  bottom=6pt,
  #1
}

\newtcolorbox{promptbox}[1][]{
  enhanced,
  width=\linewidth,
  colback=gray!1,      
  colframe=gray!60,    
  coltitle=black,      
  boxrule=2pt,
  arc=10pt,
  left=6pt, right=6pt, top=6pt, bottom=6pt,
  title={#1}, fonttitle=\bfseries,
  attach boxed title to top left={yshift*=-3mm},
  boxed title style={colback=gray!10}
}

\newtcolorbox{interferencebox}[1][]{
  enhanced,
  colback=gray!10, colframe=gray!60, coltitle=black,
  boxrule=0.6pt, arc=2pt,
  left=8pt, right=8pt, top=8pt, bottom=8pt,
  title={#1}, fonttitle=\bfseries,
  attach boxed title to top left={yshift*=-2mm},
  boxed title style={colback=gray!10}
}

\lstset{
  basicstyle=\ttfamily\scriptsize,
  breaklines=true,
  breakatwhitespace=true,
  columns=fullflexible,
  keepspaces=true,
  aboveskip=2pt,
  belowskip=2pt
}

\newcommand{\correcttag}{{\color{OkGreen}\faCheckCircle}\,Correct}
\newcommand{\wrongtag}{{\color{BadRed}\faTimesCircle}\,Wrong}

\clearpage
\appendix

\section{Comparison between GSM8K-V and Other Mathematical Benchmark}

\label{appendix:comparison_benchmark}

Existing visual mathematical benchmarks predominantly focus on structured diagrams or geometry-based layouts, where mathematical content is explicitly presented through symbols, labels, or standard notations. Their reasoning process relies on recognizing visual content and then converting it into symbolic form for calculation. In contrast, GSM8K-V evaluates whether models can extract mathematically relevant information when it is implicitly encoded across multiple images, requiring cross-scene integration and implicit visual inference. Table~\ref{tab:benchmark_comparison} summarizes the key differences between GSM8K-V and existing benchmarks across multiple dimensions.

\begin{table*}[htbp]
  \centering
  \small
  \begin{tabular}{l c c c c c}
    \toprule
    \textbf{Benchmark} & \textbf{Input} & \textbf{Math Info} & \textbf{Cross-Image} & \textbf{Implicit} & \textbf{Domain} \\
    & \textbf{Format} & \textbf{Encoding} & \textbf{Reasoning} & \textbf{Inference} & \\
    \midrule
    MathVista~\citep{lu2023mathvista} & Single-image & Explicit & \ding{55} & \ding{55} & Geometry, Algebra \\
    MathVision~\citep{wang2024measuring} & Single-image & Explicit & \ding{55} & \ding{55} & Geometry, Diagrams \\
    MathVerse~\citep{zhang2025mathverse} & Single-image & Explicit & \ding{55} & \ding{55} & Geometry, Algebra \\
    VideoMathQA~\citep{rasheed2025videomathqa} & Video & Explicit & \ding{51} & \ding{55} & Instructional \\
    \midrule
    \textbf{GSM8K-V (Ours)} & \textbf{Multi-image} & \textbf{Implicit} & \ding{51} & \ding{51} & \textbf{General} \\
    \bottomrule
  \end{tabular}
  \vspace{1mm}
  
  \raggedright
  \footnotesize
  \textbf{Math Info Encoding}: Whether mathematical information is explicitly labeled (symbols, numbers, labels) or implicitly embedded (requiring perception and inference). \\
  \textbf{Cross-Image Reasoning}: Whether solving requires integrating information across multiple images/frames. \\
  \textbf{Implicit Inference}: Whether reasoning requires inferring unstated information from visual context (e.g., actions from scene transitions). \\
  \textbf{Domain}: Geometry/Algebra = structured mathematical diagrams; Instructional = educational video content; General = diverse word problem scenarios.
    \caption{Comparison of visual mathematical benchmarks across multiple dimensions.}
  \label{tab:benchmark_comparison}
\end{table*}

\section{Detailed Evaluation Setup}

\label{appendix:Evaluation Setup}
In this section, we report the detailed evaluation settings and mode.

\subsection{Model Hyperparameters}
We set temperature = 0.2 for closed-source models and temperature = 0.1 for open-source models to reduce the randomness in the model generation. Table ~\ref{appendix:hyperparams}
displays the parameters we used for generation in VLMs.

\begin{table*}[htbp]
\centering
\small
\setlength{\tabcolsep}{6pt}

\begin{tabular}{@{\extracolsep{\fill}} l l}
\toprule
\textbf{Model} & \textbf{Hyperparameters} \\
\midrule
\rowcolor[rgb]{ .91,  .91,  .91} 
\multicolumn{2}{l}{\textit{Closed-Source VLMs (other settings follow official defaults)}} \\
\midrule
Gemini-3-Pro, Gemini-2.5-Pro \citep{comanici2025gemini} 
& temperature = 0.2, max\_tokens = 4096 \\
GPT-5 \citep{singh2025openai} 
& temperature = 1.0, max\_tokens = 4096 \\
& reasoning\_effort = minimal, verbosity = low \\
GPT-4o \citep{openai2024gpt4o} 
& temperature = 0.1, max\_tokens = 2048 \\
QVQ-Max-Latest \citep{qwen2025qvqmax} 
& temperature = 0.1, max\_tokens = 4096 \\

\midrule
\rowcolor[rgb]{ .91,  .91,  .91} 
\multicolumn{2}{l}{\textit{Open-Source VLMs}} \\
\midrule
Llama-4-17B-16E-Instruct \citep{meta2024llama4} 
& temperature = 0.1, max\_tokens = 2048 \\
Llama-4-17B-128E-Instruct \citep{meta2024llama4} 
& temperature = 0.1, max\_tokens = 2048 \\

InternVL3.5-8B \citep{wang2025internvl35} 
& temperature = 0.1, max\_tokens = 2048 \\
InternVL3.5-38B \citep{wang2025internvl35} 
& temperature = 0.1, max\_tokens = 2048 \\
InternVL3.5-30B-A3B \citep{wang2025internvl35} 
& temperature = 0.1, max\_tokens = 2048 \\
InternVL3.5-241B-A28B \citep{wang2025internvl35} 
& temperature = 0.1, max\_tokens = 2048 \\

Qwen3.6-35B-A3B \citep{qwen2026qwen36a3b} 
& temperature = 0.1, max\_tokens = 2048 \\
Qwen3.6-35B-A3B-Thinking \citep{qwen2026qwen36a3b} 
& temperature = 1.0, max\_tokens = 8192 \\

Qwen3.5-9B \citep{qwen2026qwen35} 
& temperature = 0.1, max\_tokens = 2048 \\
Qwen3.5-27B \citep{qwen2026qwen35} 
& temperature = 0.1, max\_tokens = 2048 \\
Qwen3.5-122B-A10B \citep{qwen2026qwen35} 
& temperature = 0.1, max\_tokens = 2048 \\
Qwen3.5-122B-A10B-Thinking \citep{qwen2026qwen35} 
& temperature = 1.0, max\_tokens = 8192 \\

Qwen3-VL-8B-Instruct \citep{bai2025qwen3vl} 
& temperature = 0.1, max\_tokens = 2048 \\
Qwen3-VL-8B-Thinking \citep{bai2025qwen3vl} 
& temperature = 0.1, max\_tokens = 8192 \\
Qwen3-VL-30B-A3B-Instruct \citep{bai2025qwen3vl} 
& temperature = 0.1, max\_tokens = 2048 \\
Qwen3-VL-30B-A3B-Thinking \citep{bai2025qwen3vl} 
& temperature = 0.1, max\_tokens = 8192 \\

Qwen2.5-VL-7B-Instruct \citep{bai2025qwen25vl} 
& temperature = 0.1, max\_tokens = 2048 \\
Qwen2.5-VL-32B-Instruct \citep{bai2025qwen25vl} 
& temperature = 0.1, max\_tokens = 2048 \\
Qwen2.5-VL-72B-Instruct \citep{bai2025qwen25vl} 
& temperature = 0.1, max\_tokens = 2048 \\

Ovis2.5-2B \citep{lu2025ovis25} 
& temperature = 0.1, max\_tokens = 2048 \\
Ovis2.5-9B \citep{lu2025ovis25} 
& temperature = 0.1, max\_tokens = 2048 \\

Step3 \citep{stepfun2025step3} 
& temperature = 0.1, max\_tokens = 2048 \\

Kimi-K2.6 \citep{moonshot2026kimik26} 
& temperature = 0.1, max\_tokens = 8192 \\
Kimi-VL-A3B-Thinking-2506 \citep{team2025kimivl} 
& temperature = 0.1, max\_tokens = 8192 \\

MiniCPM-V-4.6 \citep{openbmb2026minicpmv46} 
& temperature = 0.1, max\_tokens = 2048 \\
MiniCPM-V-4.6-Thinking \citep{openbmb2026minicpmv46} 
& temperature = 1.0, max\_tokens = 8192 \\
MiniCPM-V 4.5 \citep{yao2024minicpmv} 
& temperature = 0.1, max\_tokens = 2048 \\

GLM-4.6V \citep{zai2025glm46v_hf} 
& temperature = 0.1, max\_tokens = 2048 \\
GLM-4.5V \citep{team2026glm45v} 
& temperature = 0.1, max\_tokens = 2048 \\

\bottomrule
\end{tabular}
\captionsetup{width=\linewidth}
\caption{Hyperparameter settings used for all evaluated vision-language models. For closed-source VLMs, all unlisted configurations follow the official default settings.}
\label{appendix:hyperparams}
\end{table*}

\subsection{Complete Evaluation Results}
\label{appendix:complete_results}

To complement the main results presented in Section~\ref{sec:evaluation}, 
we provide the complete performance comparison of all evaluated models in 
Table~\ref{tab:appendix_full_results}. 
While Table~\ref{tab:main_results} summarizes the main evaluation results, 
Table~\ref{tab:appendix_full_results} reports the full results across all six 
visual reasoning categories, the overall GSM8K-V average accuracy, and the 
corresponding text-only GSM8K performance.

\begin{table*}[t]
\centering
\small
\setlength{\tabcolsep}{5.5pt}        %
\begin{tabular}{lcccccc>{\columncolor[gray]{0.95}}cc}  %
\toprule
\textbf{Model} &
\multicolumn{7}{c}{\textbf{GSM8K-V}} &
\textbf{GSM8K} \\
\cmidrule(lr){2-8} \cmidrule(lr){9-9}

& \makecell{Signboard \\ \& Icon} 
& \makecell{Measu- \\ rement} 
& \makecell{Physical \\ Metric} 
& \makecell{Temp- \\oral} 
& \makecell{Ratio \& \\ Percentage} 
& \makecell{Other} 
& \makecell{Avg.} 
& \makecell{Avg.}  \\
\midrule
Human & \textbf{90.27} & \textbf{86.76} & \textbf{96.05} & \textbf{91.20} & \textbf{88.57} & \textbf{93.25} & \textbf{91.15} & - \\
\midrule

\rowcolor[rgb]{ .91,  .91,  .91} \multicolumn{9}{l}{\textit{Closed-Source VLMs}} \\
\midrule

Gemini-3-Pro & \textbf{66.18} & \textbf{63.11} & \textbf{62.18} & \textbf{57.30} & \textbf{56.54} & \textbf{54.34} & \textbf{59.06} & \textbf{95.68} \\

Gemini-2.5-Pro & \underline{52.94} & \underline{52.43} & \underline{56.30} & \underline{47.57} & \underline{42.05} & \underline{41.18} & \underline{46.93} & \underline{95.22} \\

GPT-5 & 41.54 & 46.60 & 26.89 & 28.65 & 21.91 & 24.93 & 30.10 & 95.00 \\
GPT-4o & 41.91 & 42.72 & 31.93 & 28.11 & 21.91 & 22.41 & 29.57 & 94.92 \\
QVQ-Max-Latest & 30.51 & 39.81 & 21.85 & 25.95 & 16.25 & 17.37 & 23.20 & 95.00 \\

\midrule
\rowcolor[rgb]{ .91,  .91,  .91}\multicolumn{9}{l}{\textit{Open-Source VLMs}} \\
\midrule

Llama-4-17B-16E-Instruct & 36.76 & 36.89 & 30.25 & 25.41 & 19.79 & 21.01 & 26.69 & 94.16 \\
Llama-4-17B-128E-Instruct & 42.28 &  41.75 & 28.57 & 34.59 & 22.61 & 27.45 &  31.69 & \underline{95.30} \\

InternVL3.5-8B & 16.91 & 24.27 & 10.92 & 8.65 & 5.30 & 7.56 & 10.77 & 90.67 \\
InternVL3.5-38B & 36.03 & 38.84 &  31.09 & 25.41 & 17.67 & 22.97 & 26.84 & 93.86 \\
InternVL3.5-30B-A3B & 29.41 & 37.86 &  21.85 &  25.41 & 13.43 & 19.89 &  22.82 & 93.18 \\
InternVL3.5-241B-A28B & 36.76 & 46.60 & 28.57 & 27.03 & 21.20 & 21.29 & 27.90 & 95.22 \\

Qwen3.6-35B-A3B & 38.44 & 39.46 & 44.49 & 32.56 & 24.82 & 24.60 & 30.71 & 94.16 \\

Qwen3.6-35B-A3B-Thinking & 39.73 & 40.78 & 45.98 & 33.65 & 25.65 & 25.43 & 31.74 & 94.09 \\

Qwen3.5-9B & 33.90 & 44.95 & 29.02 & 28.22 & 21.21 & 21.48 & 27.82 & 94.24 \\

Qwen3.5-27B & 43.04 & 42.42 & \textbf{49.08} & 42.18 & \underline{37.79} & 34.40 & 39.88 & 94.77 \\

Qwen3.5-122B-A10B & 50.87 & 46.21 & 39.58 & 36.42 & 35.15 & 34.76 & 40.08 & 94.47 \\

Qwen3.5-122B-A10B-Thinking & \underline{54.62} & \underline{49.34} & 42.48 & 39.08 & 36.21 & \underline{37.27} & \underline{42.12} & \underline{95.30} \\

Qwen3-VL-8B-Instruct & 34.69 & 35.92 & 28.81 & 29.28 & 15.60 & 22.19 & 26.00 & 94.01 \\
Qwen3-VL-8B-Thinking & 39.71 & 35.92 & 29.66 & 27.07 & 15.25 & 21.07 & 26.46 & 90.14 \\
Qwen3-VL-30B-A3B-Instruct & 37.13 & 39.81 & 33.05 & 30.39 & 17.73 & 22.47 & 27.90 & 94.24 \\
Qwen3-VL-30B-A3B-Thinking & 37.13 & 40.78 & 31.36 & 32.60 & 21.28 & 21.35 & 28.58 & 92.42 \\
Qwen2.5-VL-7B-Instruct & 24.26 & 24.27 & 17.65 & 11.35 & 7.07 & 13.17 & 15.16 & 78.32 \\
Qwen2.5-VL-32B-Instruct & 30.51 & 38.83 & 18.49 & 20.54 & 13.07 & 18.77 & 21.76 & 88.70 \\
Qwen2.5-VL-72B-Instruct & 28.31 & 36.89 & 21.85 & 24.86 & 13.43 & 17.93 & 21.91 & 94.09 \\

Ovis2.5-2B & 19.12 & 27.18 & 15.13 & 13.51 & 6.01 & 10.64 & 13.50 & 86.28 \\
Ovis2.5-9B & 37.13 & 38.83 & 31.09 & 24.86 & 18.02 & 24.37 & 27.45 & 94.16 \\

Step3 & 34.19 & 34.95 & 33.61 & 33.51 & 15.19 & 18.77 & 25.85 & 83.09 \\

Kimi-K2.6 & \textbf{62.50 }& \textbf{51.35} & \underline{48.33} & \textbf{49.06} & \textbf{40.56} & \textbf{39.03} & \textbf{47.34} & \textbf{95.75} \\

Kimi-VL-A3B-Thinking-2506 & 17.28 & 20.39 & 10.92 & 6.49 & 6.71 & 11.48 & 11.60 & 84.15 \\

MiniCPM-V-4.6 & 15.81 & 13.59 & 11.76 & 7.03 & 4.24 & 8.12 & 9.48 & 75.51 \\

MiniCPM-V-4.6-Thinking & 17.65 & 20.39 & 12.61 & 8.11 & 4.95 & 9.04 & 11.02 & 85.11 \\

MiniCPM-V 4.5 & 26.10 & 29.13 & 18.49 & 18.92 & 12.01 & 18.49 & 19.56 & 90.83 \\

GLM-4.6V & 48.29 & 46.00 & 35.09 & \underline{42.52} & 27.91 & 32.12 & 37.78 & 94.66 \\

GLM-4.5V & 34.93 & 42.72 & 26.05 & 31.35 & 22.26 & 22.97 & 28.28 & 90.60 \\

\bottomrule
\end{tabular}
\caption{Complete performance (\%) comparison of vision-language models across six visual reasoning categories, overall average accuracy on GSM8K-V, and text-only GSM8K accuracy.}
\label{tab:appendix_full_results}
\end{table*}

\subsection{Evaluation Mode}
We provide detailed definitions and examples for each evaluation mode, and illustrate the input format, example questions, and the exact prompts used in our experiments.

\label{appendix:Evaluation Modes}

\vspace{5mm}

\subsubsection{Text Only}

\textbf{Answer Format Prompt.}
\begin{lstlisting}
When providing your final answer:
- If the answer can be expressed as a whole number (integer), provide it as an integer
- If the answer requires decimals and can be expressed with a finite decimal, use decimal format
- Otherwise, use fraction format if appropriate
Please think step by step. After your reasoning, output your 
final answer on a new line starting with "FINAL ANSWER: " 
followed by the number only.

\end{lstlisting}

\vspace{5mm}

\begin{promptbox}[{Text Only Mode}]
\textbf{Meta Data:} 
\begin{lstlisting}

{
   "question_id": "195",
   "original_question": "Violetta wants to buy new crayons. She needs them in 5 different colors and prepared $20 for this purchase. One crayon costs $2. How much change will she get?",
   "math_ground_truth": "10"
}

\end{lstlisting}

\vspace{4mm}

\textbf{Input (question):}  
\emph{``A pen costs as much as a pencil and eraser combined. A pencil costs \$1.20 and an eraser costs \$0.30. How much will 8 pens cost?''}

\vspace{4mm}

\textbf{Mode Prompt.}
\begin{lstlisting}

You are an expert at solving mathematical word problems.
Please solve the following problem step by step.
Problem: {question}
# Answer Prompt:
{Answer Format Prompt}

\end{lstlisting}
\end{promptbox}

\subsubsection{Implicit and Explicit}
\label{appendix:implicit_explicit}

\begin{figure*}[h]
    \centering
    \includegraphics[width=1\linewidth]{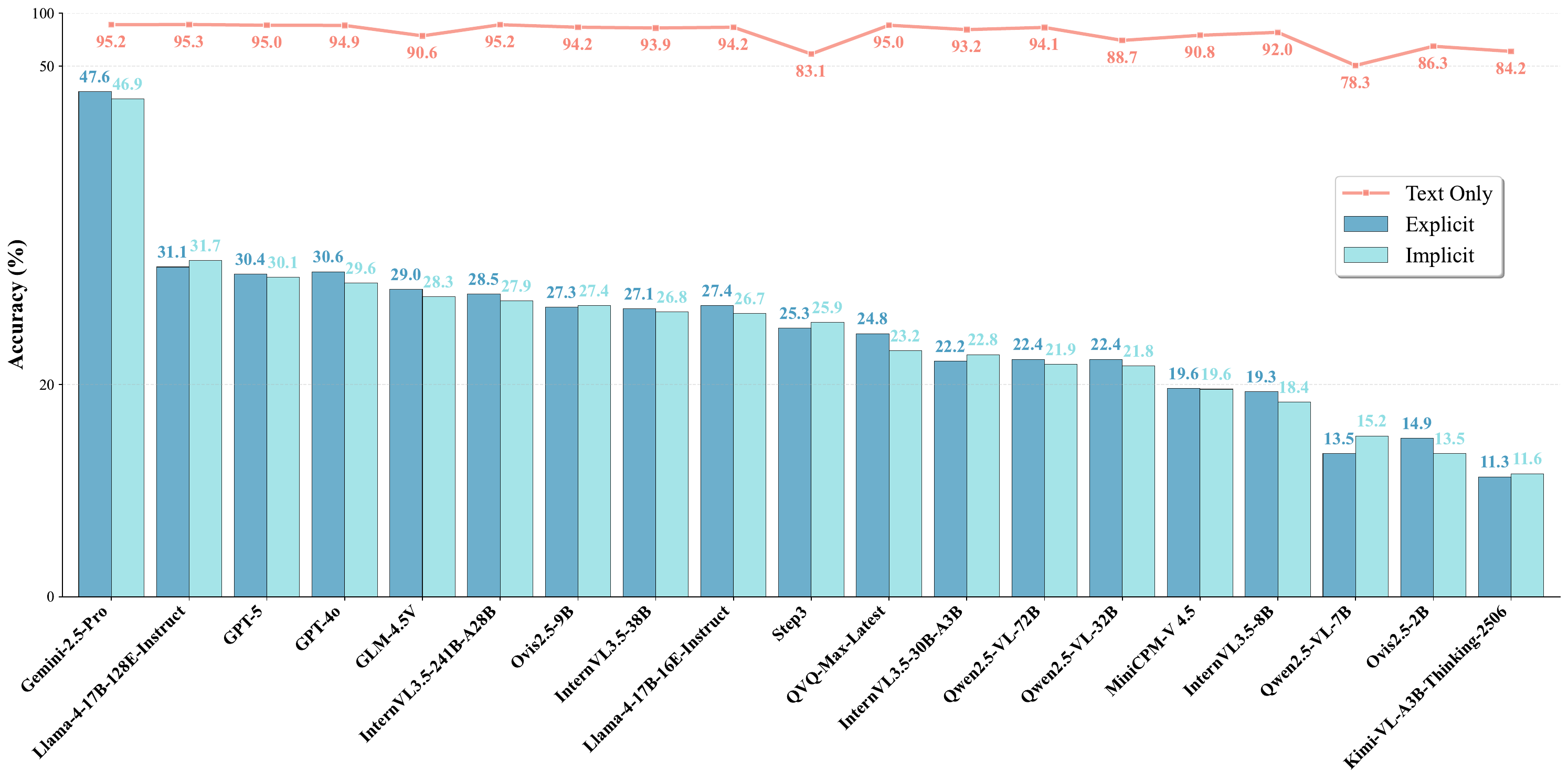}
    \caption{Illustration of performance (\%) comparison between explicit and implicit evaluation settings.}
    \label{fig:explicit_implicit_comparison}
\end{figure*}

\paragraph{Explicit and Implicit.} Traditional visual benchmarks typically treat images as feature inputs and pose questions about them through textual descriptions.
In contrast, \datasetname embeds the actual problem to be solved directly within the image. To investigate whether models can capture such problem statements from images, we conducted an ablation study to examine whether explicitly providing the question in textual form can improve model performance. We compare two input modes: \emph{implicit}, where only scene images are provided, and \emph{explicit}, 
where images are paired with a masked question linking visual entities to semantics.
To better illustrate the two evaluation results, we provide the visualization in Figure~\ref{fig:explicit_implicit_comparison}. 

\begin{table*}[htbp]
\centering
\small
\renewcommand{\arraystretch}{1.3}
\setlength{\tabcolsep}{6pt}
\begin{tabular}{l|c|c|c|c|c|c|c|c}
\toprule
\textbf{Model} &
\makecell{\textbf{Text} \\ \textbf{only}} &
\makecell{\textbf{Image} \\ \textbf{only}} &
\makecell{\textbf{Qwen-} \\ \textbf{OCR}} &
\makecell{\textbf{Paddle} \\ \textbf{OCR} \\ \textbf{-VL}} &
\makecell{\textbf{Min} \\ \textbf{eru} \\ \textbf{2.5}} &
\makecell{\textbf{Qwen} \\ \textbf{-OCR +}\\ \textbf{Image}} &
\makecell{\textbf{Paddle} \\ \textbf{OCR-VL} \\ \textbf{+ Image}} &
\makecell{\textbf{MinerU} \\ \textbf{2.5 + }\\ \textbf{Image}} \\
\midrule
GPT-4o & 94.92 & 29.57 & 11.98 & 12.05 & 12.13 & 33.06 & 39.50 & 39.12 \\
GLM-4.5V & 90.60 & 28.28 & 9.33 & 10.24 & 10.01 & 32.22 & 37.98 & 37.91 \\
Qwen2.5-VL-72B & 94.09 & 21.91 & 9.63 & 10.08 & 10.01 & 24.34 & 27.82 & 27.82 \\
Qwen2.5-VL-7B & 78.32 & 15.16 & 3.11 & 4.17 & 4.02 & 16.45 & 17.97 & 17.74 \\
\bottomrule
\end{tabular}
\centering
\newline
\captionsetup{width=\textwidth}
\caption{Performance (\%) comparison on different input settings.}

\label{tab:ocr_modality_comparison}
\end{table*}

As shown in Figure~\ref{fig:explicit_implicit_comparison}, providing explicit textual queries generally yields slightly higher accuracy across most models, as textual grounding helps reduce ambiguity in visual reference resolution. However, even with this explicit support, the performance of state-of-the-art VLMs remains remarkably limited. For instance, the top-performing Gemini-2.5-Pro achieves only  in the explicit setting, which is a staggering drop compared to its  accuracy on text-only GSM8K. 

This pervasive performance gap underscores that the core challenge of \datasetname lies not just in understanding the question, but in the complex multi-image reasoning required to solve the problem. The fact that robust textual reasoning capabilities on GSM8K do not translate to visual success highlights a critical deficiency in current VLMs' ability to handle real-world mathematical problems in visual contexts. These results validate the practical value of our benchmark as a challenging frontier for the next generation of multimodal models. In the following, we provide the detailed formats of both \textit{implicit} and \textit{explicit} inputs and prompts.

In the \textit{implicit mode}, for each mathematical problem, we only provide the model with several images and a textual prompt. The prompt does not contain any mathematical information or the actual problem statement; it solely specifies the required output format. After receiving the prompt and the input images, the model generates an output according to the prescribed format. We then parse and verify the model’s output, and finally evaluate the correctness of the predicted answer.

In the \textit{explicit mode}, for each mathematical problem, we provide the model with several images, a pre-processed textual representation of the problem (which excludes any mathematical information), and a textual prompt. The prompt also contains no mathematical information or actual problem statement; it merely specifies the required output format.

\begin{figure*}[t]
\centering
\begin{promptbox}[{Implicit Mode}]
\small

\textbf{Meta Data:}
\begin{lstlisting}
{
  "question_id": "195",
  "original_question": "xx(the same as text only mode)",
  "math_ground_truth": "10",
  "masked_question": "How much change will Violetta (long wavy brown hair, pink backpack, light green shirt) get after buying her crayons?",
  "pic_ids": ["195_1.jpg", "195_2.jpg", "195_3.jpg", "195_4.jpg"]
}
\end{lstlisting}

\vspace{1mm}
\textbf{Input(Only Images):}

\vspace{1mm}
\centering
\begin{minipage}{0.23\linewidth}
  \centering
  \includegraphics[width=\linewidth]{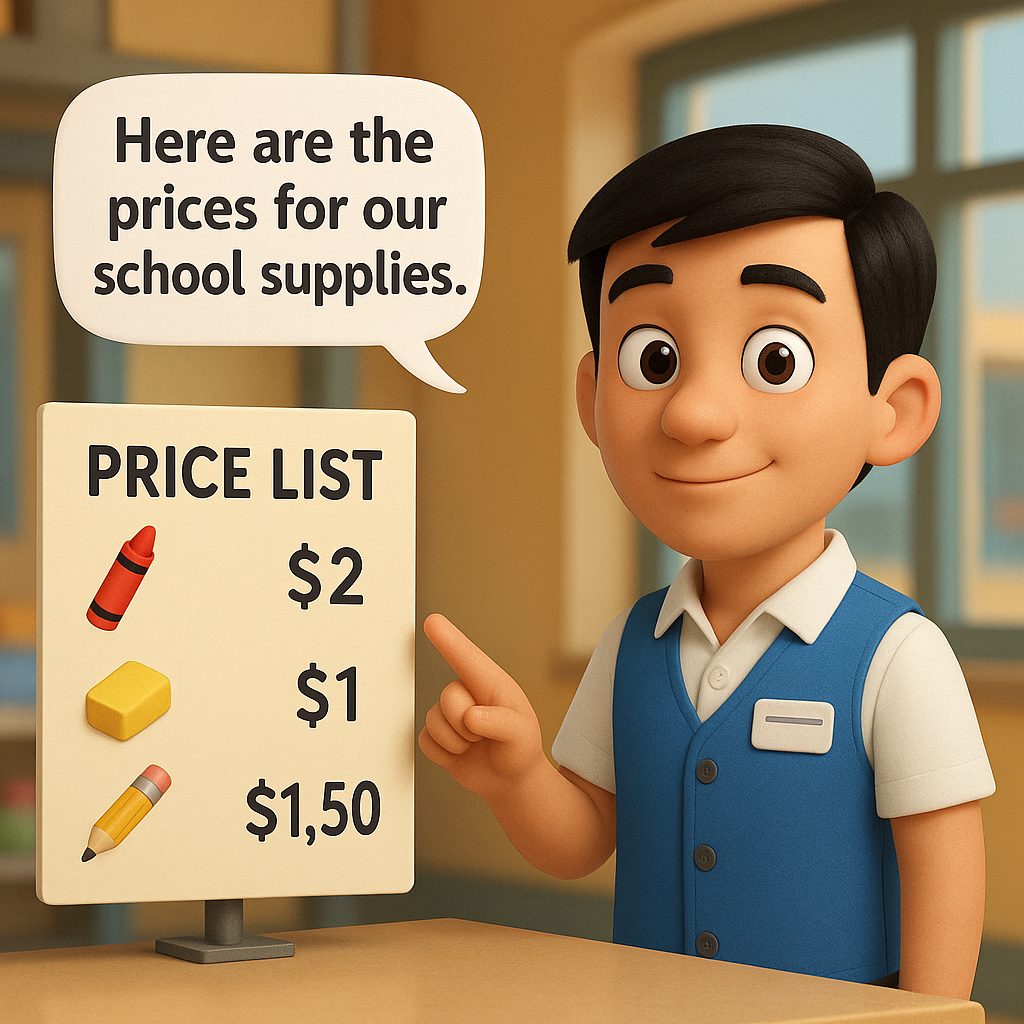}
\end{minipage}\hfill
\begin{minipage}{0.23\linewidth}
  \centering
  \includegraphics[width=\linewidth]{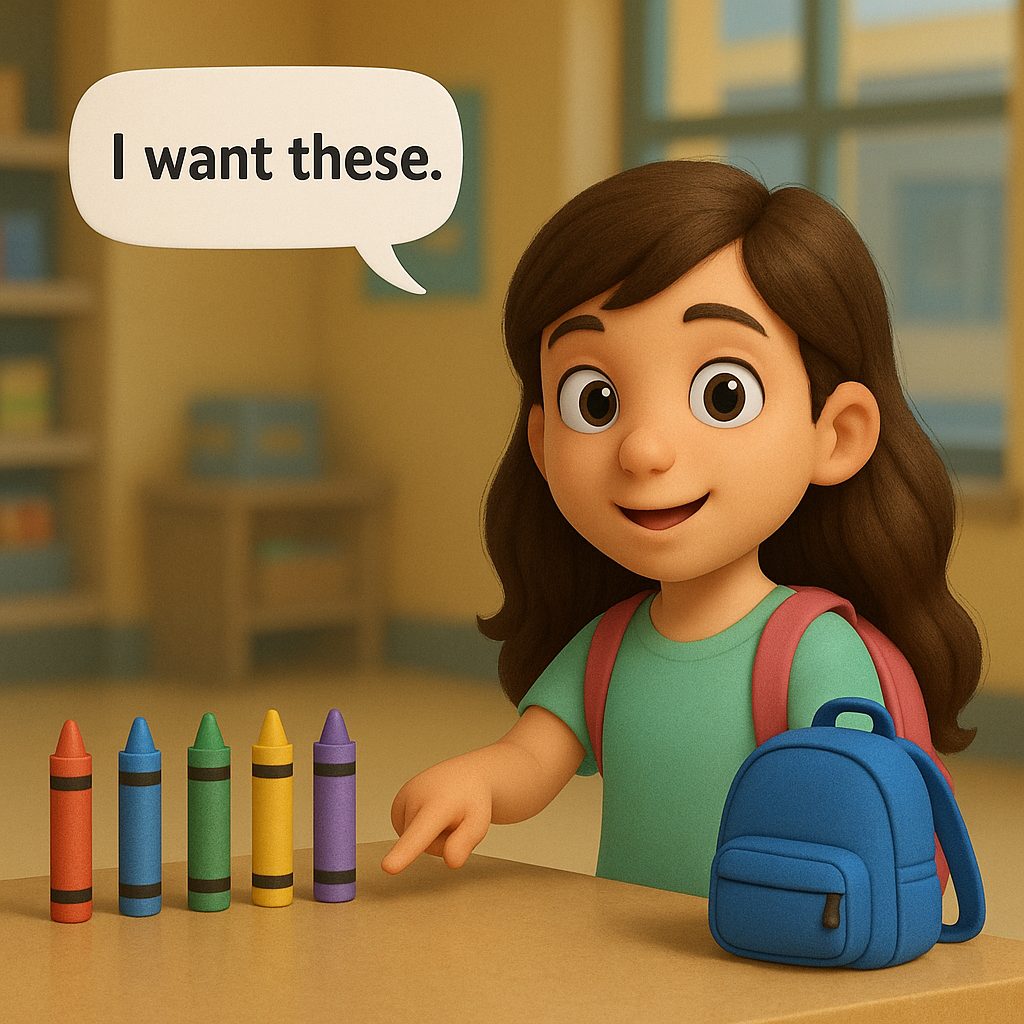}
\end{minipage}\hfill
\begin{minipage}{0.23\linewidth}
  \centering
  \includegraphics[width=\linewidth]{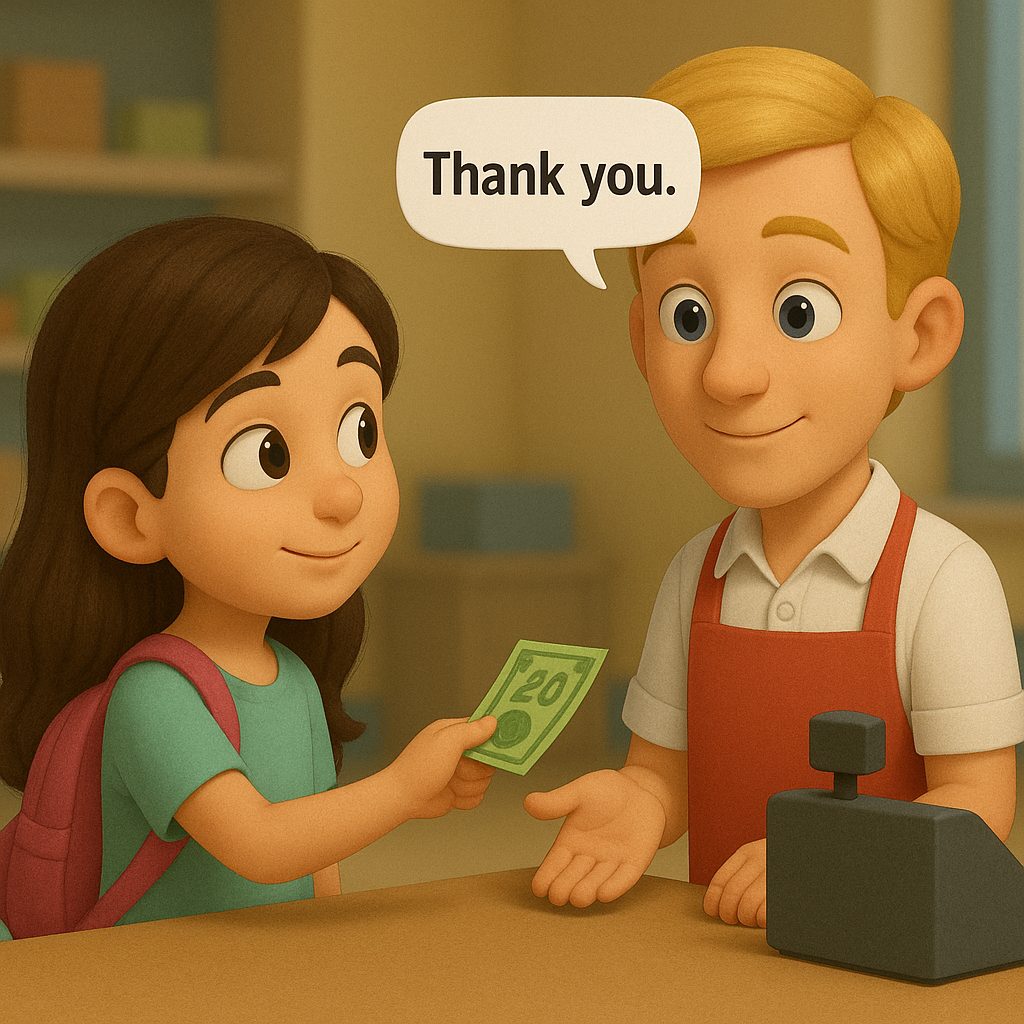}
\end{minipage}\hfill
\begin{minipage}{0.23\linewidth}
  \centering
  \includegraphics[width=\linewidth]{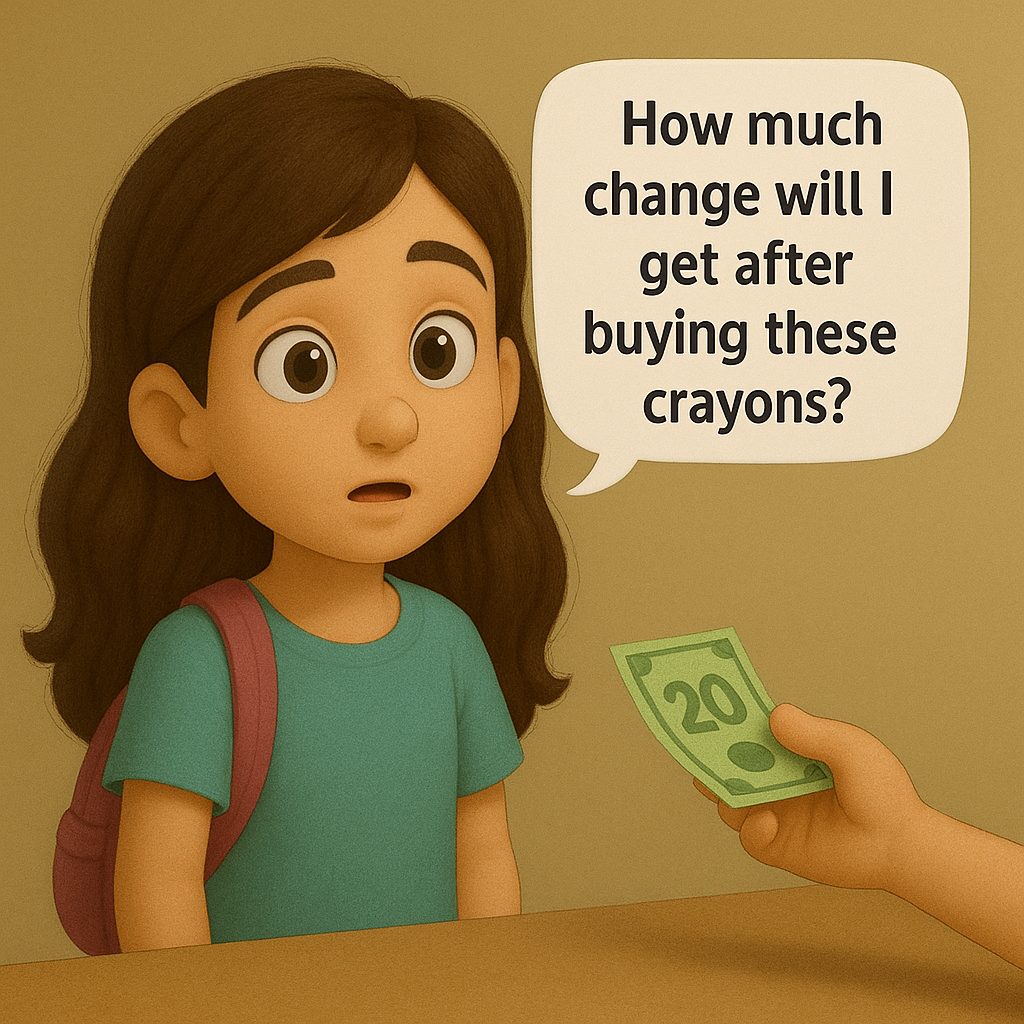}
\end{minipage}

\vspace{3mm}

\textbf{Prompt.}

\begin{lstlisting}
You are an expert at solving mathematical problems based on visual information. I'll show you some images that contain a math problem or story. Answer the question in the images.
# Answer Prompt:
{Answer Format Prompt}
\end{lstlisting}

\end{promptbox}
\end{figure*}

\begin{figure*}[t]
\centering

\begin{promptbox}[{ Explicit Mode}]
\textbf{Meta Data:} 
\begin{lstlisting}
{
   "question_id": "195",
    "original_question": "xx(the same as text only mode)",
    "math_ground_truth": "10",
    "masked_question": "How much change will Violetta (long wavy brown hair, pink backpack, light green shirt) get after buying her crayons?",
    "pic_ids": [ "195_1.jpg", "195_2.jpg", "195_3.jpg", "195_4.jpg"]
}
\end{lstlisting}

\vspace{3mm}
\textbf{Input(Images + Masked Question)}  

The model receives four scene images representing the problem, 
together with a masked version of the original text question.  

\begin{center}
\begin{tabular}{cc}
    \includegraphics[width=0.24\linewidth]{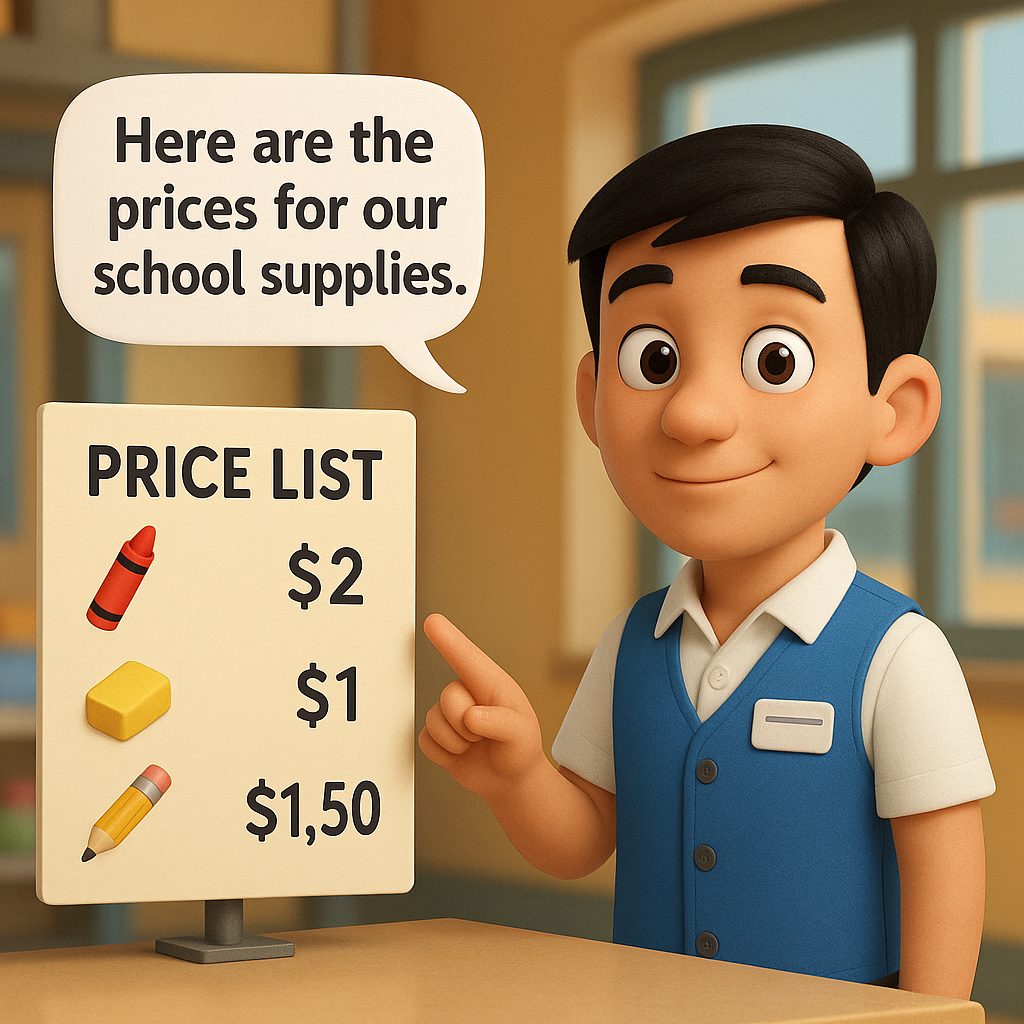}
    \includegraphics[width=0.24\linewidth]{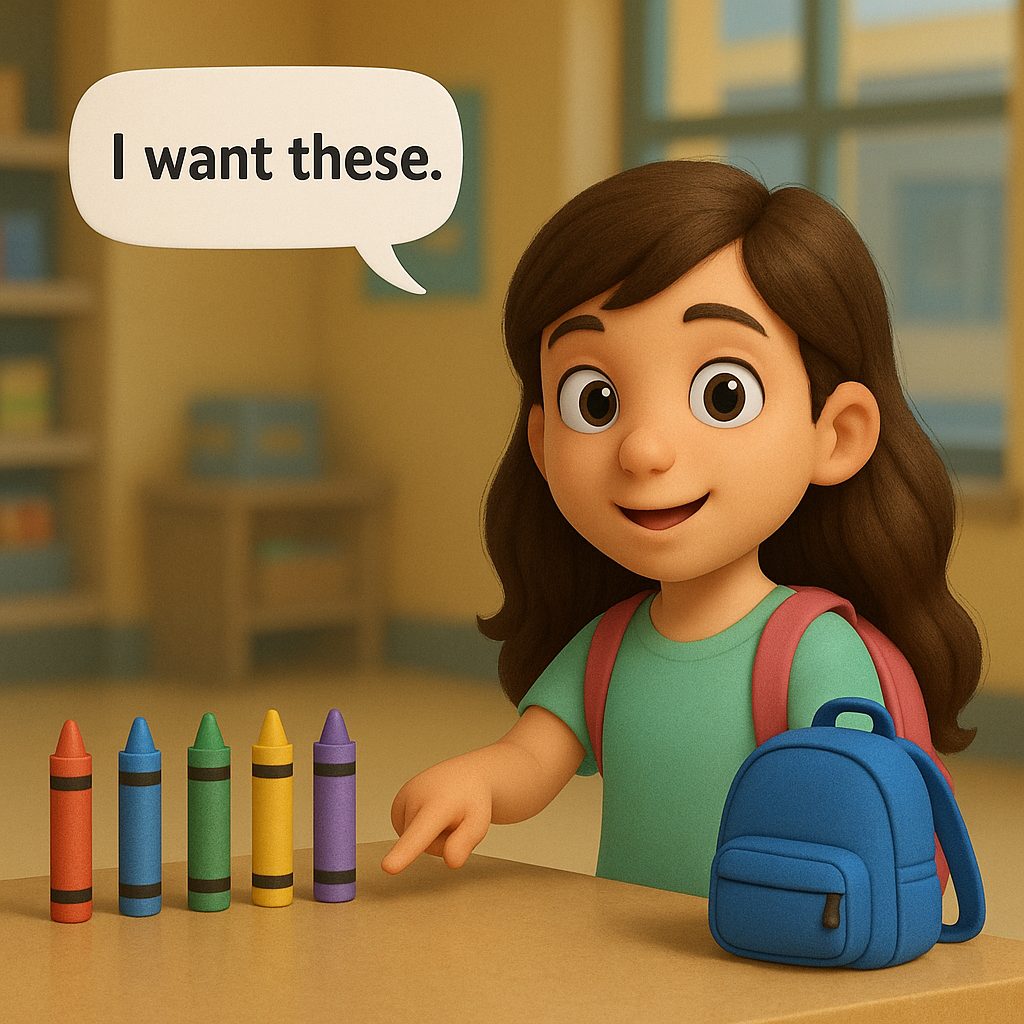} 
    \includegraphics[width=0.24\linewidth]{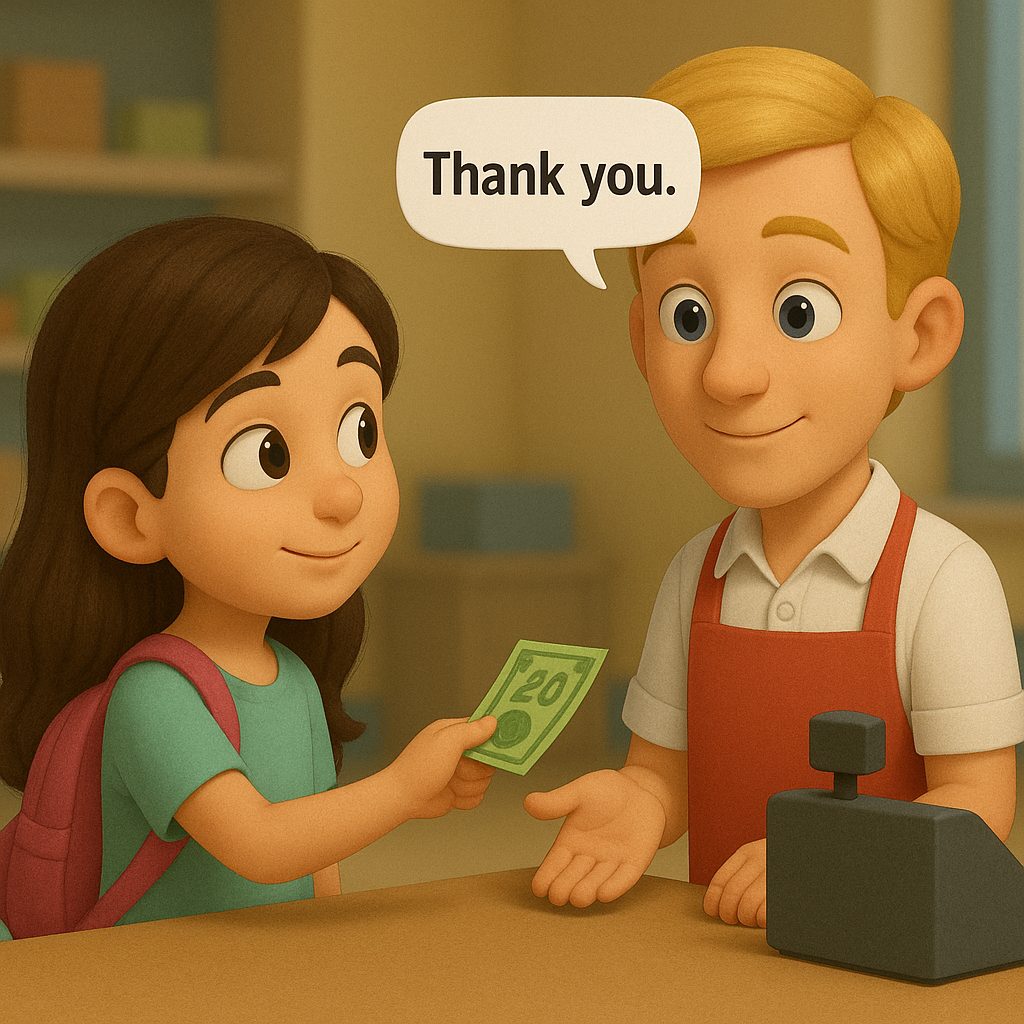}
    \includegraphics[width=0.24\linewidth]{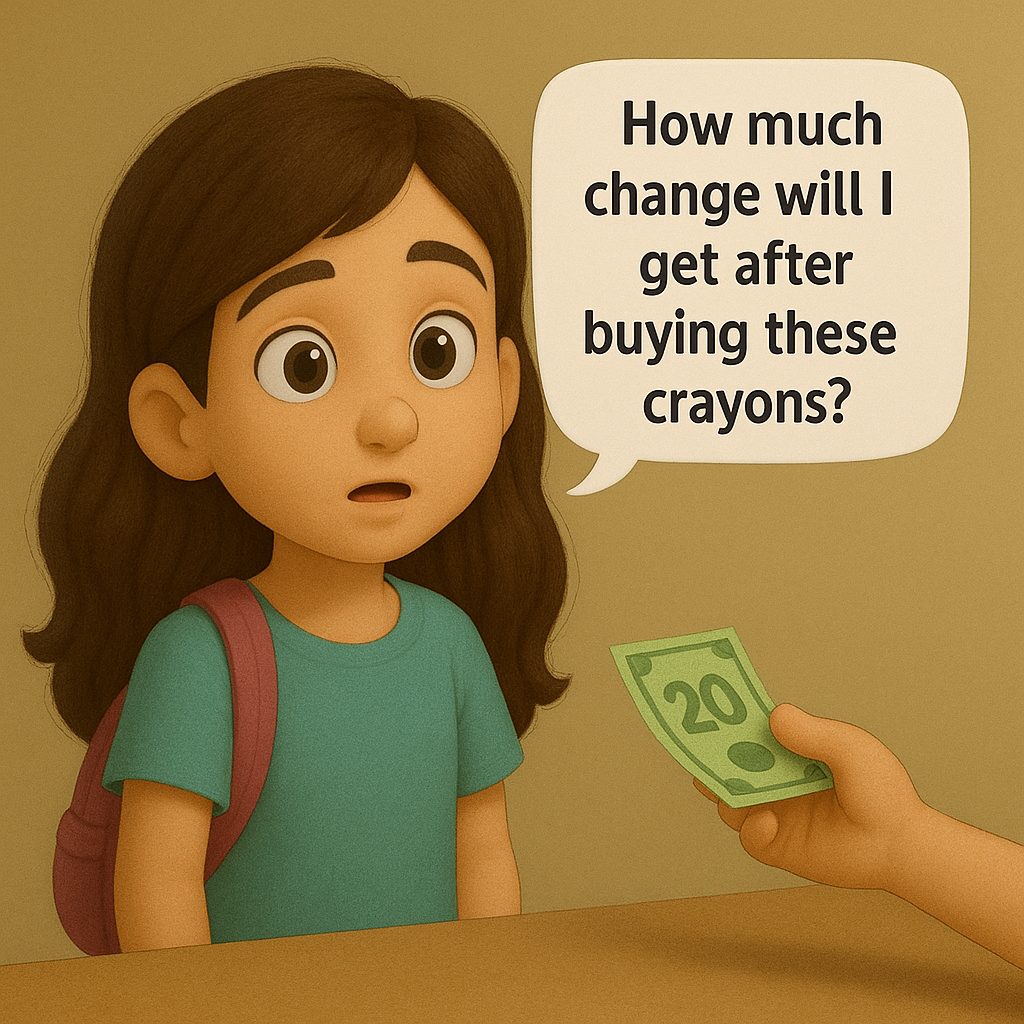} \\
\end{tabular}
\end{center}

\textbf{masked question:}  
\emph{``How much change will Violetta (long wavy brown hair, pink backpack, light green shirt) get after buying her crayons?''}

\vspace{3mm}

\textbf{Prompt.}
\begin{lstlisting}
You are an expert at solving mathematical problems based on visual information. I'll show you one or more images related to a math problem, along with a question.
Please analyze the images and answer the math question step by step. Question to answer: {masked question}
# Answer Prompt:
{Answer Format Prompt}

\end{lstlisting}
\end{promptbox}

\end{figure*}

\subsection{Modality Valid}
\label{appendix:modality_valid}

In GSM8K-V, OCR is intentionally insufficient for solving the problems. To evaluate this, we conduct the ocr experiment in Table~\ref{tab:ocr_modality_comparison}. 
Text only means the original textual question. Image only uses the images as the sole input. Qwen-OCR, PaddleOCR-VL and MinerU2.5 provide the model with the OCR-extracted text generated by Qwen-OCR, PaddleOCR-VL or MinerU2.5, respectively. Qwen-OCR + Image, PaddleOCR-VL + Image and MinerU2.5 + Image combine both the original image and the corresponding OCR results as the model input for joint reasoning.
From Table~\ref{tab:ocr_modality_comparison}, We observe that when models are given only the outputs of OCR Model, their accuracy remains very low. This indicates that GSM8K-V cannot be solved via text recognition alone and necessarily requires implicit visual reasoning. Adding the original image to the OCR results yields some improvement, but still limited. This shows that OCR capability alone is insufficient, and the small accuracy gain further indicates that additional types of errors are involved beyond text recognition. 

An illustrative instance is an event in which an agent consumes five apples. The caption may list entities such as the agent, apples, a table, spatial arrangement and gesture such as pointing toward the apples. A later scene may retain the table and agent while omitting the apples. Although coherent for guiding image generation, this structure does not explicitly encode the act of eating. The underlying semantics must be inferred through cross-scene continuity and visual logical reasoning. This behavior reflects a broader distinction between visual and textual modalities. 
Such representational differences contribute to the remaining performance gap between caption-based and text-only settings.

\begin{figure*}[t]
\centering

\begin{promptbox}[{OCR Mode}]
\textbf{Meta Data:} 
\begin{lstlisting}
{
    "question_id": "196",
    "ocr_text": "Here are the\nprices for our\nschool supplies.\nPRICE LIST\n$2\n$1\n$1,50 I want these. Thank you.\n20 How much\nchange will I\nget after\nbuying these\ncrayons?\n20",
    "math_ground_truth": "12"
}
\end{lstlisting}

\vspace{3mm}

\textbf{Input (OCR Text):}

\begin{lstlisting}
Here are the
prices for our
school supplies.
PRICE LIST
$2
$1
$1,50 I want these. Thank you.
20 How much
change will I
get after
buying these crayons?
20
\end{lstlisting}

\vspace{3mm}

\textbf{Prompt.}
\begin{lstlisting}
You are an expert at solving mathematical problems extracted from {ocr_text}.
The input is a raw OCR transcription of images. Carefully read the text, resolve formatting artifacts (e.g., misplaced line breaks, currency symbols, punctuation), and then solve the math problem it describes.
# Answer Prompt:
{Answer Format Prompt}
\end{lstlisting}
\end{promptbox}
\end{figure*}

\begin{figure*}[t]
\centering

\begin{promptbox}[{OCR + Image Mode}]

\begin{subsecbox}
\textbf{Input (OCR Text + Image):}

\begin{lstlisting}

Here are the
prices for our
school supplies.
PRICE LIST
$2
$1
$1,50 I want these. Thank you.
20 How much
change will I
get after
buying these
crayons?
20

\end{lstlisting}

\vspace{2mm}

\begin{minipage}{0.24\linewidth}
  \centering
  \includegraphics[width=\linewidth]{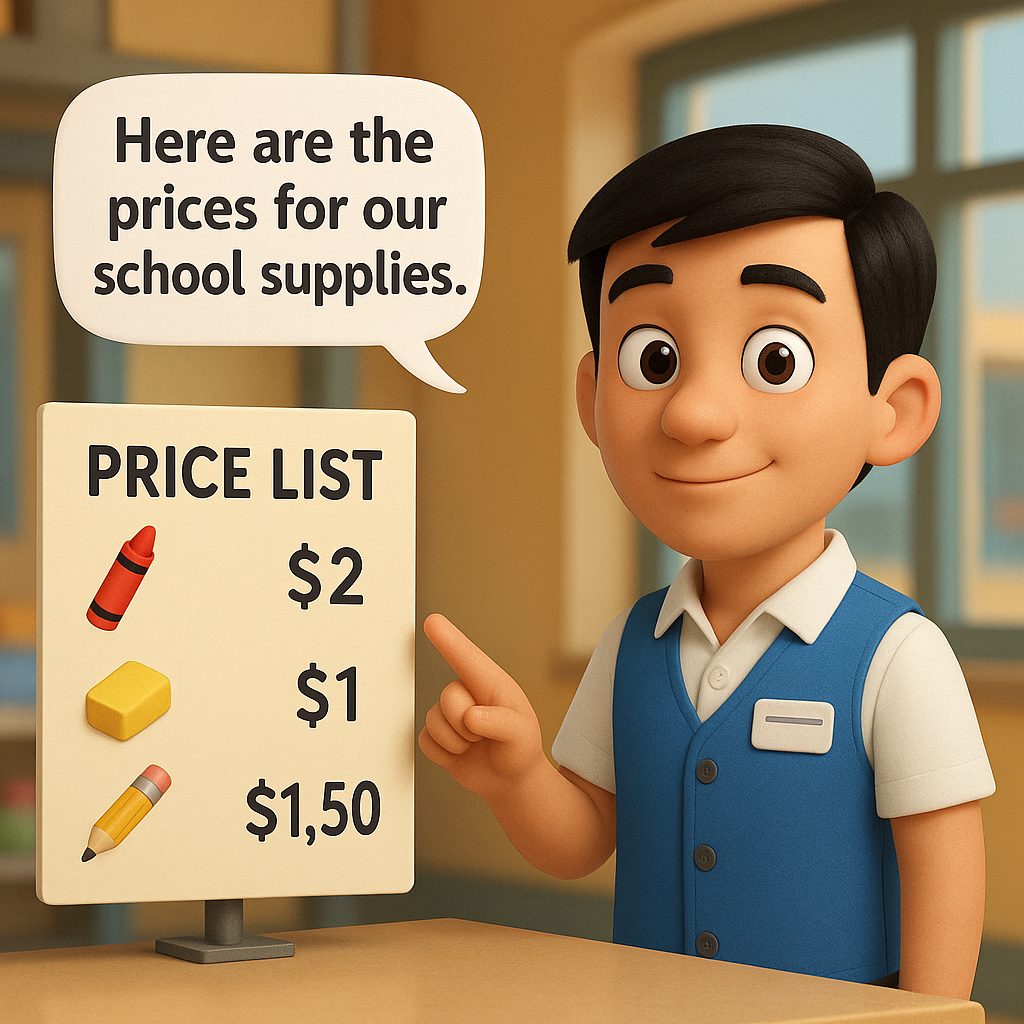}\\
  {\small Scene 1}
\end{minipage}\hfill
\begin{minipage}{0.24\linewidth}
  \centering
  \includegraphics[width=\linewidth]{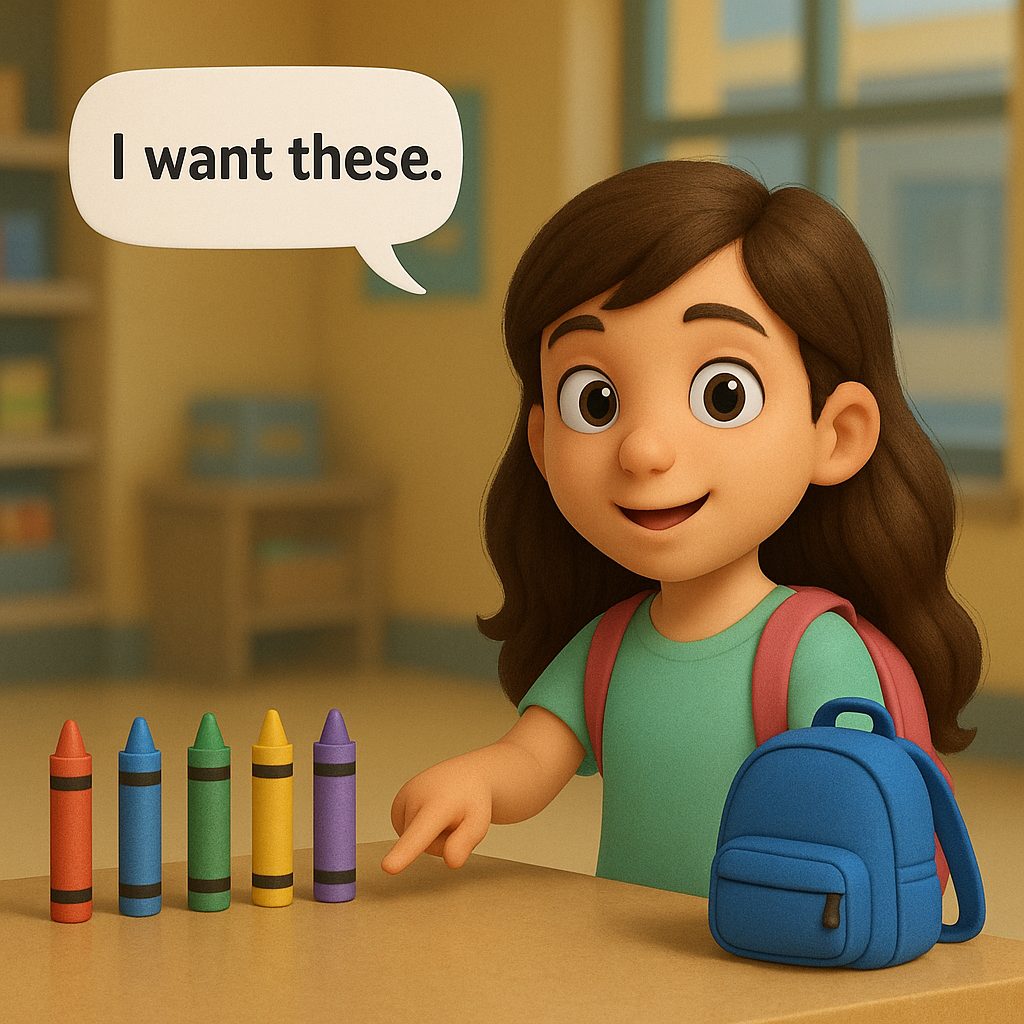}\\
  {\small Scene 2}
\end{minipage}\hfill
\begin{minipage}{0.24\linewidth}
  \centering
  \includegraphics[width=\linewidth]{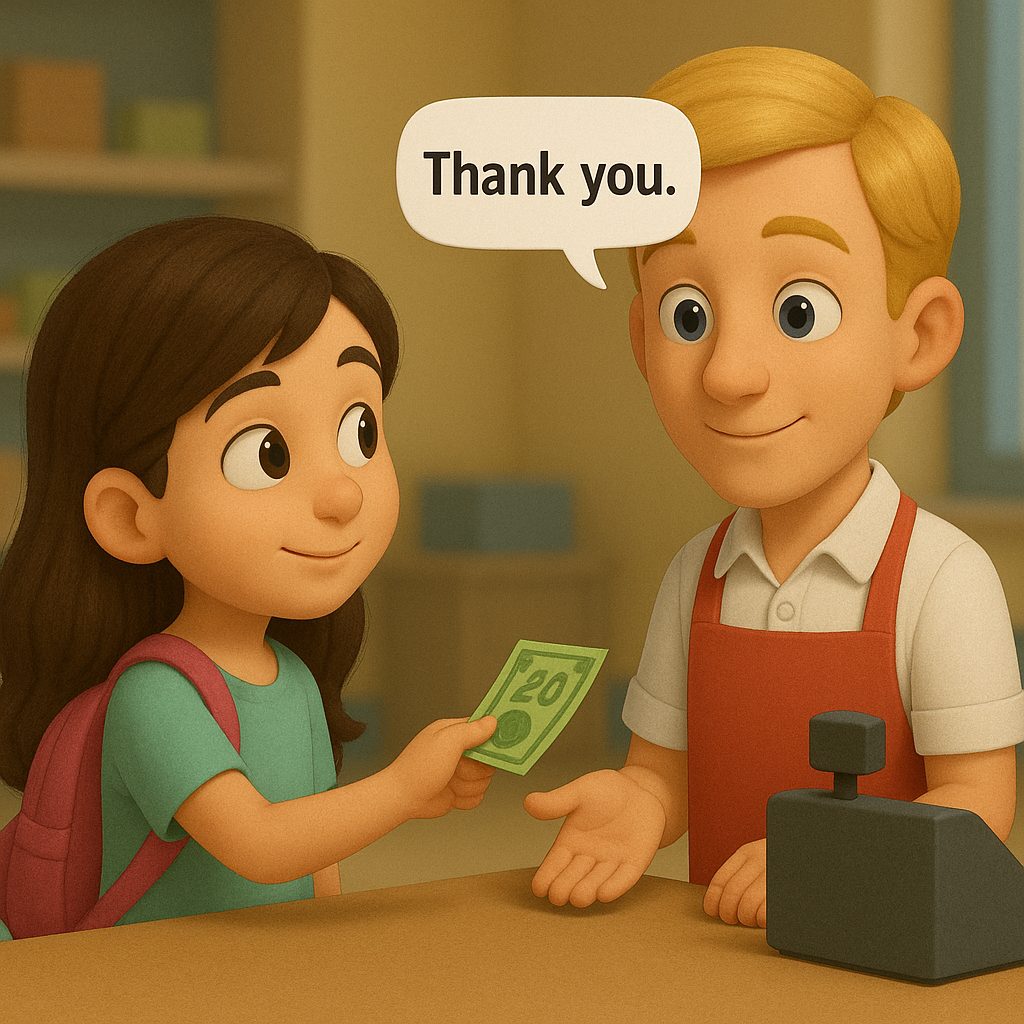}\\
  {\small Scene 3}
\end{minipage}\hfill
\begin{minipage}{0.24\linewidth}
  \centering
  \includegraphics[width=\linewidth]{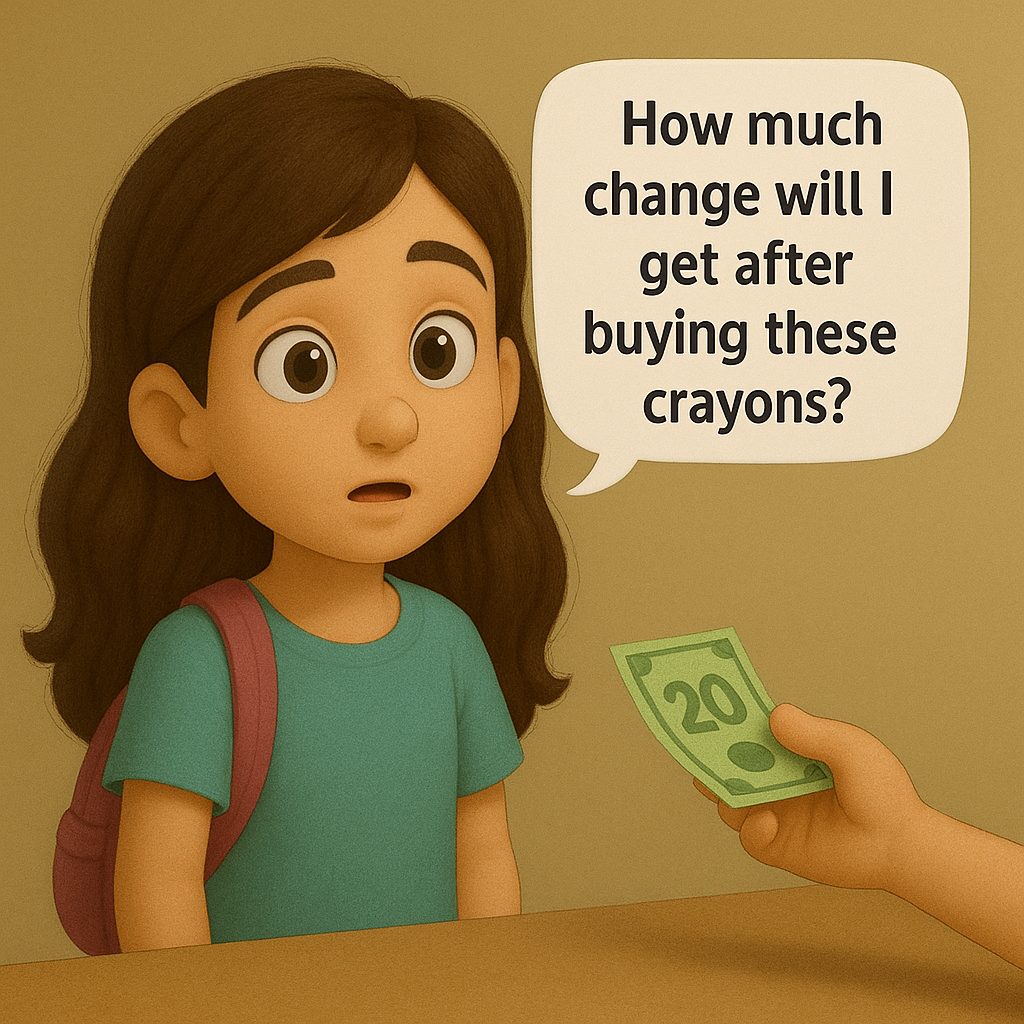}\\
  {\small Scene 4}
\end{minipage}

\end{subsecbox}

\vspace{3mm}

\textbf{Prompt.}
\begin{lstlisting}
You are an expert at solving mathematical word problems using both visual and textual information.
You will be given one or more images containing a math problem, as well as the OCR result extracted from those images.
Use the OCR content as a supplementary source and combine it with the visual information from the images to reason step by step and solve the problem.
OCR Content (from the images): {question}
# Answer Prompt:
{format_instruction}
\end{lstlisting}
\end{promptbox}

\end{figure*}

\clearpage

\begin{strip}

\centering

\begin{promptbox}[{Caption Mode}]
\textbf{Input (Scene Description):}

\begin{lstlisting}

Scene 1: 
(*@\includegraphics[height=8pt]{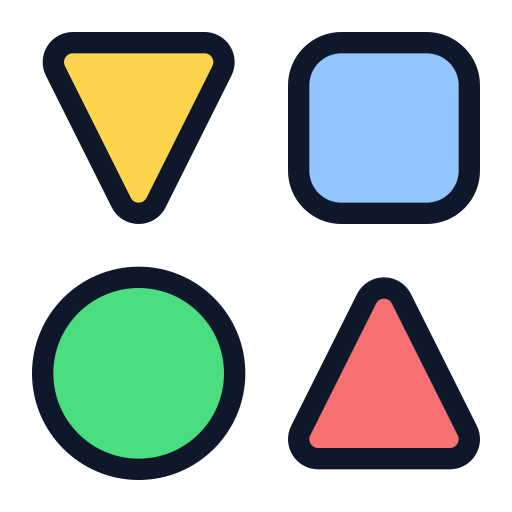}@*) Objects: 
- A visually clear and attractive price list with textual label 'price list'
- The price list line1 states: (*@\includegraphics[height=7pt]{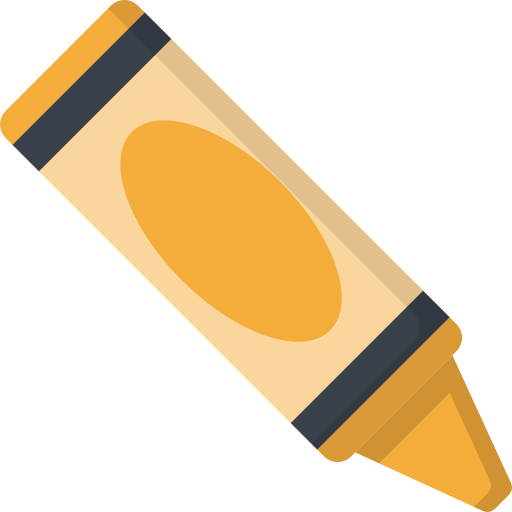}@*) $2
- The price list line2 states: (*@\includegraphics[height=7pt]{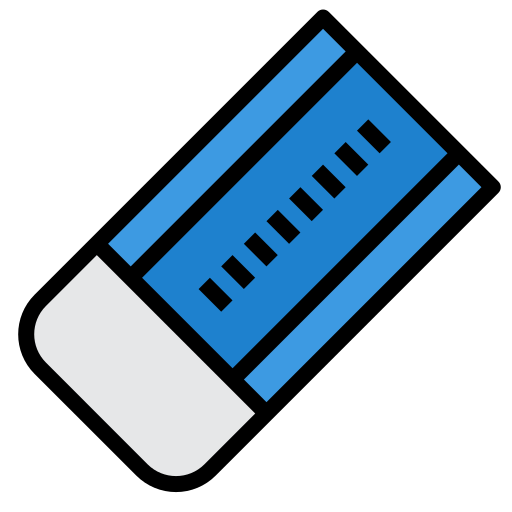}@*) $1
- The price list line3 states: (*@\includegraphics[height=7pt]{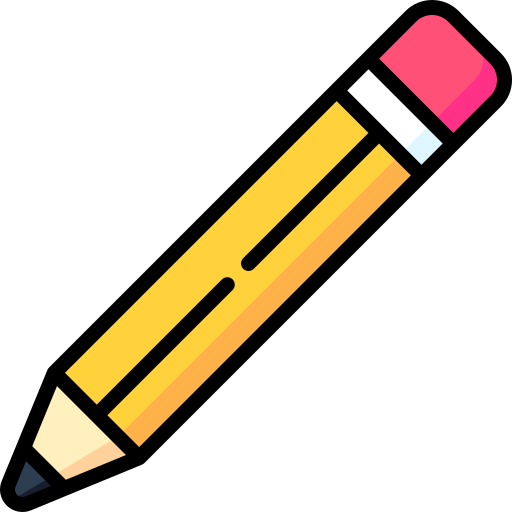}@*)  $1.50
- Shop assistant (short black hair, blue vest, white shirt, name tag).A white speech bubble extending from the shop assistant's mouth, containing the text: 'Here are the prices for our school supplies.'
(*@\includegraphics[height=8pt]{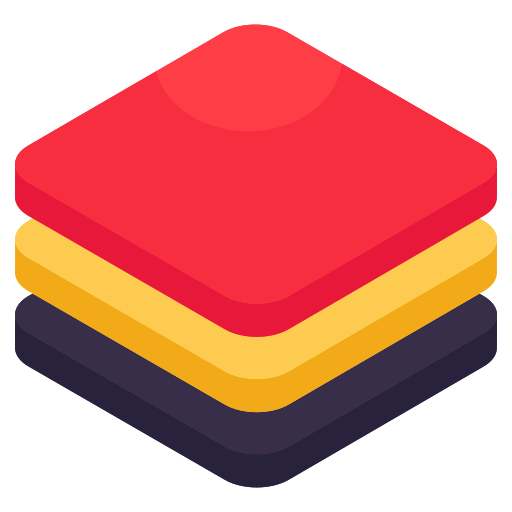}@*)Composition: 
- The price list is mounted on a stand or board near the store entrance, positioned at eye level and angled slightly toward the viewer for full visibility.

- Each line of the price list is clearly separated, with the crayon, eraser, and pencil icons on the left and their respective prices to the right, all on separate lines.
(*@\includegraphics[height=8pt]{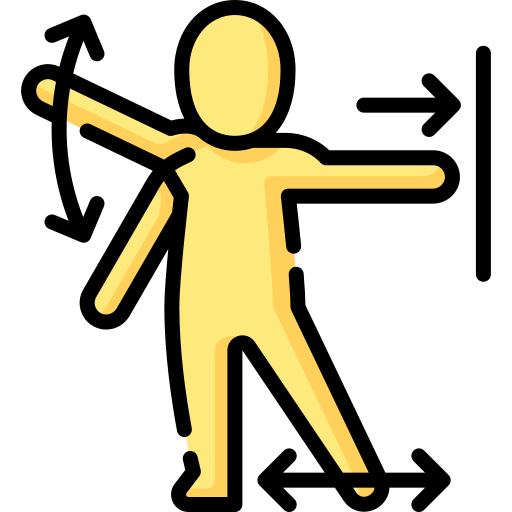}@*)Action: 
The shop assistant extends their right hand, pointing directly at the price list to visually anchor their spoken statement to the displayed prices.
Scene 2: ...
FINAL SCENE 4(Problem Demonstration) ...
A white speech bubble extending from Violetta's mouth, containing the text: 'How much change will I get after buying these crayons?'
...

\end{lstlisting}

\textbf{Prompt.}
\begin{lstlisting}

You are an expert at analyzing visual scenes and solving mathematical problems based on interconnected information across multiple scenes. Analyze the following sequence of scenes carefully. 
Each scene provides information that connects to the others, building toward the final question. You need to understand the relationships between scenes and track quantities, actions, and information flow. 
Now analyze these scenes:{scene_description}
# Answer Prompt:
{Answer Format Prompt}
\end{lstlisting}
\end{promptbox}

\end{strip}

\subsection{Single-Image(r*) vs. Multi-Image(r*)}
\label{appendix:single_multi}
To illustrate the cross-scene integration analysis in Section~\ref{subsec:image_input_format_ablation}, we provide representative examples from the \emph{Single-Image}, \emph{Single-Image(r*)}, \emph{Multi-Image}, and \emph{Multi-Image(r*)} settings. Each case shows the textual problem, the corresponding visual input(s), and model responses. In the single-image setting, all information is concatenated into one image, which can obscure temporal or sequential cues, especially under randomized order ($r^*$), leading to failures in retrieving critical details. In contrast, the multi-image setting separates the problem across multiple ordered scenes, better preserving temporal and semantic structure. The attached figures demonstrate how models like GPT-4o interpret these inputs differently, and highlight common failure modes such as misreading scene-specific numerical cues or misaligning temporal sequences. This example complements Table~\ref{tab:single_multi} by providing a concrete visualization of why cross-scene integration remains a major challenge in \datasetname .

\subsection{Image Style Effects}
This section provides qualitative examples for the image style ablation in Section~\ref{subsec:image_style_ablation}. 
We present the same GSM8K-V problem rendered in three visual styles: \emph{Pixar-Style}, \emph{Ghibli-Style}, and \emph{Real-world Style}. 
Across these variants, the underlying mathematical semantics, scene structure, and ground-truth answer remain unchanged, while only the rendering style is varied. 
The examples illustrate that model failures are not primarily caused by a specific visual style, but instead arise from the same core challenges identified in \datasetname, such as extracting relevant visual evidence and reconstructing the correct mathematical relation from the scene sequence. 
This qualitative comparison complements Table~\ref{tab:image_styles} and further supports the cross-style generalizability of \datasetname.

\label{appendix:image_styles}

\clearpage

\begin{figure*}[htbp]
\centering

\begin{casestudy}[Single-Image(r*) vs. Multi-Image(r*) (Temporal Reasoning)]
\begin{subsecbox}
\textbf{Text Problem.}
\vspace{0.5mm}

\emph{``Lloyd earns \$10 an hour on Math tutoring. He tutored 5 hours for the
first week and 8 hours for the second week. How much did he earn for the first
two weeks?''}
\end{subsecbox}

\begin{subsecbox}

\textbf{Single-Image (One Image).}
\vspace{0.5mm}

\begin{minipage}{0.45\linewidth}
  \centering
  \includegraphics[width=\linewidth]{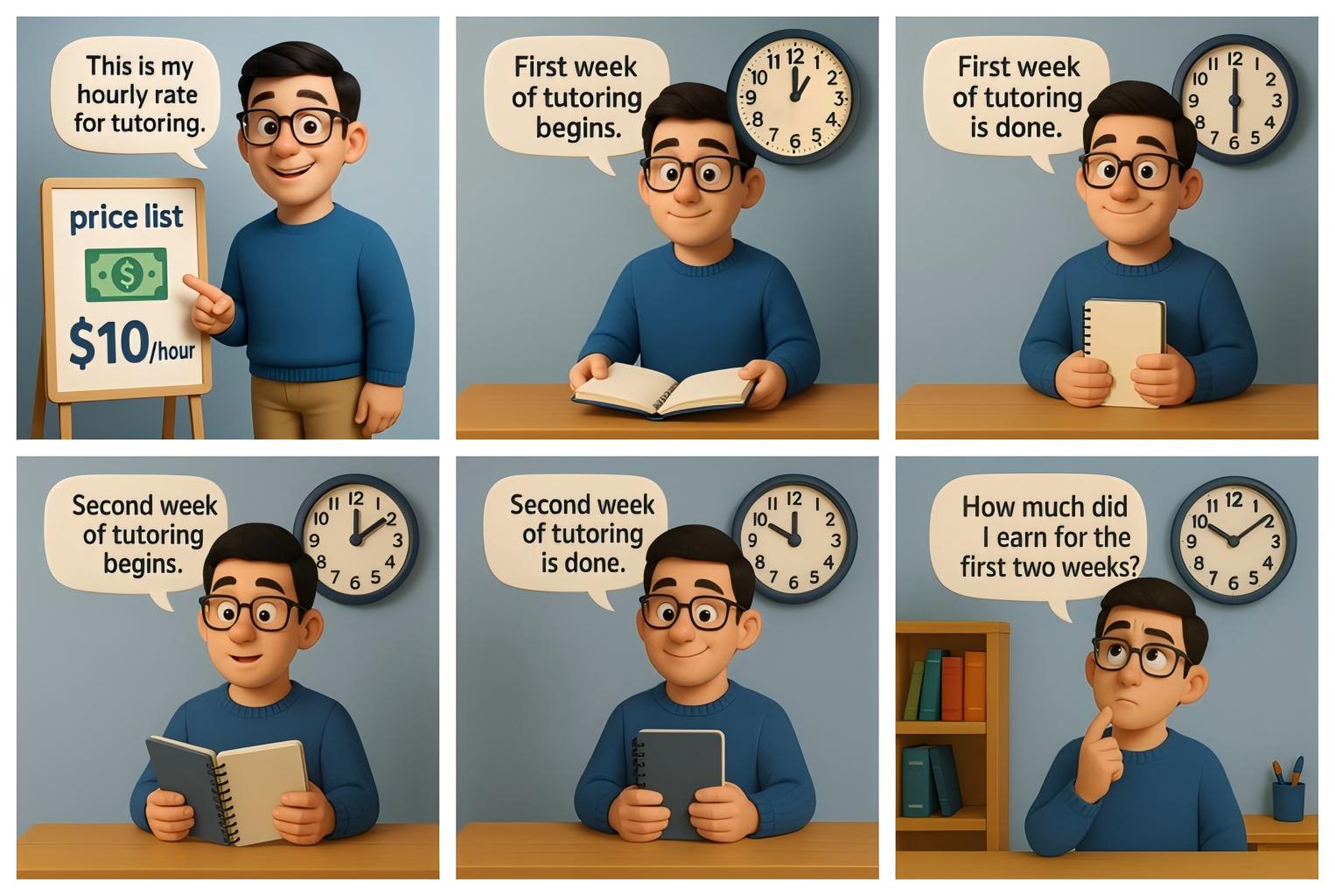}\\
  {\small Single-Image (concatenated orderly)}
\end{minipage}\hfill
\begin{minipage}{0.45\linewidth}
  \centering
  \includegraphics[width=\linewidth]{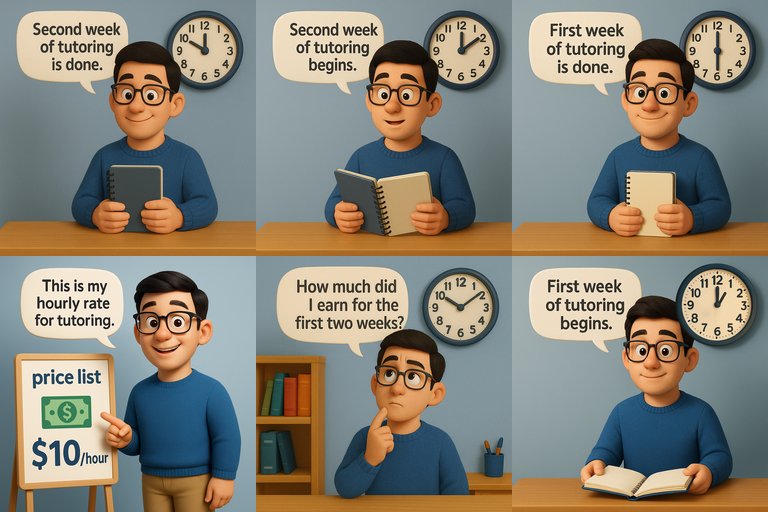}\\
  {\small Single-Image(r*) (concatenated randomly)}
\end{minipage}\hfill

\end{subsecbox}

\vspace{1mm}

\begin{subsecbox}
\textbf{Multi-Image (6 Scenes).}
\vspace{0.5mm}

\begin{minipage}{0.27\linewidth}
  \centering
  \includegraphics[width=\linewidth]{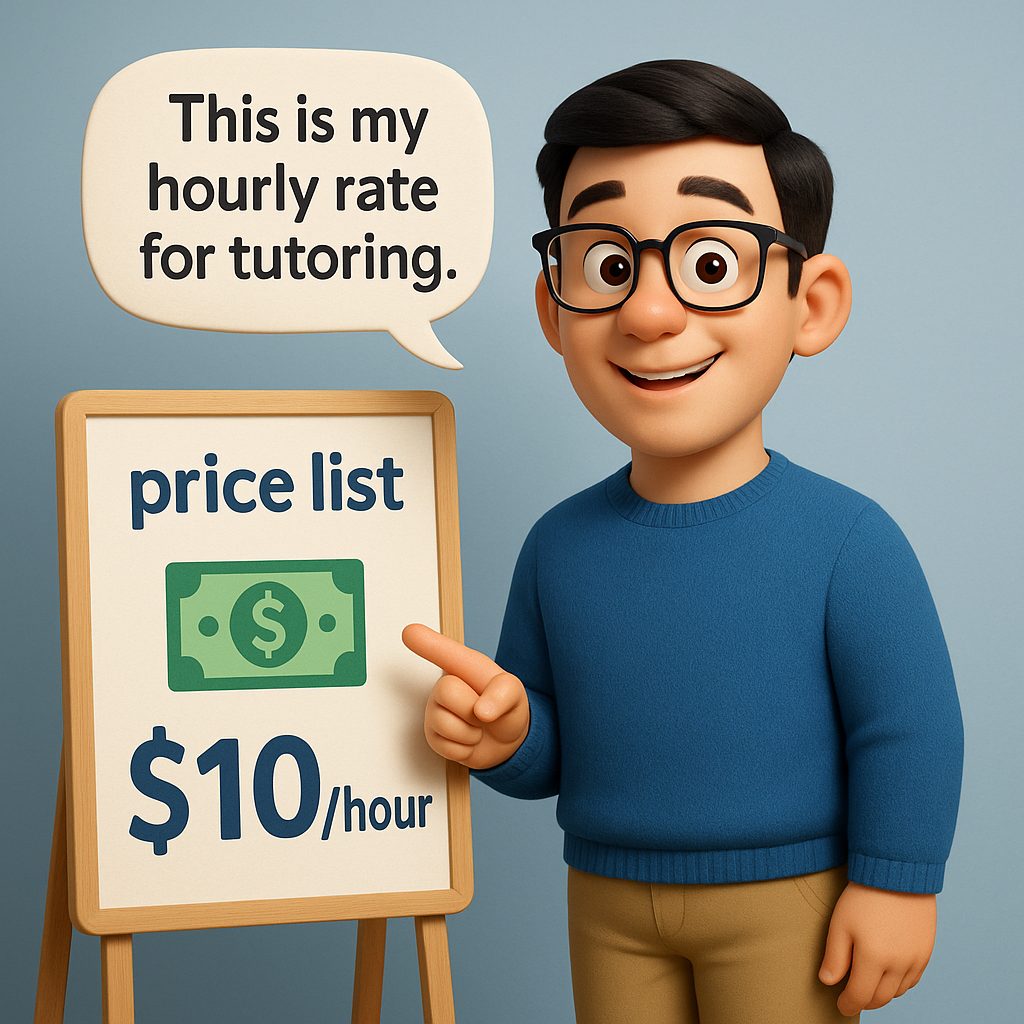}\\
  {\small Scene 1}
\end{minipage}\hfill
\begin{minipage}{0.27\linewidth}
  \centering
  \includegraphics[width=\linewidth]{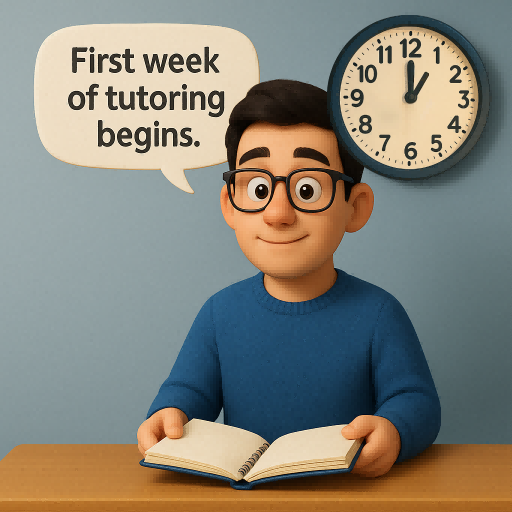}\\
  {\small Scene 2}
\end{minipage}\hfill
\begin{minipage}{0.27\linewidth}
  \centering
  \includegraphics[width=\linewidth]{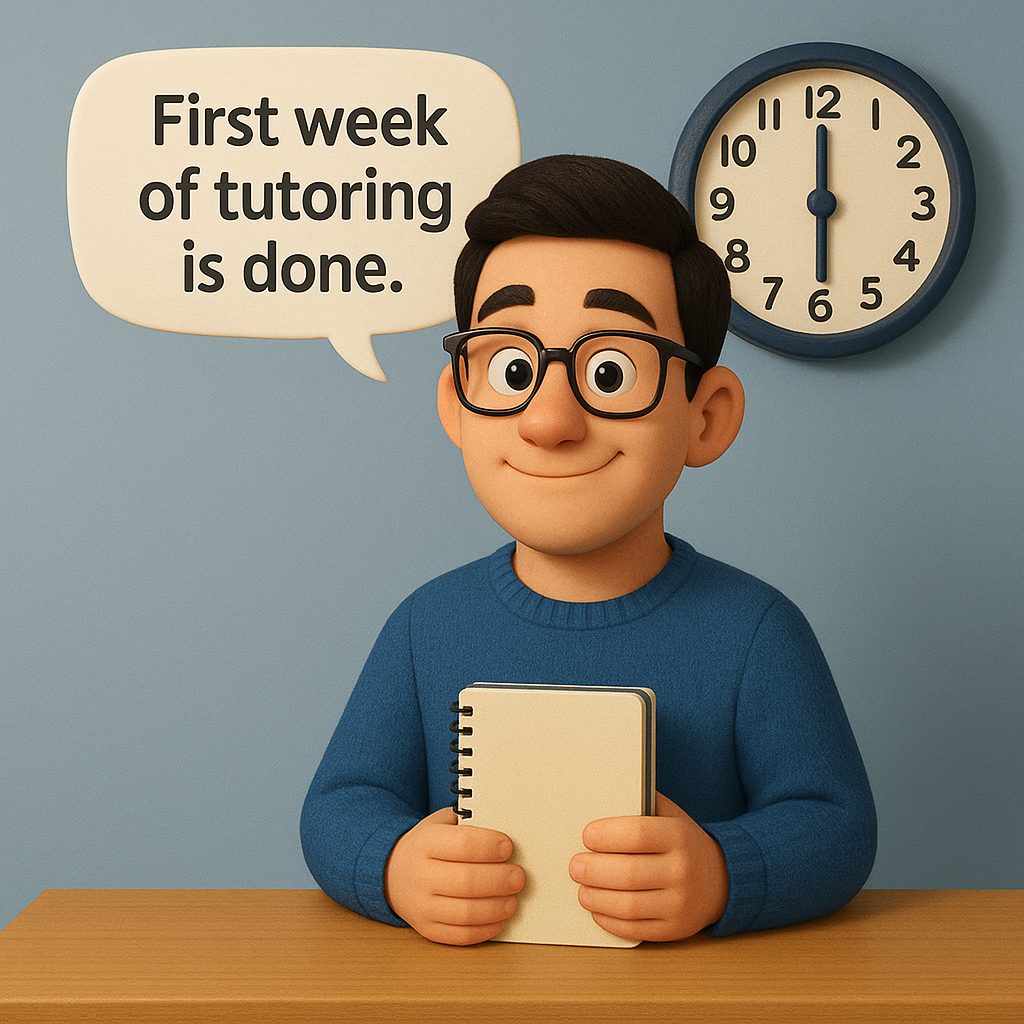}\\
  {\small Scene 3}
\end{minipage}

\vspace{2mm}

\begin{minipage}{0.27\linewidth}
  \centering
  \includegraphics[width=\linewidth]{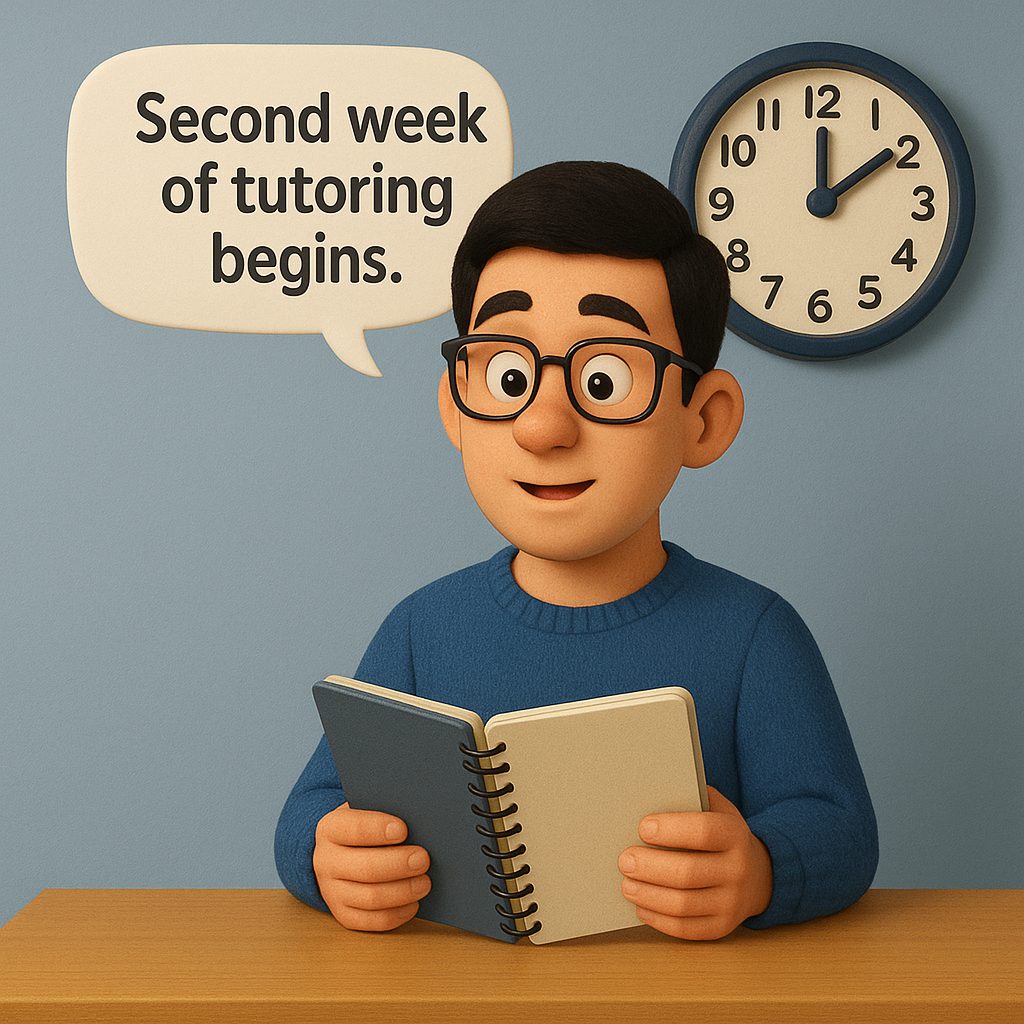}\\
  {\small Scene 4}
\end{minipage}\hfill
\begin{minipage}{0.27\linewidth}
  \centering
  \includegraphics[width=\linewidth]{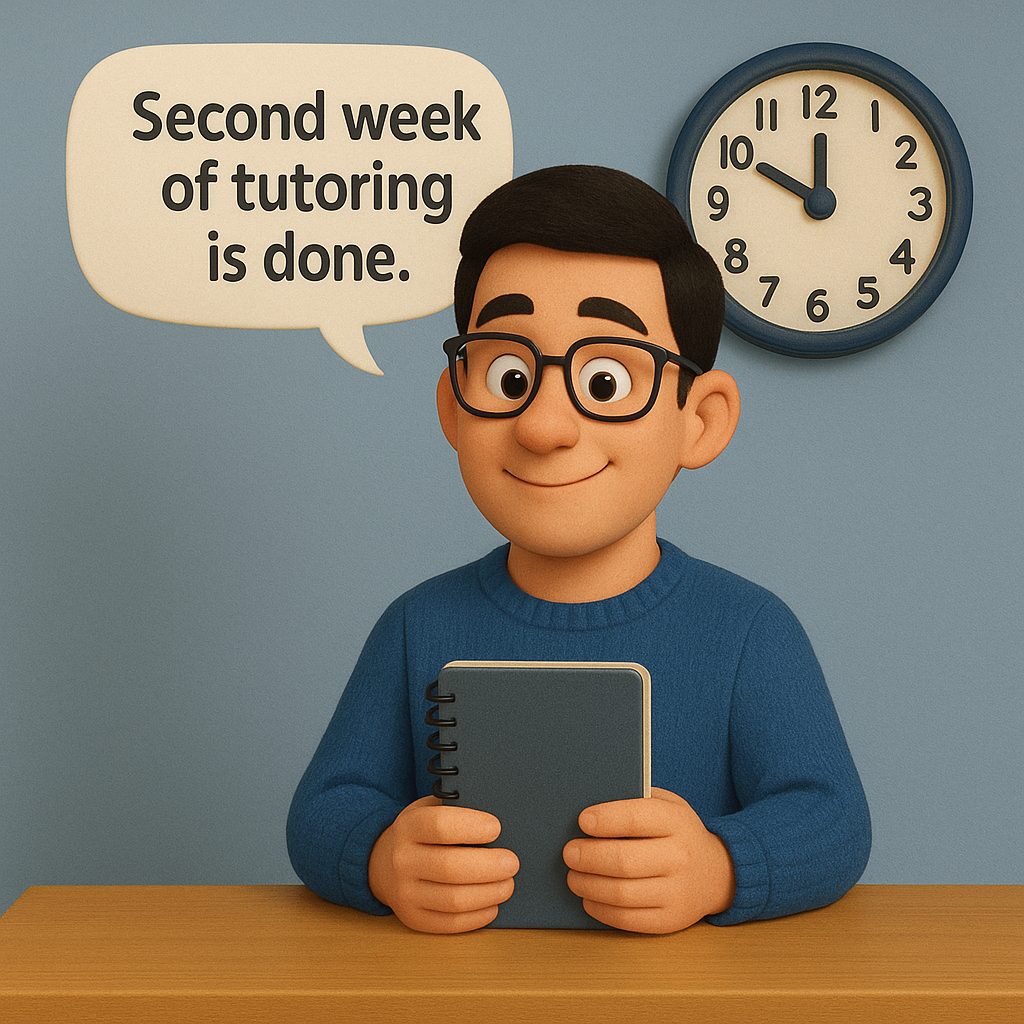}\\
  {\small Scene 5}
\end{minipage}\hfill
\begin{minipage}{0.27\linewidth}
  \centering
  \includegraphics[width=\linewidth]{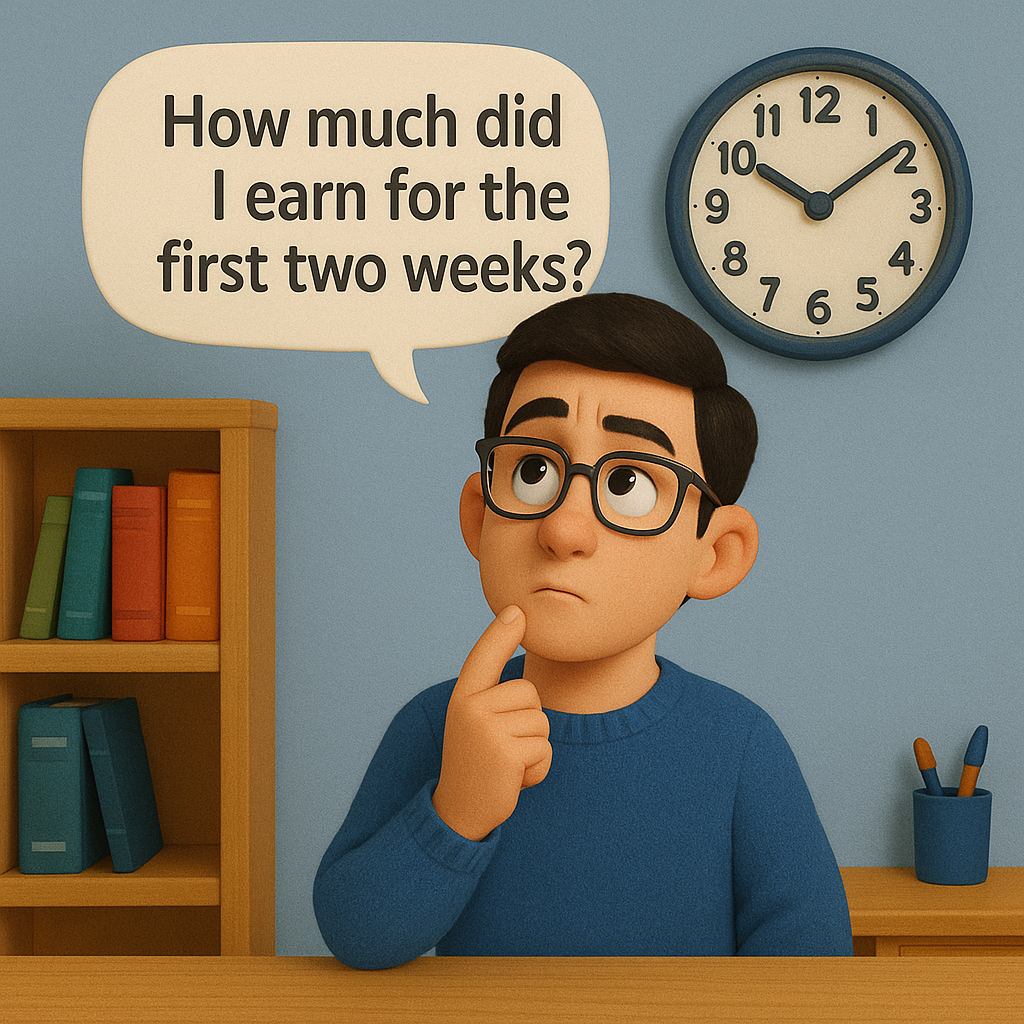}\\
  {\small Scene 6}
\end{minipage}

\end{subsecbox}

\begin{subsecbox}
\textbf{Ground Truth: } \textcolor{green!60!black}{\textbf{130}}
\vspace{0.5mm}

Hourly rate = \$10. Total hours = 5 + 8 = 13. Total earnings = \$10 $\times$ 13 = \textcolor{green!60!black}{130}.
\end{subsecbox}
\end{casestudy}

\end{figure*}

\begin{figure*}[t]
\centering

\begin{casestudy}[Single vs. Multi Images (Temporal Reasoning)]

\vspace{6pt}

\begin{modelbox}[GPT-4o, Single-Image (\wrongtag)]{BadRed}{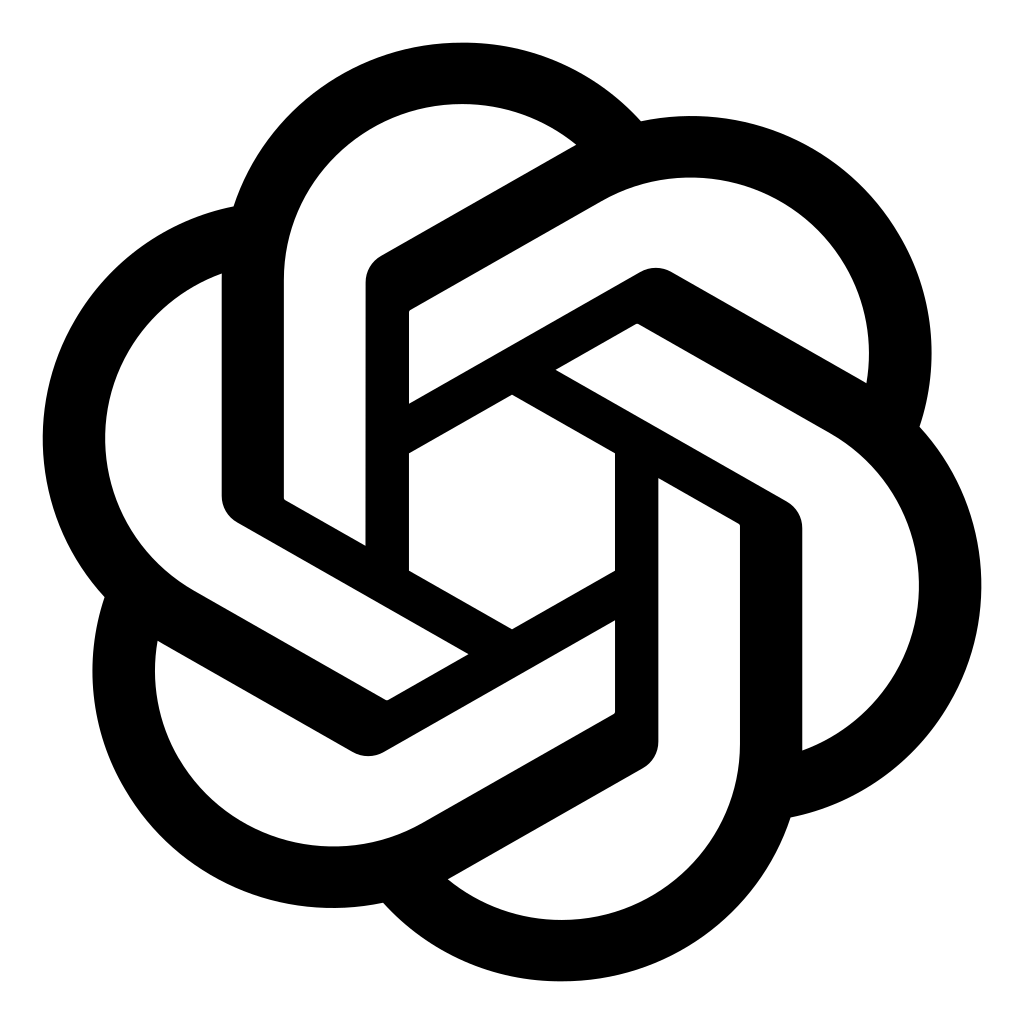}
\textbf{Answer:}

\begin{tcolorbox}[colback=LightGray!20, colframe=black!10, boxrule=0.3pt, arc=2pt, left=6pt, right=6pt, top=4pt, bottom=4pt]

To determine how much Lloyd earned for the first two weeks of tutoring, we need to follow these steps:

1. **Identify the hourly rate**: The price list shows that Lloyd charges \$10 per hour for tutoring.

2. **Determine the number of hours worked each week**:  
   - The first image indicates the start of the first week, and the third image indicates the end of the first week. However, the number of hours worked is not directly shown.  
   - The fourth image indicates the start of the second week, and the fifth image indicates the end of the second week. Again, the number of hours worked is not directly shown.  

3. **Calculate the total earnings**: Since the number of hours worked each week is not specified in the images, we cannot calculate the exact earnings without additional information.  

Given the lack of specific hours worked, we cannot provide a numerical answer based on the images alone.  

\medskip
FINAL ANSWER: \textcolor{red!60!black}{0}

\end{tcolorbox}
\end{modelbox}

\begin{modelbox}[GPT-4o, Single-Image(r*) (\wrongtag)]{BadRed}{figures/logos/openai.png}
\textbf{Answer:}

\begin{tcolorbox}[colback=LightGray!20, colframe=black!10, boxrule=0.3pt, arc=2pt, left=6pt, right=6pt, top=4pt, bottom=4pt]

To determine how much Lloyd earned for the first two weeks of tutoring, let's break down the information provided in the images:

1. **Hourly Rate**: The image shows a price list indicating that Lloyd charges \$10 per hour.

2. **Duration of Tutoring**: - The first week of tutoring is done.  - The second week of tutoring begins and is also completed by the end of the sequence.

3. **Assumption on Hours Worked**: Since the exact number of hours worked each week isn't specified, we need to make a reasonable assumption. Typically, a standard workweek might be around 40 hours. However, since this is tutoring, it could be less. Let's assume he tutored 20 hours per week, which is a common part-time schedule.

4. **Calculation**:  For the first week: 20 * 10 = 200 \$   - For the second week:20 * 10 = 200 \$ - Total earnings = \$200 + \$200 = \$400

Given these steps and assumptions, we can conclude:

FINAL ANSWER: 400

\medskip
FINAL ANSWER: \textcolor{red!60!black}{0}

\end{tcolorbox}
\end{modelbox}

\end{casestudy}

\end{figure*}

\begin{figure*}[t]
\centering
\begin{casestudy}[Single vs. Multi Images (Temporal Reasoning)]

\begin{modelbox}[GPT-4o, Multi-Image (\wrongtag)]{BadRed}{figures/logos/openai.png}
\textbf{Answer:}

\begin{tcolorbox}[colback=LightGray!20, colframe=black!10, boxrule=0.3pt, arc=2pt, left=6pt, right=6pt, top=4pt, bottom=4pt]

To determine the earnings for the first two weeks, we need to calculate the total hours worked and multiply by the hourly rate.

1. **Hourly Rate**: \$10/hour

2. **First Week**:  
   - Start time: 9:00 AM  
   - End time: 3:00 PM  
   - Total hours per day = 3:00 PM - 9:00 AM = 6 hours  

3. **Second Week**:  
   - Start time: 9:00 AM  
   - End time: 3:00 PM  
   - Total hours per day = 3:00 PM - 9:00 AM = 6 hours  
   - **Total Hours for Two Weeks**: 6 hours (first week) + 6 hours (second week) = 12 hours  

4. **Total Earnings**: 12 hours * \$10/hour = \$120  

\medskip
FINAL ANSWER: \textcolor{red!60!black}{120}

\end{tcolorbox}
\end{modelbox}

\begin{modelbox}[GPT-4o, Multi-Image(r*) (\wrongtag)]{BadRed}{figures/logos/openai.png}
\textbf{Answer:}

\begin{tcolorbox}[colback=LightGray!20, colframe=black!10, boxrule=0.3pt, arc=2pt, left=6pt, right=6pt, top=4pt, bottom=4pt]

To determine how much Lloyd earned for the first two weeks of tutoring, let's break down the information provided in the images:

Hourly Rate: From Image 6, we know that Lloyd's hourly rate for tutoring is \$10/hour.

Duration of Tutoring:

From Images 2 and 5, we see that the first week of tutoring has begun and then completed.

From Images 3 and 4, we see that the second week of tutoring has begun and then completed.

Assumption on Hours Worked Per Week: The images do not specify the exact number of hours Lloyd worked each week. However, typically, a standard workweek might be around 40 hours. We'll assume he worked 40 hours per week unless otherwise specified.

Calculation:

For the first week: (40 hours × \$10/hour = \$400)

For the second week: (40 hours × \$10/hour = \$400)

Total Earnings for Two Weeks:

Total earnings = Earnings from the first week + Earnings from the second week

Total earnings = \$400 + \$400 = \$800

Given these steps and assumptions, we can conclude: 

\medskip
FINAL ANSWER: \textcolor{red!60!black}{120}

\end{tcolorbox}
\end{modelbox}

\end{casestudy}
\end{figure*}

\begin{figure*}[t]
\centering

\begin{casestudy}[Image Style Example]
\begin{subsecbox}
\textbf{Text Problem.}
\vspace{0.5mm}

A fog bank rolls in from the ocean to cover a city.  
It takes 10 minutes to cover every 3 miles of the city.  
If the city is 42 miles across from the oceanfront to the opposite inland edge, how many minutes will it take for the fog bank to cover the whole city?

\end{subsecbox}

\begin{subsecbox}
\textbf{GSM8K-V Images (Pixar vs. Giphli vs. Real-World).}
\vspace{0.5mm}

\begin{minipage}{0.3\linewidth}
  \centering
  \includegraphics[width=\linewidth]{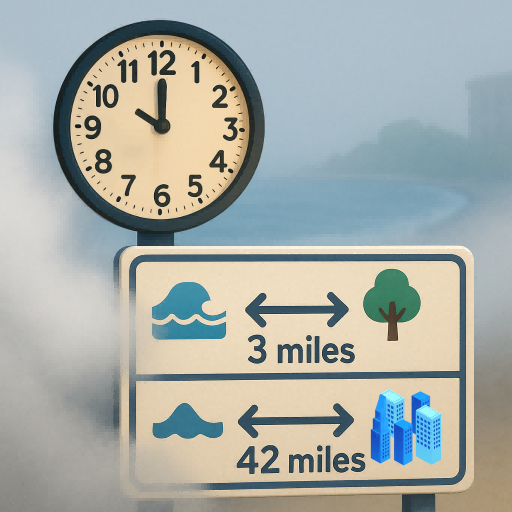}\\
  {\small Pixar Scene 1}
\end{minipage}\hfill
\begin{minipage}{0.3\linewidth}
  \centering
  \includegraphics[width=\linewidth]{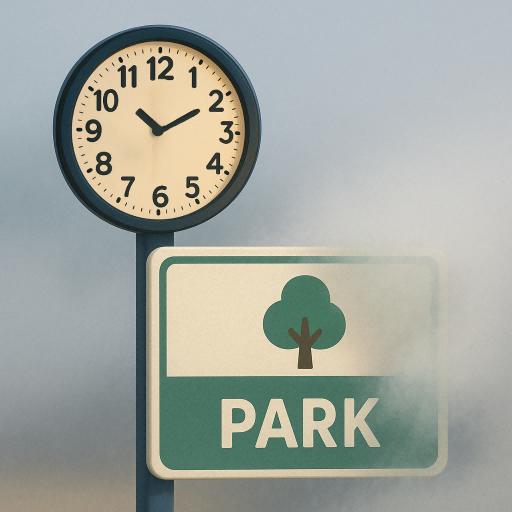}\\
  {\small Pixar Scene 2}
\end{minipage}\hfill
\begin{minipage}{0.3\linewidth}
  \centering
  \includegraphics[width=\linewidth]{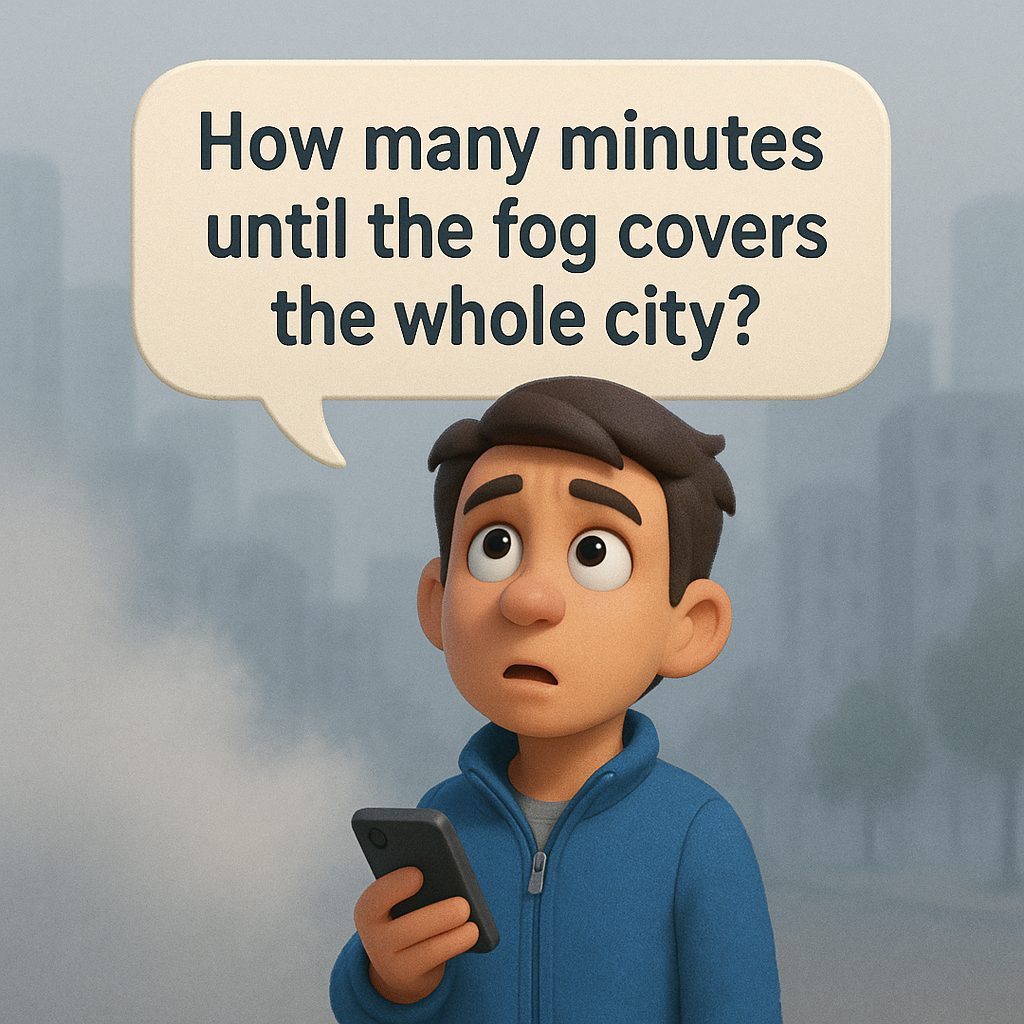}\\
  {\small Pixar Scene 3}
\end{minipage}

\vspace{3mm}

\begin{minipage}{0.3\linewidth}
  \centering
  \includegraphics[width=\linewidth]{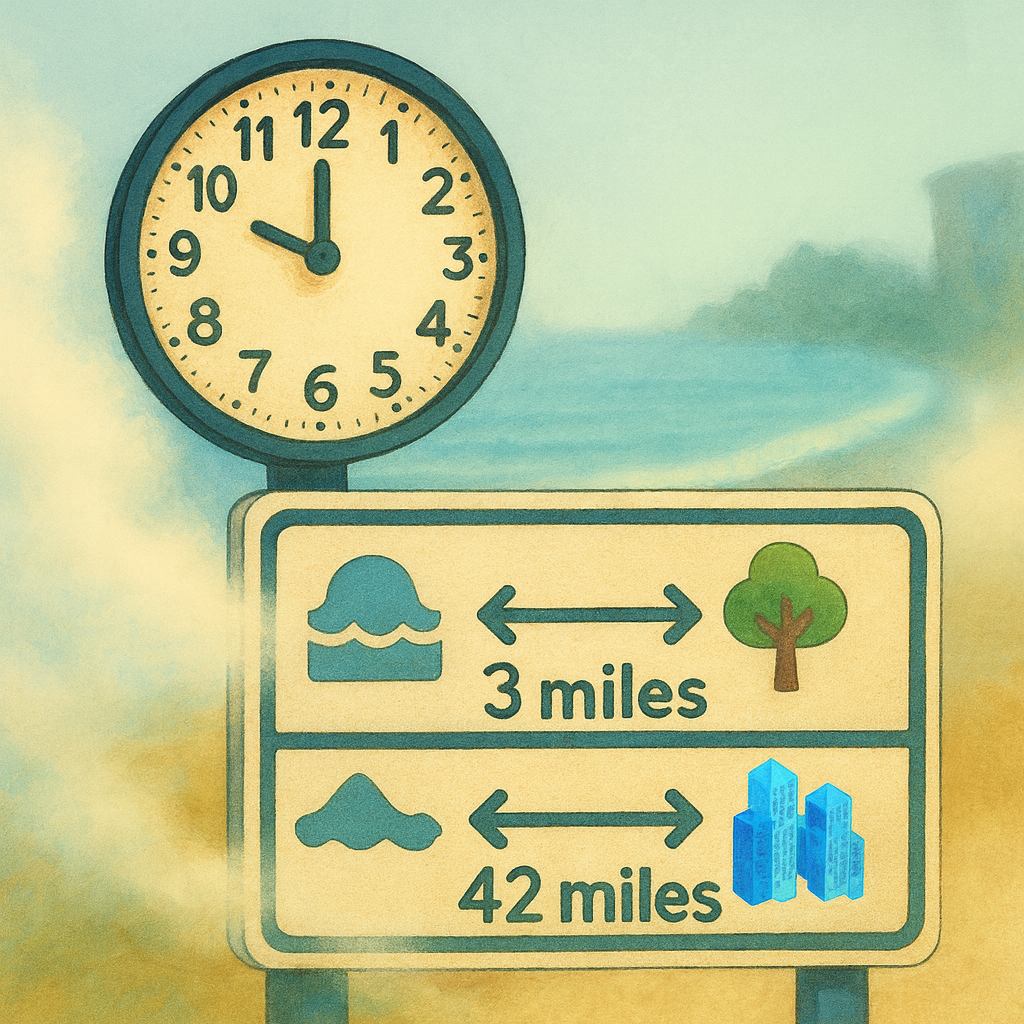}\\
  {\small Giphli Scene 1}
\end{minipage}\hfill
\begin{minipage}{0.3\linewidth}
  \centering
  \includegraphics[width=\linewidth]{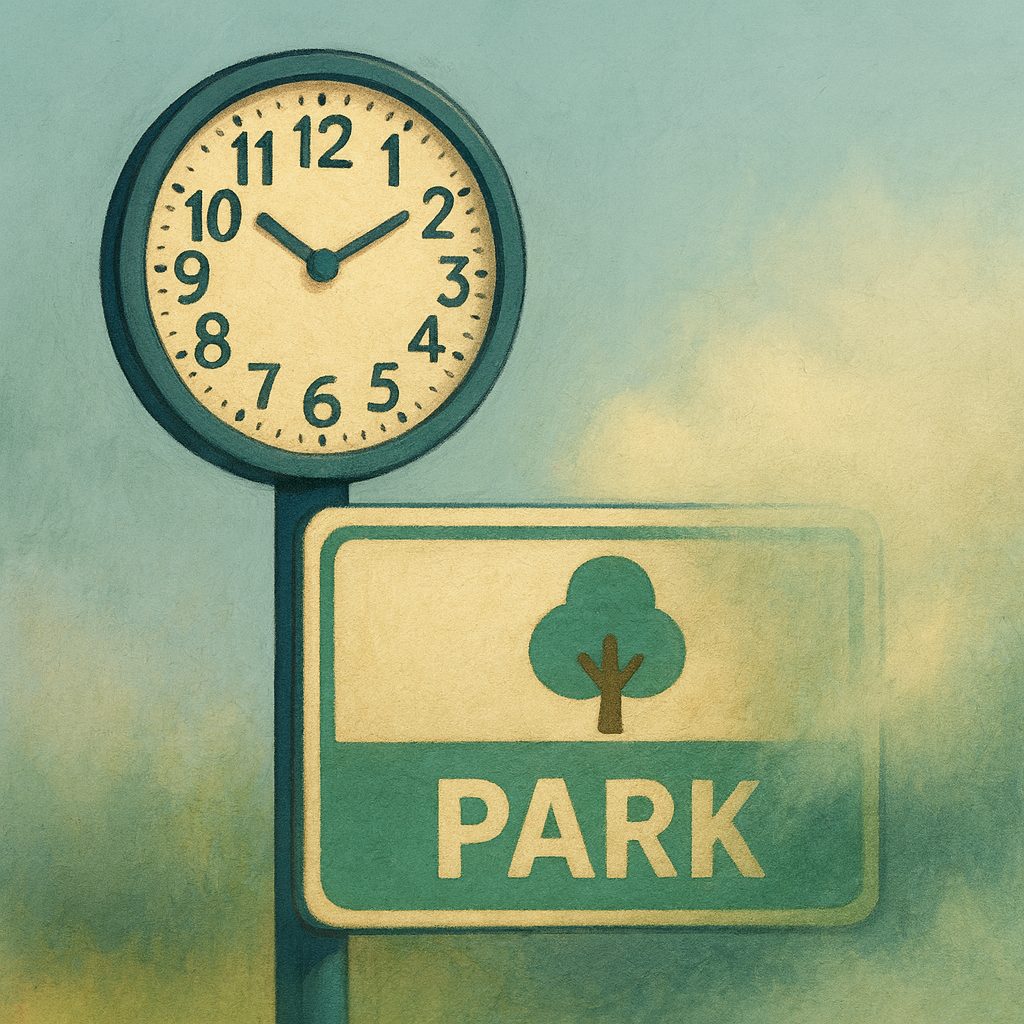}\\
  {\small Giphli Scene 2}
\end{minipage}\hfill
\begin{minipage}{0.3\linewidth}
  \centering
  \includegraphics[width=\linewidth]{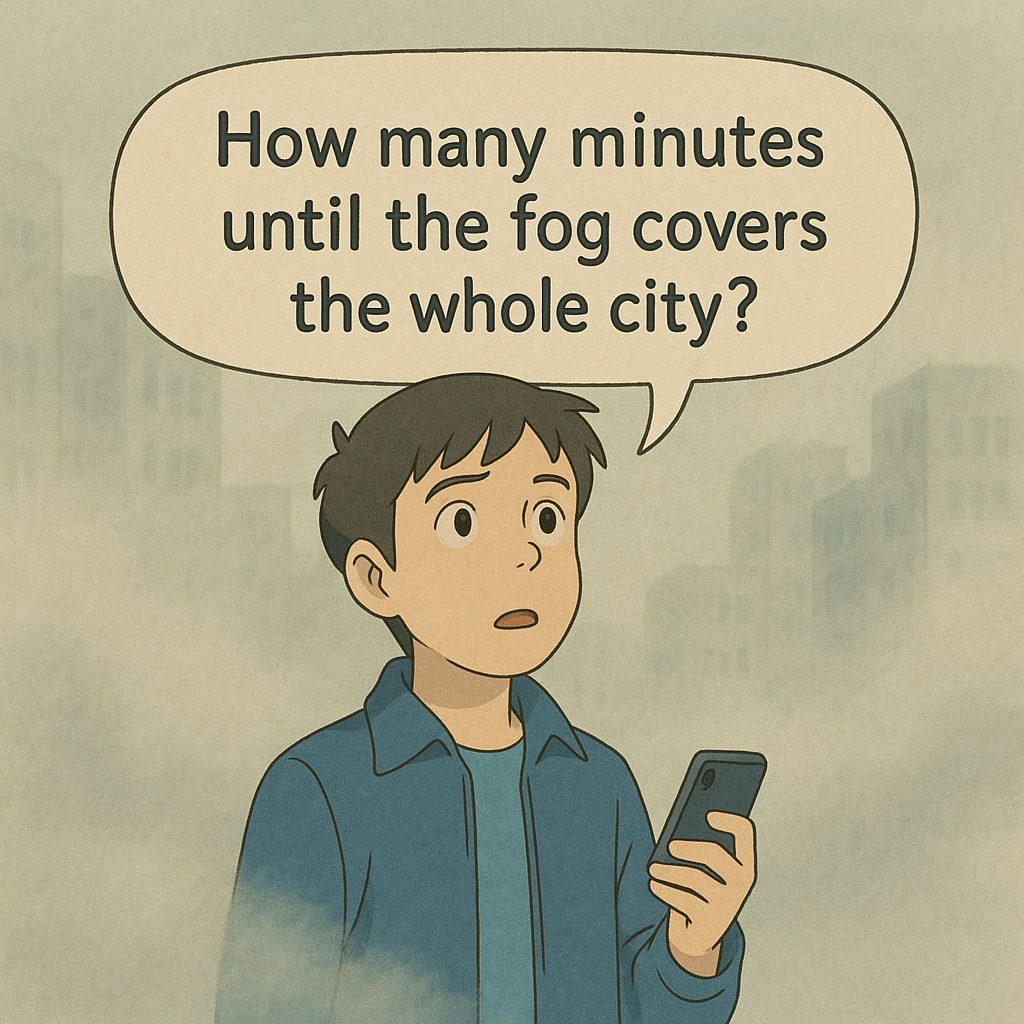}\\
  {\small Giphli Scene 3}
\end{minipage}

\vspace{3mm}

\begin{minipage}{0.3\linewidth}
  \centering
  \includegraphics[width=\linewidth]{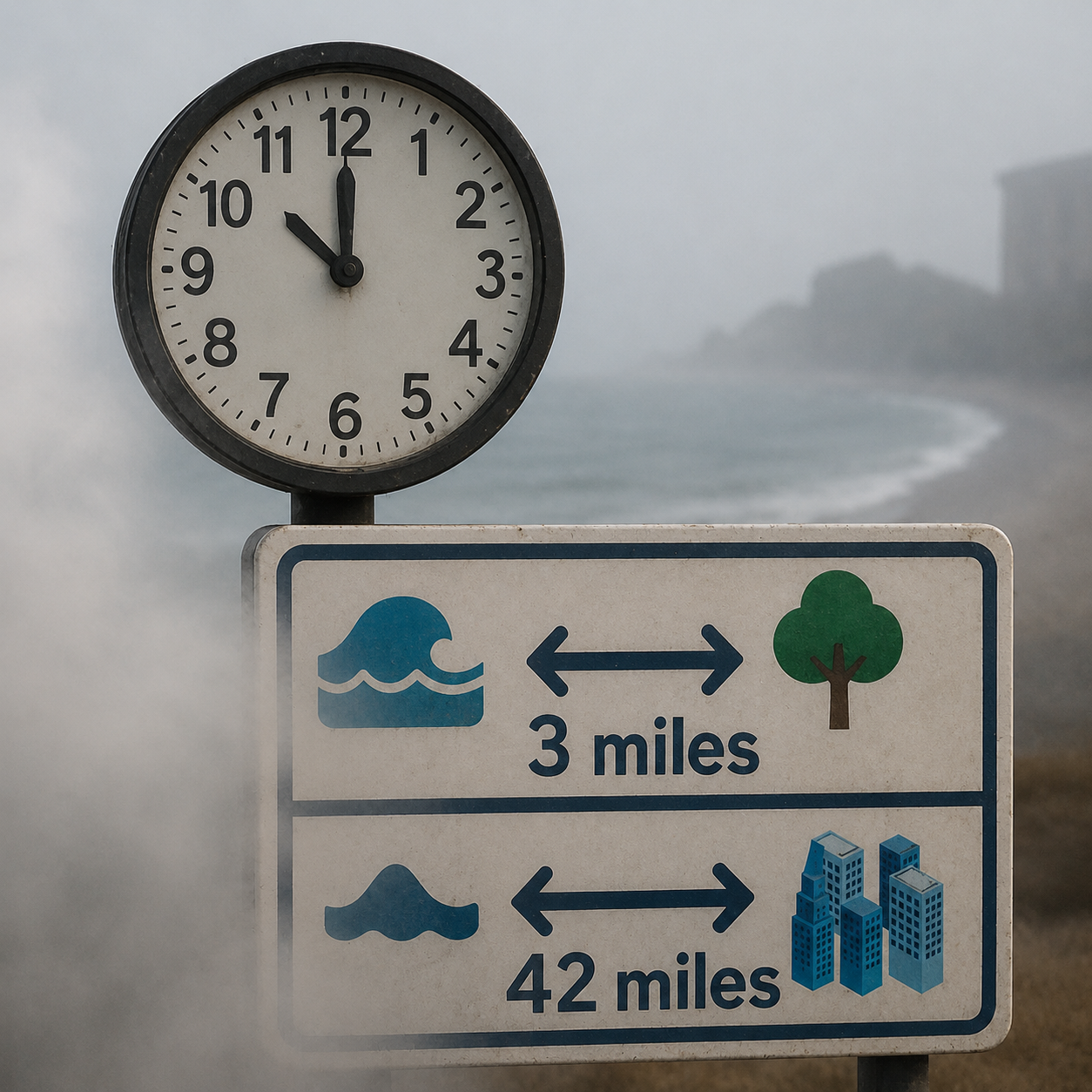}\\
  {\small Real-World Scene 1}
\end{minipage}\hfill
\begin{minipage}{0.3\linewidth}
  \centering
  \includegraphics[width=\linewidth]{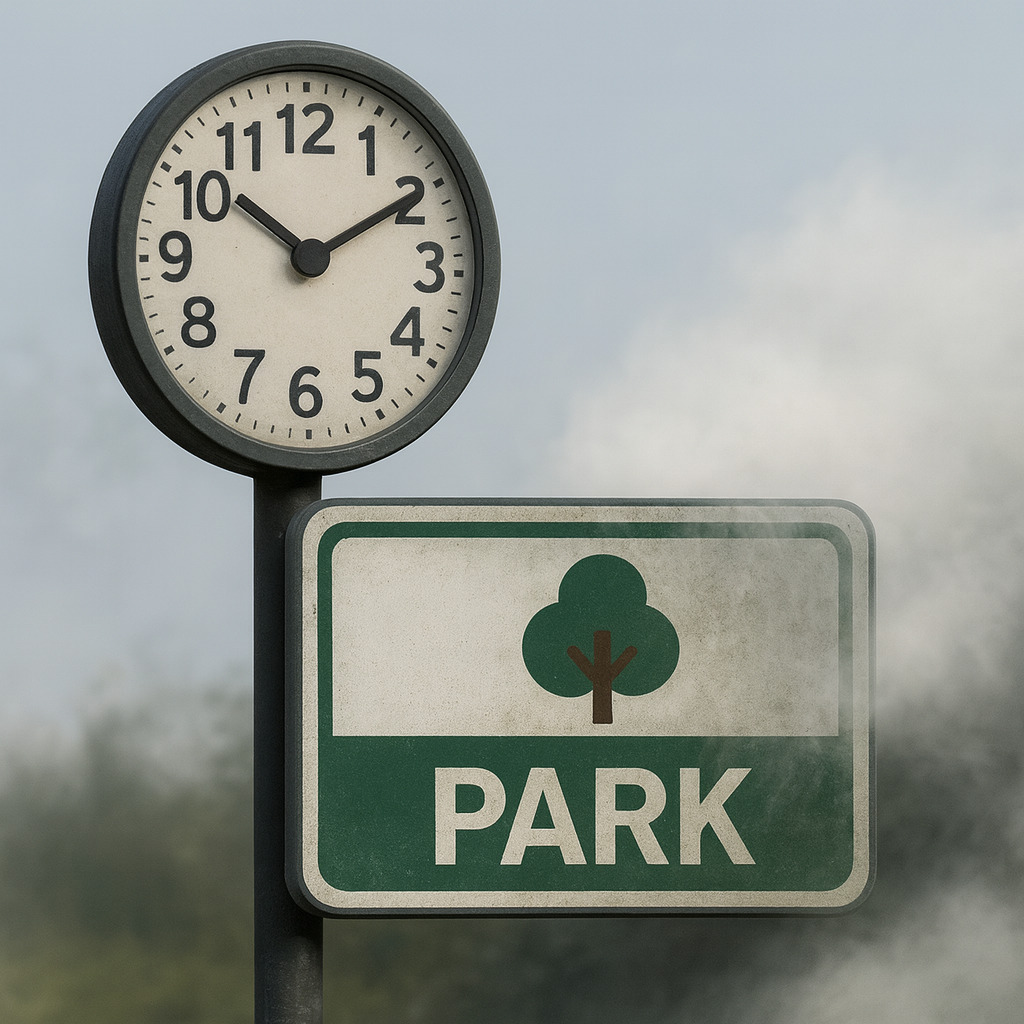}\\
  {\small Real-World Scene 2}
\end{minipage}\hfill
\begin{minipage}{0.3\linewidth}
  \centering
  \includegraphics[width=\linewidth]{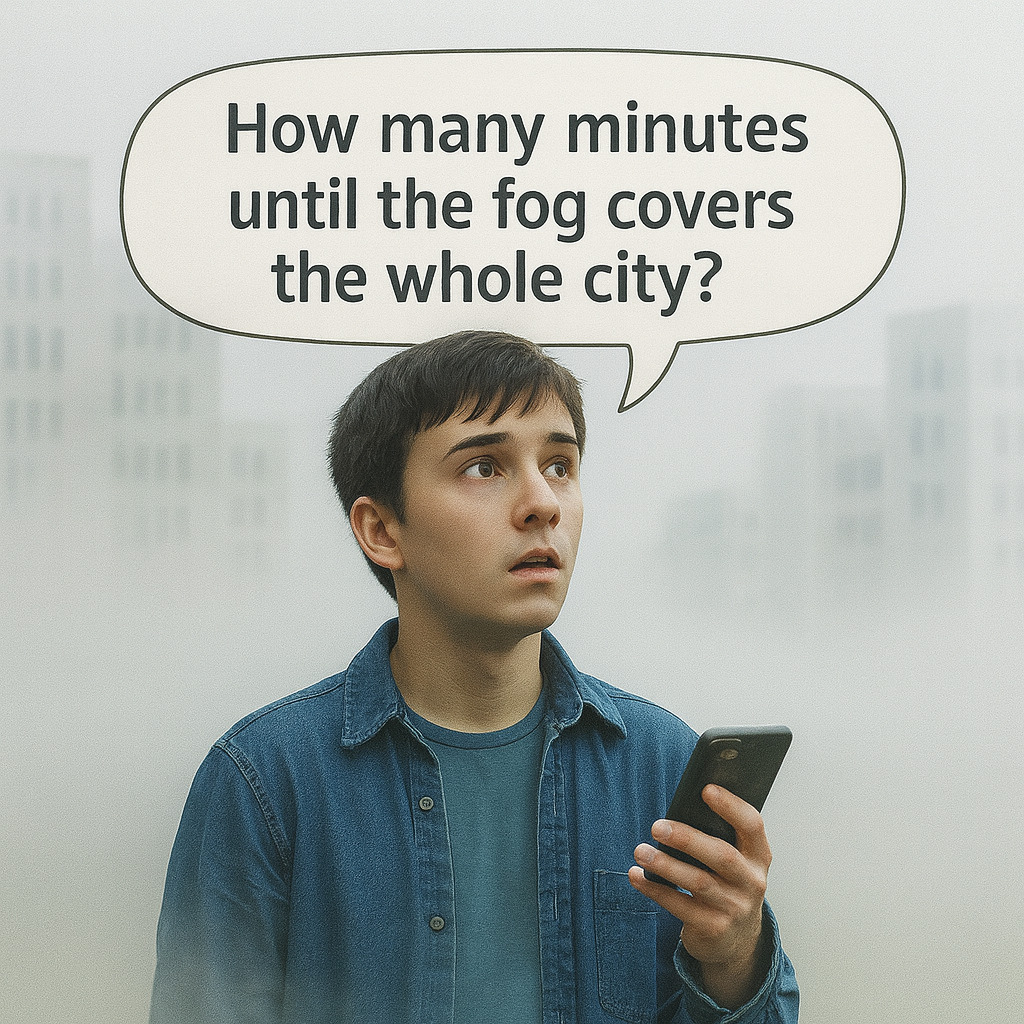}\\
  {\small Real-World Scene 3}
\end{minipage}

\end{subsecbox}

\begin{subsecbox}
\textbf{Ground Truth: } \textcolor{green!60!black}{\textbf{140}}
\vspace{0.5mm}

The fog covers $3$ miles in $10$ minutes.  
The city is $42$ miles wide.  
Thus, total time $= \frac{42}{3} \times 10 = 14 \times 10 = \textcolor{green!60!black}{\mathbf{140}}$ minutes.

\end{subsecbox}

\end{casestudy}

\end{figure*}

\begin{figure*}[t]
\centering

\begin{casestudy}[Image Style Example]

\begin{modelbox}[GPT-4o, Pixar Style (\wrongtag)]{BadRed}{figures/logos/openai.png}
\textbf{Answer:}

\begin{tcolorbox}[colback=LightGray!20, colframe=black!10, boxrule=0.3pt, arc=2pt, left=6pt, right=6pt, top=4pt, bottom=4pt]

To solve the problem, let's analyze the images step by step:

1. **First Image:**  
- The clock shows 10:00.  
- The sign indicates the beach is 3 miles away and the city is 42 miles away.  

2. **Second Image:**  
- The clock shows 11:00.  
- The sign indicates the park.  

3. **Third Image:**  
- The question asks how many minutes until the fog covers the whole city.  

**Analysis:**  

- The time difference between the first and second images is 60 minutes (from 10:00 to 11:00).  
- The fog has already covered the park by 11:00.  
- The distance from the beach to the city is 42 miles, and the park is 3 miles from the beach.  
- Assuming the fog moves at a constant speed, it took 60 minutes to cover 3 miles.  

**Calculation:**  

- Speed of fog = Distance / Time = 3 miles / 60 minutes = 0.05 miles per minute.  
- Remaining distance to cover the city = 42 miles - 3 miles = 39 miles.  
- Time to cover 39 miles = Distance / Speed = 39 miles / 0.05 miles per minute = 780 minutes.  

\medskip
FINAL ANSWER: \textcolor{red!60!black}{780}

\end{tcolorbox}
\end{modelbox}

\vspace{3pt}

\begin{modelbox}[GPT-4o, Giphli Style (\wrongtag)]{BadRed}{figures/logos/openai.png}
\textbf{Answer:}

\begin{tcolorbox}[colback=LightGray!20, colframe=black!10, boxrule=0.3pt, arc=2pt, left=6pt, right=6pt, top=4pt, bottom=4pt]

To solve the problem, we need to determine how long it will take for the fog to cover the entire city.

1. **Understand the Distances:**  
- The first image shows that the distance from the beach to the park is 3 miles.  
- The distance from the beach to the city is 42 miles.  

2. **Analyze the Clocks:**  
- The first clock shows 10:00.  
- The second clock shows 11:00.  

3. **Determine the Time Taken for Fog to Cover 3 Miles:**  
- The fog covers the distance from the beach to the park (3 miles) in 1 hour (from 10:00 to 11:00).  

4. **Calculate the Speed of the Fog:**  
- Speed of the fog = Distance / Time = 3 miles / 1 hour = 3 miles per hour.  

5. **Calculate the Time to Cover 42 Miles:**  
- Time = Distance / Speed = 42 miles / 3 miles per hour = 14 hours.  

6. **Convert Hours to Minutes:**  
- 14 hours = 14 × 60 minutes = 840 minutes.  

Therefore, it will take 840 minutes for the fog to cover the whole city.  

\medskip
FINAL ANSWER: \textcolor{red!60!black}{840}

\end{tcolorbox}
\end{modelbox}

\vspace{3pt}

\end{casestudy}

\end{figure*}

\begin{figure*}
    
\centering

\begin{casestudy}[Image Style Example]

\begin{modelbox}[GPT-4o, Real-World Style (\wrongtag)]{BadRed}{figures/logos/openai.png}
\textbf{Answer:}

\begin{tcolorbox}[colback=LightGray!20, colframe=black!10, boxrule=0.3pt, arc=2pt, left=6pt, right=6pt, top=4pt, bottom=4pt]

To solve the problem of determining how many minutes it will take for the fog bank to cover the whole city, we need to analyze the images provided:

1. First Image:
- The clock shows 10:00.
- The sign indicates the beach is 3 miles away and the city is 42 miles away.

2. Second Image:
- The clock shows 11:00.
- The sign indicates the park.

3. Third Image:
- The question asks how many minutes until the fog covers the whole city.

Analysis:

- The time difference between the first and second images is from 10:00 to 11:00.
  This is 3 hours and 50 minutes, which equals 230 minutes.
- The fog has already reached the park by 11:00.
- The distance from the beach to the city is 42 miles, and the park is 3 miles from the beach.
- Assuming the fog moves at a constant speed, it took 60 minutes to travel 3 miles.

Calculation:

- Speed of the fog = 3 miles / 60 minutes.
- Remaining distance to reach the city = 42 - 3 = 39 miles.
- Time required = 39 / (3 / 60) = 39 * (60 / 3) = 780 minutes.

\medskip
FINAL ANSWER: \textcolor{red!60!black}{780}

\end{tcolorbox}
\end{modelbox}

\end{casestudy}

\end{figure*}

\clearpage

\section{GSM8K-V Construction Details}
\subsection{Math Information Classification}
\label{appendix:math_information_classification}
This step involves the classification of mathematical information extracted from GSM8K into 13 predefined categories, ensuring the mathematical content is properly represented in the visual format. These categories include types like weight, distance, price, and count, each of which is associated with specific rendering strategies to preserve mathematical fidelity.
The 13 categories are as follows.

\begin{figure*}[t]
\centering
\begin{promptbox}[{\includegraphics[height=10pt]{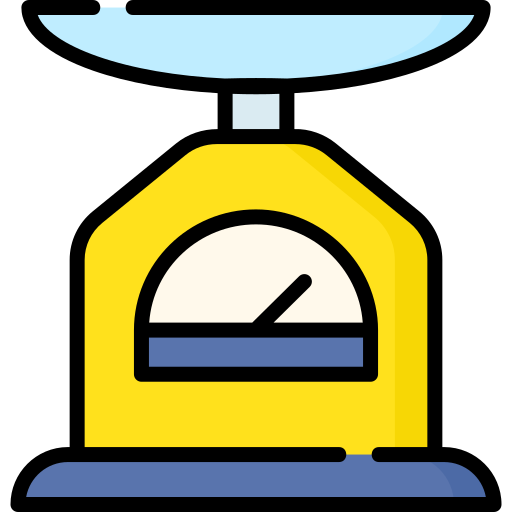} \ Weight}]
Weight refers to the measurement of how heavy an object is, usually expressed in units such as grams, kilograms, or pounds.  
\emph{Examples:} “An apple weighs 200 g”, “The suitcase shows 15 kg”.
\end{promptbox}

\begin{promptbox}[{\includegraphics[height=10pt]{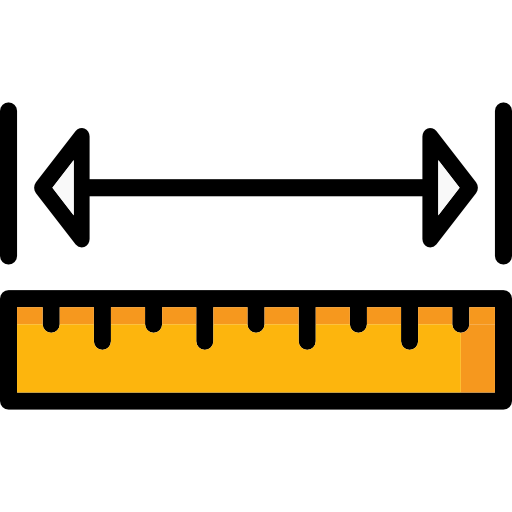} \ Length/Area/Volume}]
Length, area, and volume capture spatial dimensions of objects or regions.  
\emph{Examples:} “The pencil is 12 cm long”, “The field covers 500 m\textsuperscript{2}”, “The box has a volume of 2 liters”.
\end{promptbox}

\begin{promptbox}[{\includegraphics[height=10pt]{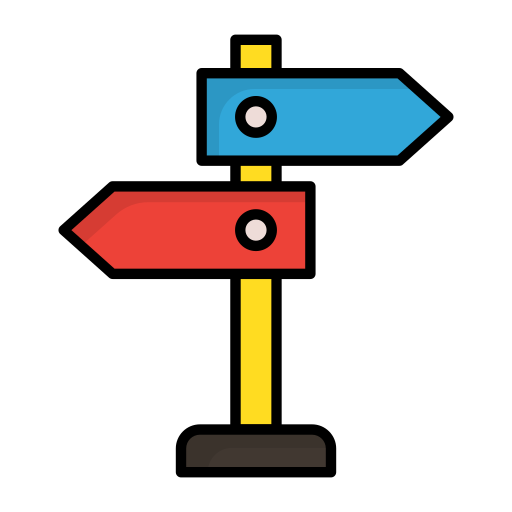} \ Distance}]
Distance denotes the separation between two locations or points.  
\emph{Examples:} “The school is 3 km from home”, “The two trees are 20 m apart”.
\end{promptbox}

\begin{promptbox}[{\includegraphics[height=10pt]{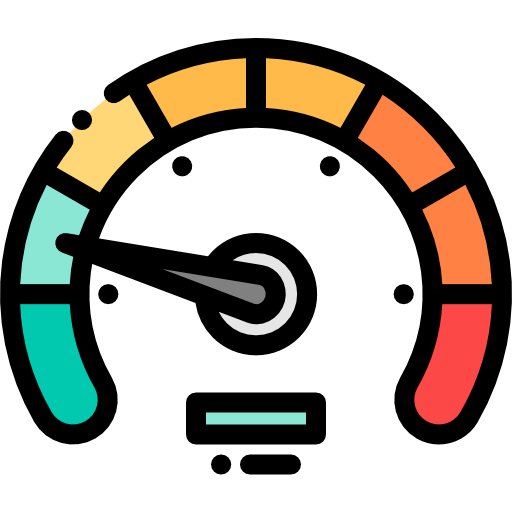} \ Speed}]
Speed refers to the rate of motion, often read from a dashboard or speedometer.  
\emph{Examples:} “The car moves at 60 km/h”, “The bike shows 15 mph”.
\end{promptbox}

\begin{promptbox}[{\includegraphics[height=10pt]{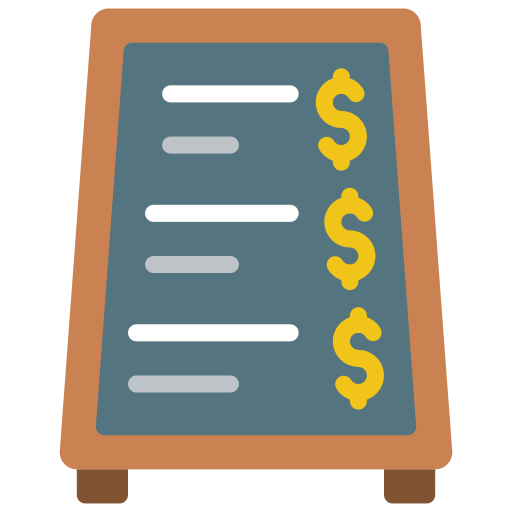} \ Price}]
Price expresses the cost of an item, usually displayed on a signboard or tag.  
\emph{Examples:} “The bread costs \$2.50”, “The toy is labeled 15 yuan”.
\end{promptbox}

\begin{promptbox}[{\includegraphics[height=10pt]{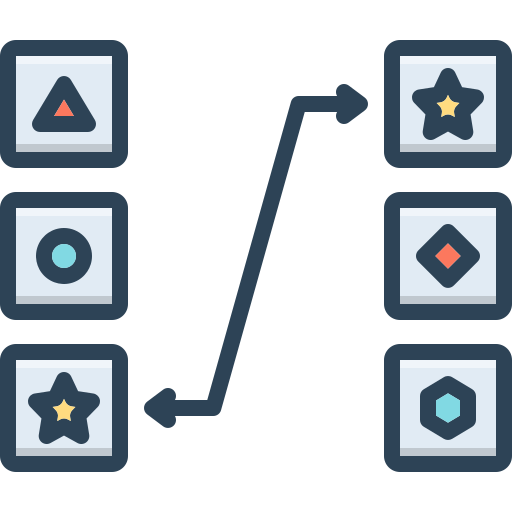} \ Group}]
Group information involves the use of icons to represent collections or categories, where each icon aligns with a specific textual meaning.  
\emph{Examples:}  
- The icon of a person holding a cloth represents the action of "washing".  
- The icon of a teacher standing in front of a board signifies the action of "teaching".  
These icons are used to represent common tasks, roles, or collections that are easily understood through visual semantics.
\end{promptbox}

\begin{promptbox}[{\includegraphics[height=10pt]{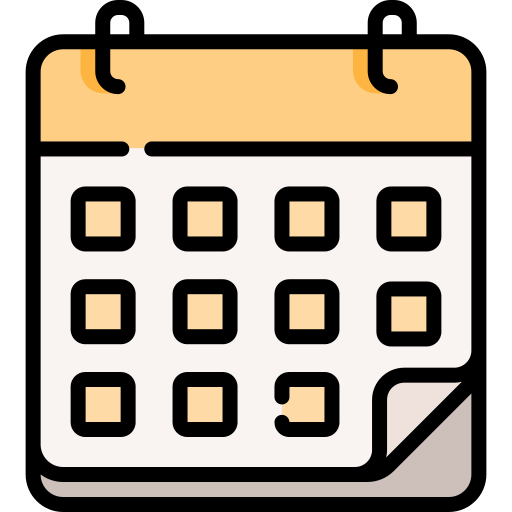} \ Calendar/Age}]
Calendar and age capture temporal aspects such as dates, years, or personal age.  
\emph{Examples:} “Today is March 15”, “The child is 8 years old”.
\end{promptbox}

\begin{promptbox}[{\includegraphics[height=10pt]{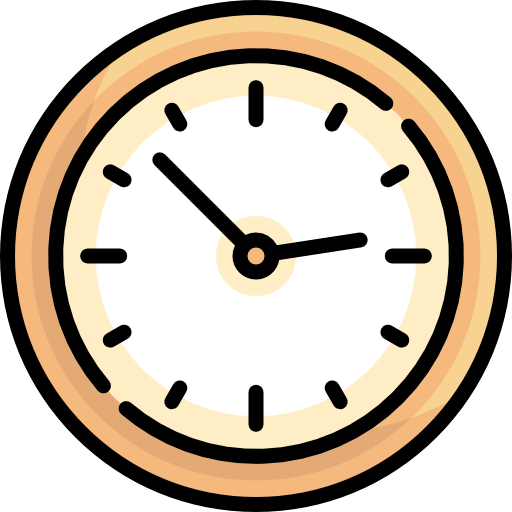} \ Clock}]
Clock information refers to specific times of day shown on analog or digital clocks.  
\emph{Examples:} “The clock shows 7:30”, “The watch displays 12:45”.
\end{promptbox}

\begin{promptbox}[{\includegraphics[height=10pt]{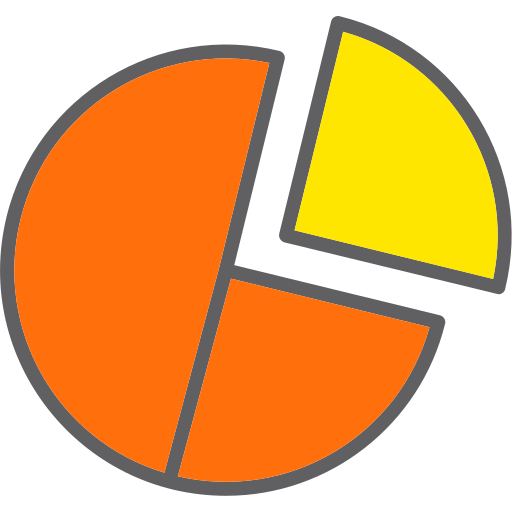} \ Graph}]
Graphs depict ratios, percentages, or quantities visually, often with bars, lines, or pie charts.  
\emph{Examples:} “40\% of the pie chart is shaded”, “The bar reaches value 10”.
\end{promptbox}

\end{figure*}

\begin{figure*}[t]
\centering

\begin{promptbox}[{\includegraphics[height=10pt]{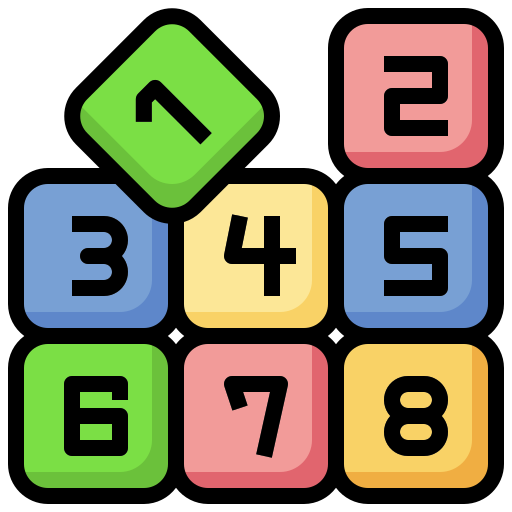} \ Statistics}]
Statistics summarize data in tables or charts, highlighting counts or distributions.  
\emph{Examples:} “The table shows 25 students in Class A”, “The chart lists 300 voters”.
\end{promptbox}

\begin{promptbox}[{\includegraphics[height=10pt]{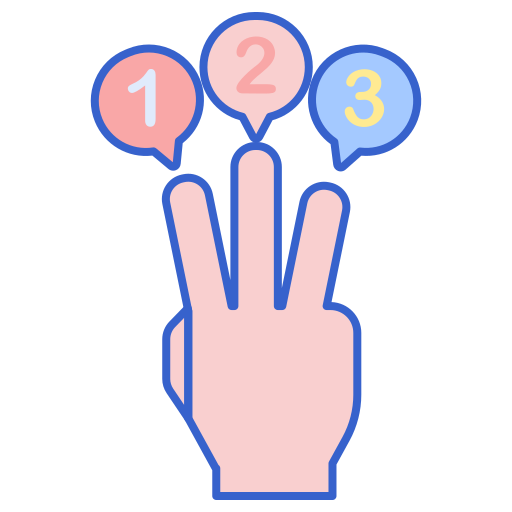} \ Count}]
Count refers to the number of discrete, visible objects in a scene.  
\emph{Examples:} “There are 8 apples on the table”, “The image shows 3 books”.
\end{promptbox}

\begin{promptbox}[{\includegraphics[height=10pt]{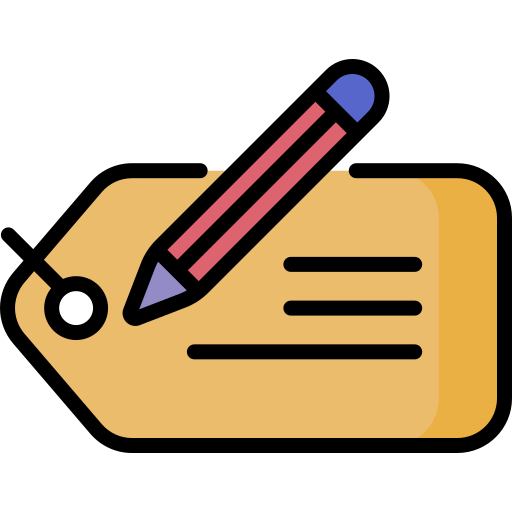} \ Label}]
Labels are direct textual annotations attached to objects, carrying mathematical values.  
\emph{Examples:} “A box labeled 10 kg”, “A jar with tag ‘2 L’”.
\end{promptbox}

\begin{promptbox}[{\includegraphics[height=10pt]{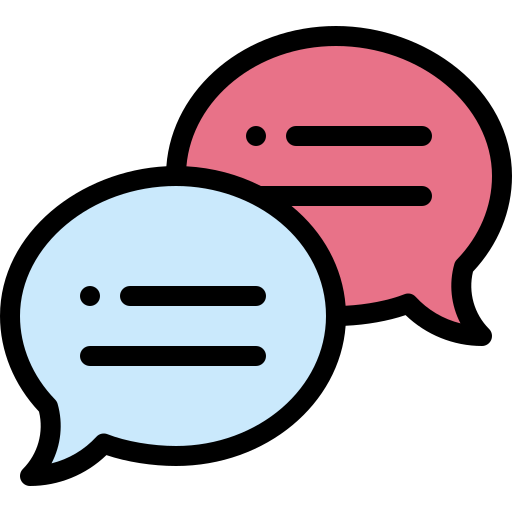} \ Dialogue}]
Dialogue involves characters communicating contextually relevant text in speech bubbles. The dialogue should never directly provide numbers.  
\emph{Examples:} “A customer says ‘I’d like a large pizza’”, “A teacher says ‘Let’s start the exam’”.
\end{promptbox}

\end{figure*}

\subsection{Assign Category}
\label{appendix:assign_category}
\begin{figure*}[h]
    \centering
    \includegraphics[width=0.9\linewidth]{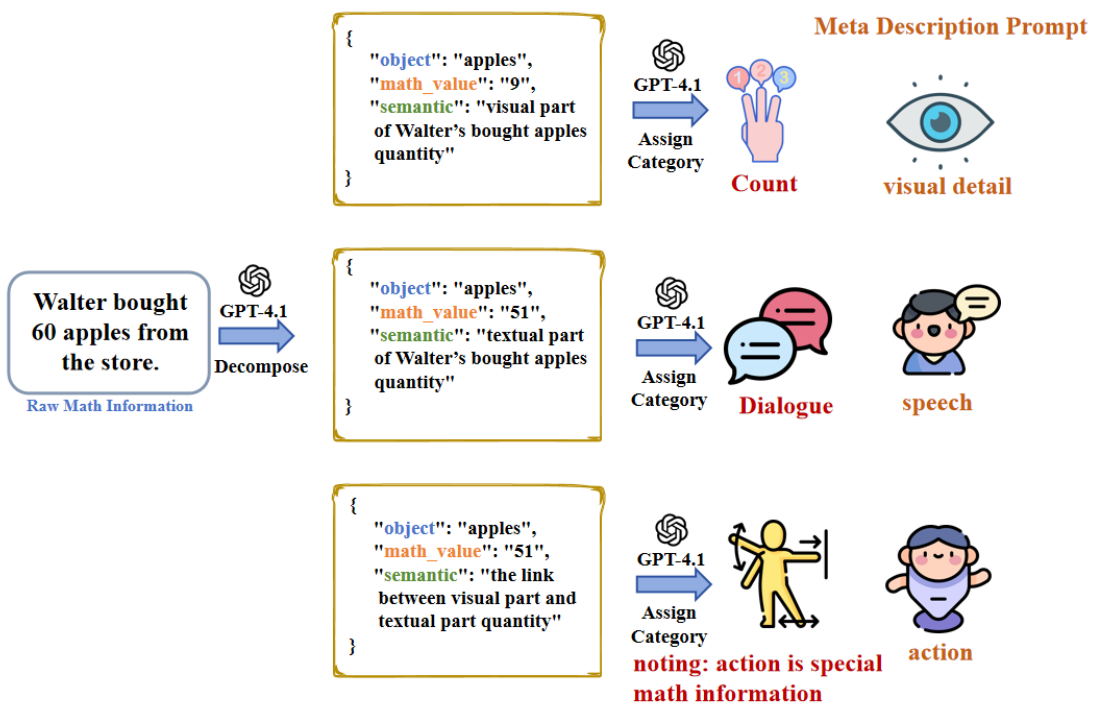}
    \caption{An detailed example that assigns the extracted mathematical information into corresponding categories.}
    \label{fig:assign_category}
\end{figure*}

The allocation process, carried out with the assistance of GPT-4.1, organizes the information in a way that facilitates accurate visual representation. This step is critical for ensuring that each element is positioned correctly in the generated scene. For a detailed look at how the extracted information is categorized and allocated, see Figure~\ref{fig:assign_category}, which illustrates the assignment process.

\subsection{API Token Usage and Cost of GSM8K-V }
\label{appendix:cost_gsm8k-v}

We summarize token usage and cost of data construction. 
Stage~1 and Stage~2 used GPT-4.1 APIs, while Stage~3 used GPT-Image-1 for rendering. 
Pricing details are based on official documentation at the time of experiment. Details are in Table~\ref{tab:token_cost}.

\subsection{Human cross check and refinement}
\label{appendix:human_verification}

\begin{figure*}[t]
    \centering
    \includegraphics[width=1\linewidth]{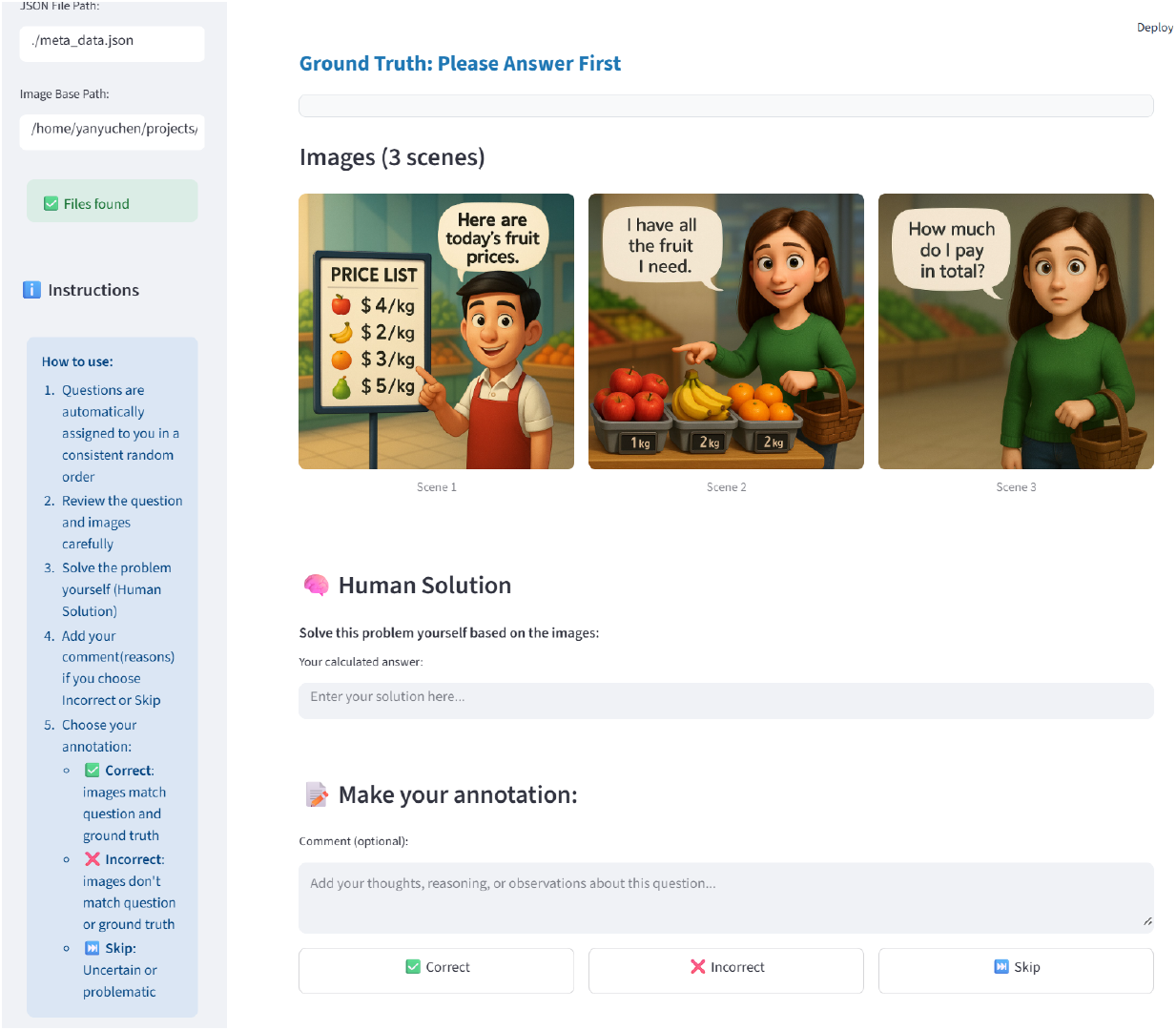}
    \caption{Illustration of the dual human cross-checking process in GSM8K-V construction. 
    Each problem is independently validated by two annotators, who verify scene fidelity under the \emph{3C principles}: 
    \textbf{Consistency} (preserving entities, quantities, and constraints from the original text), 
    \textbf{Completeness} (ensuring all problem-relevant information is visually available), and 
    \textbf{Compliance} (adhering to safety and formatting standards). 
    Annotators also solve each problem to confirm correctness of the derived answer, ensuring benchmark reliability.}
    \label{fig:human_verification}
\end{figure*}

To ensure that the visual scenes accurately reflect the original mathematical problems, we apply a dual human cross-checking process. 
As shown in Figure~\ref{fig:annotator_details}, the verification is conducted by trained annotators with sufficient mathematical and computer science background, and all annotators receive task-specific training before annotation. 
During verification, two independent annotators review each problem and its corresponding visual scenes to ensure \emph{consistency}, \emph{completeness}, and \emph{compliance} with the benchmark requirements. 
This process helps identify mismatches between the original text, generated images, and ground-truth solutions, thereby improving the semantic fidelity and overall data quality of \datasetname. 
The complete human cross-checking workflow is illustrated in Figure~\ref{fig:human_verification}.

\subsection{Robustness to Image Generation Models}
\label{appendix:image_generator_robustness}

To examine whether the observed performance gap is caused by the choice of image generator, we additionally evaluate GSM8K-V variants generated by Nano Banana 2, Midjourney v8 alpha, and Midjourney v7. 
As shown in Table~\ref{tab:image_generator_robustness}, GPT-Image-1 and Nano Banana 2 produce highly consistent results across evaluated VLMs, suggesting that the main findings do not depend on a specific image generation model. 
Performance drops under Midjourney v8 alpha and especially Midjourney v7, mainly due to weaker cross-image consistency, reduced character clarity, and poorer detail fidelity. 
These results indicate that once the image generator reaches a sufficient quality level, the underperformance mainly reflects limitations of current VLMs in implicit visual reasoning rather than artifacts of data generation.

\clearpage

\begin{table*}[htbp]
\centering
\small
\renewcommand{\arraystretch}{1}  %
\setlength{\tabcolsep}{4pt}        %

\begin{tabular}{lccccc}
\toprule
\textbf{Stage} & \makecell{\# \\ Samples} & \makecell{Avg. \\ Input Tokens} & \makecell{Avg. Output \\ Tokens / Images} & \makecell{Avg. Cost \\ / Sample} & \makecell{Total \\ Cost} \\
\midrule
Stage 1 (GPT-4.1) & 1319 & 16,582 & 946 & \$0.041 & \$54.1 \\
Stage 2 (GPT-4.1) & 1319 & 13,093 & 1,039 & \$0.035 & \$46.2 \\
Stage 3 (GPT-Image-1) & 1319 & 1,038 & 4.05 images & \$0.686 & \$904.8 \\
\midrule
\textbf{Total} & - & -- & -- & -- & \textbf{\$1005.1} \\
\bottomrule
\end{tabular}
\caption{Average token usage and estimated costs across three stages of dataset construction. Stage~1/2 use GPT-4.1 inference rates (\$2.00/1M input, \$8.00/1M output). Stage~3 uses GPT-Image-1 (\$10.00/1M input, \$0.167 per image--quality:high, size:1024 x 1024).}
\label{tab:token_cost}
\end{table*}

\begin{figure*}[t]
\centering

\begin{tcolorbox}[
  enhanced,
  colback=gray!5,
  colframe=black!40,
  title={Annotator Details},
  fonttitle=\bfseries,
  arc=2pt,
  boxrule=0.6pt,
  left=8pt,
  right=8pt,
  top=6pt,
  bottom=6pt
]

\textbf{Instructions.}  
\begin{itemize}[leftmargin=12pt]
    \item Questions are automatically assigned to you in a consistent random order.  
    \item Review the question and images carefully.  
    \item Solve the problem yourself (\emph{Human Solution}).  
    \item Add your comment (reasons) if you choose \emph{Incorrect} or \emph{Skip}.  
    \item Choose your annotation:  
    \begin{itemize}
        \item \textbf{Correct}: images match the question and ground truth.  
        \item \textbf{Incorrect}: images do not match the question or ground truth.  
        \item \textbf{Skip}: uncertain or problematic case.  
    \end{itemize}
\end{itemize}

\vspace{1mm}
\textbf{Annotator Description.}  
We employed a total of \textbf{seven annotators} to conduct human verification. Among them, \textbf{five were undergraduate students} and \textbf{two were doctoral students}. All annotators had a solid background in mathematics and computer science, ensuring that their knowledge base was sufficient to understand the problems, inspect the generated scenes, and evaluate correctness. The combination of undergraduate and doctoral perspectives provided a balance between practical annotation skills and advanced reasoning expertise, strengthening the reliability of the human evaluation process.  

\vspace{3mm}
\textbf{Training Procedure.}  
Before annotation began, all annotators underwent a \textbf{one-hour training session} specifically designed for GSM8K-V. The training introduced the project background, the dataset structure, and the end-to-end construction pipeline. Special emphasis was placed on the \emph{error taxonomy}, so that annotators could consistently identify mismatches between images, textual questions, and ground truth solutions. This preparation ensured that annotators were not only familiar with the annotation interface but also equipped to provide high-quality and consistent judgments.  

\end{tcolorbox}

\caption{Annotator information and training protocol used in the human verification process.}
\label{fig:annotator_details}

\end{figure*}

\begin{table*}[t]
\centering
\small
\setlength{\tabcolsep}{6pt}
\begin{tabular}{lcccc}
\toprule
\textbf{Model} &
\makecell{\textbf{GPT-Image-1}\\\textbf{(ours)}} &
\makecell{\textbf{Nano}\\\textbf{Banana 2}} &
\makecell{\textbf{Midjourney}\\\textbf{v8 alpha}} &
\makecell{\textbf{Midjourney}\\\textbf{v7}} \\
\midrule
Gemini-3-Pro       & 57.02 & 58.68 & 43.80 & 13.22 \\
GLM-4.6V           & 32.23 & 31.40 & 17.36 & 11.57 \\
Qwen2.5-VL-72B     & 18.18 & 17.36 & 10.74 & 7.44  \\
Qwen2.5-VL-7B      & 12.40 & 12.40 & 8.26  & 7.44  \\
\bottomrule
\end{tabular}
\caption{
Performance (\%) of VLMs on GSM8K-V variants generated by different image generation models.
}
\label{tab:image_generator_robustness}
\end{table*}

\clearpage

\section{Prompt Templates}
\label{appendix:prompts}

\subsection{Step1 Prompt}
\label{appendix:step1_prompt}

This subsection presents the prompt used for problem decomposition and scene allocation, as shown in Figure~\ref{fig:step1_prompt}. 
Given an original GSM8K textual problem, the prompt extracts the mathematical information required for solving the final question and represents it as structured triples $(\textit{object}, \textit{math value}, \textit{semantic})$. 
It further specifies classification rules, rendering strategies, scene allocation principles, and optional interference settings, producing a structured JSON output for subsequent visual construction.
\begin{figure*}[t]
\centering

\begin{promptbox}[{Step 1: problem decomposition and allocation}]
\textbf{Input (Original Textual Question).}
\vspace{2mm}
\textbf{Prompt.}
\begin{lstlisting}
You are a math information extractor. Given an math word problem, identify only the mathematical information required to solve the final question, and represent it as triples:
(object, math value, semantic)

===============================
 TRIPLE DEFINITIONS
===============================

- object: the entity described (e.g., product, character, item, tool, vehicle).
- math value: the associated numeric attribute (integer, decimal, ratio, unit).
- semantic: the contextual role of this fact (e.g., placement, relation, or action).

===============================
 CLASSIFICATION RULES
===============================
1. Concrete math information:
   - Object is a countable, physical entity.
   - Math value is a pure integer (no units/decimals/percentages).
   - Example: ("apple", 3, "in the basket")
2. Abstract math information:
   - Object is abstract (time, distance, price, weight, score, duration).
   - OR math value is not an integer (e.g., 1.5, 20%
   - Example: ("distance", "12 miles", "between A and B")
===============================
 RENDERING STRATEGY
===============================
- Small countable values (<=10):
  -> Render directly as visual elements (visual_detail).
- Large integers (>10):
  -> Decompose into m = a [op] b with a <= 10.
     Show "a" visually, "b" via speech bubble or text.
- Abstract values:
  -> Express textually or with meta-descriptions
    (e.g., sign board for prices, calendar for dates, ratio graph).
===============================
 SCENE ALLOCATION
===============================
- Group triples into scenes based on context.
- Related facts appear in the same scene.
- The final scene contains only the target question
  (no numeric values), using final_scene_description.

===============================
 INTERFERENCE (OPTIONAL)
===============================
- perception: add visually unrelated but irrelevant objects.
- semantic: insert distractors that are contextually similar
  but not required for solving the problem.

===============================
 OUTPUT FORMAT (JSON)
===============================

{
  "scenes": [
    {
      "scene_id": <int>,
      "interfere": "none" | "perception" | "semantic",
      "scene_math_information": [
        {
          "raw_math_information": <string>,
          "object": <string>,
          "math_value": <string>,
          "semantic": <string>,
          "use_strategy": Distance | Object Measurement | Weight | Graph Ratio | Cross Scene Clock | Dashboard | Day | Sign Board | Action | Visual Detail | Speech Bubble | Label 
        }, ...
      ]
    }, ...
  ]
}
\end{lstlisting}
\end{promptbox}

\caption{Prompt for decomposing GSM8K problems into structured mathematical information and allocating it across visual scenes.}
\label{fig:step1_prompt}

\end{figure*}

\subsection{Meta Description Strategy}
\label{appendix:meta_description_strategy}

This subsection provides the meta description strategies used to convert structured mathematical information into visually grounded scene descriptions. 
Each strategy defines how a specific type of mathematical information, such as distance, measurement, weight, ratio, clock time, price, action, visual detail, speech bubble, or label, should be expressed through objects, actions, and composition. 
These templates help ensure that the generated scene descriptions preserve the original mathematical semantics while remaining visually clear and renderable.

\begin{figure*}[t]
\centering
\begin{promptbox}[{\includegraphics[height=10pt]{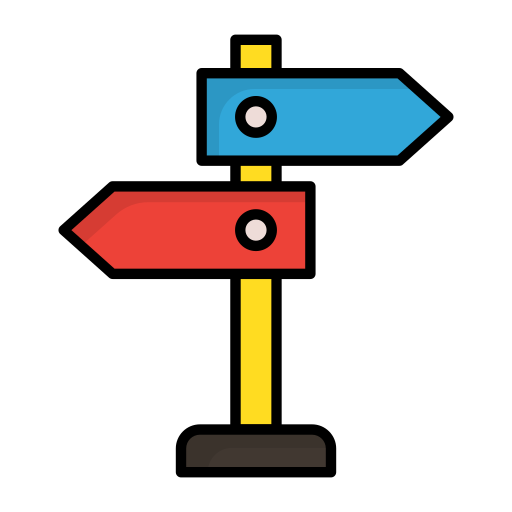} \ Distance Meta Description}]
\textbf{Input:}
\begin{verbatim}
{
  "object": "object field content",
  "math_value": "location A | location B | ...",
  "semantic": "semantic field content",
}
\end{verbatim}

\vspace{2mm}
\textbf{Output:}

\includegraphics[height=10pt]{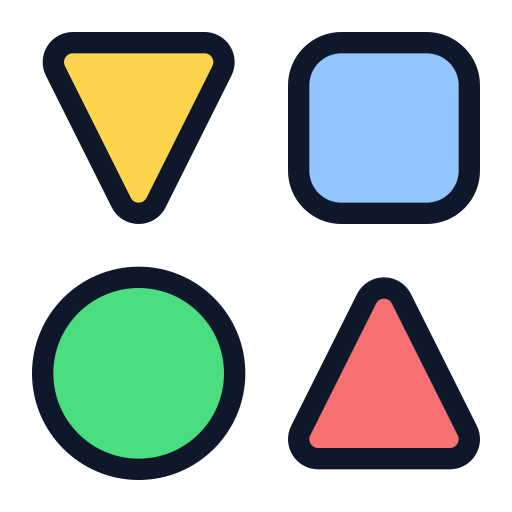}\textbf{Object :}
- A road sign board (reference to object field content).
- Two distinct real-world location icons (from math\_value: location A and B) are shown on the road sign board.
- A double-headed horizontal arrow is placed between the two icons, representing the distance.

\vspace{2mm} 
\includegraphics[height=10pt]{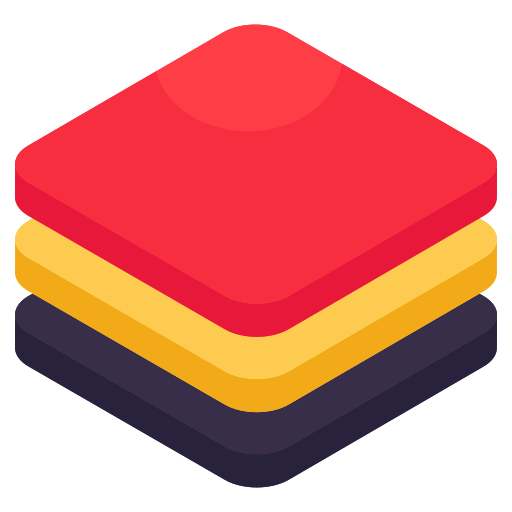}\textbf{Composition:}
- Two location icons are horizontally aligned on opposite sides of the board.
- Arrow spans between them, midpoint centered.
- Label sits exactly at the midpoint, bold and unobstructed.

\end{promptbox}
\end{figure*}

\begin{figure*}[t]
\centering
\begin{promptbox}[{\includegraphics[height=10pt]{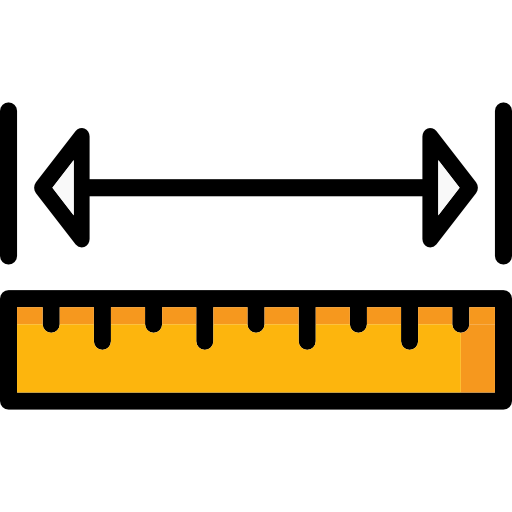} \ Object Measurement Meta Description}]
\textbf{Input:}
\begin{verbatim}
{
  "object": "object field content",
  "math_value": "numeric label",
  "semantic": "width/length/height/depth/thick",
}
\end{verbatim}

\vspace{2mm}
\textbf{Output:}

\includegraphics[height=10pt]{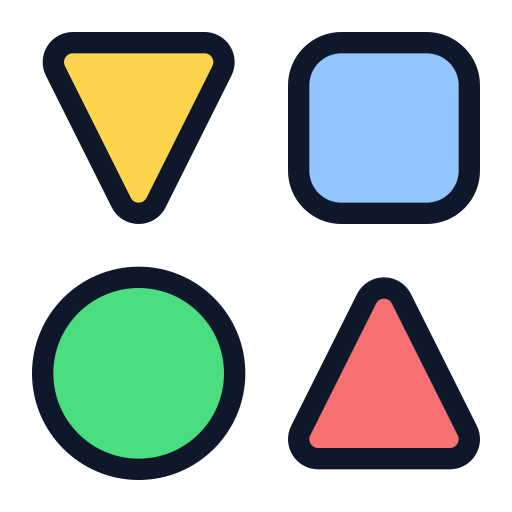}\textbf{Object :}
- A single concrete object (from object field content).
- A dimension line (based on semantic) as a bold double-headed arrow.
- Numeric label (math\_value) at the midpoint.

\vspace{2mm} \includegraphics[height=10pt]{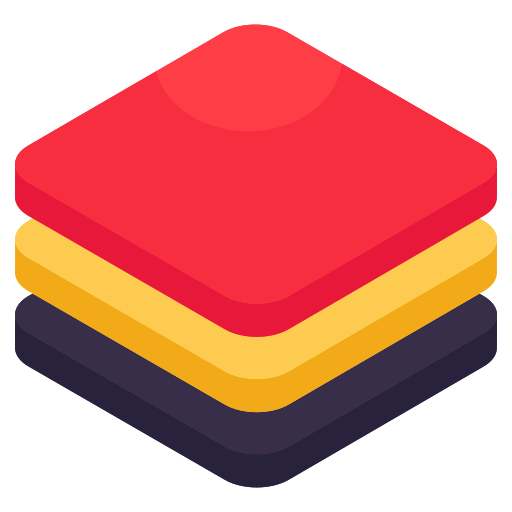}\textbf{Composition:}
- Object centered, with space around.
- Arrow aligns with relevant edges (length, width, height).
- Arrow tips flush with boundaries.
- Numeric label centered, bold, legible.
\end{promptbox}
\end{figure*}

\begin{figure*}[t]
\centering
\begin{promptbox}[{\includegraphics[height=10pt]{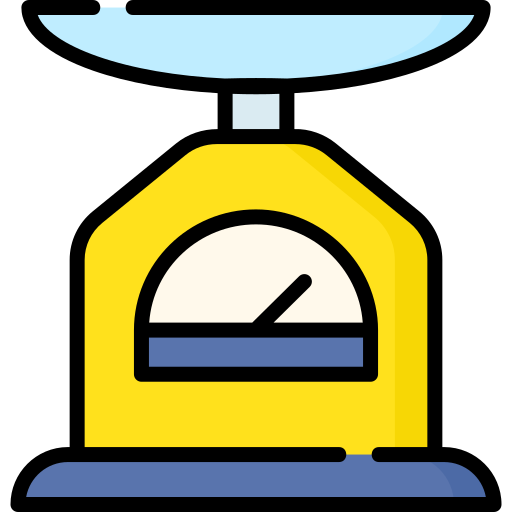} \ Weight Meta Description}]
\textbf{Input :}
\begin{verbatim}
{
  "object": "object field content",
  "math_value": "math_value field content",
  "semantic": "semantic field content",
}
\end{verbatim}

\vspace{2mm}
\textbf{Output:}

\includegraphics[height=10pt]{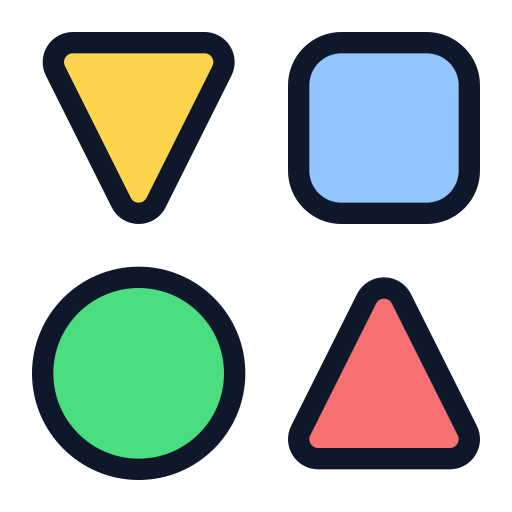}\textbf{Object :}
- Concrete object (object field content) on a digital scale.
- Scale’s screen shows weight: “[math\_value]” in black digits.

\vspace{2mm} \includegraphics[height=10pt]{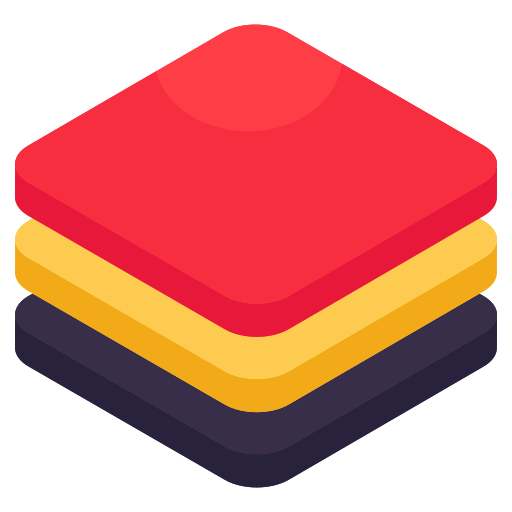}\textbf{Composition:}
- Object centered on weighing surface.
- Display at front, angled to viewer.
- Numeric weight clearly visible, no glare or obstruction.
- No overlaps or extra labels.
\end{promptbox}
\end{figure*}

\begin{figure*}[t]
\centering
\begin{promptbox}[{\includegraphics[height=10pt]{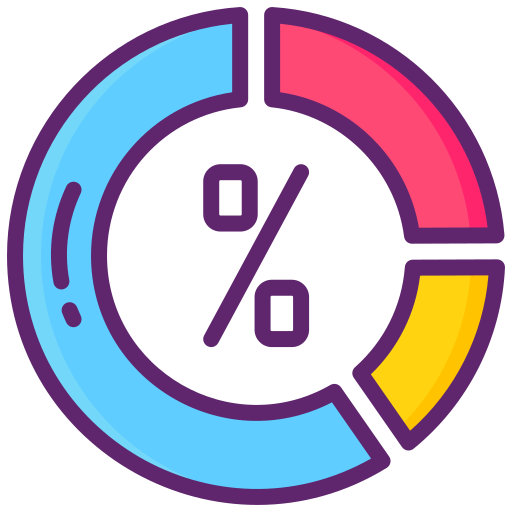} \ Graph Ratio Meta Description}]
\textbf{Input :}
\begin{verbatim}
{
  "object": "[color1] | [color2] | ...",
  "math_value": "N | M | ...",
  "semantic": "color1 legend | color2 legend | ...",
}
\end{verbatim}

\vspace{2mm}
\textbf{Output:}

\includegraphics[height=10pt]{figures/meta_description_icons/object.png}\textbf{Object:}
- A single circle visually displayed as a pie chart representing the ratio of [color1], [color2], ... segments.

- The circle is evenly divided into T equal radial segments, where T = N + M + ... . Starting at the top center and proceeding clockwise.

- Below the pie chart is a legend consisting of horizontal items:
  - A small [color1] square followed by the text '[semantic1]'.
  - A small [color2] square followed by the text '[semantic2]'.

\vspace{2mm} \includegraphics[height=10pt]{figures/meta_description_icons/composition.png}\textbf{Composition:}
- The pie chart is displayed at the position indicated by the semantic.

- All objects mentioned in the scene are clearly visible and not overlapping.
 
\end{promptbox}
\end{figure*}

\begin{figure*}[t]
\centering
\begin{promptbox}[{\includegraphics[height=10pt]{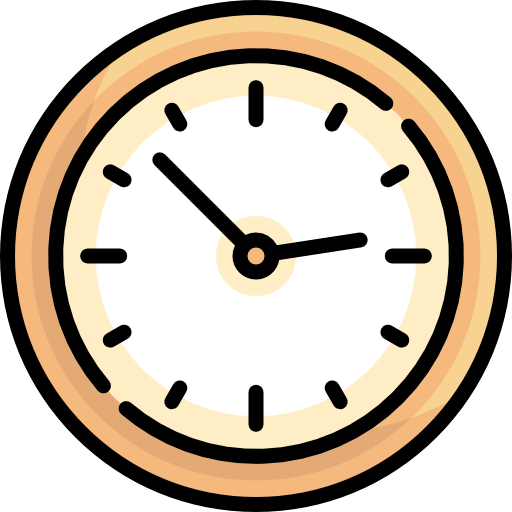} \ Cross Scene Clock Meta Description}]
\textbf{Input :}
\begin{verbatim}
{
  "object": "clock",
  "math_value": "clock time",
  "semantic": "semantic field content",
}
\end{verbatim}

\vspace{2mm}
\textbf{Output:}

\textbf{Time Rule:} If $>$ 12:00, subtract 12:00.  
Example: 15:00 → 3:00.

\includegraphics[height=10pt]{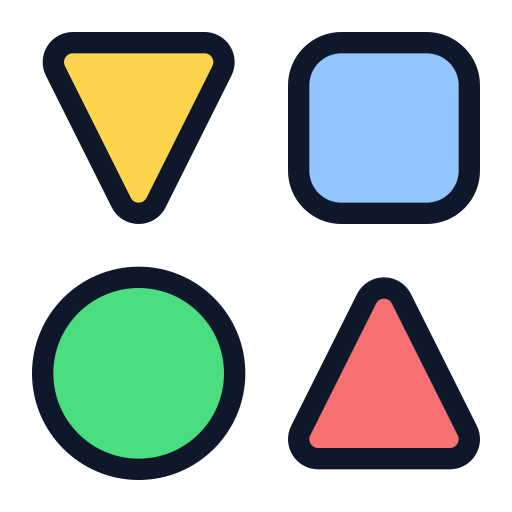}\textbf{Object:}
- Wall clock with light beige face, dark blue border, black tick marks, bold numerals 1–12.
- Distinct black hour \& minute hands (from math\_value).

\vspace{2mm} \includegraphics[height=10pt]{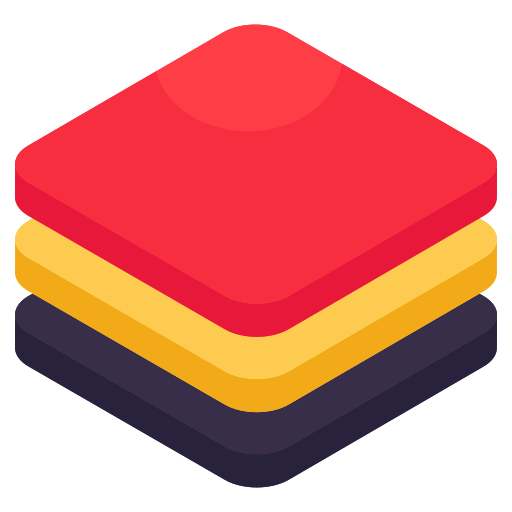}\textbf{Composition:}
- Clock placed on upper right of [semantic position].
\end{promptbox}
\end{figure*}

\begin{figure*}[t]
\centering
\begin{promptbox}[{\includegraphics[height=10pt]{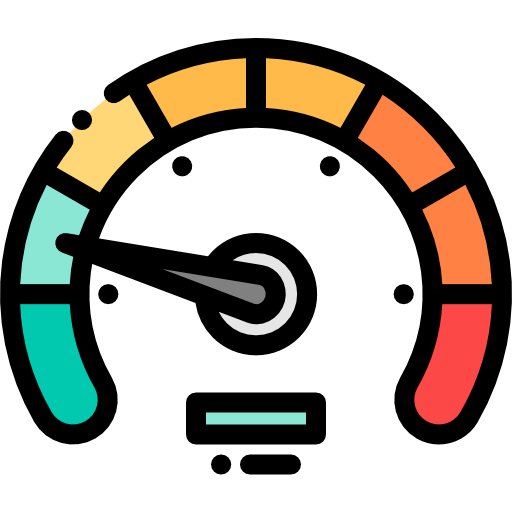} \ Dashboard Meta Description}]
\textbf{Input :}
\begin{verbatim}
{
  "object": "dashboard position",
  "math_value": "math_value field content",
  "semantic": "NUMA | NUMB | NUMC",
}
\end{verbatim}

\vspace{2mm}
\textbf{Output:}

\includegraphics[height=10pt]{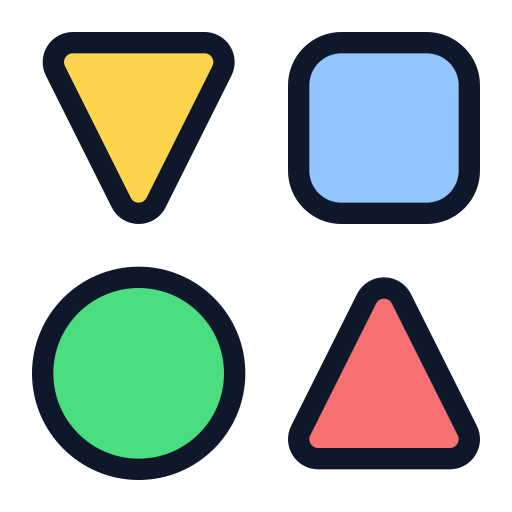}\textbf{Object :}
- Dashboard speedometer: perfect circle, no needle.
- Scale: 0–NUMA, increment NUMB, every NUMC labeled.
- Tick marks evenly spaced.
- Numbers: black sans-serif font near ticks.
- Label: bold unit (e.g. MPH, KMH), centered at bottom.

\vspace{2mm} \includegraphics[height=10pt]{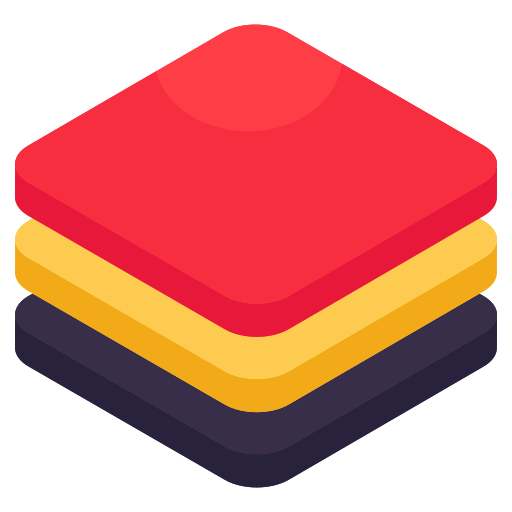}\textbf{Composition:}
- Dashboard placed on [dashboard position].
\end{promptbox}
\end{figure*}

\begin{figure*}[t]
\centering
\begin{promptbox}[{\includegraphics[height=10pt]{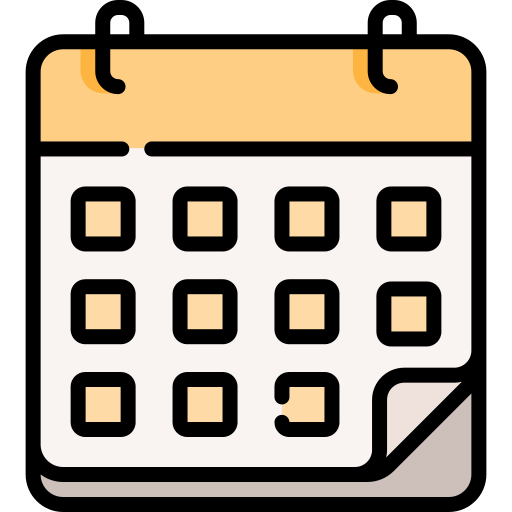} \ Day Meta Description}]
\textbf{Input :}
\begin{verbatim}
{
  "object": "calendar",
  "math_value": "specific day(s)",(day1 | day2 ...)
  "semantic": "semantic field content",
}
\end{verbatim}

\vspace{2mm}
\textbf{Output:}

\includegraphics[height=10pt]{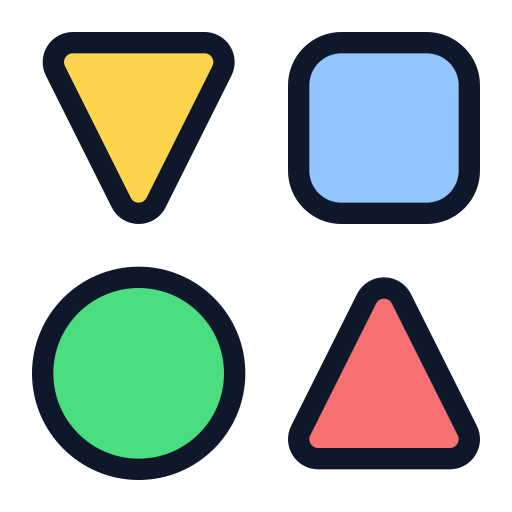}\textbf{Object :}
  - A wall-mounted monthly calendar(The background color is a light beige, while the header area (where "JULY" is written in all capital letters, centered in bold black sans-serif font) is a bright orange, creating a clear visual separation).
  
  - The calendar grid has 7 columns labeled with single-letter weekday initials: S, M, T, W, T, F, S (with Sunday on the far left).
  - calendar has 5 full rows of square date cells representing the days of the month.
  
  - Each date from '1' to '31' is shown as a bold black numeral, centered inside its square cell. The first date ('1') appears in the third cell(i.e., Tuesday column) in the first row. The last  date ('31') appears in the fifth cell(i.e., Thursday column) in the fifth row. 

\vspace{2mm} \includegraphics[height=10pt]{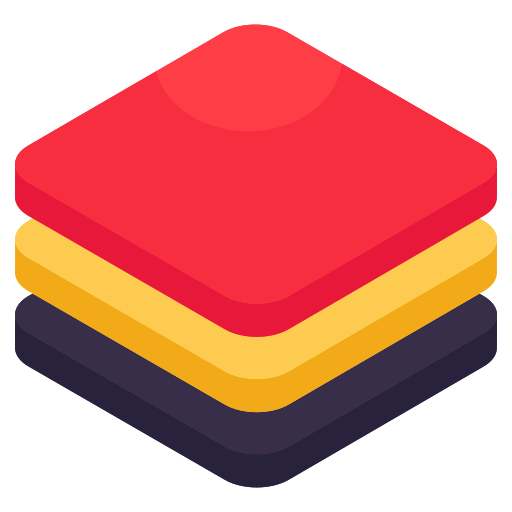}\textbf{Composition:}
composition:
  - The clock is diaplayed on the upper left of [position](based on the semantic field content).
  - The calendar is cleanly structured with even spacing and clearly separated date cells.
  - The grid is rectangular, with consistent column widths and row heights to maintain visual balance.
\end{promptbox}
\end{figure*}

\begin{figure*}[t]
\centering
\begin{promptbox}[{\includegraphics[height=10pt]{figures/meta_description_icons/signboard.png} \ Sign Board Strategy}]
\textbf{Input :}
\begin{verbatim}
{
  "object": "object field content",
  "math_value": "math_value field content",
  "semantic": "semantic field content",
}
\end{verbatim}

\vspace{2mm}
\textbf{Output:}

\includegraphics[height=10pt]{figures/meta_description_icons/object.png}\textbf{Object:}
- A [semantic field content] (add some appearance description).  
- icon (convert the [object field content] into corresponding icon) + [math\_value field content].

\vspace{2mm} \includegraphics[height=10pt]{figures/meta_description_icons/composition.png}\textbf{Composition:}
- The icon/emoji (must convert the object into corresponding icon) should be displayed on a separate line above the \texttt{math\_value} field content.  

\vspace{2mm}

\textbf{Example:}
Input:
\begin{verbatim}
{ "object": "apple", "math_value": "$3", "semantic": "price list"},
{ "object": "banana", "math_value": "$2/kg", "semantic": "price list"}
\end{verbatim}

\vspace{2mm}
\textbf{Output:}

Output:
- A visually clear and attractive price list with textual label 'price list'.  
- The price list line1 states: \includegraphics[height=7pt]{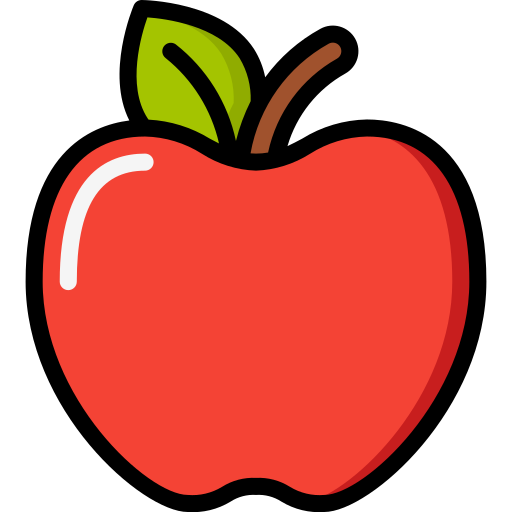} \ \$3.  
- The price list line2 states: \includegraphics[height=7pt]{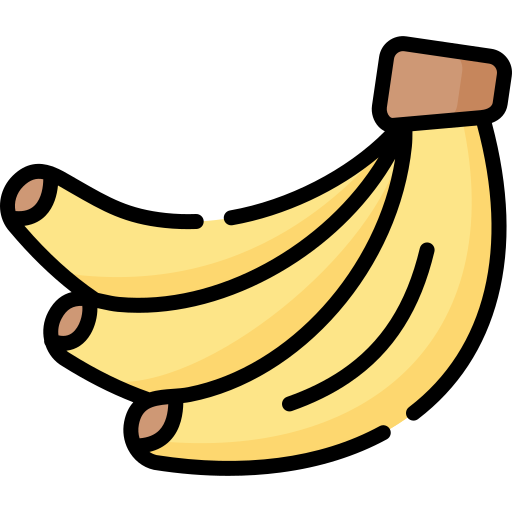} \ \$2/kg.  

Composition:  
- The \includegraphics[height=7pt]{figures/meta_description_icons/apple.png} \ \$3 and 
  \includegraphics[height=7pt]{figures/meta_description_icons/banana.png} \ \$2/kg are displayed on a separate line on the price list.
\end{promptbox}
\end{figure*}

\begin{figure*}[t]
\centering
\begin{promptbox}[{\includegraphics[height=10pt]{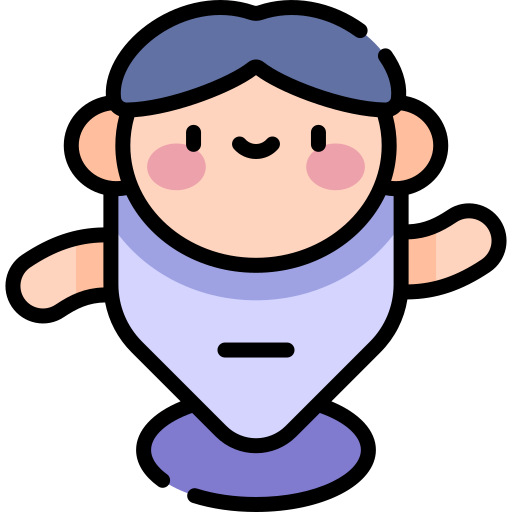} \ Action Strategy (Visual–Textual Linking)}]
\textbf{Input :}
\begin{verbatim}
{
  "object": "",
  "math_value": "",
  "semantic": "The character specific action.",
}
\end{verbatim}

\vspace{2mm}
\textbf{Output:}

\textbf{Description:}
Action is a link between visual-detail-description and textual-information-detail.

\includegraphics[height=10pt]{figures/meta_description_icons/action1.png}\textbf{Approved Cross-Modal Visual Linkage Strategies:}

1. Pointing Gesture  
   The character must clearly point toward the specific visual group.
   
2. Gaze Direction  
   The eyes of the character must align with the referenced object(s). Gaze must be directed and specific — not ambiguous or generic.
   
3. Torso Orientation  
   The character’s upper body must be turned toward the relevant visual target, reinforcing gaze or pointing.
   
4. Holding or Touching  
   The character physically holds or touches an object that semantically links to the textual math content. Use only when unambiguous.

\end{promptbox}
\end{figure*}

\begin{figure*}[t]
\centering
\begin{promptbox}[{\includegraphics[height=10pt]{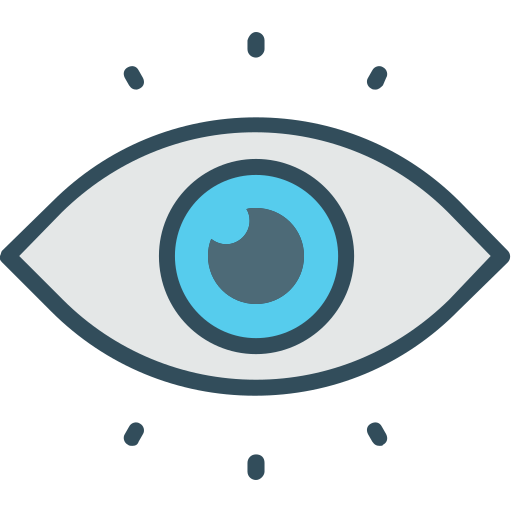} \ Visual Detail Strategy}]
\textbf{Input :}
\begin{verbatim}
{
  "object": "object field content",
  "math_value": "math_value field content",
  "semantic": "semantic field content",
}
\end{verbatim}

\vspace{2mm}
\textbf{Output:}

\includegraphics[height=10pt]{figures/meta_description_icons/object.png}
\textbf{Object:}
- “[math\_value field content] [object field content]”, each rendered with distinct visual features (e.g., color, shape, texture) are placed evenly on the where (semantic field content).

\vspace{2mm}

\includegraphics[height=10pt]{figures/meta_description_icons/composition.png}
\textbf{Composition:}
- “[math\_value field content] [object field content]” must be described in a fully visible, non-overlapping, and easily countable manner.  
- Objects must exhibit slight variation in position, size, or orientation to enhance perceptual clarity.
\end{promptbox}
\end{figure*}

\begin{figure*}[t]
\centering
\begin{promptbox}[{\includegraphics[height=10pt]{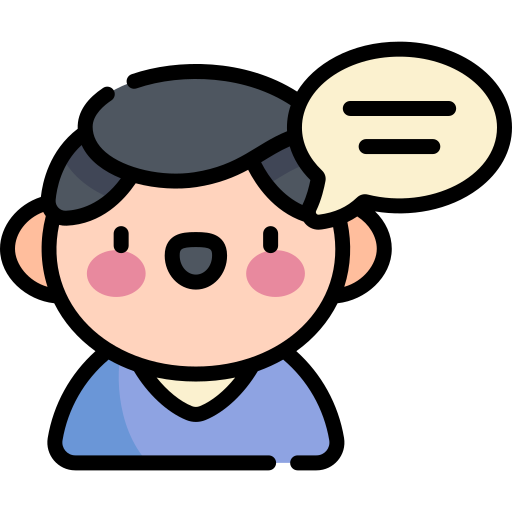} \ Speech Bubble Strategy}]
\textbf{Input :}
\begin{verbatim}
{
  "object": "object field content",
  "math_value": "",
  "semantic": "semantic field content",
}
\end{verbatim}

\vspace{2mm}
\textbf{Output:}

\includegraphics[height=10pt]{figures/meta_description_icons/object.png}
\textbf{Object:}
- A Character (must be the same as object field content) facing forward or slightly angled.  
- A clearly visible speech bubble extending from the character's mouth or face, fully and samely stating semantic field content.

\vspace{2mm} \includegraphics[height=10pt]{figures/meta_description_icons/composition.png}
\textbf{Composition:}
- The speech bubble must be connected to the speaking character with a visible tail.  
- The speech bubble should not overlap with the other object.
\end{promptbox}
\end{figure*}

\begin{figure*}[t]
\centering
\begin{promptbox}[{\includegraphics[height=10pt]{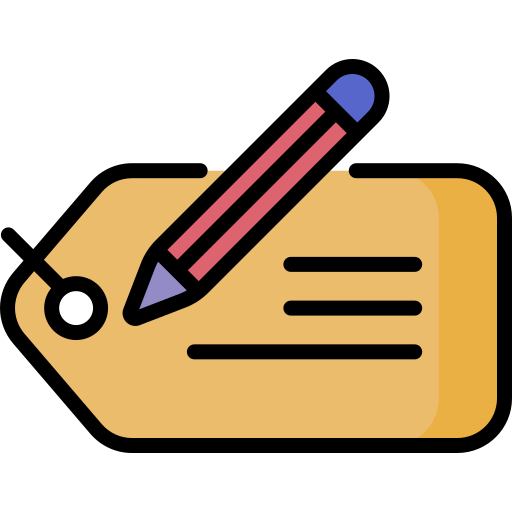} \ Label Strategy}]
\textbf{Input :}
\begin{verbatim}
{
  "object": "object field content",
  "math_value": "math_value field content",
  "semantic": "",
}
\end{verbatim}

\vspace{2mm}
\textbf{Output:}

\textbf{Object:}
- The object (must be the same as 'object' field content).  
- The label stating must be the same as 'math\_value' field content.

\textbf{Composition:}
- The label must be attached to the surface of an object (must be the same as 'object' field content) or be part of the inherent surface marking of the object.
\end{promptbox}
\end{figure*}

\subsection{Image Generation Prompts}
\label{appendix:image_generation_prompts}

This subsection reports the image generation prompt used to render the structured scene descriptions into GSM8K-V images, as shown in Figure~\ref{fig:image_generation_prompt}. 
The prompt requires the image model to generate exactly one image for each scene, follow the given scene order, maintain consistent visual style and character design, and strictly preserve all numerical and descriptive details. 
These constraints are designed to improve generation consistency and ensure that each visual scene faithfully corresponds to its assigned mathematical information.

\begin{figure*}[t]
\centering
\begin{promptbox}[{Image Generation Prompt}]

\begin{lstlisting}
You are a **skilled and consistent image illustrator**. You 
will continuously generate {num_scenes} distinct and separate 
images, each corresponding to exactly one of the scenes 
described below. You must follow the detailed instructions and 
generation process with no exception. 

***STRICT GENERATION RULES:***
- You MUST generate **exactly one image per scene**.
- You MUST generate the images in the exact given order: from **Scene 1** to **Scene {num_scenes}**.
- You MUST NOT generate multiple images for the same scene.
- You MUST NOT skip any scenes.
- You MUST NOT repeat or regenerate previously generated scenes.

***IMAGE STYLE & FORMAT RULES:***
- All {num_scenes} images must follow a **Pixar-style** with clear, sharp, and naturally lit rendering, no warm tints, soft color haze, textures, or yellow glow.
- Use consistent character design, background tone, and visual quality across all scenes.
- Every image must strictly match all numeric and descriptive details from its corresponding scene.
- Each image must be square format (exactly 1024 x 1024 pixels (square, 1:1)).

---

Now begin generating the images **scene by scene**, in strict order.
Generate the image corresponding to the current scene.
Each new image must correspond to the **next numbered scene** 
and must not duplicate any earlier scenes.

Scene 1 Special Start Instruction:
Before generating Scene 1 image, explicitly output:
"Now generating Scene 1 image (Pixar-style, at exactly 1024 x 1024 pixels (square, 1:1))."
Then explicitly output the Scene 1 description in full.
After that, generate Scene 1 image.

After finishing any scene image, explicitly output:
"Scene x image finished, now generate Scene x+1 image (Pixar-style, at exactly 1024 x 1024 pixels (square, 1:1))."

Then explicitly output the Scene X+1 description in full.
After that, generate Scene X+1 image.

After all scenes images generated, explicitly output: "All finished!"

#######################################

The {num_scenes} scenes are described below:
...

\end{lstlisting}
\end{promptbox}
\caption{Prompt for generating multi-scene GSM8K-V images from structured scene descriptions.}
\label{fig:image_generation_prompt}
\end{figure*}

\subsection{Controlled Interference}
\label{appendix:controlled_interference}

This subsection provides additional details on the controlled interference mechanism introduced in Section~\ref{sec:method_step1}. 
The goal of interference in \datasetname is not simply to increase difficulty, but to make the visualized GSM8K problems better reflect realistic visual scenarios, where task-relevant evidence often appears together with irrelevant objects, background clutter, or semantically related distractors. 
During scene allocation, we introduce interference in a controlled manner while preserving the original mathematical semantics and answer recoverability. 
Specifically, the original problem-relevant mathematical information remains unchanged, and all added interference is explicitly marked in the structured scene representation.

We consider two types of interference:
\begin{itemize}[leftmargin=15pt]
    \item \textbf{Perception interference}: visually salient but mathematically irrelevant objects are added to enrich the scene and simulate natural visual clutter.
    \item \textbf{Semantic interference}: contextually related distractors are introduced alongside the target information. 
\end{itemize}

\begin{figure*}[t]
\centering
\begin{interferencebox}[Controlled Interference Example]

\textbf{Original Problem:}  
\emph{``Violetta wants to buy new crayons. She needs them in 5 different colors and prepared \$20 for this purchase. One crayon costs \$2. How much change will she get?''}  

\medskip
\textbf{Structured JSON:}
\begin{lstlisting}
{
  "scenes": [
    {
      "scene_id": 1,
      "interfere": "semantic",
      "scene_math_information": [
        {"raw_math_information": "One crayon costs $2.", 
        "object": "crayon", 
        "math_value": "$2", 
        "semantic": "one crayon's price",
        },
        {
          "raw_math_information": "interference: One eraser costs $1.",
          "object": "eraser",
          "math_value": "$1",
          "semantic": "one eraser's price",
        },
        {
          "raw_math_information": "interference: One pencil costs $1.50.",
          "object": "pencil",
          "math_value": "$1.50",
          "semantic": "one pencil's price",
        }
      ]
    },
    {
      "scene_id": 2,
      "interfere": "perception",
      "scene_math_information": [{ "raw_math_information": "She needs them in 5 different colors.", "object": "crayon", "math_value": "5", "semantic": "Violetta's need for different color crayon",},{ "raw_math_information": "interference: There is a blue backpack on the counter.", "object": "backpack", "math_value": "1", "semantic": "on the counter next to the crayons",}]
    },
    {
      "scene_id": 3,
      "interfere": "none",
      "scene_math_information": [{ "raw_math_information": "Violetta prepared $20 for this purchase.", "object": "us bill", "math_value": "$20", "semantic": "Violetta prepared money",}]
    },
    {
      "scene_id": 4,
      "interfere": "none",
      "scene_math_information": [{"raw_math_information": "How much change will she get?", "object": "Violetta", "math_value": "", "semantic": "How much change will I get after buying these crayons?",}]
    }
  ]
}
\end{lstlisting}

\medskip
\textbf{Rendered Scenes:}

\begin{center}
\begin{tabular}{cccc}
    \begin{minipage}{0.22\linewidth}
        \centering
        \includegraphics[width=\linewidth]{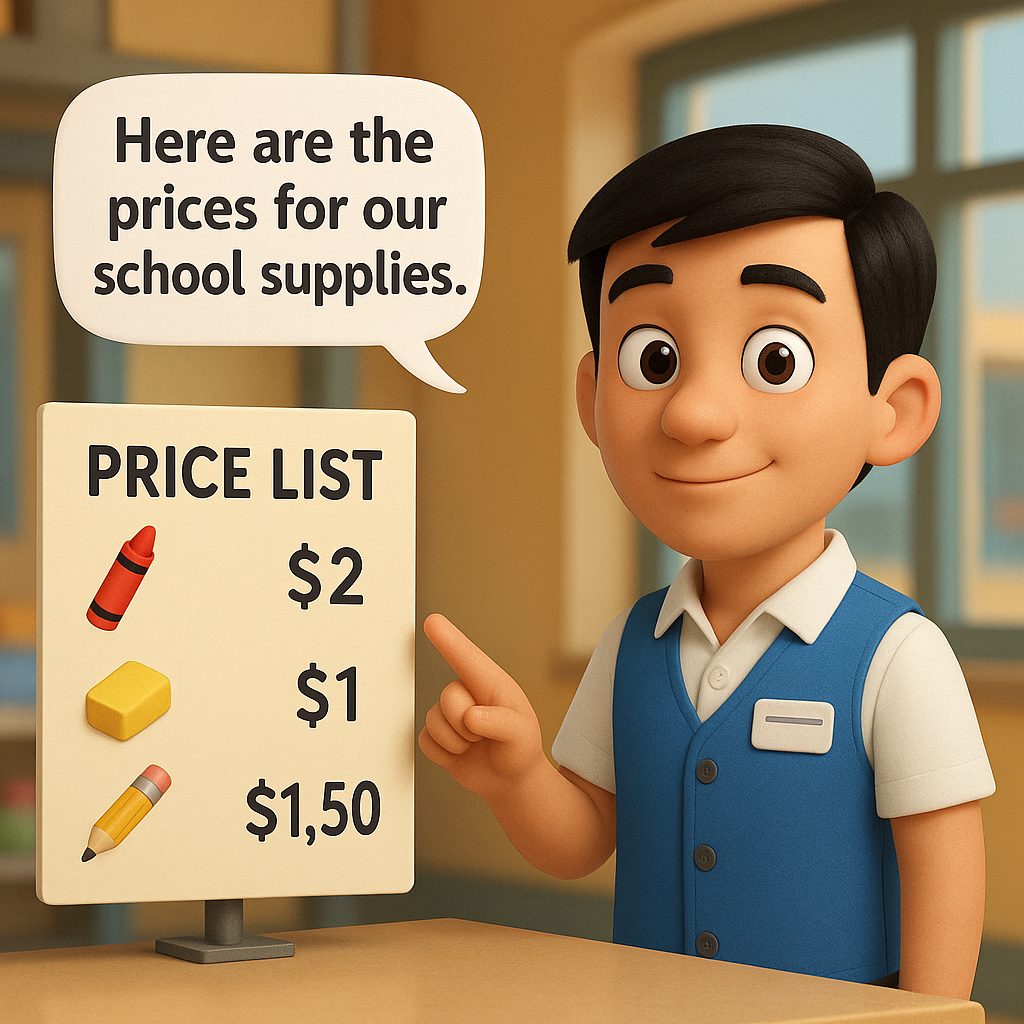} \\
        (a) Scene 1: \textcolor{red}{\textbf{Semantic interference}}
    \end{minipage} &
    \begin{minipage}{0.22\linewidth}
        \centering
        \includegraphics[width=\linewidth]{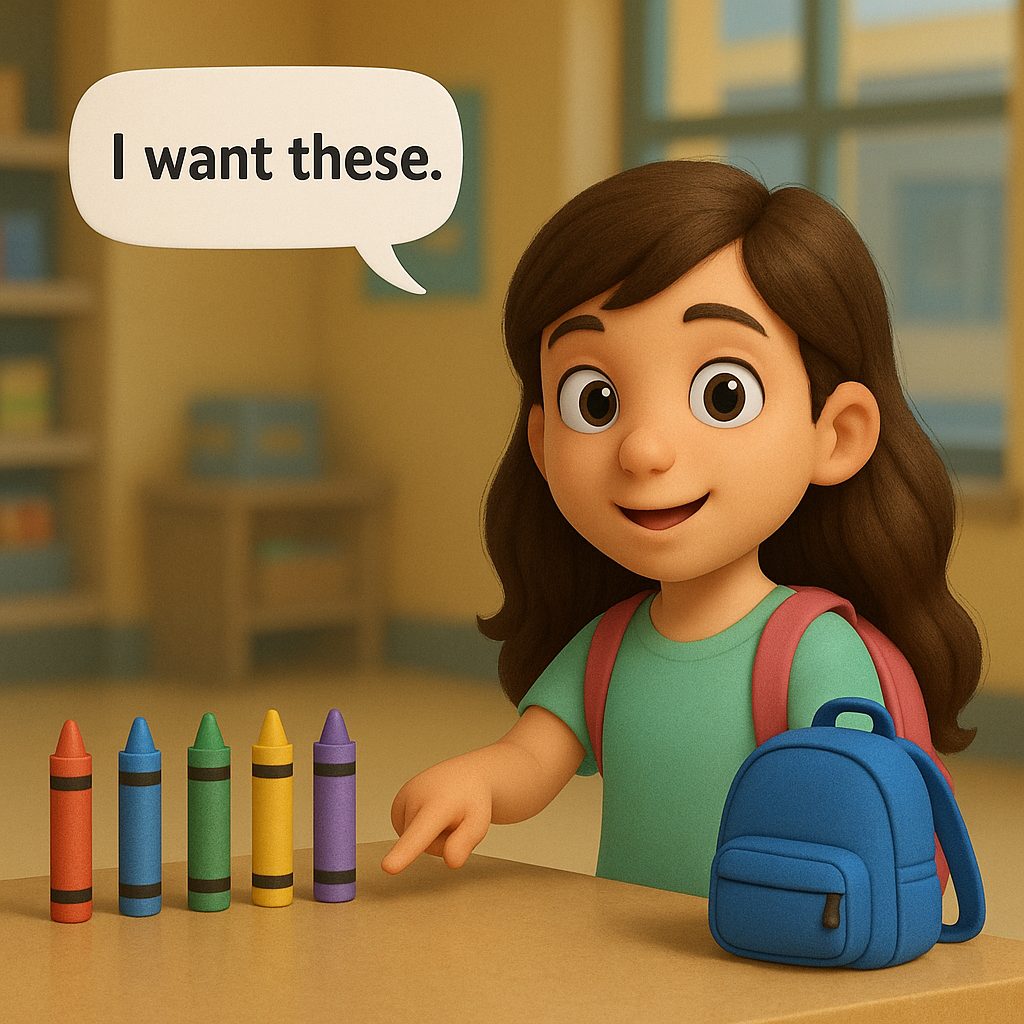} \\
        (b) Scene 2: \textcolor{blue}{\textbf{Perception interference}}
    \end{minipage} &
    \begin{minipage}{0.22\linewidth}
        \centering
        \includegraphics[width=\linewidth]{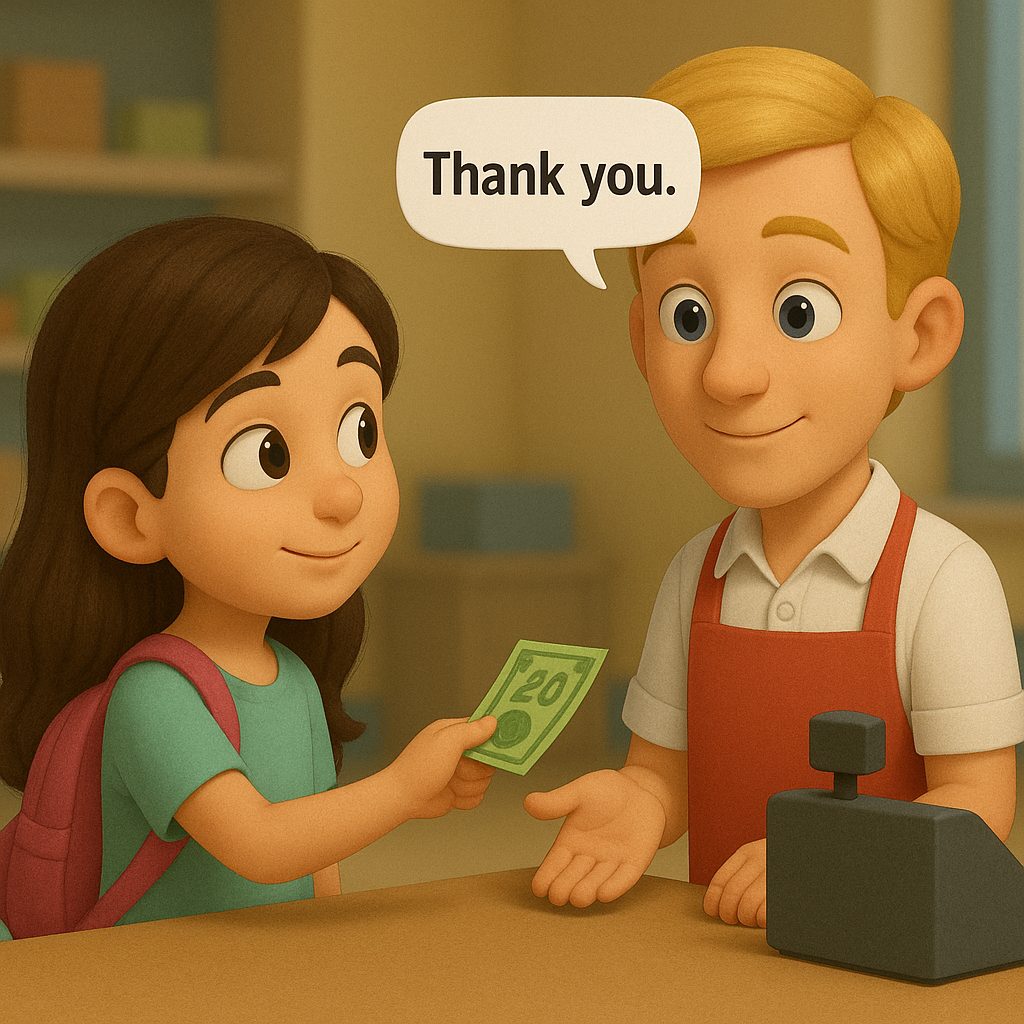} \\
        (c) Scene 3: No interference
    \end{minipage} &
    \begin{minipage}{0.22\linewidth}
        \centering
        \includegraphics[width=\linewidth]{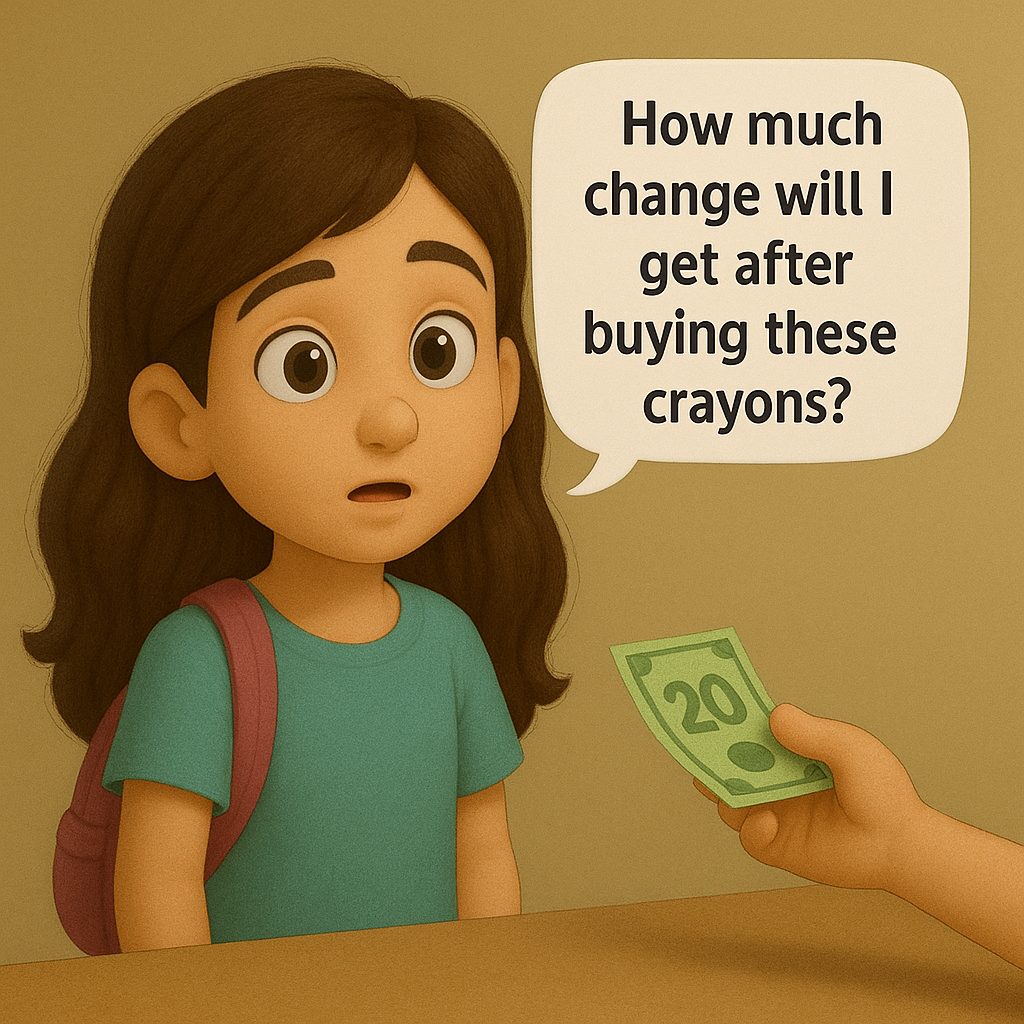} \\
        (d) Scene 4: No interference
    \end{minipage} \\
\end{tabular}
\end{center}

\medskip
\textbf{Explanation.}  
Scene~1 inserts semantically close distractors (eraser, pencil) as \textbf{semantic interference}.  
Scene~2 adds irrelevant but visually salient objects (backpack) as \textbf{perception interference}.  
Scenes~3 and~4 contain no interference and faithfully encode the original problem.

\end{interferencebox}
\caption{Controlled interference example}
\label{fig:controlled_interference_example}

\end{figure*}

Figure~\ref{fig:controlled_interference_example} shows a representative example. As discussed in Section~\ref{subsec:error_analysis}, interference-related errors such as Entity Identity Confusion (EIConf) account for only a small fraction of failures, suggesting that the main bottleneck of \datasetname remains implicit visual inference rather than distractor sensitivity alone.

\subsection{Image Generation Prompts and Verification Principles}
The prompts for GPT-Image-1 \citep{openai2025gptimage1} are constructed from the structured scene descriptions, ensuring that each $(\textit{object}, \textit{action}, \textit{composition})$ tuple is faithfully rendered. To maintain benchmark quality, we apply iterative human-in-the-loop verification guided by the \textbf{3C principles}:
\begin{itemize}
    \item \textbf{Consistency}: generated images must preserve the entities, quantities, and constraints of the original text, with any modification remaining semantics-preserving.
    \item \textbf{Completeness}: all information required to solve the problem must be visually available without relying on external assumptions.
    \item \textbf{Compliance}: images must satisfy safety and formatting standards, avoid sensitive content, and ensure clarity of objects and legible numerals.
\end{itemize}
Items violating these principles are refined or, in severe cases, re-rendered with modified scene descriptions.

\clearpage

\twocolumn[{%
\begin{center}
\includegraphics[
  width=0.93\textwidth,
  height=0.41\textheight,
  keepaspectratio
]{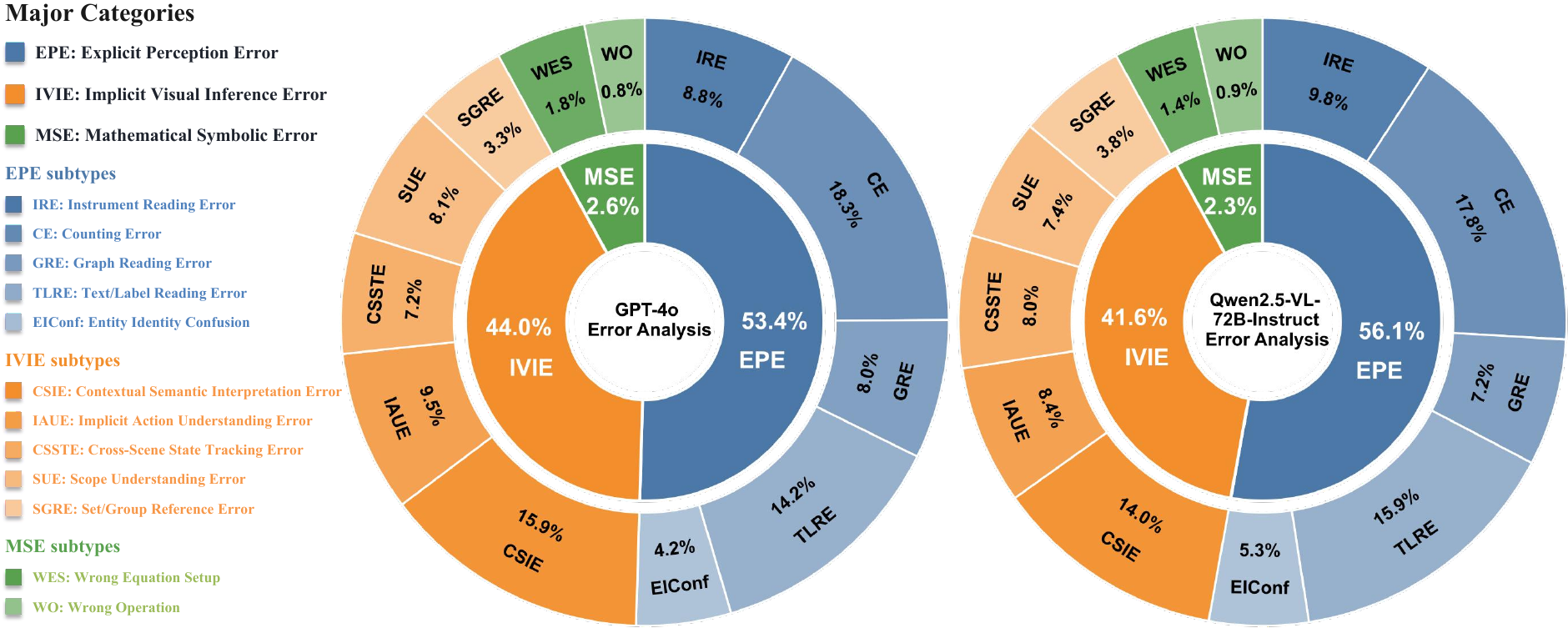}

\refstepcounter{figure}
\label{fig:error_distribution}

\vspace{1mm}
\noindent\small
\textbf{Figure~\thefigure:}
Fine-grained error distributions of GPT-4o and Qwen2.5-VL-72B on GSM8K-V.
The inner ring shows EPE, IVIE, and MSE, while the outer ring decomposes each category into fine-grained subtypes.
\end{center}

\vspace{2mm}
}]

\section{Error Analysis}
\label{appendix:error_analysis}

This section provides additional error analysis for GPT-4o and Qwen2.5-VL-72B, complementing the Gemini-3-Pro analysis in Section~\ref{subsec:error_analysis}. 
Following the same taxonomy, we categorize model failures into three major types: Explicit Perception Error (EPE), Implicit Visual Inference Error (IVIE), and Mathematical Symbolic Error (MSE). 
As shown in Figure~\ref{fig:error_distribution}, both models exhibit similar error patterns: most failures occur before the final symbolic calculation stage, with EPE and IVIE accounting for the vast majority of errors.

\paragraph{Explicit Perception Error (EPE).}
EPE refers to failures in reading explicit visual facts, such as object counts, text or labels, instrument readings, graphs, and entity identities. 
For GPT-4o and Qwen2.5-VL-72B, EPE accounts for $53.4\%$ and $56.1\%$ of analyzed failures, respectively. 
The most frequent EPE subtypes are Counting Error and Text/Label Reading Error, indicating that both models still struggle to reliably extract directly visible numerical and textual information from visual scenes. 
Instrument Reading Error (IRE), Graph Reading Error (GRE), and Entity Identity Confusion (EIConf) further show that even explicit visual evidence can be difficult to bind correctly to the corresponding mathematical entities.

\paragraph{Implicit Visual Inference Error (IVIE).}
IVIE refers to failures in recovering information implied by visual context, including actions, groupings, state changes, scope constraints, and cross-scene relations. 
This category accounts for $44.0\%$ of GPT-4o errors and $41.6\%$ of Qwen2.5-VL-72B errors, making it the second dominant failure source for both models. 
Among the fine-grained subtypes, Contextual Semantic Interpretation Error (CSIE) is the largest component, followed by Implicit Action Understanding Error (IAUE), Cross-Scene State Tracking Error (CSSTE), Scope Understanding Error (SUE), and Set/Group Reference Error (SGRE). 
These errors suggest that models often fail not because the relevant information is absent, but because they cannot infer the intended semantic relation from visual actions, scene transitions, or contextual constraints.

\paragraph{Mathematical Symbolic Error (MSE).}
MSE refers to errors in equation construction or numerical computation after the relevant visual information has been recovered. 
Compared with EPE and IVIE, MSE is rare, accounting for only $2.6\%$ of GPT-4o errors and $2.3\%$ of Qwen2.5-VL-72B errors. 
The two subtypes, WES and WO, occur much less frequently than perception and visual-inference errors. 
This confirms that GSM8K-V primarily challenges models in visual grounding and implicit visual inference, rather than in the final arithmetic step.

\paragraph{Key observations.}
The appendix results are consistent with the main error analysis of Gemini-3-Pro. 
Although GPT-4o and Qwen2.5-VL-72B show slightly higher EPE than IVIE, IVIE remains a major source of failure for both models. 
This indicates that GSM8K-V is not merely testing OCR, counting, or instrument reading; instead, a substantial portion of errors comes from the inability to recover visual semantics that are implied rather than explicitly stated. 
Overall, the results further support our conclusion that current VLMs struggle to transform distributed visual evidence into coherent mathematical reasoning chains.

\begin{figure*}[t]
\centering

\begin{casestudy}[Error Case Study: EPE-CE]

\begin{subsecbox}
\textbf{Text Problem.}

\vspace{0.5mm}
Vincent can buy flowers in packages of 3 for \$2.50 or in packages of 2 for \$1. 
How much money does he save by buying 18 flowers at the better price?
\end{subsecbox}

\begin{subsecbox}
\textbf{GSM8K-V Images.}

\vspace{1mm}
\begin{minipage}{0.44\linewidth}
  \centering
  \includegraphics[width=\linewidth,height=0.2\textheight,keepaspectratio]{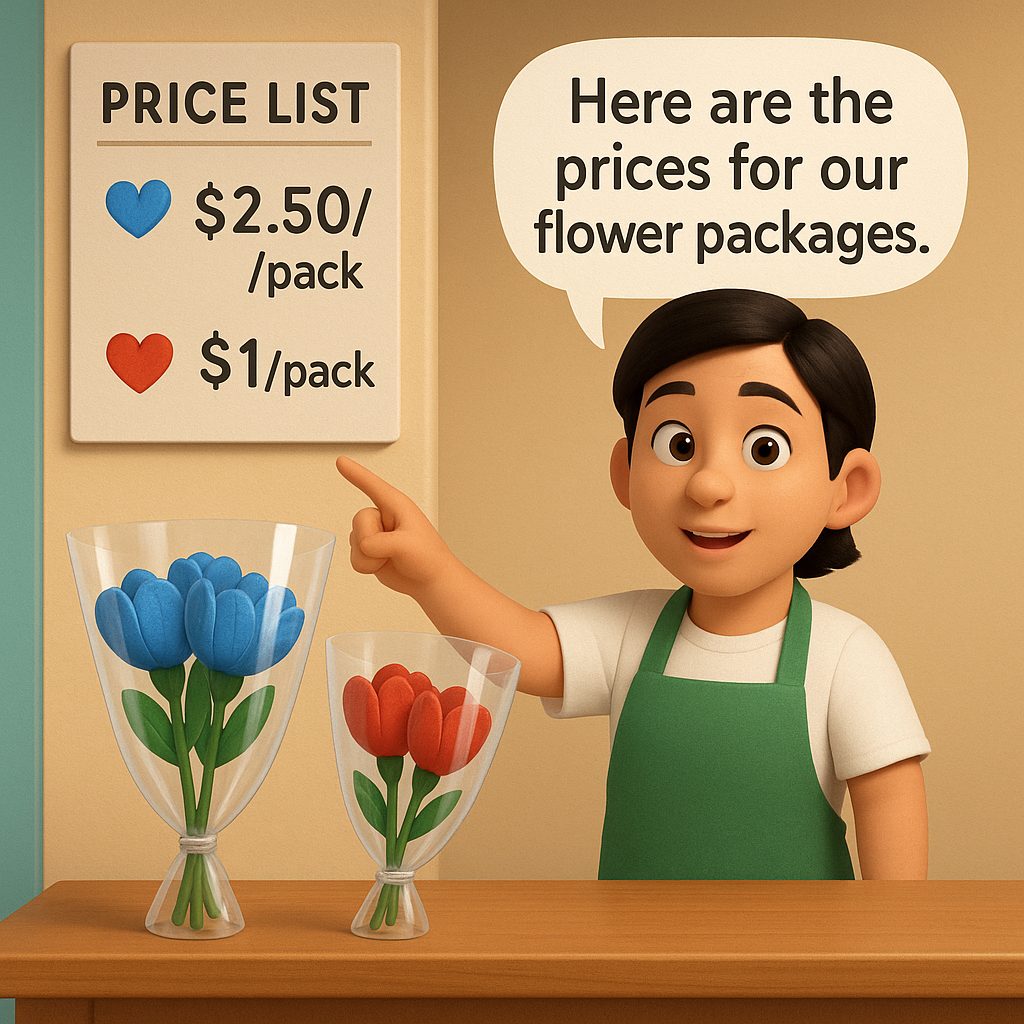}\\
  {\small Scene 1}
\end{minipage}\hfill
\begin{minipage}{0.44\linewidth}
  \centering
  \includegraphics[width=\linewidth,height=0.2\textheight,keepaspectratio]{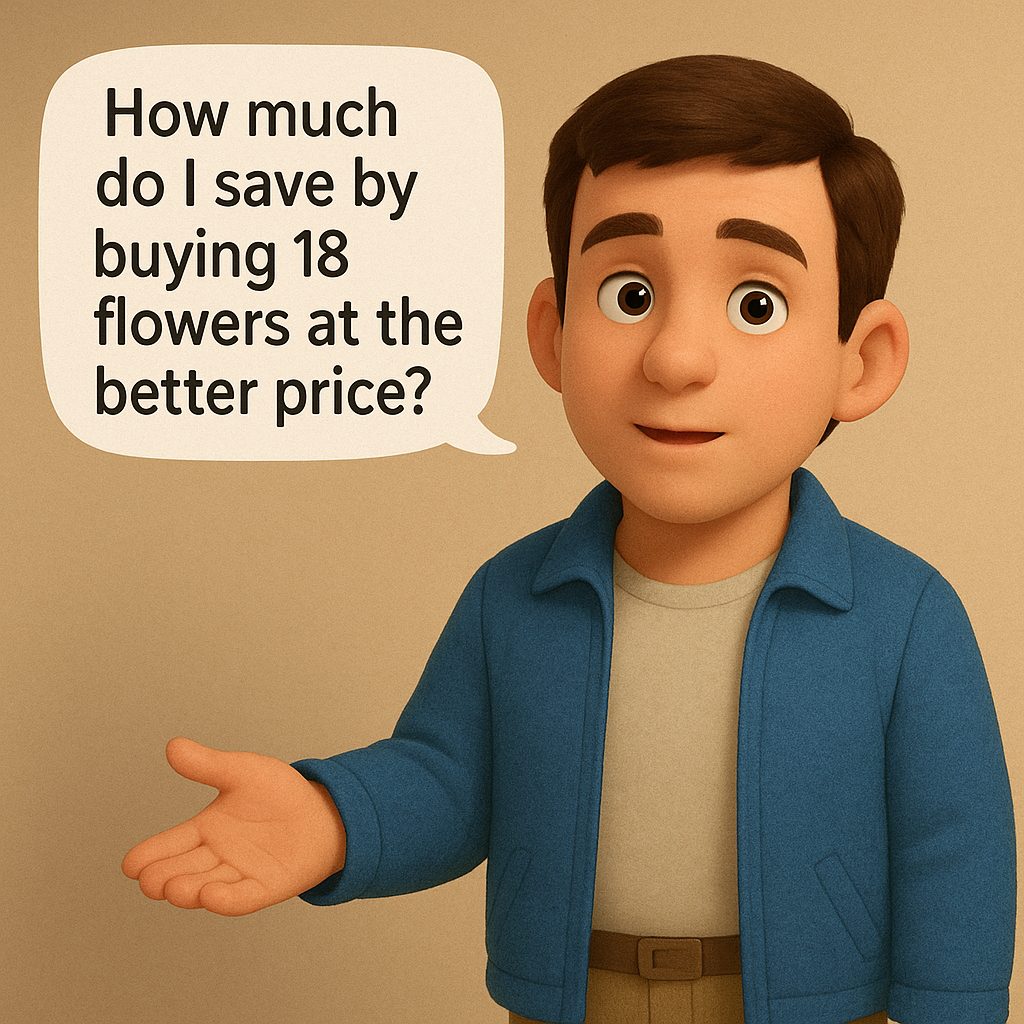}\\
  {\small Scene 2}
\end{minipage}
\end{subsecbox}

\begin{subsecbox}
\textbf{Ground Truth:} \textcolor{green!60!black}{\textbf{6}}

\vspace{0.5mm}
Better option is buying 9 packs of 2 flowers at \$1 each (total \$9) instead of 6 packs of 3 flowers at \$2.50 each (total \$15). 
Thus, saving = \$15 - \$9 = \textcolor{green!60!black}{\textbf{6}}.
\end{subsecbox}

\begin{modelbox}[Gemini-2.5-Pro (\wrongtag)]{BadRed}{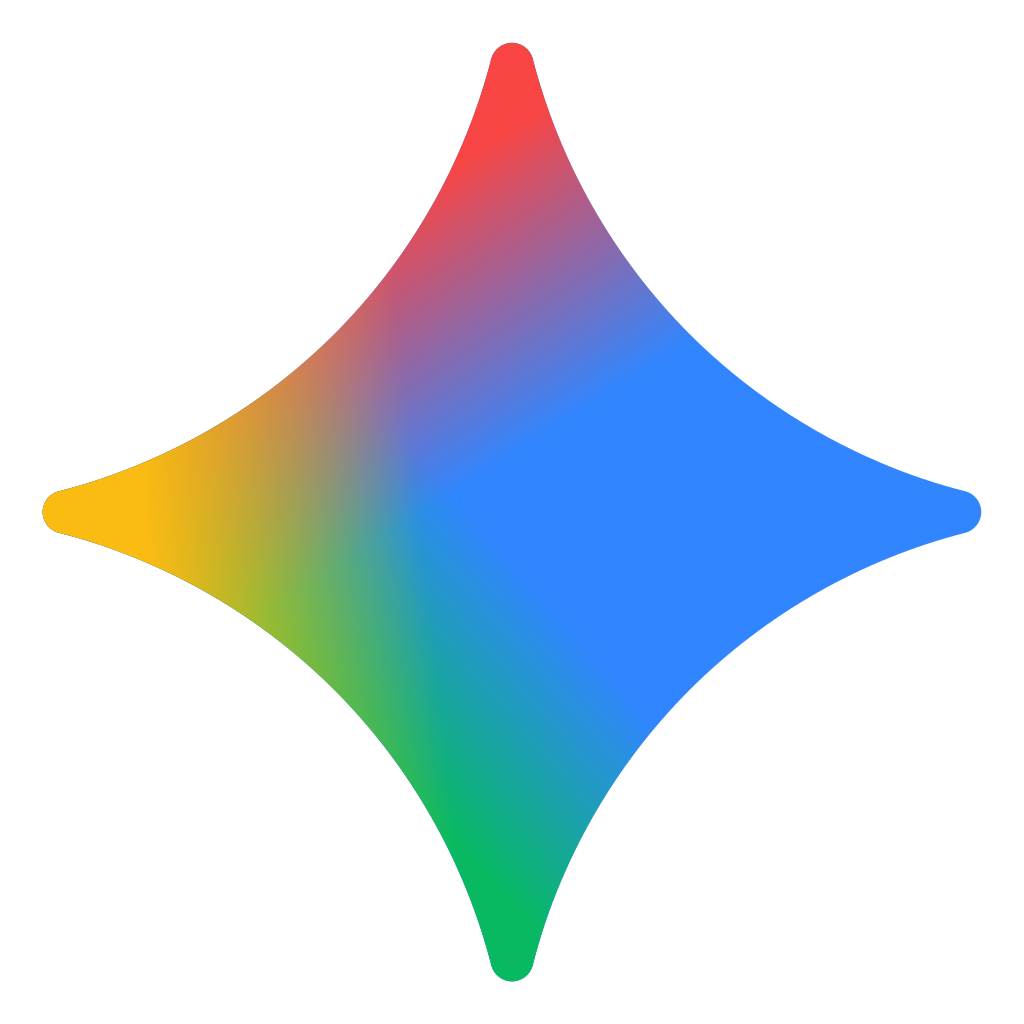}
\textbf{Answer:}
\begin{tcolorbox}[colback=LightGray!20, colframe=black!10, boxrule=0.3pt, arc=2pt, left=6pt, right=6pt, top=4pt, bottom=4pt]
Here is the step-by-step solution:

1.  **Determine the price and quantity per pack for each flower type.**  
    *   The price list shows blue flowers (blue heart) cost \$2.50 per pack and red flowers (red heart) cost \$1 per pack.  
    *   The images of the flower arrangements show that there are 3 flowers in each pack (3 blue flowers in the larger vase and 3 red flowers in the smaller vase).  

2.  **Identify the better price.**  
    *   The cost for a pack of 3 blue flowers is \$2.50.  
    *   The cost for a pack of 3 red flowers is \$1.  
    *   Comparing the two, the red flowers are cheaper and therefore the "better price".  

3.  **Calculate the cost of 18 flowers at the more expensive price.**  
    *   The more expensive flowers are the blue ones.  
    *   To buy 18 flowers, you would need 18 / 3 = 6 packs.  
    *   The total cost for 6 packs of blue flowers is 6 * \$2.50 = \$15.  

4.  **Calculate the cost of 18 flowers at the better price.**  
    *   The better priced flowers are the red ones.  
    *   To buy 18 flowers, you would need 18 / 3 = 6 packs.  
    *   The total cost for 6 packs of red flowers is 6 * \$1 = \$6.  

5.  **Calculate the total savings.**  
    *   The savings are the difference between the more expensive option and the cheaper option.  
    *   Savings = \$15 - \$6 = \$9.  

FINAL ANSWER: 9

\end{tcolorbox}

\end{modelbox}

\end{casestudy}
\end{figure*}

\begin{figure*}[t]
\centering

\begin{casestudy}[Error Case Study: EPE-CE]

\begin{modelbox}[GPT-4o (\wrongtag)]{BadRed}{figures/logos/openai.png}
\textbf{Answer:}
\begin{tcolorbox}[colback=LightGray!20, colframe=black!10, boxrule=0.3pt, arc=2pt, left=6pt, right=6pt, top=4pt, bottom=4pt]
To determine how much is saved by buying 18 flowers at the better price, we need to calculate the cost of buying 18 flowers at each price and find the difference.

1. **Price for blue flowers**: \$2.50 per pack  
2. **Price for red flowers**: \$1.00 per pack  

Assuming each pack contains the same number of flowers, buying at the cheaper price (red flowers) will save money.  

**Calculate the cost for 18 flowers:**  

- **Cost at \$2.50 per pack**:  
  - Total cost = 18 flowers × \$2.50 = \$45.00  

- **Cost at \$1.00 per pack**:  
  - Total cost = 18 flowers × \$1.00 = \$18.00  

**Calculate the savings**:  
- Savings = Cost at \$2.50 per pack - Cost at \$1.00 per pack  
- Savings = \$45.00 - \$18.00 = \$27.00  

FINAL ANSWER: 27
\end{tcolorbox}

\end{modelbox}

\begin{modelbox}[Qwen2.5-VL-72B (\wrongtag)]{BadRed}{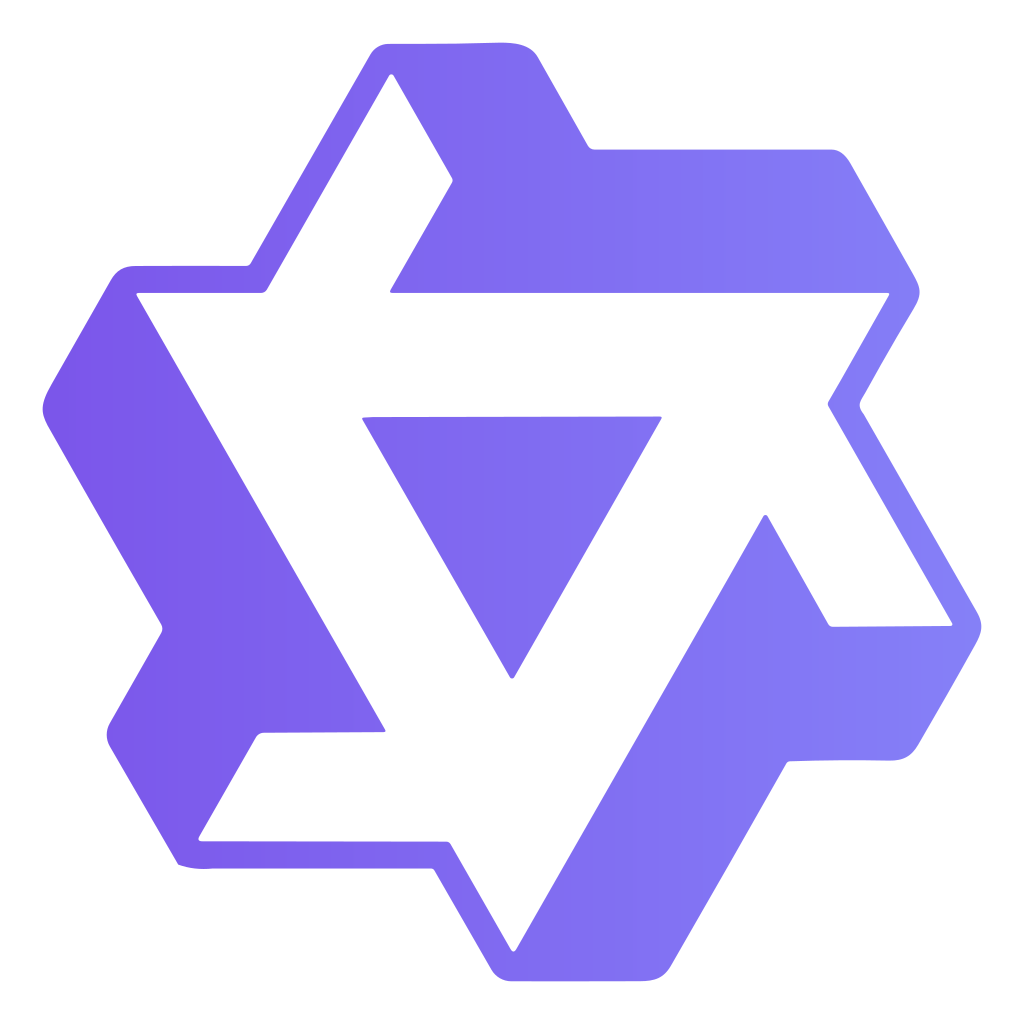}
\textbf{Answer:}
\begin{tcolorbox}[colback=LightGray!20, colframe=black!10, boxrule=0.3pt, arc=2pt, left=6pt, right=6pt, top=4pt, bottom=4pt]
To solve the problem, we need to determine how much money is saved by buying 18 flowers at the better price. Let's break it down step by step.

1. Identify the prices of the flower packages:  
   - Blue flowers: \$2.50 per pack  
   - Red flowers: \$1.00 per pack  

2. Determine which price is better:  
   - The red flowers are cheaper at \$1.00 per pack compared to \$2.50 per pack for blue flowers.  

3. Calculate the cost of buying 18 flowers at the higher price (blue flowers):  
   - Cost = 18 packs * \$2.50/pack = \$45.00  

4. Calculate the cost of buying 18 flowers at the lower price (red flowers):  
   - Cost = 18 packs * \$1.00/pack = \$18.00  

5. Determine the savings by subtracting the lower cost from the higher cost:  
   - Savings = \$45.00 - \$18.00 = \$27.00  

Thus, the amount saved by buying 18 flowers at the better price is:  

FINAL ANSWER: 27
\end{tcolorbox}

\textbf{Analysis:}  
This case is categorized as \textbf{EPE-CE} because the key failure occurs when models read explicit countable visual facts from the images. 
The problem requires identifying that one option contains packages of 3 flowers for \$2.50, while the better option contains packages of 2 flowers for \$1. 
Gemini-2.5-Pro incorrectly treats both options as packages of 3 flowers, leading to an incorrect comparison between the two package prices. 
GPT-4o and Qwen2.5-VL-72B further fail to recover the package-size information and instead treat the listed prices as if they were per-flower prices or directly multiply them by 18. 
Thus, the error is not mainly a symbolic calculation mistake: once the wrong package count is perceived or ignored, the subsequent arithmetic follows from an incorrect visual grounding. 
This matches \textbf{Counting Error (CE)} under \textbf{Explicit Perception Error (EPE)}, where the model fails to correctly read or bind explicitly visible counts to the corresponding mathematical entities.

\end{modelbox}

\end{casestudy}
\end{figure*}

\begin{figure*}[t]
\centering
\begin{casestudy}[Error Case Study: IVIE-CSIE]

\begin{subsecbox}
\textbf{Text Problem.}
\vspace{0.5mm}

Boris has 11 apples. Beck has 5 apples at first, but she ate 2 apples. If Boris gives Beck 3 apples, how many fewer apples does Beck have than Boris now?

\end{subsecbox}

\begin{subsecbox}
\textbf{GSM8K-V Images.}
\vspace{0.5mm}

\begin{minipage}{0.19\linewidth}
  \centering
  \includegraphics[width=\linewidth]{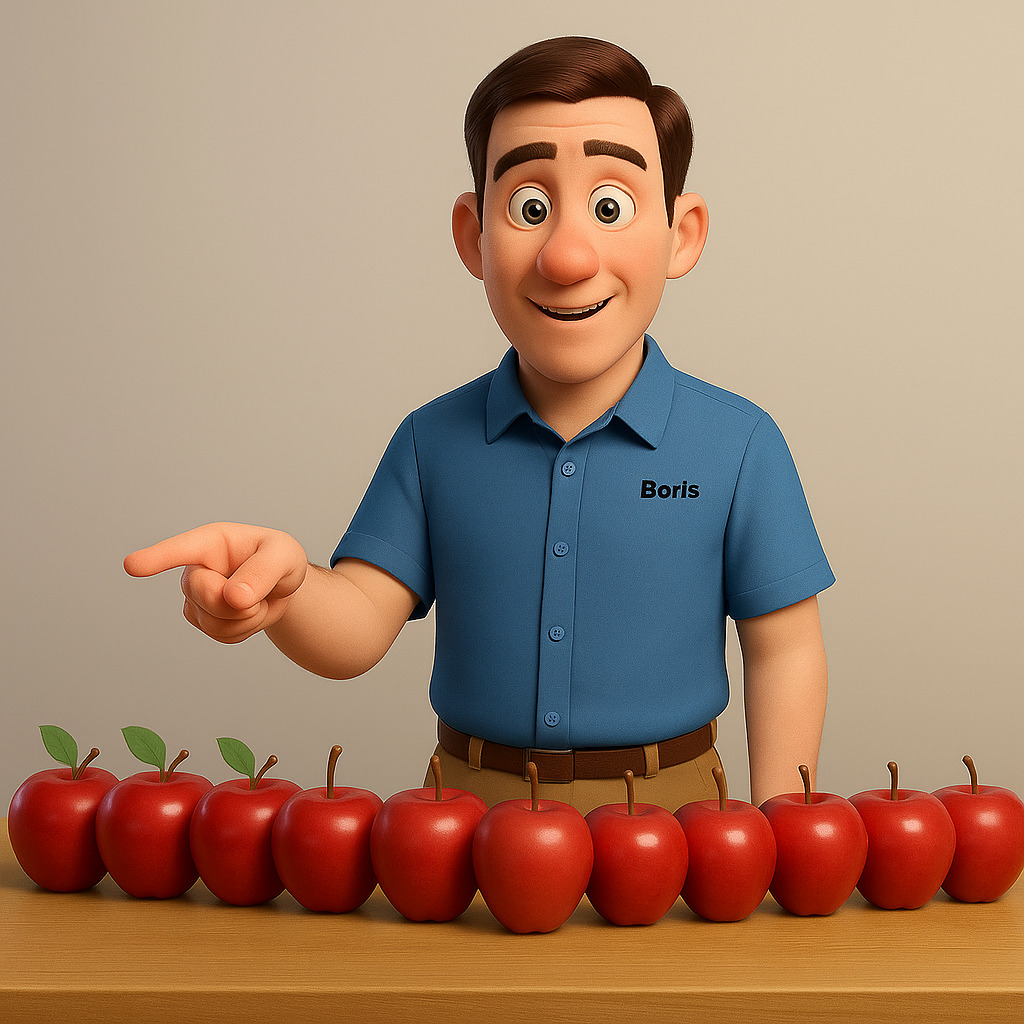}\\
  {\small Scene 1}
\end{minipage}\hfill
\begin{minipage}{0.19\linewidth}
  \centering
  \includegraphics[width=\linewidth]{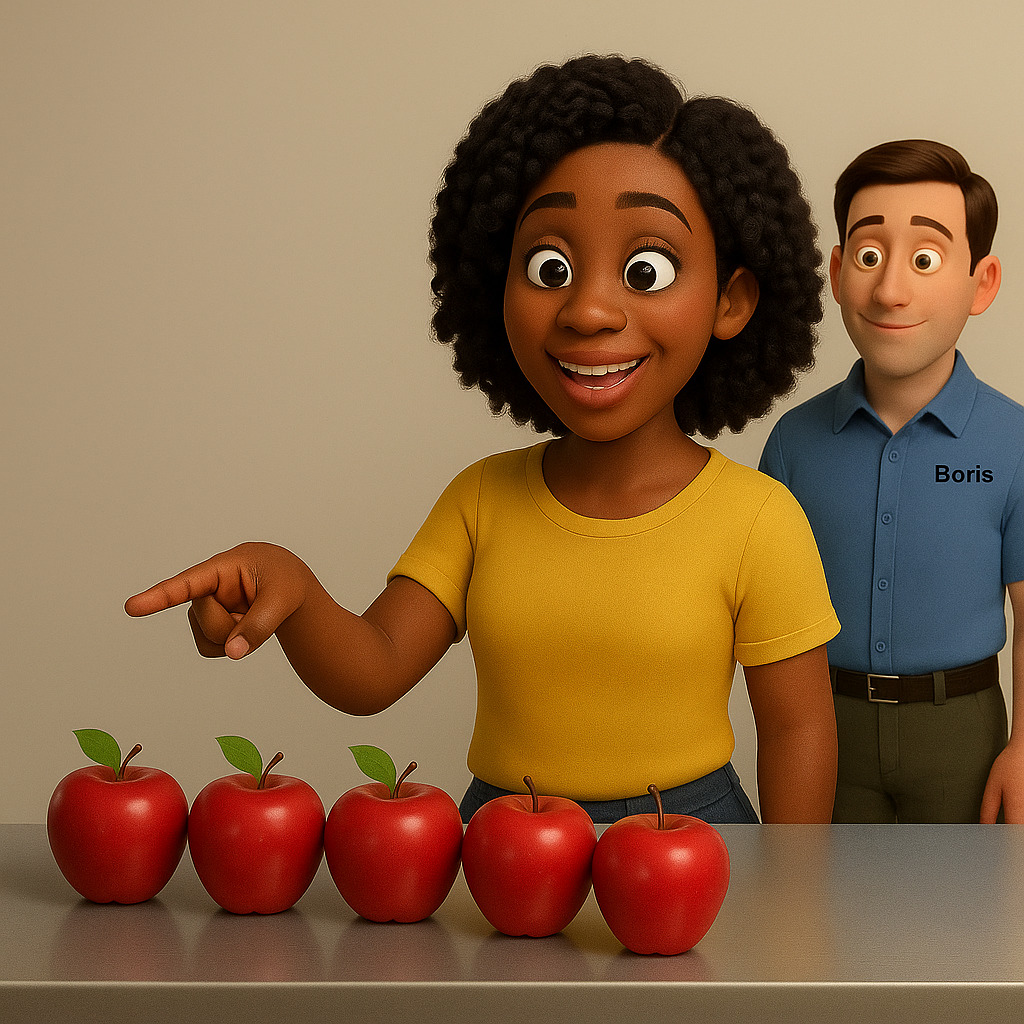}\\
  {\small Scene 2}
\end{minipage}\hfill
\begin{minipage}{0.19\linewidth}
  \centering
  \includegraphics[width=\linewidth]{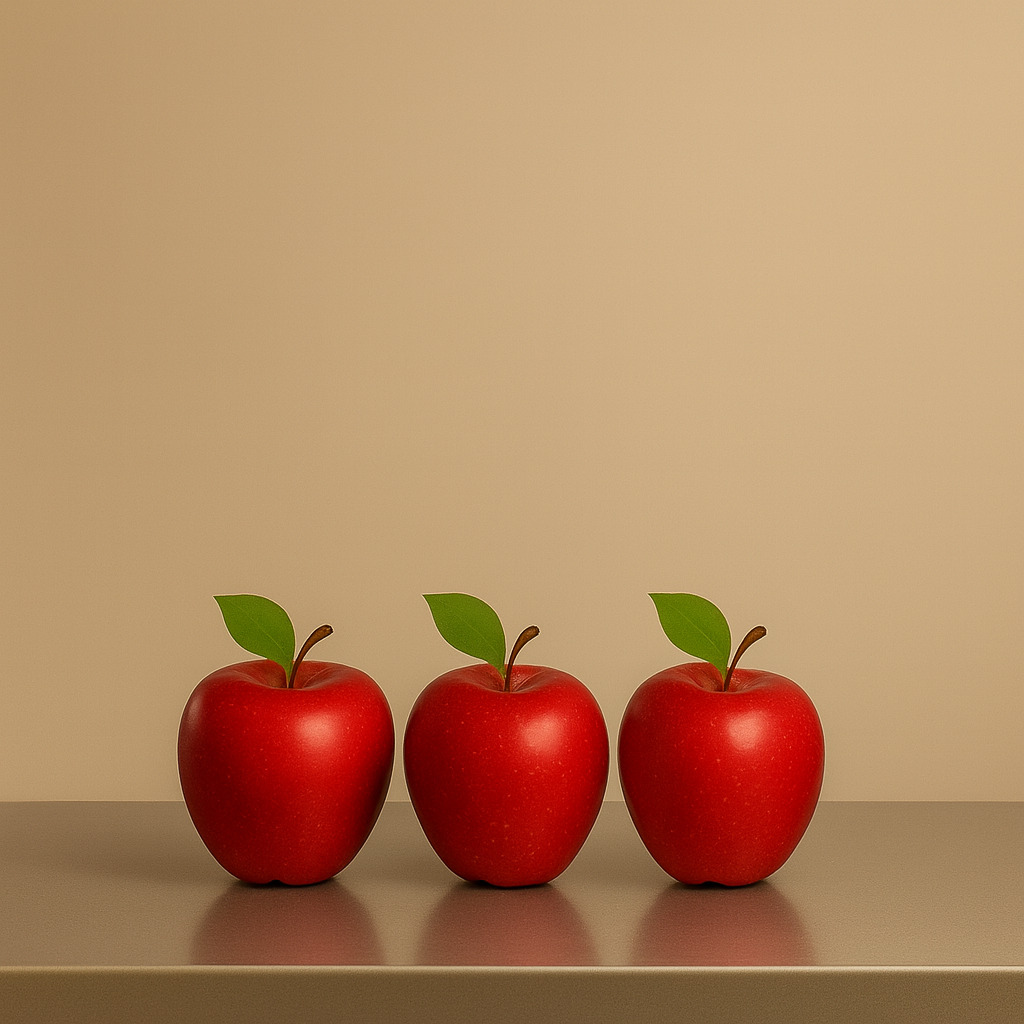}\\
  {\small Scene 3}
\end{minipage}\hfill
\begin{minipage}{0.19\linewidth}
  \centering
  \includegraphics[width=\linewidth]{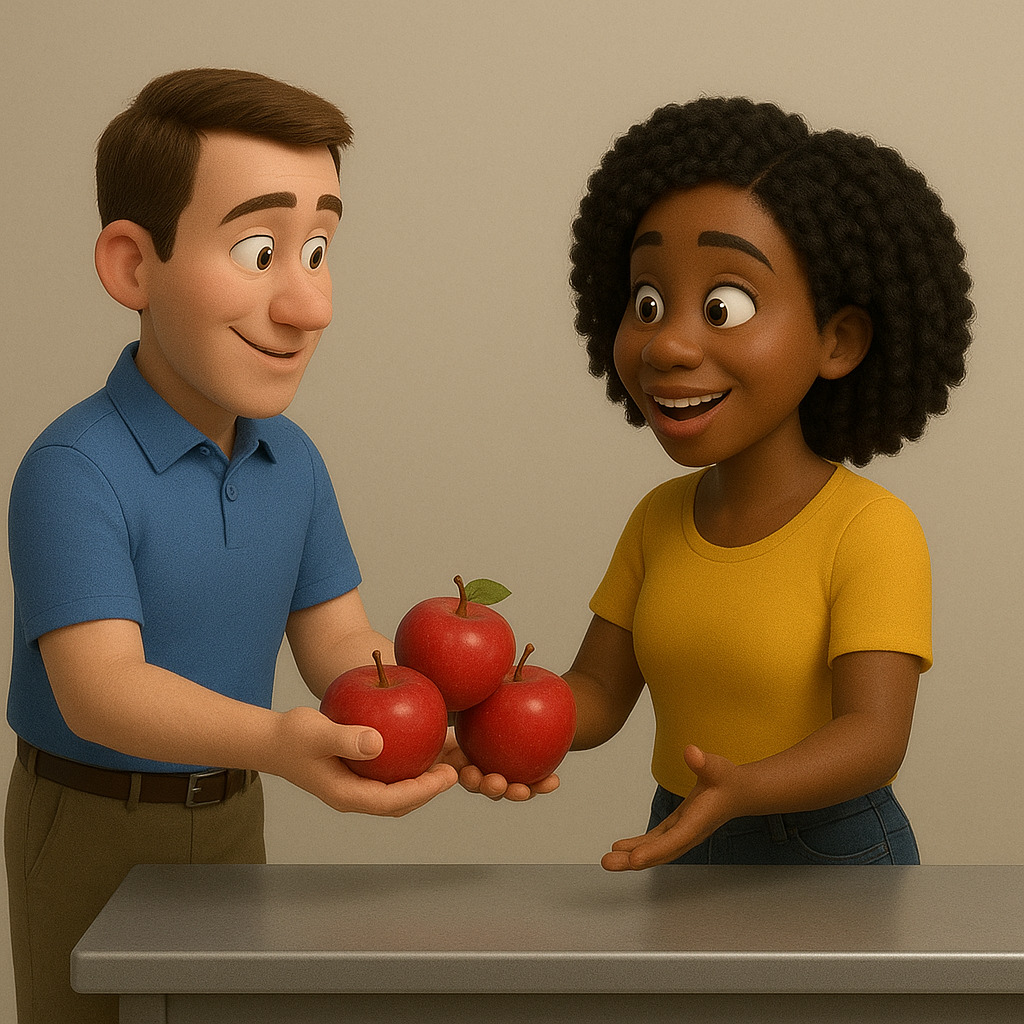}\\
  {\small Scene 4}
\end{minipage}\hfill
\begin{minipage}{0.19\linewidth}
  \centering
  \includegraphics[width=\linewidth]{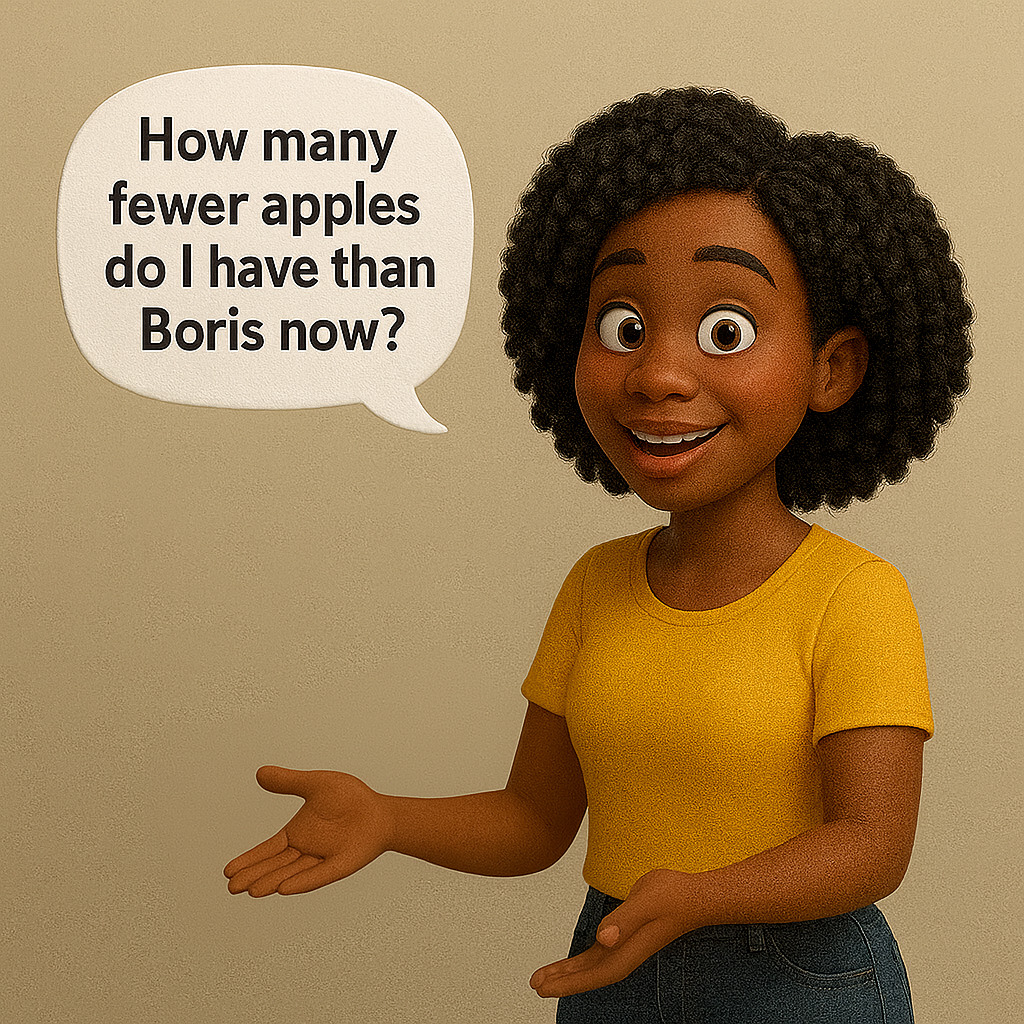}\\
  {\small Scene 5}
\end{minipage}

\end{subsecbox}

\begin{subsecbox}
\textbf{Ground Truth: } \textcolor{green!60!black}{\textbf{2}}
\vspace{0.5mm}

Boris has 11 apples. Beck has 5 apples at first, but she ate 2 apples. If Boris gives Beck 3 apples, how many fewer apples does Beck have than Boris now?

Beck's apples quantity: $5 - 2 + 3 = \langle\!\langle 5-2+3=6 \rangle\!\rangle 6$. 
Boris's apples quantity: $11 - 3 = \langle\!\langle 11-3=8 \rangle\!\rangle 8$. Thus, the difference is $8 - 6 = \langle\!\langle 8-6=2 \rangle\!\rangle 2$.

Answer: \textcolor{green!60!black}{\textbf{2}} apples.
\end{subsecbox}

\begin{modelbox}[Gemini-2.5-Pro (\wrongtag)]{BadRed}{figures/logos/gemini-color.png}
\textbf{Answer:}

\begin{tcolorbox}[colback=LightGray!20, colframe=black!10, boxrule=0.3pt, arc=2pt, left=6pt, right=6pt, top=4pt, bottom=4pt]
First, count Boris's apples in the first image: there are 11 apples.
Next, count the girl's apples in the second image: she has 5 apples.

In the middle images, Boris gives her 3 apples (we see 3 apples being passed from Boris to the girl).

After this:
* Boris: (11 - 3 = 8) apples
* Girl: (5 + 3 = 8) apples

The difference between their amounts is:
(8 - 8 = 0).
So she and Boris differ by 0 apple.

FINAL ANSWER: 0

\end{tcolorbox}

\end{modelbox}

\begin{modelbox}[GPT-4o (\wrongtag)]{BadRed}{figures/logos/openai.png}
\textbf{Answer:}

\begin{tcolorbox}[colback=LightGray!20, colframe=black!10, boxrule=0.3pt, arc=2pt, left=6pt, right=6pt, top=4pt, bottom=4pt]
Step 1 — Count Boris’s apples at the start: in the first picture Boris has 11 apples.

Step 2 — Count the girl's apples at the start: in the second picture she has 5 apples. 

Step 3 — See the transfer: Boris gives the girl 3 apples (fourth picture). 

Step 4 — Compute new totals: Boris now has \(11-3=8\) apples; the girl now has \(5+3=8\) apples. 

Step 5 — Compare: the girl has \(8-8=0\) fewer apples than Boris now. FINAL ANSWER: 0

\end{tcolorbox}

\end{modelbox}

\end{casestudy}
\end{figure*}

\begin{figure*}[t]
\centering
\begin{casestudy}[Error Case Study: IVIE-CSIE]

\begin{modelbox}[Qwen2.5-VL-72B (\wrongtag)]{BadRed}{figures/logos/qwen-color.png}
\textbf{Answer:}
\begin{tcolorbox}[colback=LightGray!20, colframe=black!10, boxrule=0.3pt, arc=2pt, left=6pt, right=6pt, top=4pt, bottom=4pt]
Let's solve this step by step.

1.  **Initial Count:**
    - In the first image, Boris is pointing to a row of apples. Let's count them: there are 10 apples in front of him.
    - In the second image, the woman (let's call her Maya) is pointing to her apples. There are 5 apples in front of her.

2.  **The Transaction:**
    - The fourth image shows Boris giving Maya 3 apples. He is holding them out to her, and she is reaching for them.

3.  **Final Counts:**
    - After receiving 3 apples, Maya's total becomes her initial 5 plus the 3 she received: 5 + 3 = 8 apples.
    - After giving away 3 apples, Boris's total becomes his initial 10 minus the 3 he gave away: 10 - 3 = 7 apples.

4.  **Answering the Question:**
    - The final image asks, "How many fewer apples do I have than Boris now?"
    - This question is asked by Maya. So, we need to find the difference between Boris's current number of apples and Maya's current number of apples.
    - Boris has 7 apples.
    - Maya has 8 apples.
    - Since Maya has *more* apples than Boris (8 > 7), she does not have *fewer* apples. The question is asking for a positive number representing how many fewer she has, which would be zero in this case, as she actually has more.

    However, let's re-read the question carefully: "How many fewer apples do I have than Boris now?". If we interpret this literally, it's asking for the value of (Boris's apples - Maya's apples). That would be 7 - 8 = -1. But since the question asks for "how many fewer," and the answer should be a whole number (as per the instructions), it implies a positive difference or zero.

    Given the context, it seems the question might be phrased incorrectly, or it's testing if we notice that Maya now has more. But logically, if she has more, she has 0 fewer apples.

    Alternatively, perhaps the question meant to ask "How many more apples do I have than Boris now?", which would be 8 - 7 = 1.

    But sticking strictly to the text: "How many fewer apples do I have than Boris now?" Since Maya has 8 and Boris has 7, Maya has 1 *more*, not fewer. Therefore, the number of apples she has *fewer* than Boris is 0.

FINAL ANSWER: 0
\end{tcolorbox}

\textbf{Analysis:}  
This case is categorized as \textbf{IVIE-CSIE} because the models fail to recover an implicit semantic event from the visual context. 
The images explicitly show the initial apple counts and the later transfer of 3 apples from Boris to Beck, but they also imply an intermediate state change: Beck eats 2 apples before receiving the 3 apples. 
Gemini-2.5-Pro and GPT-4o correctly read the visible counts for Boris and Beck and correctly identify the transfer action, but they omit the eating event, so they compute Beck's final apples as $5+3$ instead of $5-2+3$. 
Qwen2.5-VL-72B additionally miscounts Boris's apples, but the shared failure across models is the missed contextual meaning of the eating scene. 
Therefore, the main error is not simple counting or arithmetic, but a failure in \textbf{Contextual Semantic Interpretation Error (CSIE)} under \textbf{Implicit Visual Inference Error (IVIE)}: the models do not infer the intended semantic change encoded by the scene sequence.

\end{modelbox}

\end{casestudy}
\end{figure*}

\begin{figure*}[t]
\centering
\begin{casestudy}[Error Case Study: IVIE-SUE]

\begin{subsecbox}
\textbf{Text Problem.}
\vspace{0.5mm}

The rug is $5$ feet wider than the chair. The couch is $2$ feet longer than twice the width of the rug. If the chair is $3$ feet wide. How many feet long is the couch?

\end{subsecbox}

\begin{subsecbox}
\textbf{GSM8K-V Images.}
\vspace{0.5mm}

\begin{minipage}{0.3\linewidth}
  \centering
  \includegraphics[width=\linewidth]{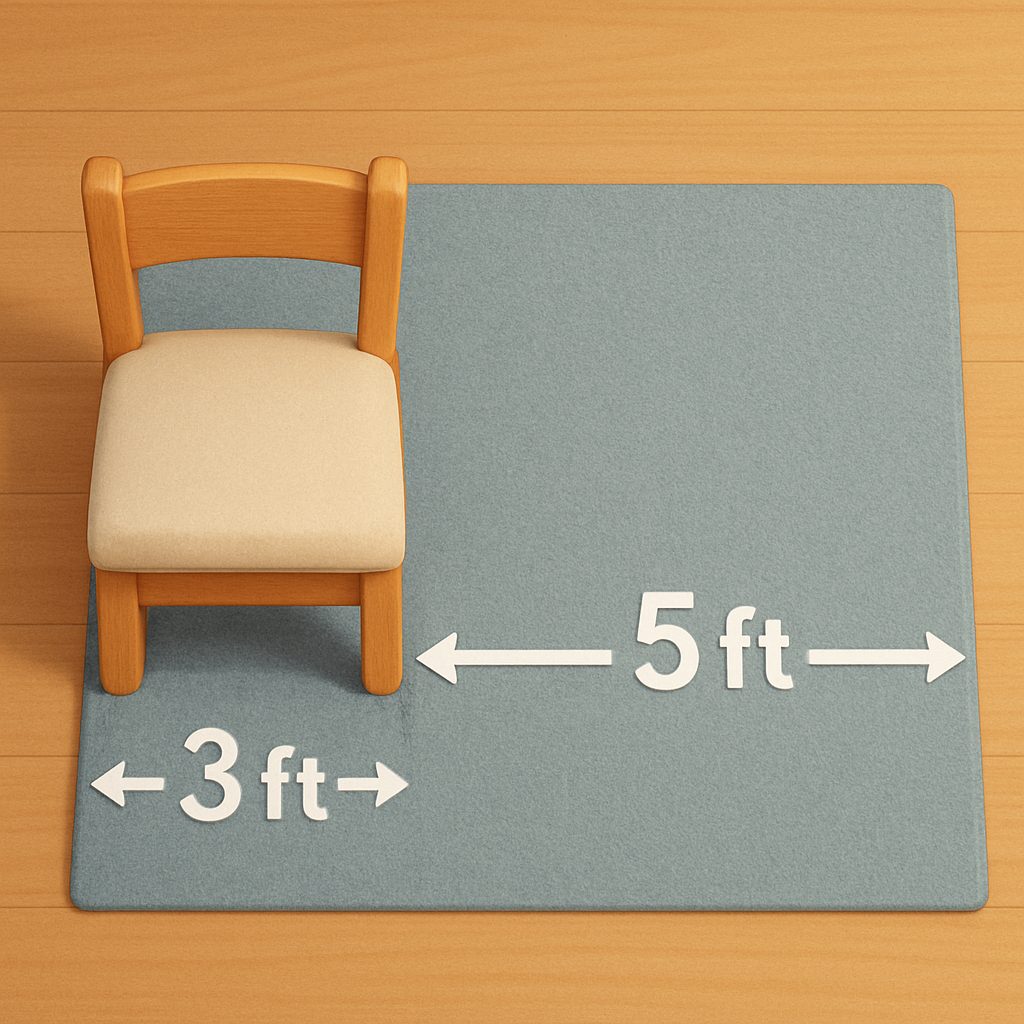}\\
  {\small Scene 1}
\end{minipage}\hfill
\begin{minipage}{0.3\linewidth}
  \centering
  \includegraphics[width=\linewidth]{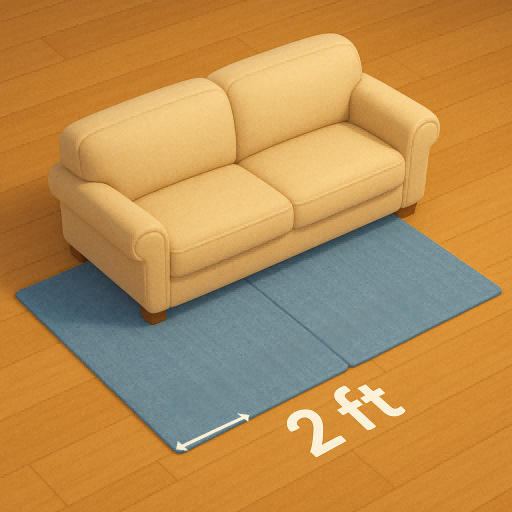}\\
  {\small Scene 2}
\end{minipage}\hfill
\begin{minipage}{0.3\linewidth}
  \centering
  \includegraphics[width=\linewidth]{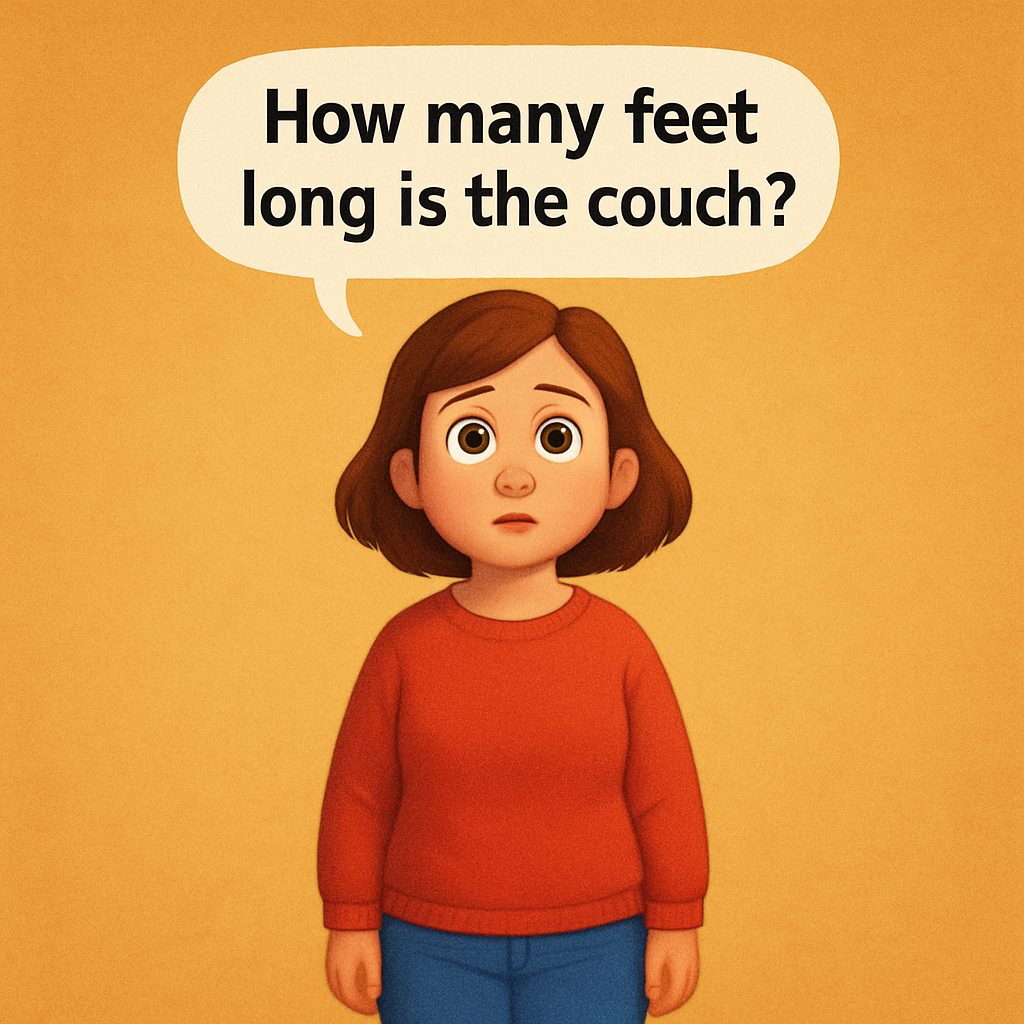}\\
  {\small Scene 3}
\end{minipage}

\end{subsecbox}

\begin{subsecbox}
\textbf{Ground Truth: }  \textcolor{green!60!black}{\textbf{18}}
\vspace{0.5mm}

Chair $= \langle\!\langle 3=3 \rangle\!\rangle 3$ feet; \#\#\# \hspace{0.5mm}
Rug $= 3 + 5 = \langle\!\langle 3+5=8 \rangle\!\rangle 8$ feet; \#\#\# \hspace{0.5mm}
Couch $= 2 \times 8 + 2 = \langle\!\langle 2 \times 8 + 2 = 18 \rangle\!\rangle 18$ feet; \#\#\# \hspace{0.5mm}
The couch is $\langle\!\langle 18=18 \rangle\!\rangle \textcolor{green!60!black}{\mathbf{18}}$ feet long.

\end{subsecbox}

\begin{modelbox}[Gemini-2.5-Pro (\wrongtag)]{BadRed}{figures/logos/gemini-color.png}
\textbf{Answer:}

\begin{tcolorbox}[colback=LightGray!20, colframe=black!10, boxrule=0.3pt, arc=2pt, left=6pt, right=6pt, top=4pt, bottom=4pt]

\begin{enumerate}[leftmargin=*,label=\textbf{Step \arabic*:}]
    \item \textbf{Determine the length of one rug.}
    \begin{itemize}
        \item The first image shows a single blue rug. A chair is placed on the left side of it.
        \item The image shows the rug is composed of two sections: The length covered by the chair, which is $3$ feet. The remaining length of the rug, which is $5$ feet.
        \item Total length of one rug: $3 + 5 = 8$ feet.
    \end{itemize}

    \item \textbf{Determine the total length of the two rugs combined.}
    \begin{itemize}
        \item The second image shows two identical rugs placed side-by-side along their length.
        \item Total length: $2 \times 8 = 16$ feet.
    \end{itemize}

    \item \textbf{Calculate the length of the couch.}
    \begin{itemize}
        \item \textcolor{red!60!black} {There is a space of $2$ feet on the left and $2$ feet on the right. }
        \item \textcolor{red!60!black} {Couch length $= 16 - 2 - 2 = 12$ feet.}
    \end{itemize}
\end{enumerate}

\medskip
\textbf{Final Answer: } \textcolor{red!60!black}{\textbf{12}}

\end{tcolorbox}

\end{modelbox}

\end{casestudy}
\end{figure*}

\begin{figure*}[t]
\centering
\begin{casestudy}[Error Case Study: IVIE-SUE]

\begin{modelbox}[GPT-4o (\wrongtag)]{BadRed}{figures/logos/openai.png}

\begin{tcolorbox}[colback=LightGray!20, colframe=black!10, boxrule=0.3pt, arc=2pt, 
left=6pt, right=6pt, top=2pt, bottom=2pt]

\begin{enumerate}[
  leftmargin=3.5em,
  label=\textbf{Step \arabic*:},
  itemsep=0pt,
  topsep=1pt,
  parsep=0pt,
  partopsep=0pt
]
    \item \textbf{Analyze the first image.}
    \begin{itemize}[
      leftmargin=-0.5em,
      itemsep=0pt,
      topsep=0pt,
      parsep=0pt,
      partopsep=0pt
    ]
        \item The chair width is labeled $3$ feet.
        
        \item \textcolor{red!60!black} {The visible rug width (horizontal, beside the chair) is labeled $5$ feet.}
        
        \item \textcolor{red!60!black} { Therefore, the length of the rug is also $5$ feet.}
    \end{itemize}

    \item \textbf{Analyze the second image.}
    \begin{itemize}[
      leftmargin=-0.5em,
      itemsep=0pt,
      topsep=0pt,
      parsep=0pt,
      partopsep=0pt
    ]
        \item A couch is on two rugs, \textcolor{red!60!black} { each measuring $2$ feet wide.}
        \item Total width under the couch: \textcolor{red!60!black} {$2 + 2 = 4$ feet.}
    \end{itemize}

    \item \textbf{Analyze the third image.}
    \begin{itemize}[
      leftmargin=-0.5em,
      itemsep=0pt,
      topsep=0pt,
      parsep=0pt,
      partopsep=0pt
    ]
        \item The question explicitly asks for the length of the couch.
    \end{itemize}
\end{enumerate}

\textcolor{red!60!black} {Since the rugs are placed side by side under the couch, the model concludes that the length of the couch equals the total width of the rugs.  }

\medskip
\textbf{Final Answer: } \textcolor{red!60!black}{\textbf{4}}

\end{tcolorbox}
\end{modelbox}

\begin{modelbox}[Qwen2.5-VL-72b (\wrongtag)]{BadRed}{figures/logos/qwen-color.png}
\textbf{Answer:}

\begin{tcolorbox}[colback=LightGray!20, colframe=black!10, boxrule=0.3pt, arc=2pt, 
left=6pt, right=6pt, top=4pt, bottom=4pt]

\begin{enumerate}[
  leftmargin=3.5em,
  label=\textbf{Step \arabic*:},
  itemsep=0pt,
  topsep=1pt,
  parsep=0pt,
  partopsep=0pt
]

\item \textbf{Determine.}
    \begin{itemize}[
      leftmargin=-0.5em,
      itemsep=0pt,
      topsep=0pt,
      parsep=0pt,
      partopsep=0pt
    ]
       \item To determine the length of the couch, we need to analyze the given images and the information provided.
    \end{itemize}

    \item \textbf{Images Content Analyze}
        \begin{itemize}[
      leftmargin=-0.5em,
      itemsep=0pt,
      topsep=0pt,
      parsep=0pt,
      partopsep=0pt
    ]
        \item The first image shows a chair with dimensions $3$ ft on \textcolor{red!60!black} {a rug with dimensions by $5$ ft.}
        \item The second image shows a couch on \textcolor{red!60!black} {a rug with a dimension of $2$ ft along one side.}
        \item The third image asks how many feet long the couch is.
    \end{itemize}

    \item \textbf{Calculate}
    \begin{itemize}[
      leftmargin=-0.5em,
      itemsep=0pt,
      topsep=0pt,
      parsep=0pt,
      partopsep=0pt
    ]
        \item 
        From the second image, we can see that the couch is placed on a rug, and the rug's dimension along the \textcolor{red!60!black} {length of the couch is $2$ ft. }
        \item \textcolor{red!60!black} {Since the couch fits within this dimension, it implies that the length of the couch is equal to or less than $2$ ft. Given that the couch appears to fully occupy this dimension, we can conclude that the length of the couch is $2$ ft.}
    \end{itemize}
\end{enumerate}

\medskip
\textbf{Final Answer: } \textcolor{red!60!black}{\textbf{2}}

\end{tcolorbox}

\textbf{Analysis:}  
This case is categorized as \textbf{IVIE-SUE} because the models fail to understand the scope of the visual relation encoded by the spatial layout. 
The first scene establishes that the rug is 5 feet wider than the 3-foot chair, so the rug width should be inferred as $3+5=8$. 
The second scene then indicates that the couch is 2 feet longer than twice the rug width, requiring the model to apply the additional 2 feet to the whole expression $2 \times 8$, rather than interpreting the visible 2-foot mark as a local gap, margin, or direct couch length. 
Gemini-2.5-Pro correctly derives the rug width and the doubled rug length, but incorrectly treats the 2-foot relation as space to subtract on both sides. 
GPT-4o and Qwen2.5-VL-72B similarly restrict the meaning of the visible measurement to a local spatial segment instead of the intended global relation defining the couch length. 
This is therefore a \textbf{Scope Understanding Error (SUE)} under \textbf{Implicit Visual Inference Error (IVIE)}: the models perceive some local visual measurements, but fail to infer the correct semantic scope over which the measurement relation should apply.

\end{modelbox}

\end{casestudy}
\end{figure*}

\clearpage

\onecolumn

\section{Category Case Study}
\label{category_case_study}

\begin{center}
\centering
\begin{casestudy}[Case Study 1: Measurement (Length / Area / Volume)]

\begin{subsecbox}
\textbf{Text Problem.}
\vspace{0.5mm}

The rug is $5$ feet wider than the chair. The couch is $2$ feet longer than twice the width of the rug. If the chair is $3$ feet wide. How many feet long is the couch?

\end{subsecbox}

\begin{subsecbox}
\textbf{GSM8K-V Images.}
\vspace{0.5mm}

\begin{minipage}{0.3\linewidth}
  \centering
  \includegraphics[width=\linewidth]{figures/cases/504_1.jpg}\\
  {\small Scene 1}
\end{minipage}\hfill
\begin{minipage}{0.3\linewidth}
  \centering
  \includegraphics[width=\linewidth]{figures/cases/504_2.png}\\
  {\small Scene 2}
\end{minipage}\hfill
\begin{minipage}{0.3\linewidth}
  \centering
  \includegraphics[width=\linewidth]{figures/cases/504_3.jpg}\\
  {\small Scene 3}
\end{minipage}

\end{subsecbox}

\begin{subsecbox}
\textbf{Ground Truth: }  \textcolor{green!60!black}{\textbf{18}}
\vspace{0.3mm}

Chair $= \langle\!\langle 3=3 \rangle\!\rangle 3$ feet; \#\#\# \hspace{0.3mm}
Rug $= 3 + 5 = \langle\!\langle 3+5=8 \rangle\!\rangle 8$ feet; \#\#\# \hspace{0.3mm}
Couch $= 2 \times 8 + 2 = \langle\!\langle 2 \times 8 + 2 = 18 \rangle\!\rangle 18$ feet; \#\#\# \hspace{0.5mm}
The couch is $\langle\!\langle 18=18 \rangle\!\rangle \textcolor{green!60!black}{\mathbf{18}}$ feet long.

\end{subsecbox}

\begin{modelbox}[Gemini-2.5-Pro (\wrongtag)]{BadRed}{figures/logos/gemini-color.png}
\textbf{Answer:}

\begin{tcolorbox}[colback=LightGray!20, colframe=black!10, boxrule=0.3pt, arc=2pt, left=6pt, right=6pt, top=4pt, bottom=4pt]

\begin{enumerate}[leftmargin=*,label=\textbf{Step \arabic*:}]
    \item \textbf{Determine the length of one rug.}
    \begin{itemize}
        \item The first image shows a single blue rug. A chair is placed on the left side of it.
        \item The image shows the rug is composed of two sections: The length covered by the chair, which is $3$ feet. The remaining length of the rug, which is $5$ feet.
        \item Total length of one rug: $3 + 5 = 8$ feet.
    \end{itemize}

    \item \textbf{Determine the total length of the two rugs combined.}
    \begin{itemize}
        \item The second image shows two identical rugs placed side-by-side along their length.
        \item Total length: $2 \times 8 = 16$ feet.
    \end{itemize}

    \item \textbf{Calculate the length of the couch.}
    \begin{itemize}
        \item \textcolor{red!60!black} {There is a space of $2$ feet on the left and $2$ feet on the right. }
        \item \textcolor{red!60!black} {Couch length $= 16 - 2 - 2 = 12$ feet.}
    \end{itemize}
\end{enumerate}

\medskip
\textbf{Final Answer: } \textcolor{red!60!black}{\textbf{12}}

\end{tcolorbox}

\end{modelbox}

\end{casestudy}

\end{center}

\begin{figure*}[t]

\begin{casestudy}[Case Study 1: Measurement (Length / Area / Volume)]

\begin{modelbox}[GPT-4o (\wrongtag)]{BadRed}{figures/logos/openai.png}
\textbf{Answer:}

\begin{tcolorbox}[colback=LightGray!20, colframe=black!10, boxrule=0.3pt, arc=2pt, 
left=6pt, right=6pt, top=4pt, bottom=4pt]

\begin{enumerate}[leftmargin=*,label=\textbf{Step \arabic*:}]
    \item \textbf{Analyze the first image.}
    \begin{itemize}
        \item The chair width is labeled $3$ feet.
        
        \item \textcolor{red!60!black} {The visible rug width (horizontal, beside the chair) is labeled $5$ feet.}
        
        \item \textcolor{red!60!black} { Therefore, the length of the rug is also $5$ feet.}
    \end{itemize}

    \item \textbf{Analyze the second image.}
    \begin{itemize}
        \item A couch is on two rugs, \textcolor{red!60!black} { each measuring $2$ feet wide.}
        \item Total width under the couch: \textcolor{red!60!black} {$2 + 2 = 4$ feet.}
    \end{itemize}

    \item \textbf{Analyze the third image.}
    \begin{itemize}
        \item The question explicitly asks for the length of the couch.
    \end{itemize}
\end{enumerate}

\textcolor{red!60!black} {Since the rugs are placed side by side under the couch, the model concludes that the length of the couch equals the total width of the rugs.  }

\medskip
\textbf{Final Answer: } \textcolor{red!60!black}{\textbf{4}}

\end{tcolorbox}
\end{modelbox}

\vspace{6pt}

\begin{modelbox}[Qwen2.5-VL-72b (\wrongtag)]{BadRed}{figures/logos/qwen-color.png}
\textbf{Answer:}

\begin{tcolorbox}[colback=LightGray!20, colframe=black!10, boxrule=0.3pt, arc=2pt, 
left=6pt, right=6pt, top=4pt, bottom=4pt]

\begin{enumerate}[leftmargin=*,label=\textbf{Step \arabic*:}]

    \item \textbf{Determine.}
    \begin{itemize}
       \item To determine the length of the couch, we need to analyze the given images and the information provided.
    \end{itemize}

    \item \textbf{Images Content Analyze}
    \begin{itemize}
        \item The first image shows a chair with dimensions $3$ ft on \textcolor{red!60!black} {a rug with dimensions by $5$ ft.}
        \item The second image shows a couch on \textcolor{red!60!black} {a rug with a dimension of $2$ ft along one side.}
        \item The third image asks how many feet long the couch is.
    \end{itemize}

    \item \textbf{Calculate}
    \begin{itemize}
        \item 
        From the second image, we can see that the couch is placed on a rug, and the rug's dimension along the \textcolor{red!60!black} {length of the couch is $2$ ft. }
        \item \textcolor{red!60!black} {Since the couch fits within this dimension, it implies that the length of the couch is equal to or less than $2$ ft. Given that the couch appears to fully occupy this dimension, we can conclude that the length of the couch is $2$ ft.}
    \end{itemize}
\end{enumerate}

\medskip
\textbf{Final Answer: } \textcolor{red!60!black}{\textbf{2}}

\end{tcolorbox}

\end{modelbox}

\end{casestudy}
\end{figure*}

\begin{figure*}[t]
\centering
\begin{casestudy}[Case Study 2: Signboard \& Icon (Price)]

\begin{subsecbox}
\textbf{Text Problem.}
\vspace{0.5mm}

Tyler wants to buy a dictionary that costs \$18, a dinosaur book that costs \$13, and a children's cookbook that costs \$8. He has saved \$14 from his allowance. If Tyler earns \$5 per hour by washing, how many hours does he have to work to afford his books?

\end{subsecbox}

\begin{subsecbox}
\textbf{GSM8K-V Images.}
\vspace{0.5mm}

\begin{minipage}{0.18\linewidth}
  \centering
  \includegraphics[width=\linewidth]{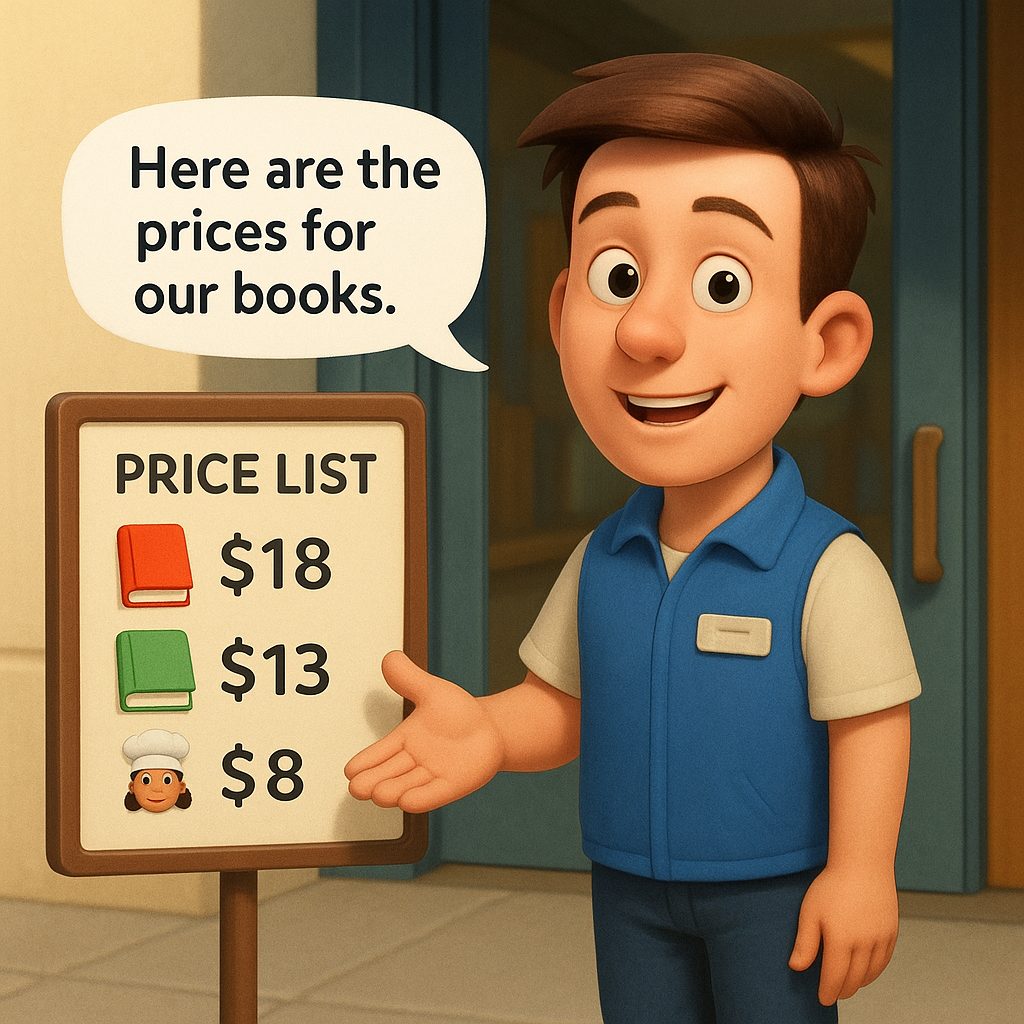}\\
  {\small Scene 1}
\end{minipage}\hfill
\begin{minipage}{0.18\linewidth}
  \centering
  \includegraphics[width=\linewidth]{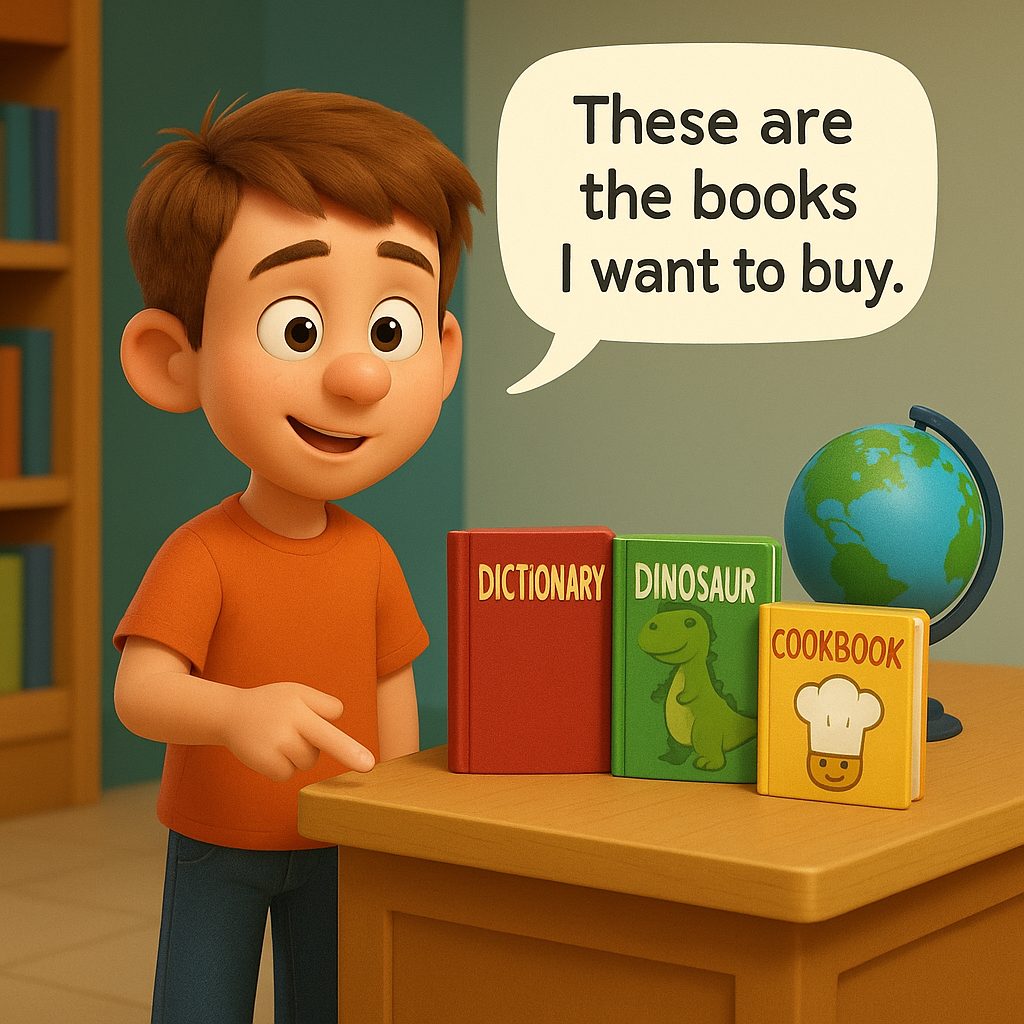}\\
  {\small Scene 2}
\end{minipage}\hfill
\begin{minipage}{0.18\linewidth}
  \centering
  \includegraphics[width=\linewidth]{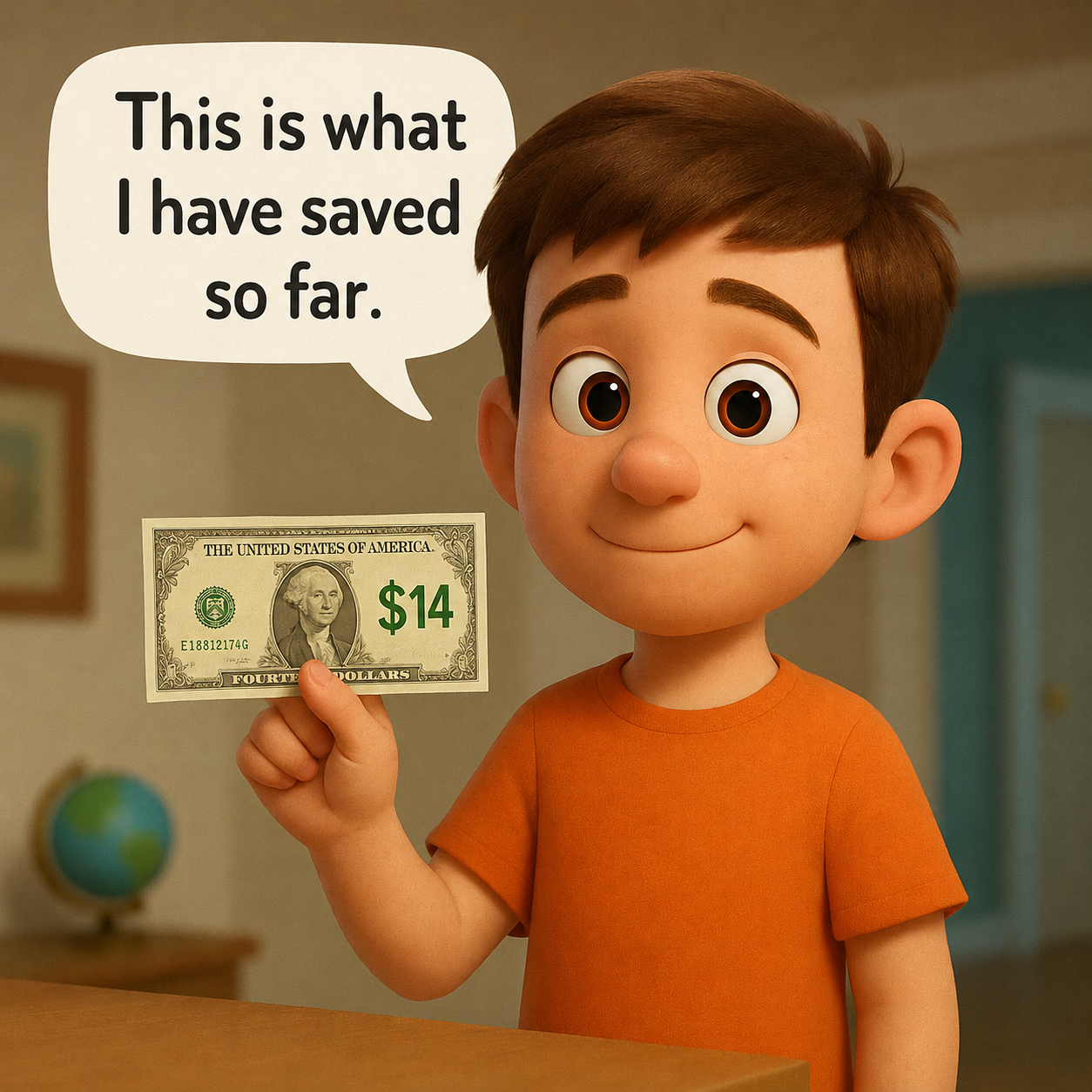}\\
  {\small Scene 3}
\end{minipage}\hfill
\begin{minipage}{0.18\linewidth}
  \centering
  \includegraphics[width=\linewidth]{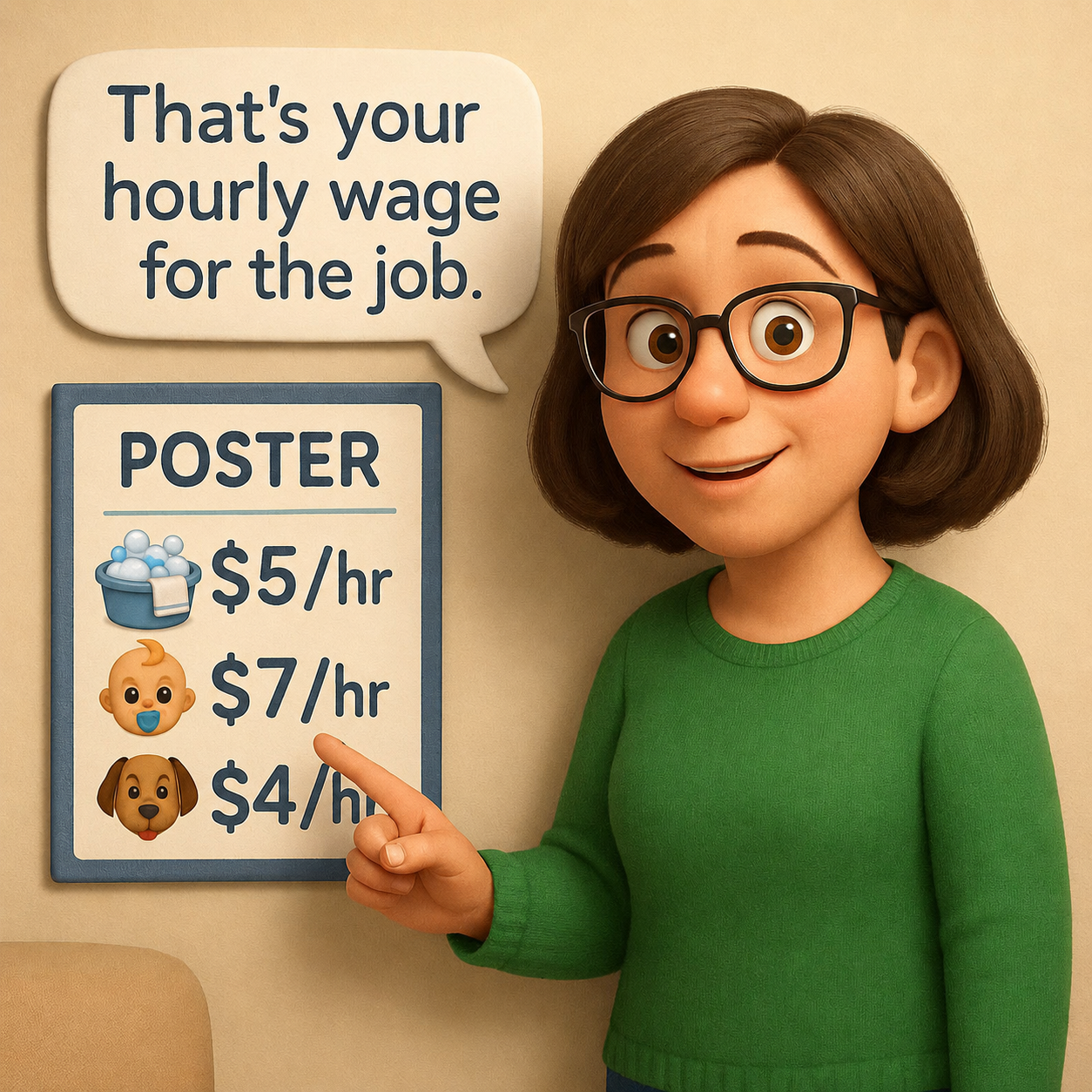}\\
  {\small Scene 4}
\end{minipage}\hfill
\begin{minipage}{0.18\linewidth}
  \centering
  \includegraphics[width=\linewidth]{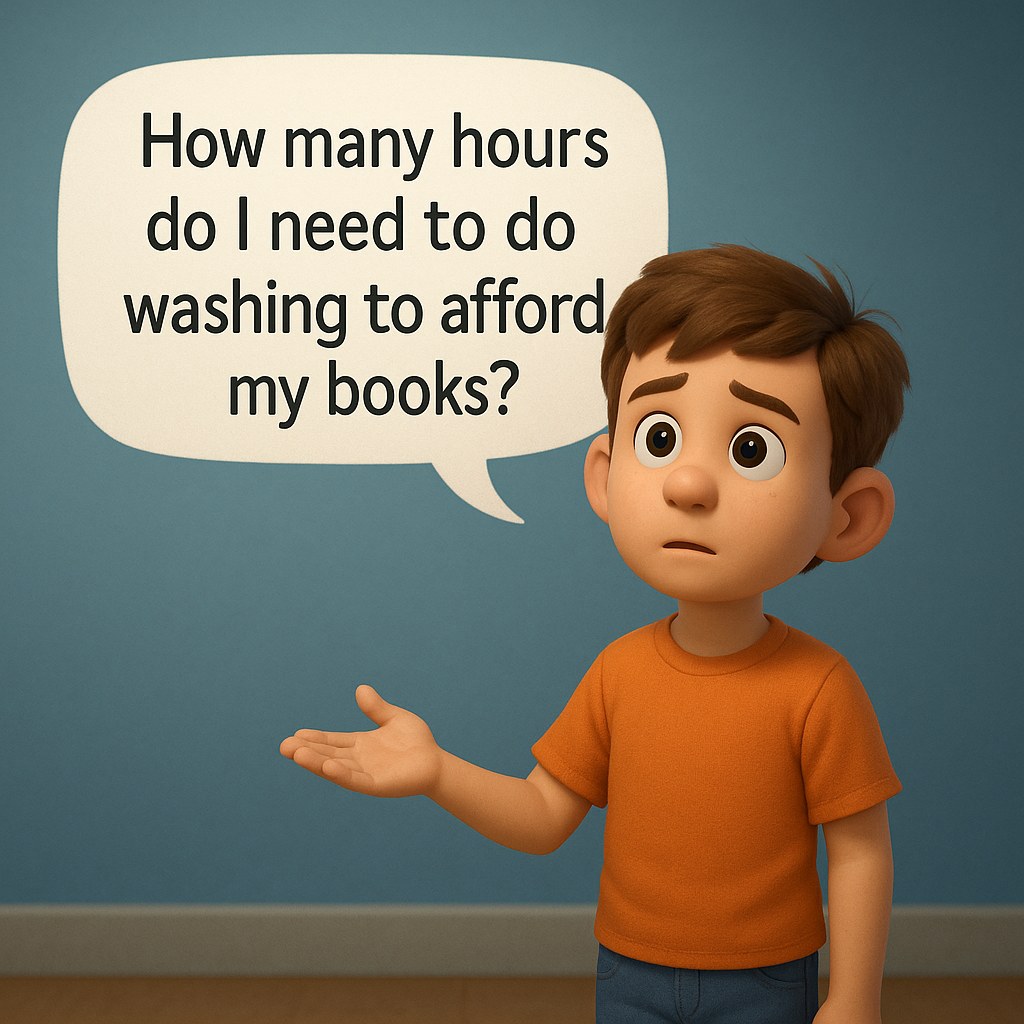}\\
  {\small Scene 5}
\end{minipage}

\end{subsecbox}

\begin{subsecbox}
\textbf{Ground Truth: }  \textcolor{green!60!black}{\textbf{5}}
\vspace{0.5mm}

Total cost $= 18 + 13 + 8 = \langle\!\langle 18+13+8=39 \rangle\!\rangle 39$; \#\#\# \hspace{0.5mm}
Money still needed $= 39 - 14 = \langle\!\langle 39-14=25 \rangle\!\rangle 25$; \#\#\# \hspace{0.5mm}
Hours needed $= 25 \div 5 = \langle\!\langle 25/5=5 \rangle\!\rangle \textcolor{green!60!black}{\mathbf{5}}$.

\end{subsecbox}

\begin{modelbox}[Gemini-2.5-Pro (\wrongtag)]{BadRed}{figures/logos/gemini-color.png}
\textbf{Answer:}

\begin{tcolorbox}[colback=LightGray!20, colframe=black!10, boxrule=0.3pt, arc=2pt, left=6pt, right=6pt, top=4pt, bottom=4pt]

Here is the step-by-step solution:

1.  **Calculate the total cost of the books:**
    *   The boy wants to buy three books: a dictionary, a dinosaur book, and a cookbook.
    *   The price list shows:
        *   Red book (Dictionary): \$18  
        *   Green book (Dinosaur): \$13  
        *   Yellow book (Cookbook): \$8  
    *   Total cost = \$18 + \$13 + \$8 = \$39.

2.  **Determine how much more money is needed:**
    *   The boy has already saved \$14.  
    *   Money still needed = Total cost - Money saved  
    *   Money still needed = \$39 - \$14 = \$25.

3.  **Identify the hourly wage for the job:**
    *   \textcolor{red!60!black}{In the fourth image, the woman is pointing to the job associated with the dog emoji, which pays \$4 per hour. She confirms this is the boy's wage.}
    *   \textcolor{red!60!black}{The boy then asks about the hours needed for "washing". Combining these facts, we can conclude his job is washing dogs for \$4 per hour.}

4.  **Calculate the number of hours needed to work:**
    *   To find out how many hours the boy needs to work, divide the amount of money he needs by his hourly wage.  
    *   \textcolor{red!60!black}{Hours to work = \$25 / \$4 = 6.25 hours.}

FINAL ANSWER: \textcolor{red!60!black}{6.25}

\end{tcolorbox}
\end{modelbox}

\end{casestudy}

\end{figure*}

\begin{figure*}[t]

\begin{casestudy}[Case Study 2: Signboard \& Icon (Price)]

\begin{modelbox}[GPT-4o (\wrongtag)]{BadRed}{figures/logos/openai.png}

\begin{tcolorbox}[colback=LightGray!20, colframe=black!10, boxrule=0.3pt, arc=2pt, 
left=6pt, right=6pt, top=1pt, bottom=1pt]

To determine how many hours Tyler needs to work, we need to follow these steps:

1. **Identify the Cost of the Books:**
   - Dictionary: \$18  
   - Dinosaur book: \$13  
   - Cookbook: \$8  

2. **Calculate the Total Cost of the Books:**
   \[
   18 + 13 + 8 = 39
   \]
   The total cost of the books is \$39.

3. **Determine Tyler's Current Savings:**
   - Tyler has saved \$14.

4. **Calculate the Additional Money Needed:**
   \[
   39 - 14 = 25
   \]
   Tyler needs an additional \$25.

5. **Identify Tyler's Hourly Wage for Washing:**
   - \textcolor{red!60!black}{Washing pays \$4 per hour.}

6. **Calculate the Number of Hours Needed:**
   \[
   \textcolor{red!60!black}{\frac{25}{4} = 6.25}
   \]
   \textcolor{red!60!black}{Since Tyler cannot work a fraction of an hour, he will need to work 7 hours to afford his books.}

FINAL ANSWER: \textcolor{red!60!black}{7}

\end{tcolorbox}
\end{modelbox}

\begin{modelbox}[Qwen2.5-VL-72B (\correcttag)]{OkGreen}{figures/logos/qwen-color.png}

\begin{tcolorbox}[colback=LightGray!20, colframe=black!10, boxrule=0.3pt, arc=2pt, 
left=6pt, right=6pt, top=1pt, bottom=1pt]

To solve this problem, we need to follow these steps:

1. Calculate the total cost of the books Tyler wants to buy.  
   The dictionary costs \$18, the dinosaur book costs \$13, and the children's cookbook costs \$8.  
   Adding these amounts together gives us the total cost of the books:  
   \[ 18 + 13 + 8 = 39 \]  
   So, the total cost of the books is \$39.

2. Determine how much more money Tyler needs.  
   Tyler has already saved \$14 from his allowance.  
   \[ 39 - 14 = 25 \]  
   So, Tyler needs an additional \$25.

3. Calculate how many hours Tyler needs to work.  
   Tyler earns \$5 per hour.  
   \[ 25 \div 5 = 5 \]  
   So, Tyler needs to work 5 hours.

FINAL ANSWER: \textcolor{green!60!black}{\textbf{5}}

\end{tcolorbox}
\end{modelbox}

\end{casestudy}
\end{figure*}

\begin{figure*}[t]
\centering
\begin{casestudy}[Case Study 3: Physical Metric (Weight)]

\begin{subsecbox}
\textbf{Text Problem.}
\vspace{0.5mm}

The basketball team went to the steakhouse to eat dinner. The first player ate a 6-ounce steak. The second player ate beef tips, containing 12 beef tips, each an ounce in size. The third player ate a one-pound steak. And the fourth and fifth players ordered vegetarian meals. In total, how many ounces of meat were consumed by the team?

\end{subsecbox}

\begin{subsecbox}
\textbf{GSM8K-V Images.}
\vspace{0.5mm}

\begin{minipage}{0.19\linewidth}
  \centering
  \includegraphics[width=\linewidth]{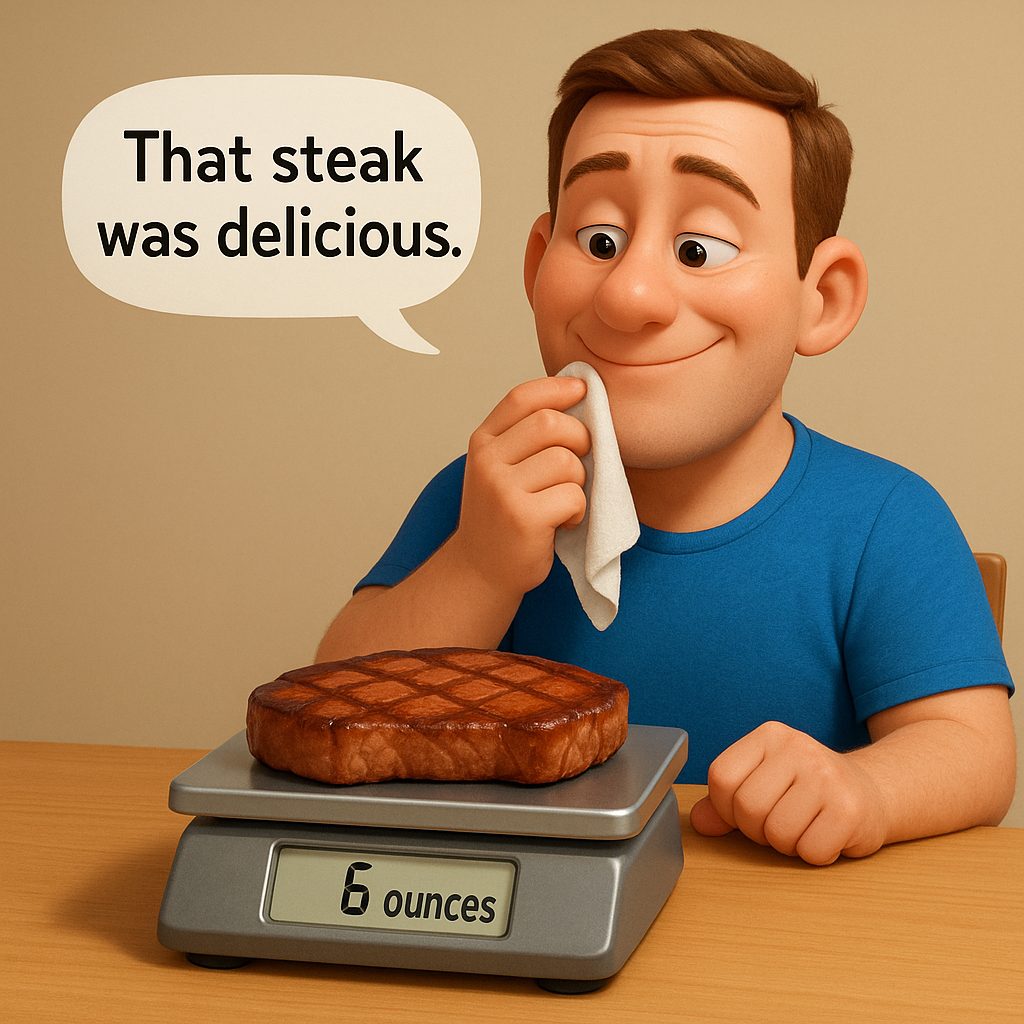}\\
  {\small Scene 1}
\end{minipage}\hfill
\begin{minipage}{0.19\linewidth}
  \centering
  \includegraphics[width=\linewidth]{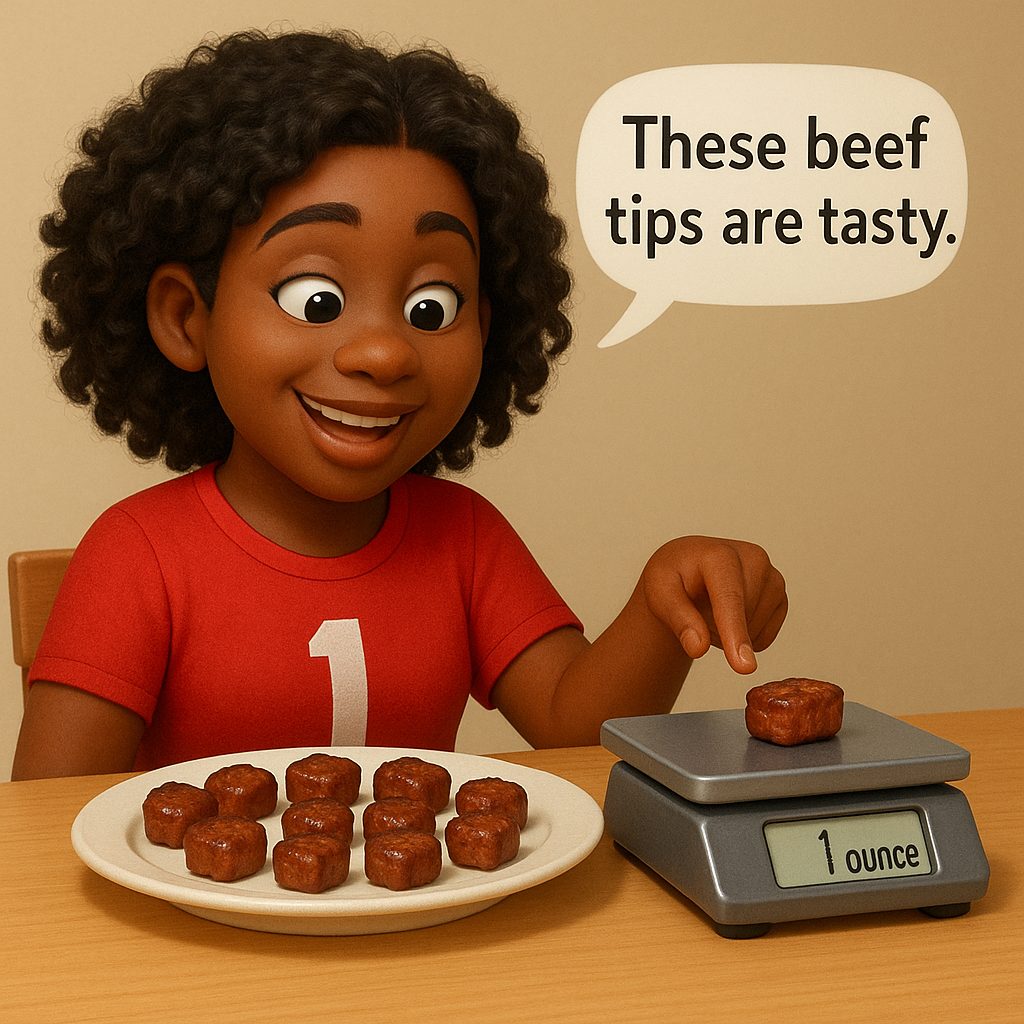}\\
  {\small Scene 2}
\end{minipage}\hfill
\begin{minipage}{0.19\linewidth}
  \centering
  \includegraphics[width=\linewidth]{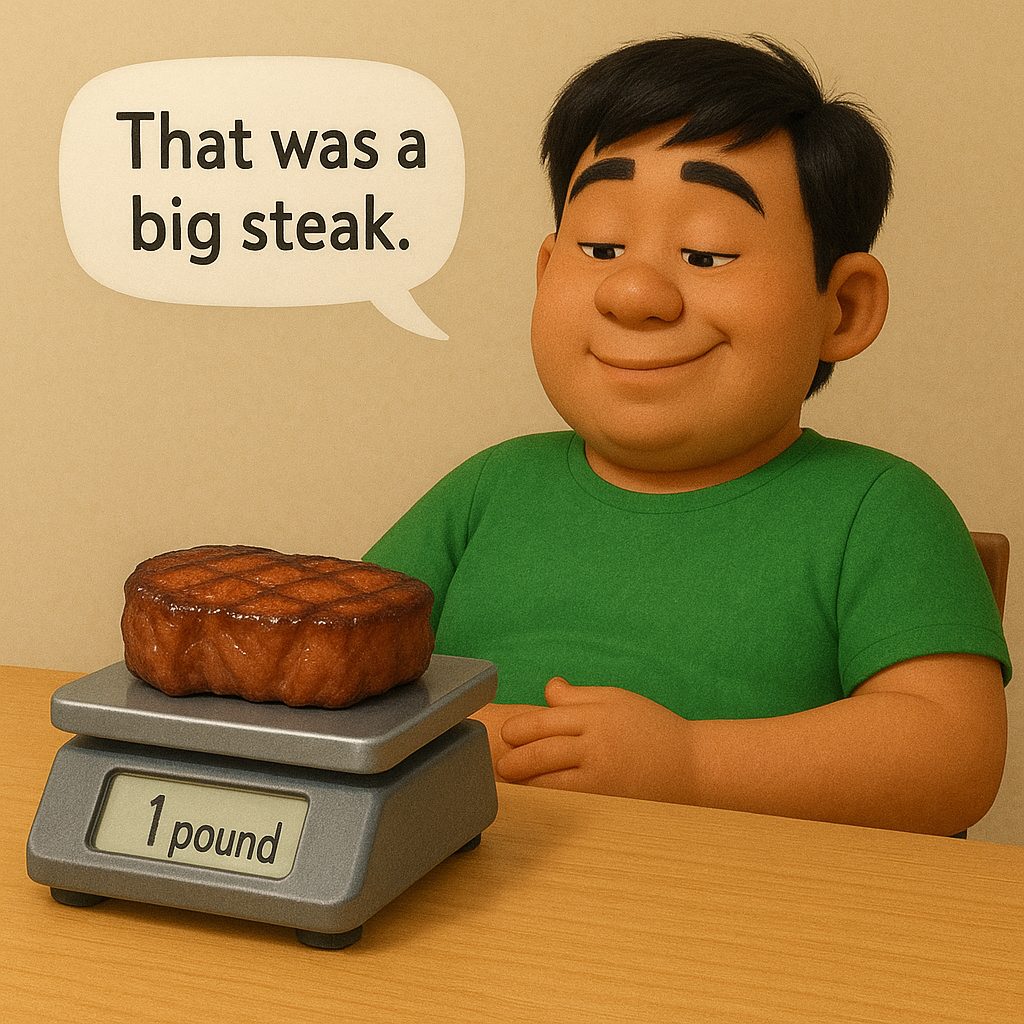}\\
  {\small Scene 3}
\end{minipage}\hfill
\begin{minipage}{0.19\linewidth}
  \centering
  \includegraphics[width=\linewidth]{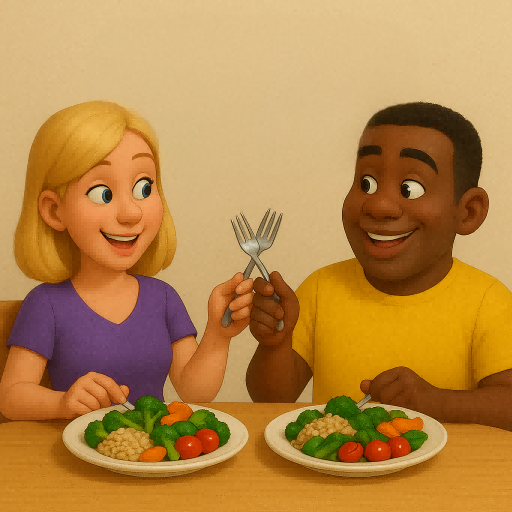}\\
  {\small Scene 4}
\end{minipage}\hfill
\begin{minipage}{0.19\linewidth}
  \centering
  \includegraphics[width=\linewidth]{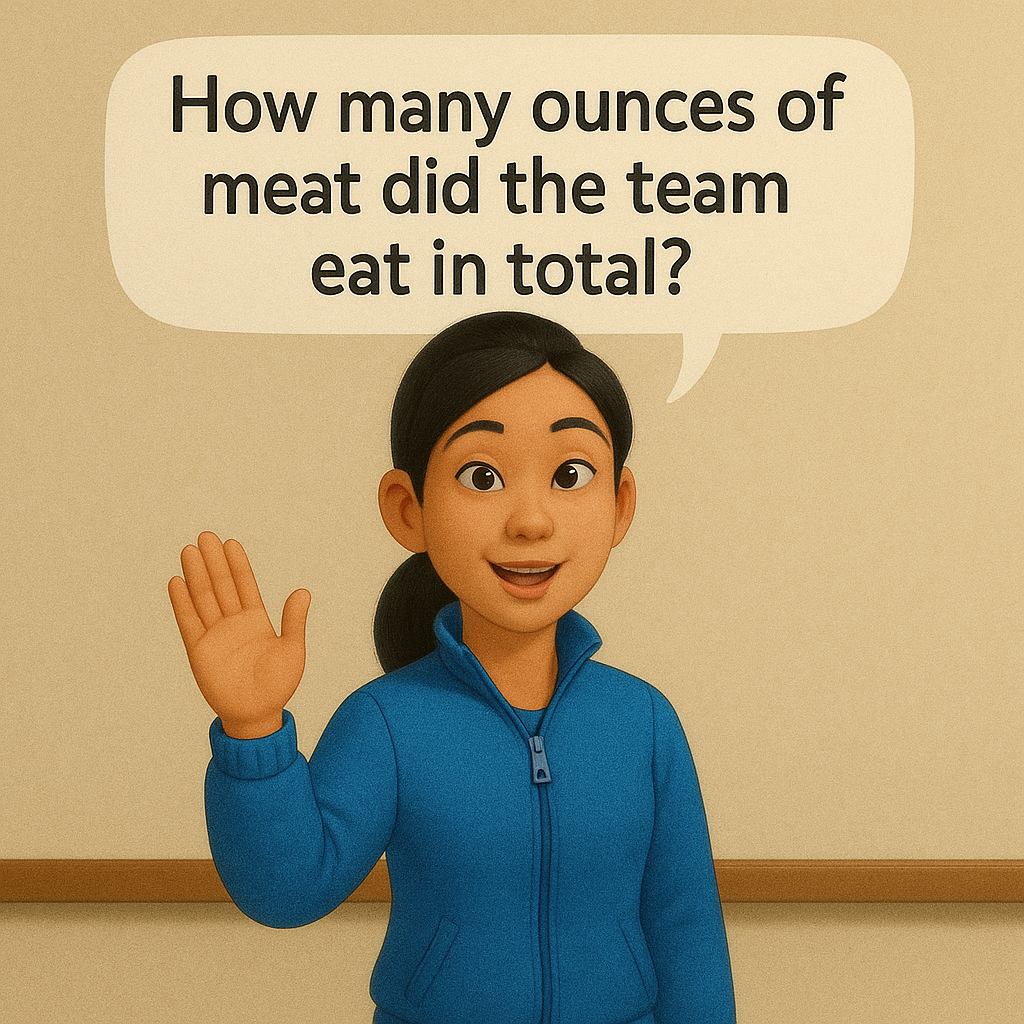}\\
  {\small Scene 5}
\end{minipage}

\end{subsecbox}

\begin{subsecbox}
\textbf{Ground Truth: }  \textcolor{green!60!black}{\textbf{34}}
\vspace{0.5mm}

Player 1: $6$ ounces; \#\#\# \hspace{0.5mm}
Player 2: $12 \times 1 = \langle\!\langle 12 \times 1 = 12 \rangle\!\rangle 12$ ounces; \#\#\# \hspace{0.5mm}
Player 3: $1$ pound $= \langle\!\langle 1 \text{ lb} = 16 \text{ oz}\rangle\!\rangle 16$ ounces; \#\#\# \hspace{0.5mm}
Total $= 6 + 12 + 16 = \langle\!\langle 34 = 34 \rangle\!\rangle \textcolor{green!60!black}{\mathbf{34}}$ ounces.

\end{subsecbox}

\begin{modelbox}[Gemini-2.5-Pro (\wrongtag)]{BadRed}{figures/logos/gemini-color.png}
\textbf{Answer:}

\begin{tcolorbox}[colback=LightGray!20, colframe=black!10, boxrule=0.3pt, arc=2pt, left=6pt, right=6pt, top=4pt, bottom=4pt]

*   **Step 1: Analyze the first image.** The first image shows a man who has eaten a steak. The steak is on a scale, and the scale reads 6 ounces. So, the first person ate 6 ounces of meat.

*   **Step 2: Analyze the second image.** The second image shows a girl eating beef tips. One beef tip is on a scale, which reads 1 ounce. On the plate, there are 9 more beef tips. In total, there are 1 + 9 = 10 beef tips. Since each beef tip weighs 1 ounce, the total weight of the beef tips is \textcolor{red!60!black}{10 * 1 = 10 ounces}. So, the second person ate \textcolor{red!60!black}{10 ounces} of meat.

*   **Step 3: Analyze the third image.** The third image shows a boy who has eaten a large steak. The steak is on a scale, and the scale reads 1 pound. The question asks for the total weight in ounces, so we need to convert pounds to ounces. There are 16 ounces in 1 pound. So, the third person ate 16 ounces of meat.

*   **Step 4: Analyze the fourth image.** The fourth image shows two people eating salads. Their meals consist of vegetables and grains, with no visible meat. Therefore, they did not eat any meat. Their contribution to the total is 0 ounces.

*   **Step 5: Calculate the total amount of meat eaten.**  
    Person 1: 6 ounces  
    Person 2: \textcolor{red!60!black}{10 ounces}  
    Person 3: 16 ounces  
    Total = \textcolor{red!60!black}{6 + 10 + 16 = 32}

FINAL ANSWER: \textcolor{red!60!black}{32}

\end{tcolorbox}
\end{modelbox}

\end{casestudy}

\end{figure*}

\begin{figure*}[t]
\begin{casestudy}[Case Study 3: Physical Metric (Weight)]

\begin{modelbox}[GPT-4o (\wrongtag)]{BadRed}{figures/logos/openai.png}
\textbf{Answer:}

\begin{tcolorbox}[colback=LightGray!20, colframe=black!10, boxrule=0.3pt, arc=2pt, 
left=6pt, right=6pt, top=4pt, bottom=4pt]

To find the total ounces of meat eaten by the basketball team, let's analyze each image:

1. **First Image**: The steak weighs 6 ounces.

2. **Second Image**: Each beef tip weighs 1 ounce. There are \textcolor{red!60!black}{10 beef tips} on the plate, so the total weight is:  
   \[
   \textcolor{red!60!black}{10 \times 1 = 10 \text{ ounces}}
   \]

3. **Third Image**: The steak weighs 1 pound. Since 1 pound equals 16 ounces, the weight is:  
   \[
   16 \text{ ounces}
   \]

4. **Fourth Image**: There is no meat shown, so we assume no meat was consumed by these individuals.

Now, add up all the ounces of meat:  
\[
\textcolor{red!60!black}{6 + 10 + 16 = 32 \text{ ounces}}
\]

FINAL ANSWER: \textcolor{red!60!black}{32}

\end{tcolorbox}
\end{modelbox}

\vspace{6pt}

\begin{modelbox}[Qwen2.5-VL-72B (\wrongtag)]{BadRed}{figures/logos/qwen-color.png}
\textbf{Answer:}

\begin{tcolorbox}[colback=LightGray!20, colframe=black!10, boxrule=0.3pt, arc=2pt, 
left=6pt, right=6pt, top=4pt, bottom=4pt]

To solve the problem, we need to determine the total amount of meat consumed by the team. Let's break down the information provided in the images:

1. The first image shows a person who ate a steak weighing 6 ounces.  
2. The second image shows a person who ate beef tips weighing \textcolor{red!60!black}{1 ounce}.  
3. The third image shows a person who ate a steak weighing 1 pound (which is equivalent to 16 ounces).

Now, let's add up the weights of the meat consumed:

- Steak from the first image: 6 ounces  
- Beef tips from the second image: \textcolor{red!60!black}{1 ounce}  
- Steak from the third image: 16 ounces  

Adding these together:  
\[
\textcolor{red!60!black}{6 + 1 + 16 = 23 \text{ ounces}}
\]

So, the total amount of meat consumed by the team is \textcolor{red!60!black}{23 ounces}.

FINAL ANSWER: \textcolor{red!60!black}{23}

\end{tcolorbox}
\end{modelbox}

\end{casestudy}
\end{figure*}

\newpage

\begin{figure*}[t]
\centering
\begin{casestudy}[Case Study 4: Temporal(Clock)]

\begin{subsecbox}
\textbf{Text Problem.}
\vspace{0.5mm}

Every day Charisma meditates for 15 minutes when she first wakes up and again before she goes to sleep. 
5 days a week she practices 1 hour of yoga. In 4 weeks, how much time has she spent on meditation/yoga practice?

\end{subsecbox}

\begin{subsecbox}
\textbf{GSM8K-V Images.}
\vspace{0.5mm}

\begin{minipage}{0.23\linewidth}
  \centering
  \includegraphics[width=\linewidth]{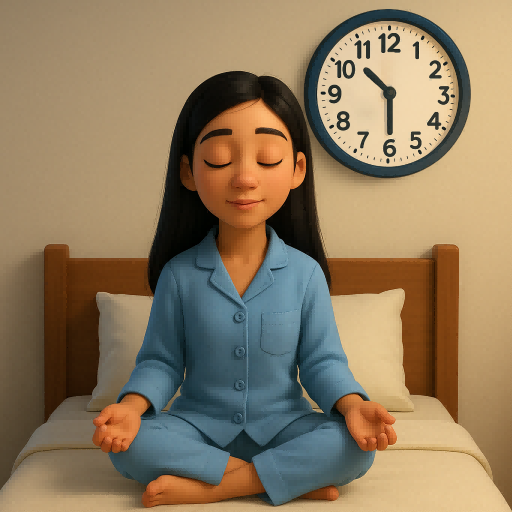}\\
  {\small Scene 1}
\end{minipage}\hfill
\begin{minipage}{0.23\linewidth}
  \centering
  \includegraphics[width=\linewidth]{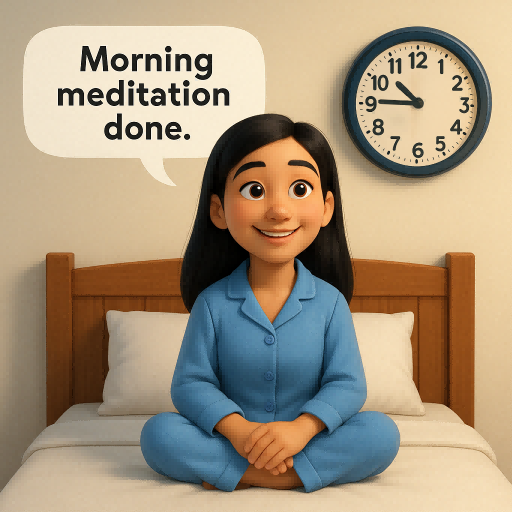}\\
  {\small Scene 2}
\end{minipage}\hfill
\begin{minipage}{0.23\linewidth}
  \centering
  \includegraphics[width=\linewidth]{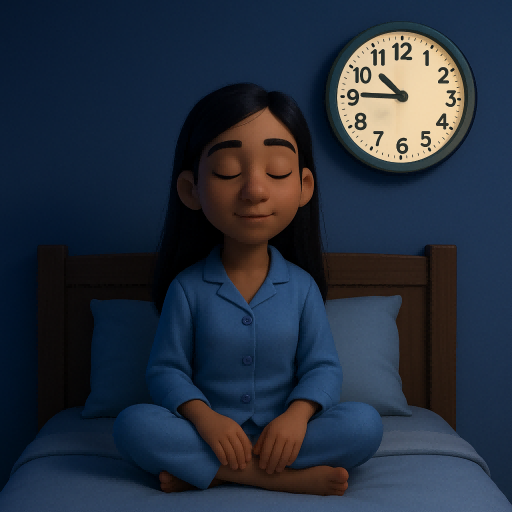}\\
  {\small Scene 3}
\end{minipage}\hfill
\begin{minipage}{0.23\linewidth}
  \centering
  \includegraphics[width=\linewidth]{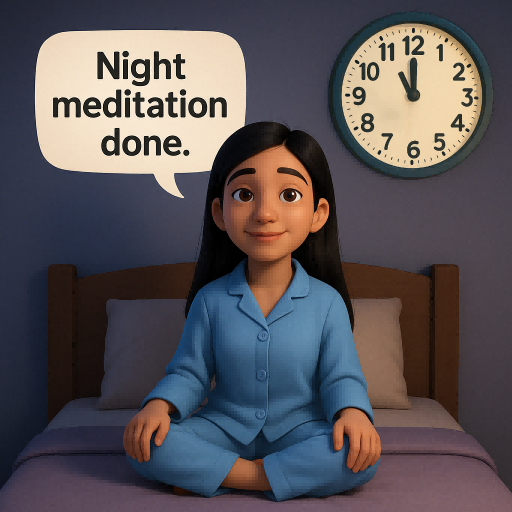}\\
  {\small Scene 4}
\end{minipage}

\vspace{2mm}

\begin{minipage}{0.3\linewidth}
  \centering
  \includegraphics[width=\linewidth]{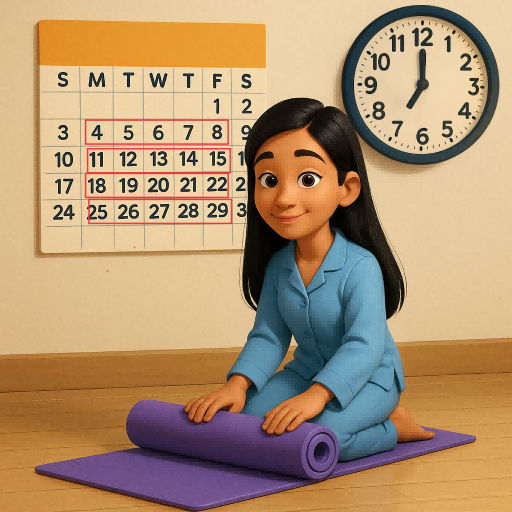}\\
  {\small Scene 5}
\end{minipage}\hfill
\begin{minipage}{0.3\linewidth}
  \centering
  \includegraphics[width=\linewidth]{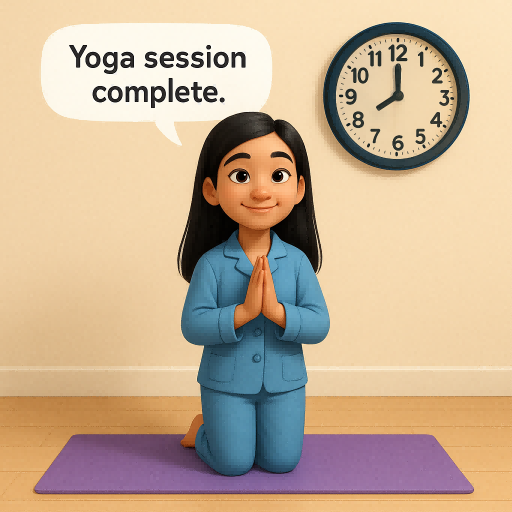}\\
  {\small Scene 6}
\end{minipage}\hfill
\begin{minipage}{0.3\linewidth}
  \centering
  \includegraphics[width=\linewidth]{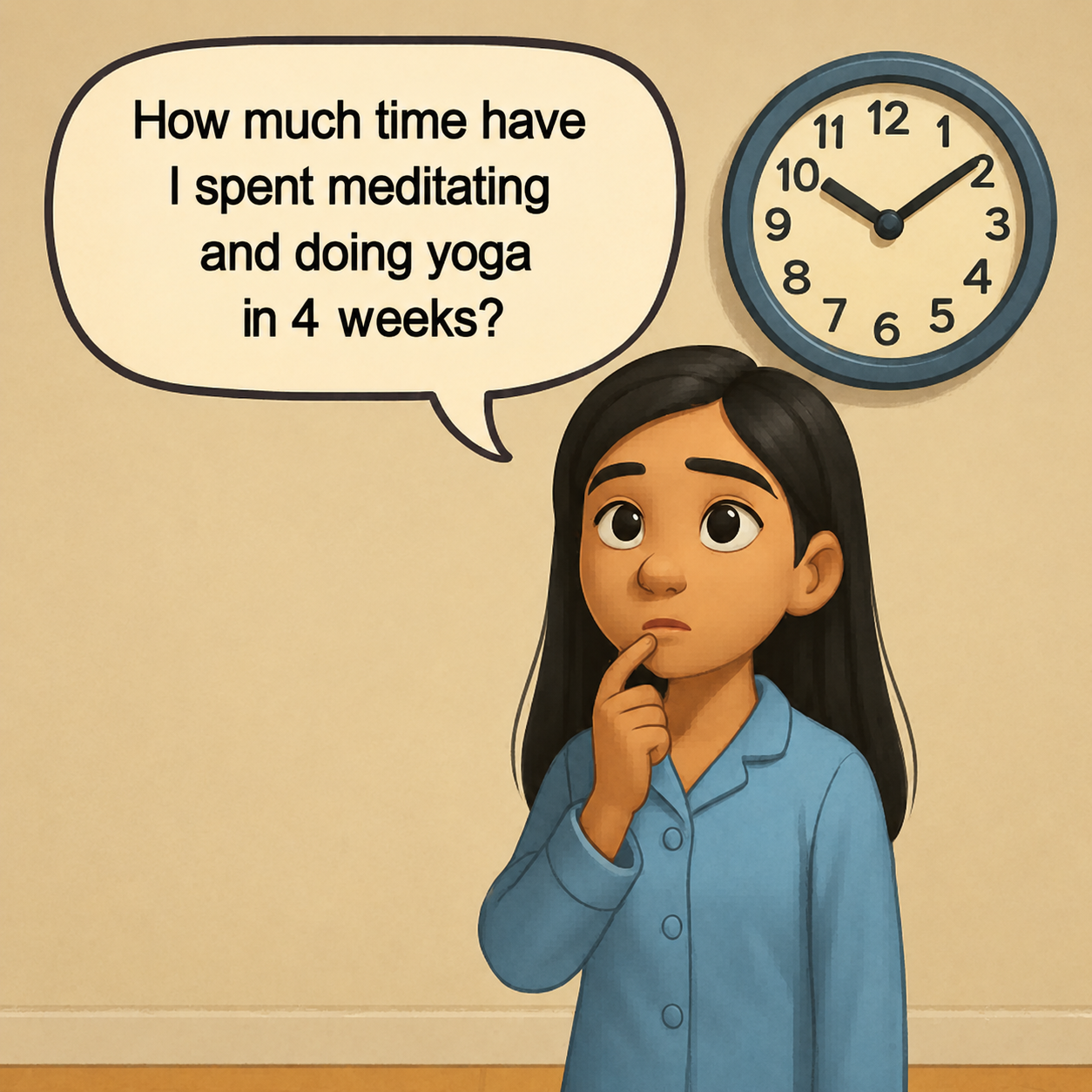}\\
  {\small Scene 7}
\end{minipage}

\end{subsecbox}

\begin{subsecbox}
\textbf{Ground Truth: }  \textcolor{green!60!black}{\textbf{34}}
\vspace{0.5mm}

She meditates for 15 minutes in the morning and 15 minutes at night for a total of $15+15 = \langle\!\langle 15+15=30 \rangle\!\rangle 30$ minutes.  
She meditates 30 minutes daily so in 7 days/1 week she meditates $30 \times 7 = \langle\!\langle 30 \times 7=210 \rangle\!\rangle 210$ minutes.  
She meditates 210 minutes in 1 week so in 4 weeks she meditates $210 \times 4 = \langle\!\langle 210 \times 4=840 \rangle\!\rangle 840$ minutes.  
$60$ minutes are in an hour so she meditates for $840/60 = \langle\!\langle 840/60=14 \rangle\!\rangle 14$ hours.  
She practices yoga for 1 hour 5 days a week so that's $1 \times 5 = \langle\!\langle 1 \times 5=5 \rangle\!\rangle 5$ hours.  
Over 4 weeks she practices yoga for $4 \times 5 = \langle\!\langle 4 \times 5=20 \rangle\!\rangle 20$ hours.  
She spends 14 hours meditating and 20 hours practicing yoga for a total of $14+20 = \langle\!\langle 14+20=34 \rangle\!\rangle \textcolor{green!60!black}{\mathbf{34}}$ hours.

\end{subsecbox}

\end{casestudy}

\end{figure*}

\begin{figure*}[t]

\begin{casestudy}[Case Study 4: Temporal(Clock)]

\begin{modelbox}[Gemini-2.5-Pro (\wrongtag)]{BadRed}{figures/logos/gemini-color.png}
\textbf{Answer:}

\begin{tcolorbox}[colback=LightGray!20, colframe=black!10, boxrule=0.3pt, arc=2pt, left=6pt, right=6pt, top=4pt, bottom=4pt]

Here is the step-by-step solution:

1.  **Calculate the duration of each morning meditation session:**  
    *   Start time: 6:00 (from the first image).  
    *   End time: 6:15 (from the second image).  
    *   Duration: 6:15 - 6:00 = \textcolor{green!60!black}{15 minutes}.  

2.  **Calculate the duration of each night meditation session:**  
    *   Start time: 9:45 (from the third image).  
    *   End time: 10:00 (from the fourth image).  
    *   Duration: 10:00 - 9:45 = \textcolor{green!60!black}{15 minutes}.  

3.  **Calculate the total daily meditation time:**  
    *   Total daily time = Morning meditation + Night meditation  
    *   Total daily time = 15 minutes + 15 minutes = \textcolor{green!60!black}{30 minutes}.  

4.  **Calculate the total meditation time over 4 weeks:**  
    *   Meditation is performed every day.  
    *   Number of days in 4 weeks = 4 weeks * 7 days/week = 28 days.  
    *   Total meditation time = 30 minutes/day * 28 days = \textcolor{green!60!black}{840 minutes}.  

5.  **Calculate the duration of each yoga session:**  
    *   Start time: 6:15 (from the fifth image).  
    *   End time: 6:45 (from the sixth image).  
    *   Duration: 6:45 - 6:15 = \textcolor{red!60!black}{30 minutes}.  

6.  **Calculate the total yoga time over 4 weeks:**  
    *   The calendar shows yoga is done on weekdays (Monday to Friday).  
    *   Number of yoga sessions in 4 weeks = 5 days/week * 4 weeks = 20 sessions.  
    *   Total yoga time = 30 minutes/session * 20 sessions = \textcolor{red!60!black}{600 minutes}.  

7.  **Calculate the total time spent on all activities:**  
    *   Grand total time = Total meditation time + Total yoga time  
    *   Grand total time = 840 minutes + 600 minutes = \textcolor{red!60!black}{1440 minutes}.  

8.  **Convert the total time from minutes to hours:**  
    *   There are 60 minutes in 1 hour.  
    *   Total hours = 1440 minutes / 60 minutes/hour = \textcolor{red!60!black}{24 hours}.  

The total time spent meditating and doing yoga in 4 weeks is \textcolor{red!60!black}{24 hours}.

\medskip
FINAL ANSWER: \textcolor{red!60!black}{24}

\end{tcolorbox}
\end{modelbox}

\end{casestudy}

\end{figure*}

\begin{figure*}[t]

\begin{casestudy}[Case Study 4: Temporal(Clock)]

\begin{modelbox}[GPT-4o (\wrongtag)]{BadRed}{figures/logos/openai.png}
\textbf{Answer:}

\begin{tcolorbox}[colback=LightGray!20, colframe=black!10, boxrule=0.3pt, arc=2pt, left=6pt, right=6pt, top=4pt, bottom=4pt]

To solve this problem, we need to calculate the total time spent meditating and doing yoga over 4 weeks.

1. **Morning Meditation:**  
   - Starts at 6:00 AM and ends at 6:30 AM.  
   - Duration: \textcolor{red!60!black}{30 minutes}.  

2. **Night Meditation:**  
   - Starts at 9:00 PM and ends at 9:30 PM.  
   - Duration: \textcolor{red!60!black}{30 minutes}.  

3. **Yoga Session:**  
   - Starts at 5:00 PM and ends at 5:30 PM.  
   - Duration: \textcolor{red!60!black}{30 minutes}.  

Each activity is done daily.

**Total Daily Time:**  
- Morning Meditation: 30 minutes  
- Night Meditation: 30 minutes  
- Yoga Session: 30 minutes  

Total daily time spent = \textcolor{red!60!black}{30 + 30 + 30 = 90 minutes}  

**Total Time for 4 Weeks:**  
- 4 weeks = 28 days  
- Total time = 90 minutes/day × 28 days = \textcolor{red!60!black}{2520 minutes}  

Convert minutes to hours:  
- 2520 minutes ÷ 60 = \textcolor{red!60!black}{42 hours}  

\medskip
FINAL ANSWER: \textcolor{red!60!black}{42}

\end{tcolorbox}
\end{modelbox}

\end{casestudy}

\begin{casestudy}[Case Study 4: Temporal(Clock)]

\begin{modelbox}[Qwen2.5-VL-72b (\wrongtag)]{BadRed}{figures/logos/qwen-color.png}
\textbf{Answer:}

\begin{tcolorbox}[colback=LightGray!20, colframe=black!10, boxrule=0.3pt, arc=2pt, left=6pt, right=6pt, top=4pt, bottom=4pt]

To solve the problem, we need to determine how much time the person has spent meditating and doing yoga over a period of 4 weeks.

\textbf{Step-by-Step Reasoning:}

1. **Identify the daily meditation and yoga times:**  
   - Morning meditation: The clock shows 7:00 AM.  
   - Night meditation: The clock shows 9:00 PM.  
   - Yoga session: The clock shows 6:00 AM.  

2. **Calculate the duration of each activity:**  
   - Morning meditation: Assume it lasts for \textcolor{red!60!black}{30 minutes (common duration)}.  
   - Night meditation: Assume it also lasts for \textcolor{red!60!black}{30 minutes}.  
   - Yoga session: Assume it lasts for \textcolor{green!60!black}{60 minutes (common duration)}.  

3. **Total daily time spent on activities:**  
   - Morning meditation: 30 minutes  
   - Night meditation: 30 minutes  
   - Yoga session: 60 minutes  
   - Total daily time = \textcolor{red!60!black}{30 + 30 + 60 = 120 minutes}  

4. **Convert daily time to hours:**  
   - 120 minutes = \textcolor{red!60!black}{2 hours}  

5. **Calculate the total time over 4 weeks:**  
   - Number of days in 4 weeks = 4 * 7 = 28 days  
   - Total time = 2 hours/day * 28 days = \textcolor{red!60!black}{56 hours}  

\medskip
FINAL ANSWER: \textcolor{red!60!black}{56}

\end{tcolorbox}
\end{modelbox}

\end{casestudy}
\end{figure*}

\begin{figure*}[t]
\centering
\begin{casestudy}[Case Study 5: Ratio \& Percentage (Graph)]

\newpage

\begin{subsecbox}
\textbf{Text Problem.}
\vspace{0.5mm}

There are 1.5 times as many boys as girls at Dr. Wertz's school. If there are 60 girls and 5 students to every teacher, how many teachers are there?

\end{subsecbox}

\begin{subsecbox}
\textbf{GSM8K-V Images.}
\vspace{0.5mm}

\begin{minipage}{0.24\linewidth}
  \centering
  \includegraphics[width=\linewidth]{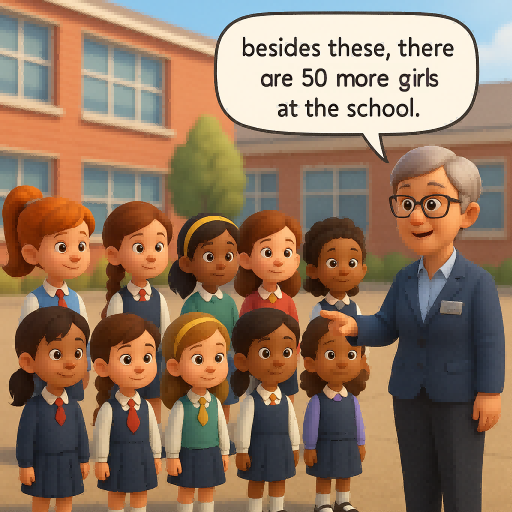}\\
  {\small Scene 1}
\end{minipage}\hfill
\begin{minipage}{0.24\linewidth}
  \centering
  \includegraphics[width=\linewidth]{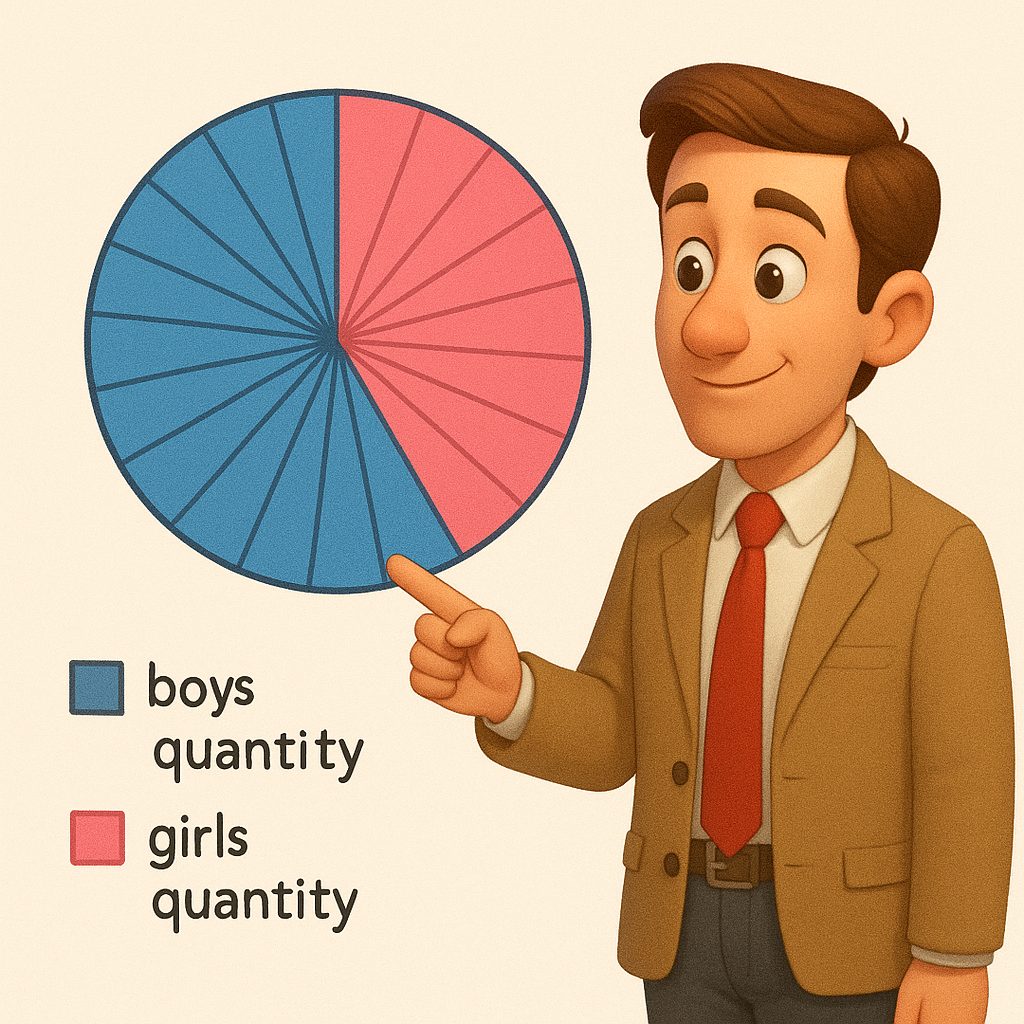}\\
  {\small Scene 2}
\end{minipage}\hfill
\begin{minipage}{0.24\linewidth}
  \centering
  \includegraphics[width=\linewidth]{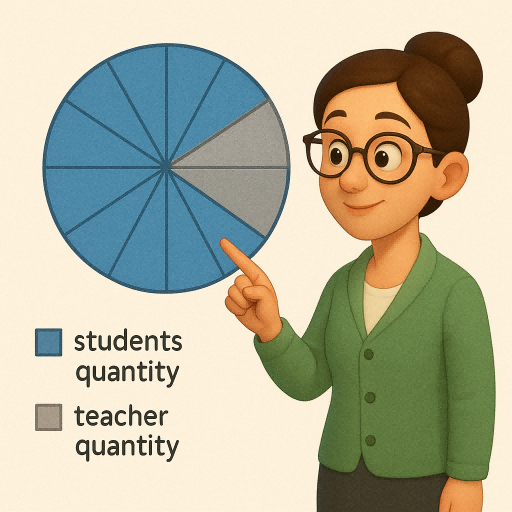}\\
  {\small Scene 3}
\end{minipage}\hfill
\begin{minipage}{0.24\linewidth}
  \centering
  \includegraphics[width=\linewidth]{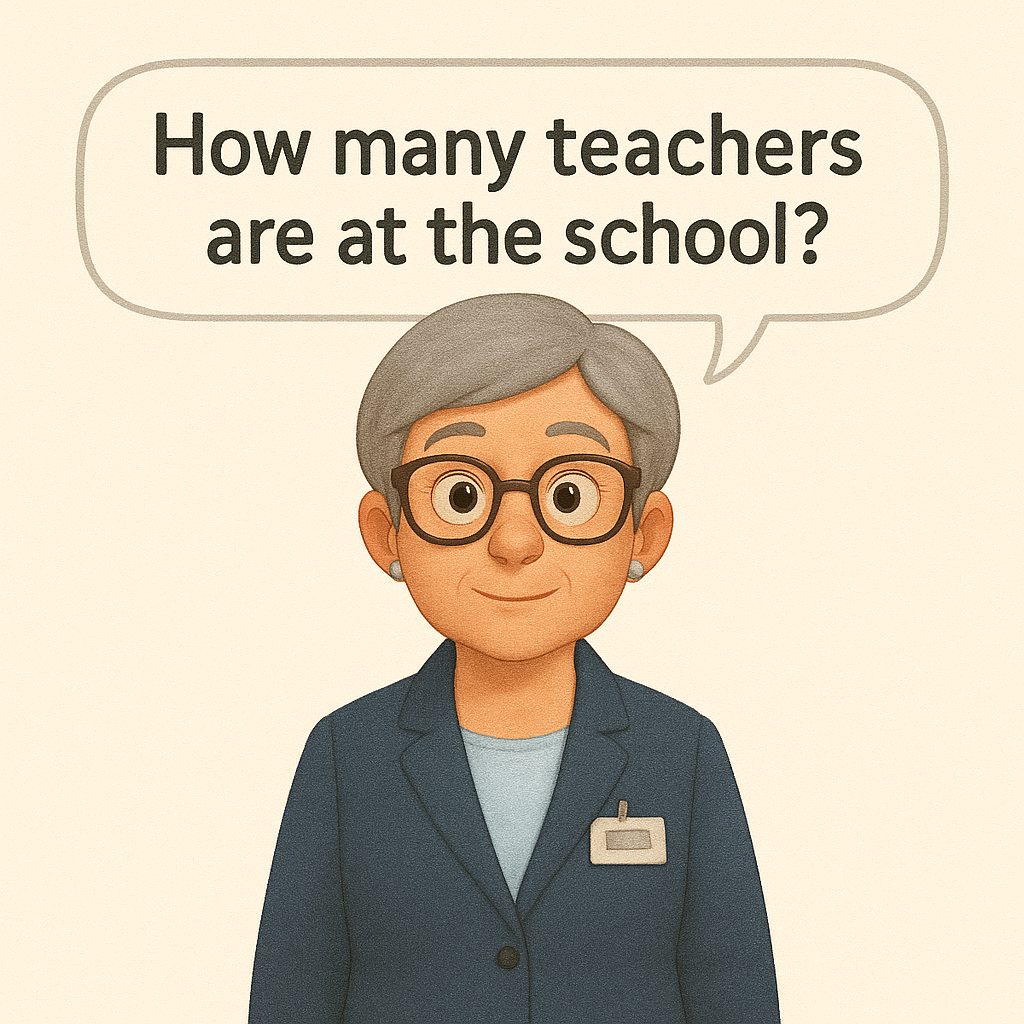}\\
  {\small Scene 4}
\end{minipage}

\end{subsecbox}

\begin{subsecbox}
\textbf{Ground Truth: }  \textcolor{green!60!black}{\textbf{30}}
\vspace{0.5mm}

There are $1.5 \times 60 = \langle\!\langle 1.5 \times 60 = 90 \rangle\!\rangle 90$ boys.  
Total students $= 90 + 60 = \langle\!\langle 90+60=150 \rangle\!\rangle 150$.  
Each teacher has 5 students, so teachers $= 150/5 = \langle\!\langle 150/5=30 \rangle\!\rangle \textcolor{green!60!black}{\mathbf{30}}$.

\end{subsecbox}

\begin{modelbox}[Gemini-2.5-Pro (\correcttag)]{GoodGreen}{figures/logos/gemini-color.png}
\textbf{Answer:}

\begin{tcolorbox}[colback=LightGray!20, colframe=black!10, boxrule=0.3pt, arc=2pt, left=6pt, right=6pt, top=4pt, bottom=4pt]

Here is a step-by-step solution to the problem:

1.  **Find the total number of girls.**  
    *   The first image shows a group of girls. By counting them, we see there are 10 girls in the picture (5 in the back row and 5 in the front row).  
    *   The teacher says, "besides these, there are 50 more girls at the school."  
    *   So, the total number of girls is the sum of the girls in the picture and the additional girls: $10 + 50 = \textcolor{green!60!black}{60}$ girls.  

2.  **Find the total number of students.**  
    *   The second image shows a pie chart representing the proportion of boys and girls at the school.  
    *   The chart is divided into slices. Let's count them:  
        *   Pink slices (girls): 8  
        *   Blue slices (boys): 12  
    *   The total number of slices is $8 + 12 = 20$.  
    *   This means that girls make up $8/20 = 2/5$ of the student population.  
    *   We know from step 1 that there are 60 girls.  
    *   $(2/5) \times S = 60 \;\Rightarrow\; S = 150$.  
    *   There are \textcolor{green!60!black}{150 students} in total at the school.  

3.  **Find the number of teachers.**  
    *   The third image shows a pie chart representing the proportion of students and teachers at the school.  
    *   Let's count the slices in this chart:  
        *   Blue slices (students): 10  
        *   Gray slices (teachers): 2  
    *   This shows that the ratio of students to teachers is 10 to 2, which simplifies to 5 to 1.  
    *   We know from step 2 that there are 150 students.  
    *   $150 / T = 5/1 \;\Rightarrow\; T = 30$.  

4.  **Answer the final question.**  
    *   The last image asks, "How many teachers are at the school?"  
    *   Based on our calculation, there are \textcolor{green!60!black}{30 teachers} at the school.  

\medskip
FINAL ANSWER: \textcolor{green!60!black}{30}

\end{tcolorbox}
\end{modelbox}

\end{casestudy}

\end{figure*}

\begin{figure*}[t]

\begin{casestudy}[Case Study 5: Ratio \& Percentage (Graph)]

\begin{modelbox}[GPT-4o (\wrongtag)]{BadRed}{figures/logos/openai.png}
\textbf{Answer:}

\begin{tcolorbox}[colback=LightGray!20, colframe=black!10, boxrule=0.3pt, arc=2pt, left=6pt, right=6pt, top=4pt, bottom=4pt]

To solve the problem, let's analyze the images step by step:

1. **First Image**: There are 10 girls shown, and the speech bubble mentions there are 50 more girls at the school. So, the total number of girls is:  
\[
10 + 50 = 60
\]

2. **Second Image**: The pie chart shows the ratio of boys to girls. There are \textcolor{red!60!black}{12 sections in total, with 3 sections for girls and 9 sections for boys}.  
This means the ratio of boys to girls is \textcolor{red!60!black}{9:3 = 3:1}.  
Therefore, the number of boys is:  
\[
\textcolor{red!60!black}{3 \times 60 = 180}
\]

3. **Third Image**: The pie chart shows the ratio of students to teachers. There are \textcolor{red!60!black}{10 sections in total, with 8 sections for students and 2 sections for teachers}.  
This means the ratio of students to teachers is \textcolor{red!60!black}{8:2 = 4:1}.

4. **Calculate Total Students**:  
\[
\textcolor{red!60!black}{180 + 60 = 240}
\]

5. **Calculate Number of Teachers**:  
\[
\textcolor{red!60!black}{240 / 4 = 60}
\]

\medskip
FINAL ANSWER: \textcolor{red!60!black}{60}

\end{tcolorbox}
\end{modelbox}

\begin{modelbox}[Qwen2.5-VL-72b (\wrongtag)]{BadRed}{figures/logos/qwen-color.png}
\textbf{Answer:}

\begin{tcolorbox}[colback=LightGray!20, colframe=black!10, boxrule=0.3pt, arc=2pt, left=6pt, right=6pt, top=4pt, bottom=4pt]

Let's break down the problem step by step using the information provided in the images.

Step 1: Determine the number of girls and boys at the school.
- From the first image, we see a group of 10 girls. The speech bubble indicates that there are 50 more girls at the school.  
  - Total number of girls = 10 (in the image) + 50 (additional girls) = \textcolor{green!60!black}{60} girls.  
- The second image shows a pie chart with sections for boys and girls. The section for girls is larger than the section for boys, but it doesn't give exact numbers.  
  - Let's assume the total number of students is $T$.  
  - If the pie chart is divided into 10 segments and girls occupy 6 segments, then boys occupy 4 segments.  
  - This means the ratio of girls to boys is \textcolor{red!60!black}{6:4 = 3:2}.  
  - So, if girls are 60, boys would be \textcolor{red!60!black}{$(2/3) \times 60 = 40$}.  
  - Total number of students $T = \textcolor{red!60!black}{60 + 40 = 100}$.  

Step 2: Determine the number of teachers.
- The third image shows a pie chart with a small grey section representing teachers.  
- If the grey section is \textcolor{red!60!black}{$1/10$ of the pie chart}, then the number of teachers is \textcolor{red!60!black}{$(1/10) \times 100 = 10$}.  

\medskip
FINAL ANSWER: \textcolor{red!60!black}{10}

\end{tcolorbox}
\end{modelbox}

\end{casestudy}
\end{figure*}

\begin{figure*}[t]
\centering
\begin{casestudy}[Case Study 6: Other (Count)]

\begin{subsecbox}
\textbf{Text Problem.}
\vspace{0.5mm}

Janet’s ducks lay 16 eggs per day. She eats three for breakfast every morning and bakes muffins for her friends every day with four. She sells the remainder at the farmers' market daily for \$2 per fresh duck egg. How much in dollars does she make every day at the farmers' market?

\end{subsecbox}

\begin{subsecbox}
\textbf{GSM8K-V Images.}
\vspace{0.5mm}

\begin{minipage}{0.19\linewidth}
  \centering
  \includegraphics[width=\linewidth]{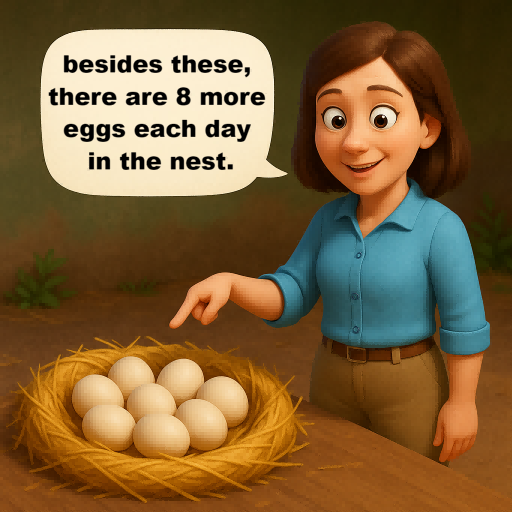}\\
  {\small Scene 1}
\end{minipage}\hfill
\begin{minipage}{0.19\linewidth}
  \centering
  \includegraphics[width=\linewidth]{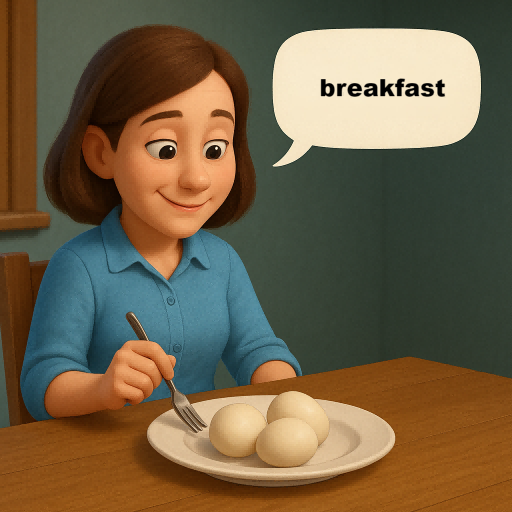}\\
  {\small Scene 2}
\end{minipage}\hfill
\begin{minipage}{0.19\linewidth}
  \centering
  \includegraphics[width=\linewidth]{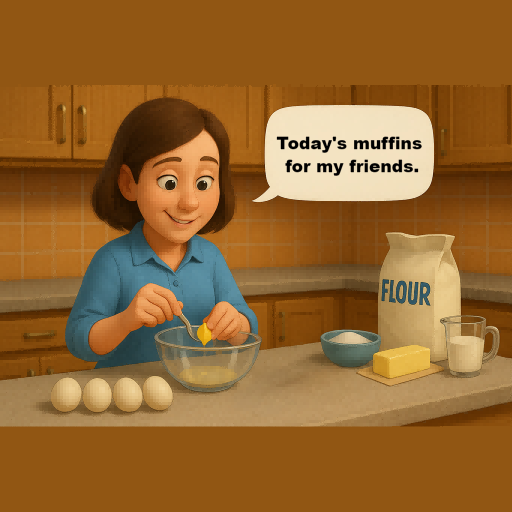}\\
  {\small Scene 3}
\end{minipage}\hfill
\begin{minipage}{0.19\linewidth}
  \centering
  \includegraphics[width=\linewidth]{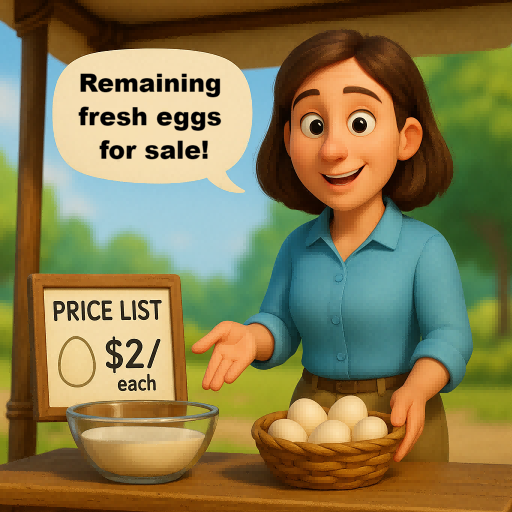}\\
  {\small Scene 4}
\end{minipage}\hfill
\begin{minipage}{0.19\linewidth}
  \centering
  \includegraphics[width=\linewidth]{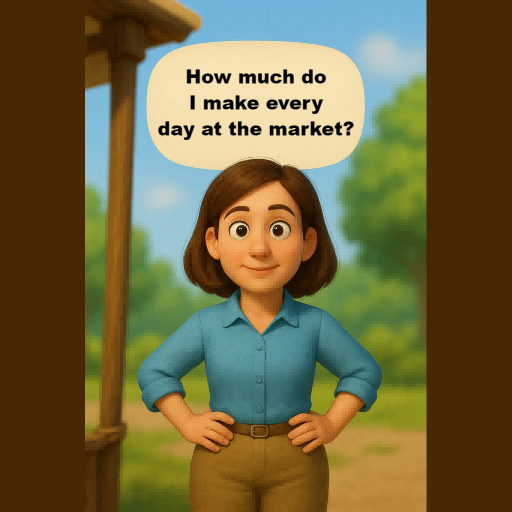}\\
  {\small Scene 5}
\end{minipage}

\end{subsecbox}

\begin{subsecbox}
\textbf{Ground Truth: }  \textcolor{green!60!black}{\textbf{18}}
\vspace{0.5mm}

Janet sells $16 - 3 - 4 = \langle\!\langle 16-3-4=9 \rangle\!\rangle 9$ duck eggs a day.  
She makes $9 \times 2 = \$\langle\!\langle 9*2=18 \rangle\!\rangle \textcolor{green!60!black}{\mathbf{18}}$ every day at the farmers' market.

\end{subsecbox}

\begin{modelbox}[Gemini-2.5-Pro (\wrongtag)]{BadRed}{figures/logos/gemini-color.png}
\textbf{Answer:}

\begin{tcolorbox}[colback=LightGray!20, colframe=black!10, boxrule=0.3pt, arc=2pt, left=6pt, right=6pt, top=4pt, bottom=4pt]

Here is a step-by-step solution based on the information in the images:

1.    **Calculate the total number of eggs collected each day.  **  
    *   In the first image, there are \textcolor{red!60!black}{7 eggs} in the nest.  
    *   The woman states that besides those, there are \textcolor{green!60!black}{8 more eggs} each day.  
    *   Total daily eggs = \textcolor{red!60!black}{7 + 8 = 15 eggs}.  

2.    **Calculate the number of eggs used each day.  **  
    *   In the second image, \textcolor{green!60!black}{she eats 3 eggs for breakfast.}
    *   In the third image, \textcolor{green!60!black}{she uses 4 eggs to bake muffins. }
    *   \textcolor{green!60!black}{Total eggs used = 3 + 4 = 7 eggs.  }

3.    **Determine the number of eggs available to sell.  **  
    *   Eggs for sale = \textcolor{red!60!black}{15 - 7 = 8 eggs}.  

4.    **Calculate the total earnings from the market.  **  
    *   Each egg is sold for \$2.  
    *   Total earnings = \textcolor{red!60!black}{8 × 2 = 16}.  

FINAL ANSWER: \textcolor{red!60!black}{16}

\end{tcolorbox}
\end{modelbox}

\end{casestudy}

\end{figure*}

\begin{figure*}[t]

\begin{casestudy}[Case Study 6: Other (Count)]

\begin{modelbox}[GPT-4o (\wrongtag)]{BadRed}{figures/logos/openai.png}
\textbf{Answer:}

\begin{tcolorbox}[colback=LightGray!20, colframe=black!10, boxrule=0.3pt, arc=2pt, left=6pt, right=6pt, top=4pt, bottom=4pt]

To solve the problem, let's analyze the images step by step:

1.  **First Image**: There are 8 eggs in the nest, \textcolor{red!60!black}{ and 8 more eggs are added each day.}

2.  **Second Image**: \textcolor{green!60!black}{She eats 3 eggs for breakfast.}

3.  **Third Image**: \textcolor{green!60!black}{She uses 4 eggs to make muffins.}

4.  **Fourth Image**: She sells eggs at \$2 each.

5.  **Fifth Image**: We need to calculate how much she makes every day at the market. 

**Calculations**:

- Total eggs produced each day: \textcolor{red!60!black}{8 eggs}.  
- Eggs used for breakfast: 3 eggs.  
- Eggs used for muffins: 4 eggs.  
- Total eggs used: 3 + 4 = 7 eggs.  
- Eggs available for sale: \textcolor{red!60!black}{8 - 7 = 1 egg}.  
- Income from selling eggs: \textcolor{red!60!black}{1 × \$2 = \$2}.  

FINAL ANSWER: \textcolor{red!60!black}{2}

\end{tcolorbox}
\end{modelbox}

\vspace{6pt}

\begin{modelbox}[Qwen2.5-VL-72b (\wrongtag)]{BadRed}{figures/logos/qwen-color.png}
\textbf{Answer:}

\begin{tcolorbox}[colback=LightGray!20, colframe=black!10, boxrule=0.3pt, arc=2pt, left=6pt, right=6pt, top=4pt, bottom=4pt]

To solve the problem, let's break down the information given in the images step by step:

1. **Eggs in the nest**: The first image states that there are \textcolor{red!60!black}{8 more eggs each day} in the nest.  
2. **Breakfast**: The second image shows that some eggs are used for breakfast, \textcolor{red!60!black}{but it doesn't specify how many.}  
3. **Muffins**: The third image indicates that some eggs are used to make muffins for friends, \textcolor{red!60!black}{but again, it doesn't specify the number.}
4. **Selling eggs**: The fourth image shows that she sells fresh duck eggs for \$2 each.  
5. **Daily earnings**: The fifth image asks how much she makes every day at the market.  

\textcolor{red!60!black}{Since the exact number of eggs used for breakfast and muffins is not specified, we'll assume all 8 eggs are sold each day for simplicity.}

- Each egg is sold for \$2.  
- She has \textcolor{red!60!black}{8 eggs} each day.  

Now, calculate the daily earnings:  
\[
\text{Daily earnings} = 8 \times 2 = \textcolor{red!60!black}{16}
\]

Thus, her daily earnings from selling eggs are \textcolor{red!60!black}{\$16}.

FINAL ANSWER: \textcolor{red!60!black}{16}

\end{tcolorbox}
\end{modelbox}

\end{casestudy}

\end{figure*}

\end{document}